\documentclass{zlab}

\usepackage[T1]{fontenc}
\usepackage{tgpagella}
\usepackage{mathpazo}
\usepackage{inconsolata}

\usepackage{xspace}
\usepackage{graphicx,amsmath,amssymb,hyperref}
\usepackage[authoryear,round]{natbib}

\usepackage{graphicx}
\usepackage{booktabs}

\usepackage[export]{adjustbox}
\usepackage{subcaption}
\usepackage{url}
\usepackage{booktabs}
\usepackage[table,xcdraw,dvipsnames]{xcolor}
\usepackage{makecell}
\usepackage{tabularx}
\usepackage{graphicx}
\usepackage{array}
\usepackage{multirow}
\usepackage{float}
\usepackage{etoc}
\usepackage{amsmath}
\usepackage{wrapfig}
\usepackage{adjustbox}

\usepackage[scaled]{helvet}

\renewcommand{\paragraph}[1]{\vspace{1.25mm}\noindent\textbf{#1}}

\newcolumntype{x}[1]{>{\centering\arraybackslash}p{#1pt}}
\newcolumntype{y}[1]{>{\raggedright\arraybackslash}p{#1pt}}
\newcolumntype{z}[1]{>{\raggedleft\arraybackslash}p{#1pt}}

\newlength\savewidth\newcommand\shline{\noalign{\global\savewidth\arrayrulewidth
  \global\arrayrulewidth 1pt}\hline\noalign{\global\arrayrulewidth\savewidth}}

\newcommand{\tablestyle}[2]{\setlength{\tabcolsep}{#1}\renewcommand{\arraystretch}{#2}\centering}

\usepackage{pifont}

\usepackage{tcolorbox}
\tcbuselibrary{breakable, skins, listings}
\usepackage{enumitem}
\newcounter{findingcounter}
\newcommand{\finding}[2]{
    \refstepcounter{findingcounter}
    \begin{tcolorbox}[
        colback=white!90!gray,     
        colframe=teal!60!black,     
        arc=5pt,                    
        boxsep=5pt,                 
        left=10pt,                  
        right=10pt,                 
        top=2pt,                    
        bottom=2pt,                 
        boxrule=0.8pt,              
        drop shadow=gray!50!white,  
        enhanced jigsaw             
    ]
    \vspace{-0.1cm}
        \paragraph{\textbf{\textit{Finding \thefindingcounter.}}} #2
    \end{tcolorbox}
    \vspace{0.2cm}
}

\newtcblisting{promptbox}{
    breakable,
    listing only,
    colback=cyan!5!white,
    colframe=teal!60!black,    
    arc=3pt,                    
    boxsep=2pt,                
    left=5pt,                  
    right=5pt,                 
    top=0pt,                    
    bottom=0pt,                 
    boxrule=0.8pt,              
    enhanced jigsaw,
    listing options={
        basicstyle=\ttfamily\scriptsize,
        breaklines=true,       
        literate={-}{-}1,      
    }
}

\newcommand{\cmark}{\ding{51}} 
\newcommand{\xmark}{\ding{55}} 

\RequirePackage{xspace}
\makeatletter
\DeclareRobustCommand\onedot{\futurelet\@let@token\@onedot}
\def\@onedot{\ifx\@let@token.\else.\null\fi\xspace}
\def\eg{\emph{e.g}\onedot} 
\def\ie{\emph{i.e}\onedot} 
 
 \def\vs{\emph{vs}\onedot}

\definecolor{green}{HTML}{009000} 
\definecolor{red}{HTML}{ea4335}
\newcommand{\hlg}[1]{\textcolor{green}{#1}}
\newcommand{\hlr}[1]{\textcolor{red}{#1}}
\newcommand{\better}[1]{\scriptsize\hlg{$\uparrow\,$#1}}
\newcommand{\worse}[1]{\scriptsize\hlr{$\downarrow\,$#1}}
\newcommand{\same}[1]{\scriptsize{$\phantom{\uparrow}\,$#1}}

\definecolor{baselinecolor}{gray}{.9}
\newcommand{\baseline}[1]{\cellcolor{baselinecolor}{#1}}

\hypersetup{
  colorlinks=true,
  urlcolor=magenta,
  linkcolor=black,
  citecolor=link,
  filecolor=black,
  pdfborder={0 0 0}
}

\definecolor{link}{HTML}{0063BE}
\providecommand{\equationname}{Equation}
\providecommand{\sectionname}{Section}

\makeatletter
\renewcommand\@makefnmark{%
  \hbox{\@textsuperscript{\normalfont\textcolor{link}{\@thefnmark}}}%
}
\makeatother

\newcommand{\blueref}[1]{\hyperref[#1]{\textcolor{link}{\ref*{#1}}}}

\newcommand{\figref}[1]{%
  \figurename~\hyperref[#1]{\textcolor{link}{\ref*{#1}}}%
}
\newcommand{\tabref}[1]{%
  \tablename~\hyperref[#1]{\textcolor{link}{\ref*{#1}}}%
}
\newcommand{\eqrefc}[1]{%
  \equationname~\hyperref[#1]{\textcolor{link}{\ref*{#1}}}%
}
\newcommand{\secrefc}[1]{%
  \sectionname~\hyperref[#1]{\textcolor{link}{\ref*{#1}}}%
}
\newcommand{\appendixref}[1]{%
  Appendix~\hyperref[#1]{\textcolor{link}{\ref*{#1}}}%
}

\newcommand{\github}{\raisebox{-1.3pt}{\includegraphics[height=1.05em]{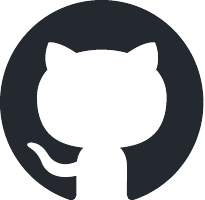}}\xspace}
\newcommand{\worldwideweb}{\raisebox{-1.3pt}{\includegraphics[height=1.05em]{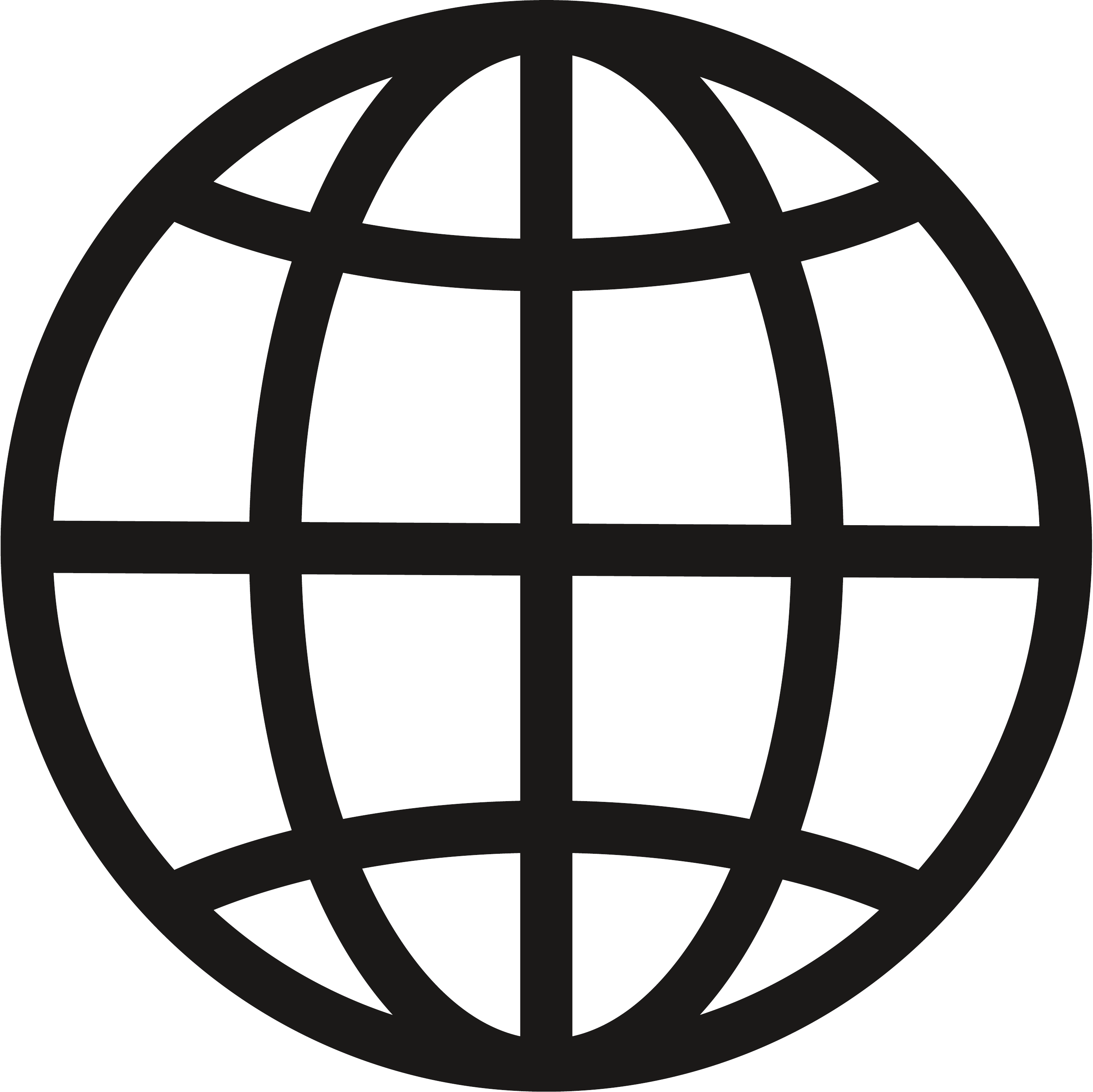}}\xspace}
\newcommand{\hf}{\raisebox{-1.3pt}{\includegraphics[height=1.05em]{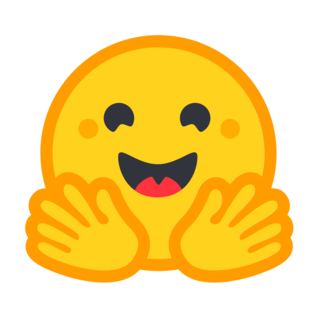}}\xspace}

\makeatletter
\fancypagestyle{firststyle}{
  \fancyhead[L]{}
  \fancyhead[C]{}
  \fancyhead[R]{}
  \fancyfoot[L]{\footerfont \the\correspondingauthor}
  \fancyfoot[R]{}
}
\makeatother

\DeclareTextFontCommand{\textbf}{\bfseries}
\title{\centering \fontsize{18}{24}\selectfont{i1: A Simple and Fully Open Recipe for Strong\\Text-to-Image Models}}

\author{
    \vspace{.2cm}
    \parbox{\textwidth}{\centering
        Boya Zeng \quad
        Tianze Luo \quad
        Shu Pu \quad
        Jucheng Shen \\
        \vspace{-0.5em}\Authfont
        Taiming Lu \quad
        Gabriel Sarch \quad
        Zhuang Liu\textsuperscript{\dag}
    }
    \\
    \vspace{.3cm}
    {\normalfont\fontsize{11}{15}\selectfont {Princeton University}}
}

\makeatletter
\renewcommand{\abscontent}{}%
\makeatother

\newenvironment{abstractblock}{%
  {\centering\large\bfseries Abstract\par}
  \vspace{0.2em}
  \begin{list}{}{%
      \setlength{\leftmargin}{2em}
      \setlength{\rightmargin}{2em}
      \setlength{\topsep}{0pt}
      \setlength{\parsep}{0pt}
  }
  \item[]
}{%
  \end{list}
  \par\normalfont\vspace{1em}
}

\begin{document}

\begingroup
\vspace*{-1.23cm}
\makeatletter
\let\raggedright\centering
\makeatother

\maketitle
\endgroup

\setlength{\skip\footins}{.5cm}
\begingroup\renewcommand\thefootnote{}\footnotetext{\fontsize{9}{11}\selectfont\begin{tabular}[t]{@{}l@{}}$^\dag$\,Corresponding Author\end{tabular}}\addtocounter{footnote}{0}\endgroup

\vspace{-1cm}
\begin{center}
    \worldwideweb \href{https://zlab-princeton.github.io/i1}{\textbf{Website}} \quad
    \github \href{https://github.com/zlab-princeton/i1}{\textbf{Code}} \quad
    \hf \href{https://huggingface.co/zlab-princeton/i1-3B}{\textbf{Model}} \quad
    \hf \href{https://huggingface.co/datasets/zlab-princeton/i1-captions}{\textbf{Data}}
\end{center}

\newcommand{\abstractcontent}{%
Diffusion models have consistently driven progress in text-to-image generation. However, it is challenging to attribute recent progress to specific modeling and data choices: state-of-the-art open-weight models provide limited ablations, and do not disclose their training data and full training details. The research community needs fully open (weights, data, and code) models as a foundation for further research; yet existing fully open models still fall significantly short of leading models in performance.
In this project, we conduct a systematic investigation of the modeling and data design choices in text-to-image diffusion training and inference with 300+ controlled experiments totaling 700K+ TPU v6e hours.
Our experiments highlight several empirical findings (\eg, equal weighting is a strong default for mixing curated datasets) and simple design decisions (\eg, larger text encoder adapters improve performance with minimal added parameters) for training strong models.
Guided by these insights, we train \textbf{i1}, a 3B-parameter text-to-image diffusion model using only publicly available datasets. \textbf{i1} is competitive with leading models on five representative benchmarks (GenEval, DPG, PRISM, CVTG-2K, and LongText), and outperforms the best existing fully open model by 29.5 absolute percentage points on average. We provide the \textbf{i1} checkpoints, training and inference code, and the data processing pipeline. Together, our findings and the \textbf{i1} recipe establish a practical foundation for future open research in text-to-image diffusion models.
}

\vspace{-0.1cm}

\begin{figure}[!ht]
    \centering
    \includegraphics[width=0.9\textwidth]{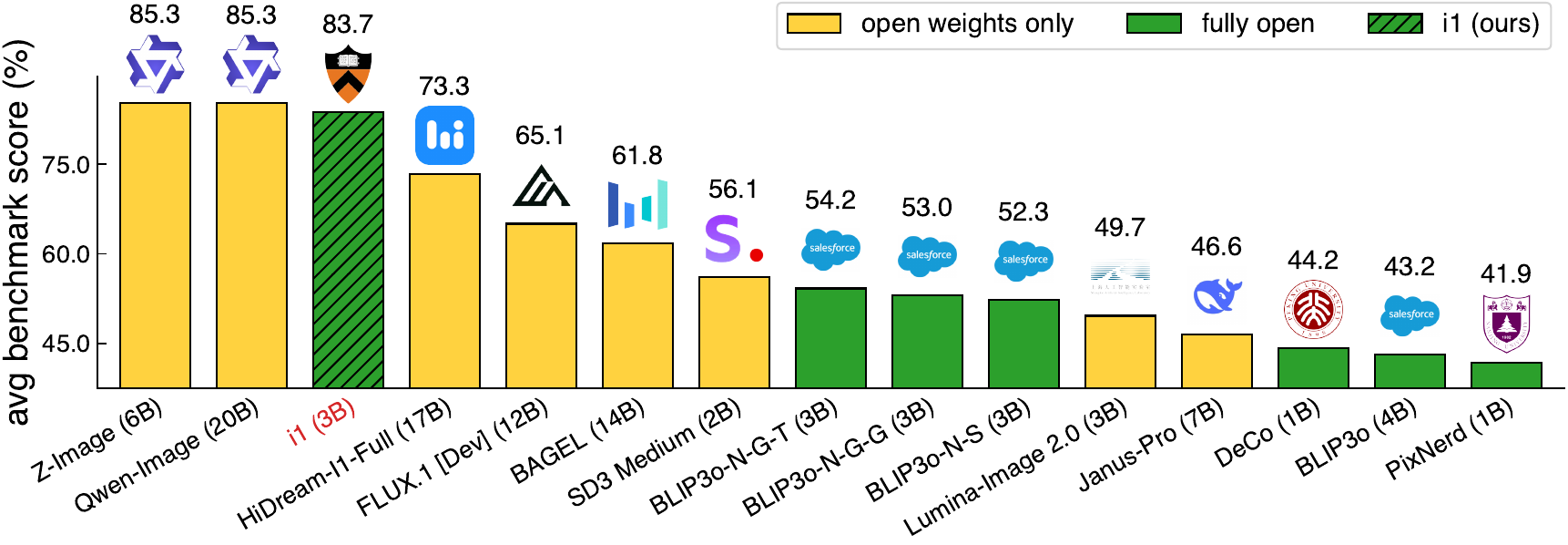}
    \caption{We investigate the design space of text-to-image diffusion models to understand how modeling and data choices affect model capabilities. This exploration culminates in \textbf{i1}, a 3B-parameter model that performs competitively with leading models at 1024-resolution, as measured by the average percentage score across GenEval, DPG-Bench, PRISM, CVTG-2K, and LongText-Bench. We open-source our model, code, and data to support future research.}
    \label{fig:teaser}
\end{figure}
\vspace{0.3cm}

\begin{abstractblock}
\abstractcontent
\end{abstractblock}

\begin{figure}[H]
\centering
\setlength{\tabcolsep}{0pt}
\renewcommand{\arraystretch}{0}

\begin{tabular}{@{}ccc@{}}

\includegraphics[width=0.32\linewidth]{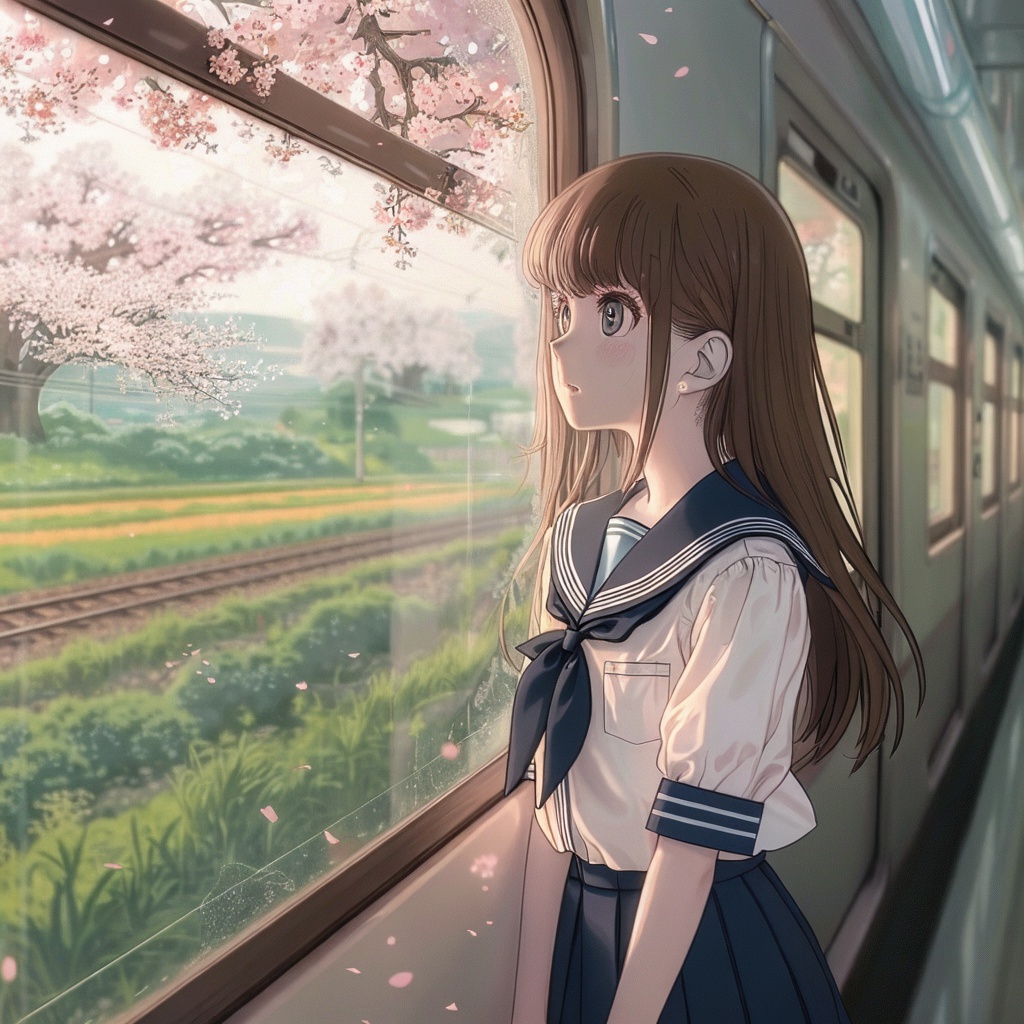} &
\includegraphics[width=0.32\linewidth]{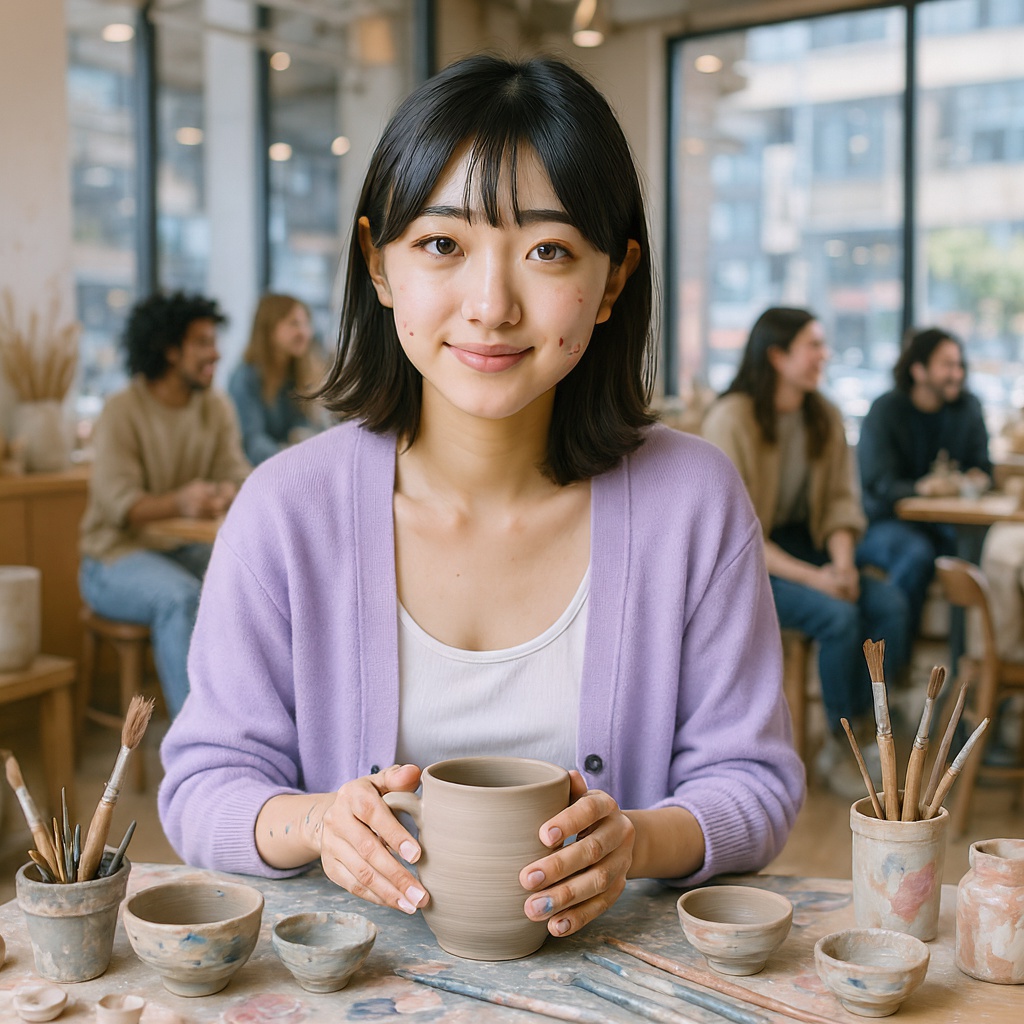} &
\includegraphics[width=0.32\linewidth]{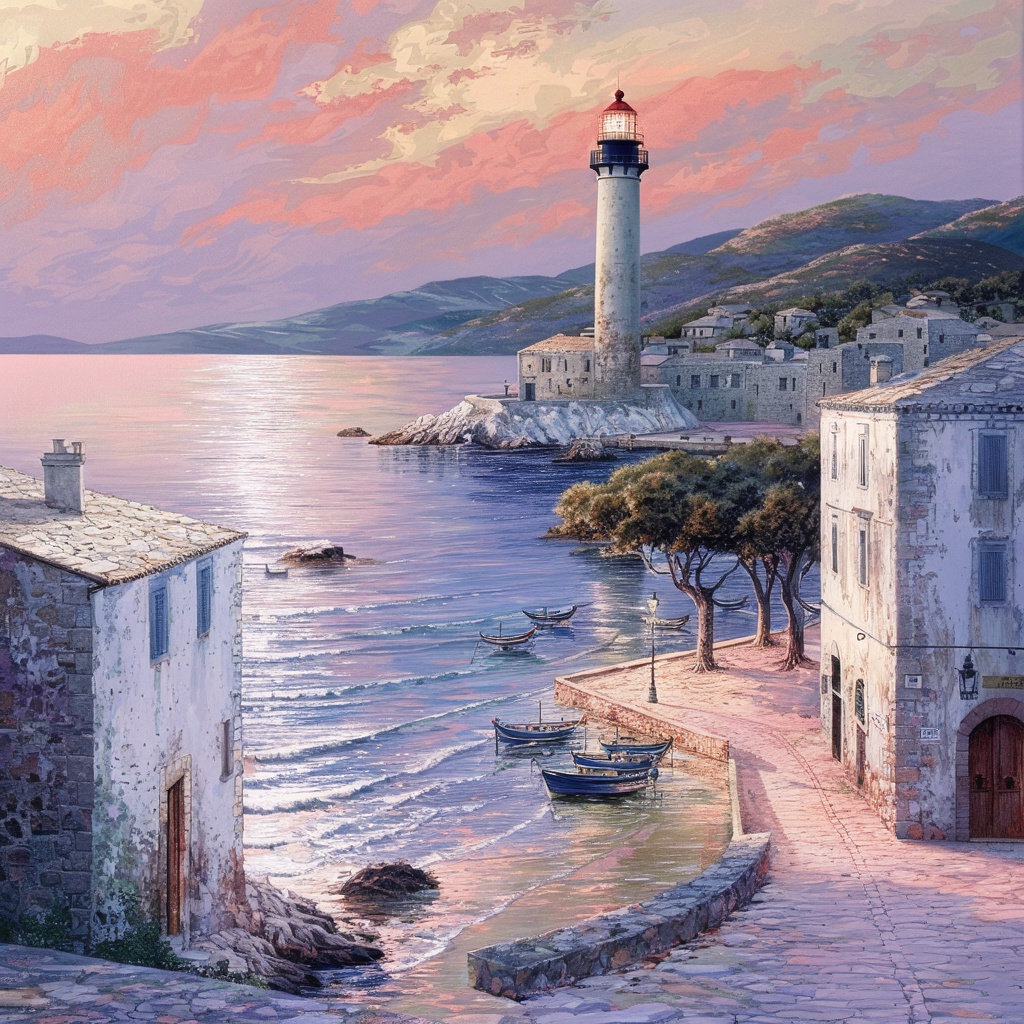} \\

\includegraphics[width=0.32\linewidth]{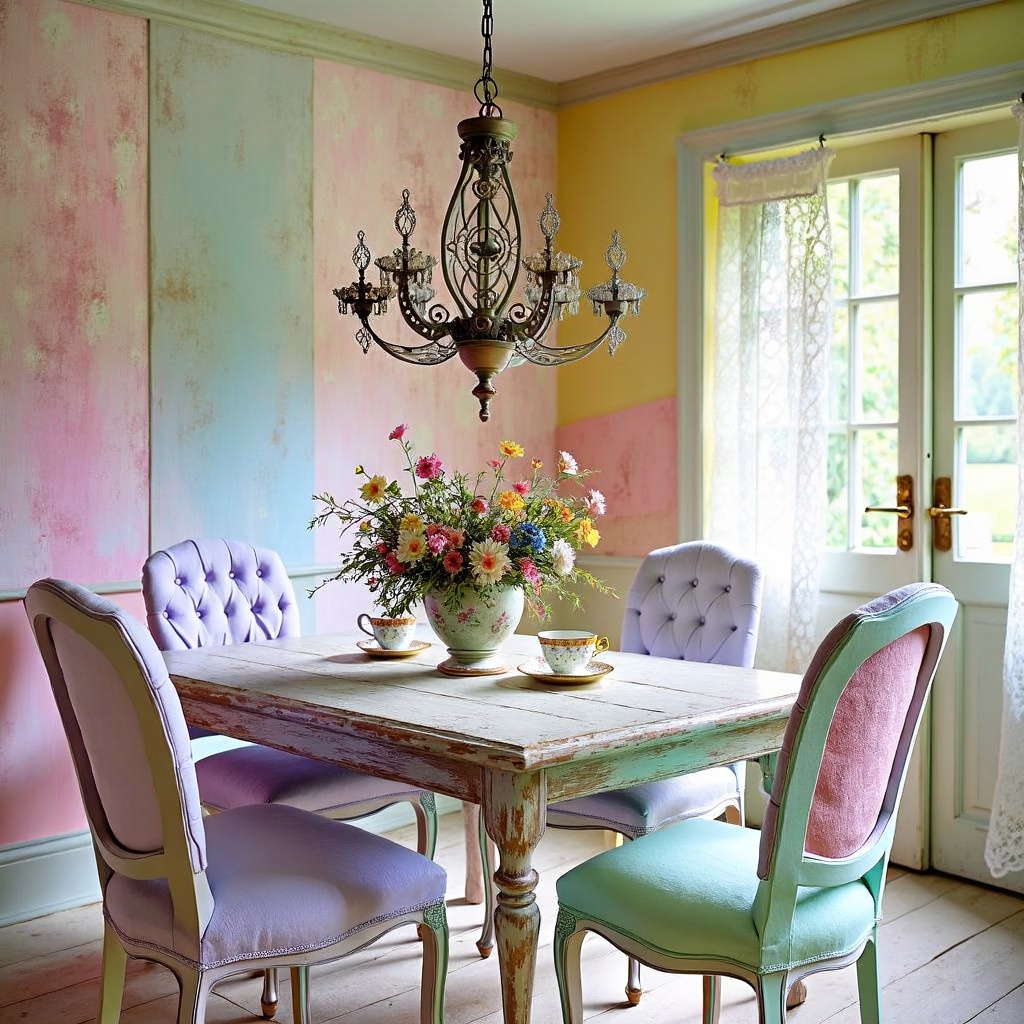} &
\includegraphics[width=0.32\linewidth]{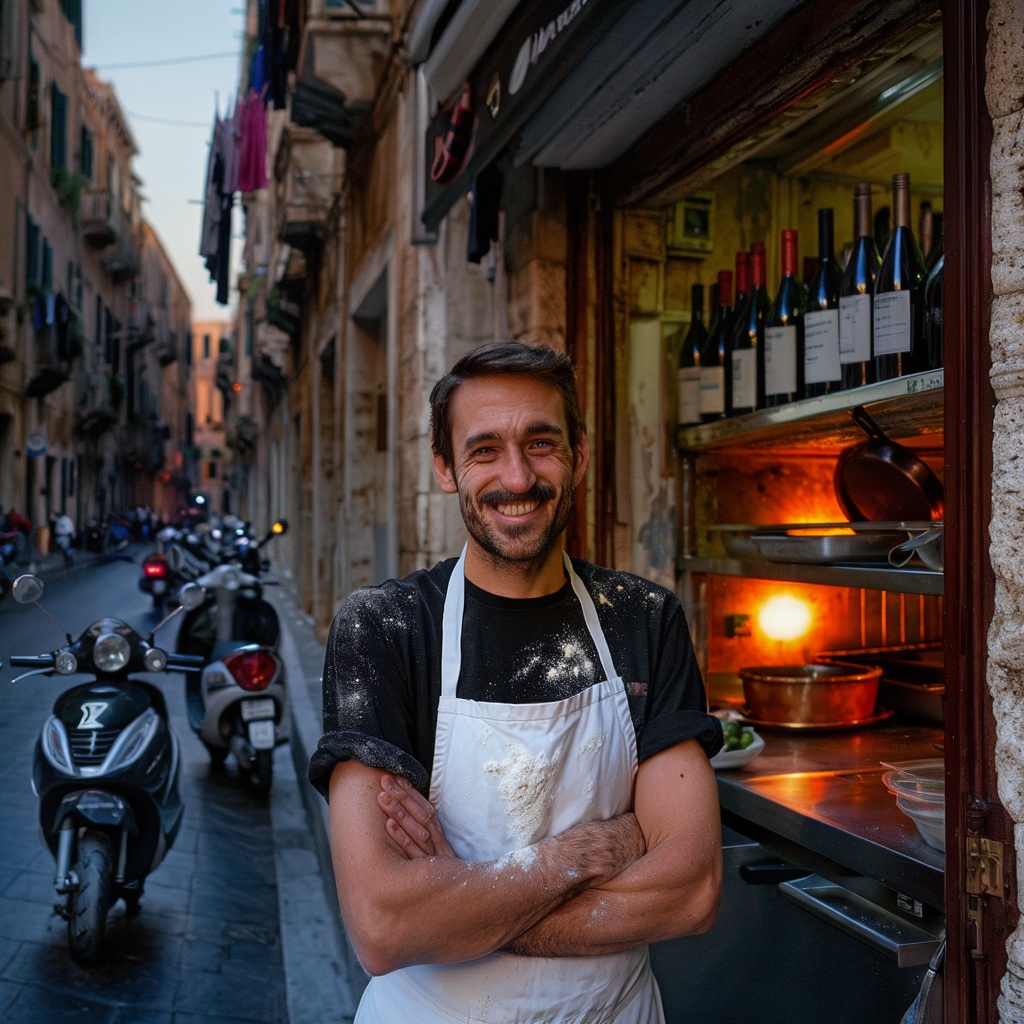} &
\includegraphics[width=0.32\linewidth]{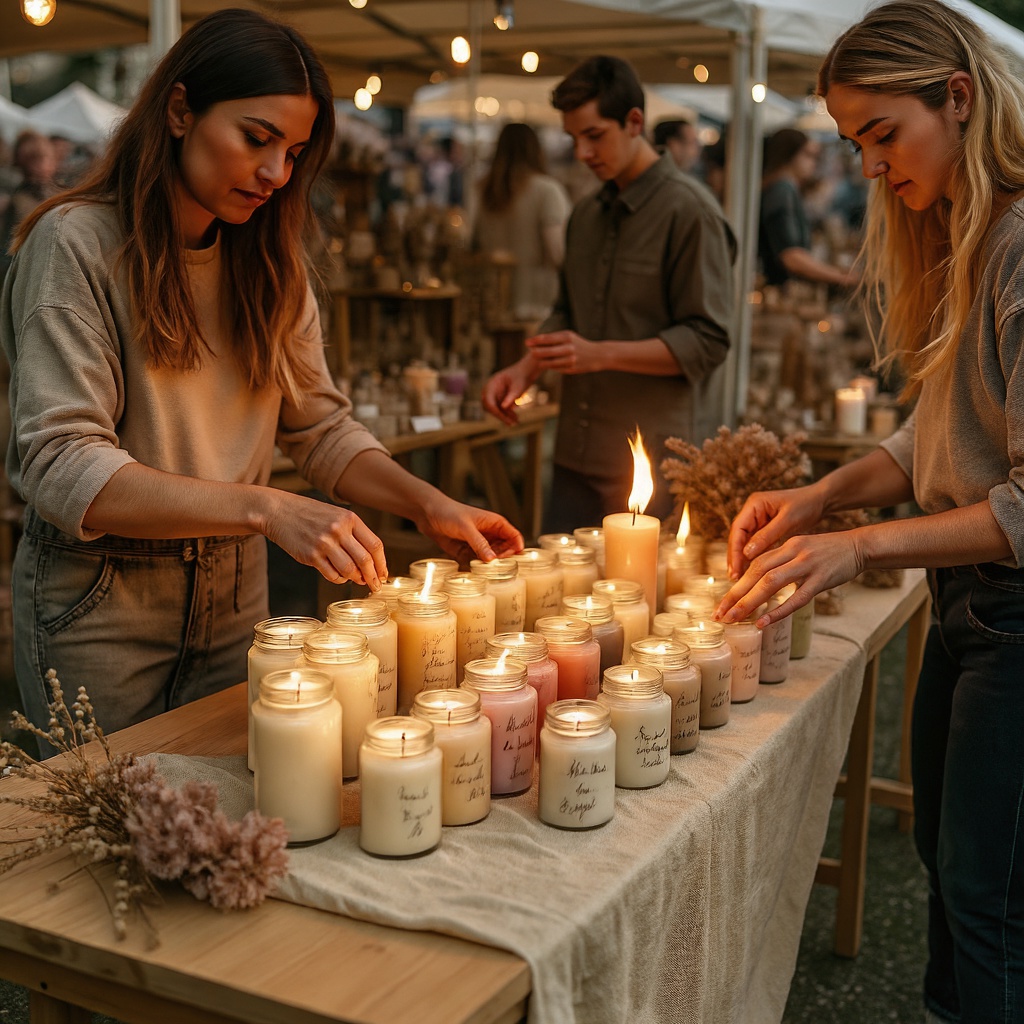} \\

\includegraphics[width=0.32\linewidth]{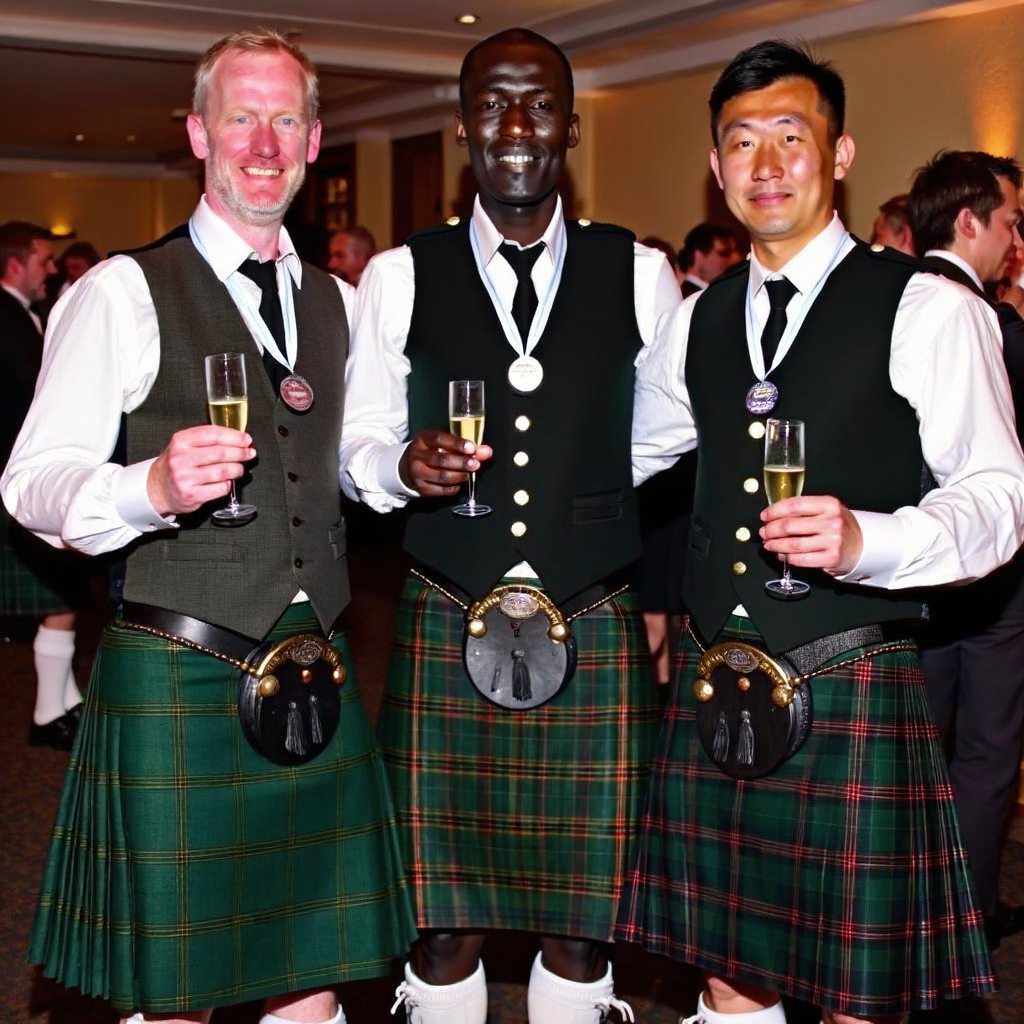} &
\includegraphics[width=0.32\linewidth]{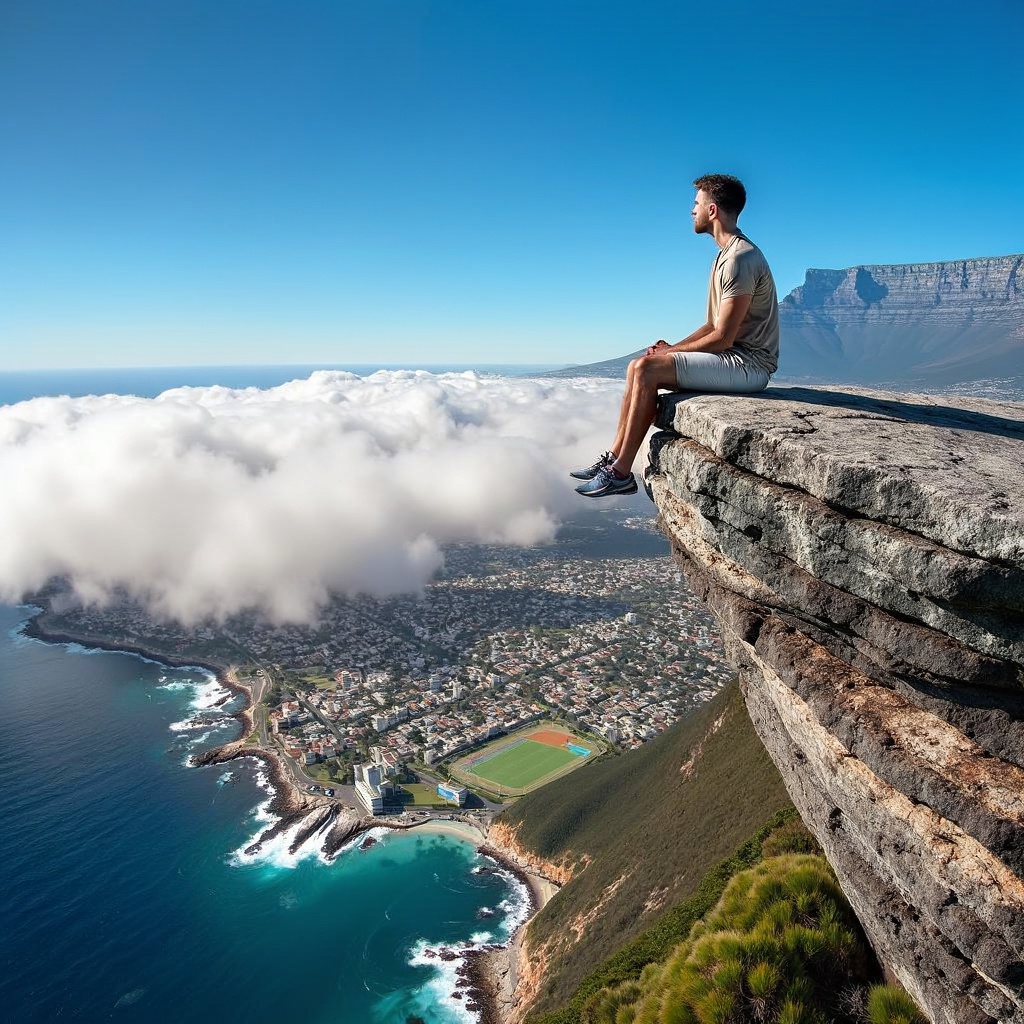} &
\includegraphics[width=0.32\linewidth]{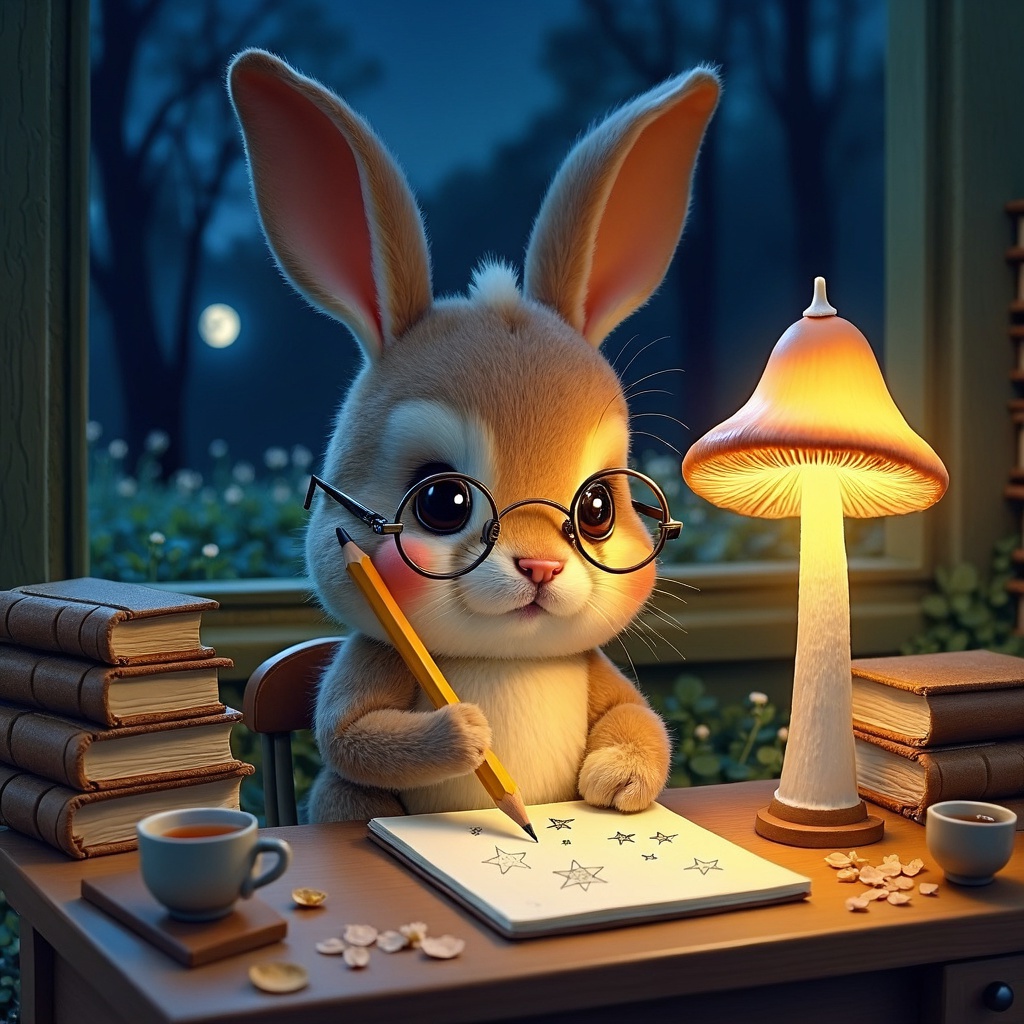} \\

\includegraphics[width=0.32\linewidth]{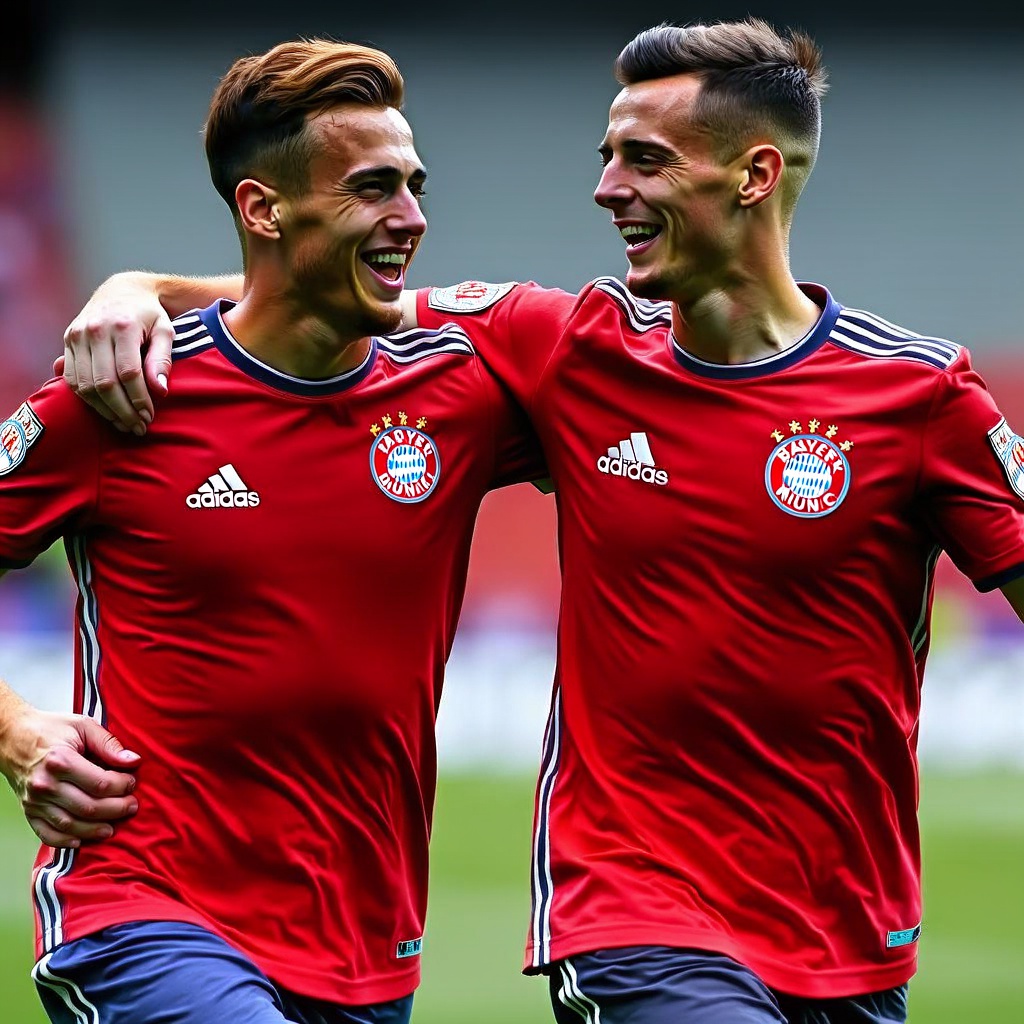} &
\includegraphics[width=0.32\linewidth]{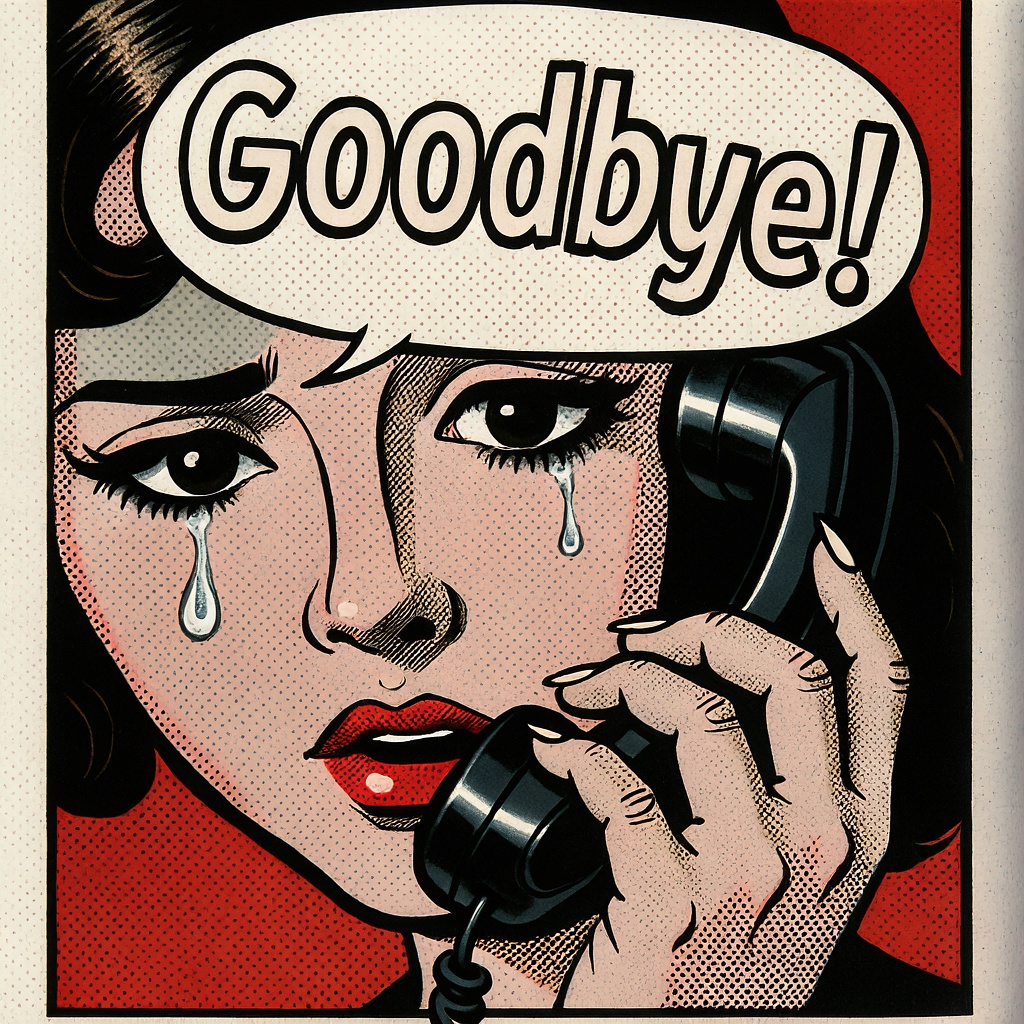} &
\includegraphics[width=0.32\linewidth]{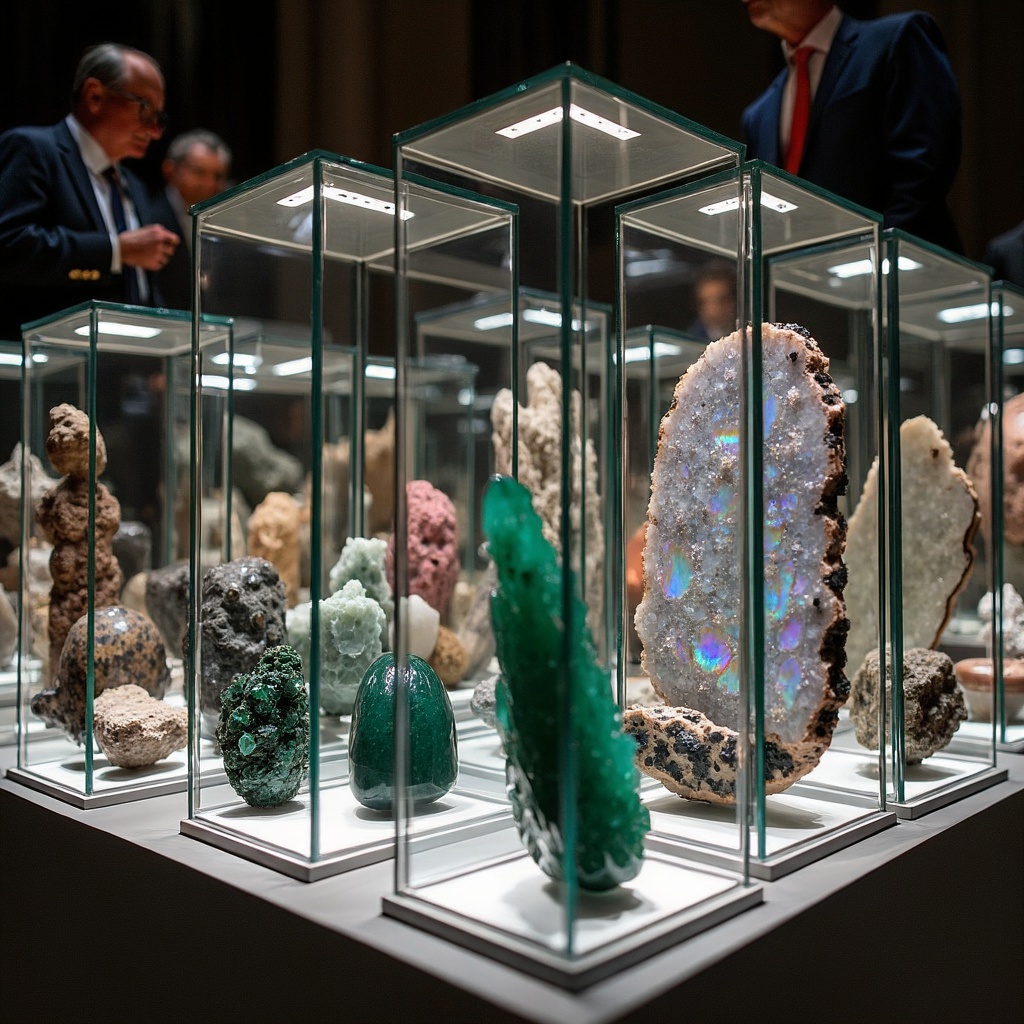}
\end{tabular}
\caption{Curated showcase of \textbf{i1} in general image generation (more examples in~\appendixref{appendix:qualitative_comparison}).}
\label{fig:showcase_general}
\end{figure}

\newpage

\begin{figure}[H]
\centering
\setlength{\tabcolsep}{0pt}
\renewcommand{\arraystretch}{0}

\begin{tabular}{@{}ccc@{}}

\includegraphics[width=0.32\linewidth]{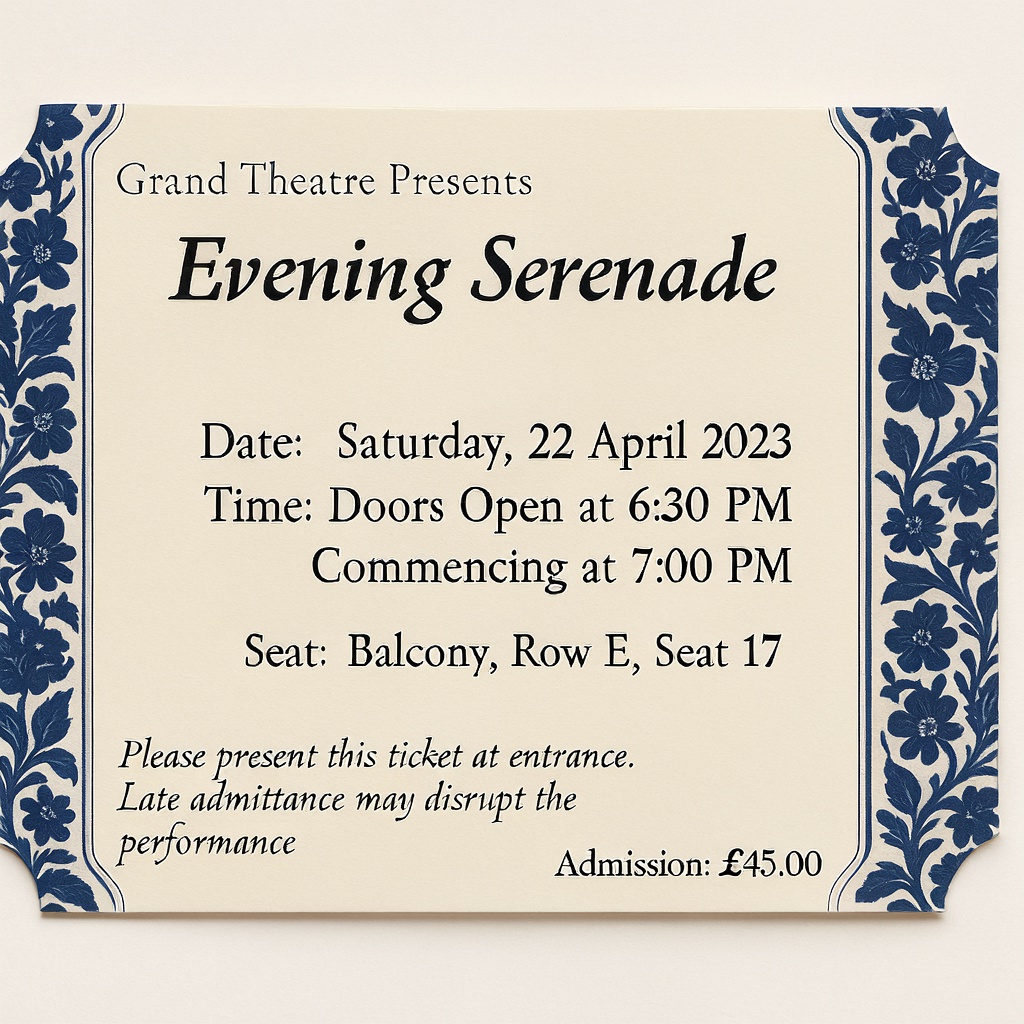} &
\includegraphics[width=0.32\linewidth]{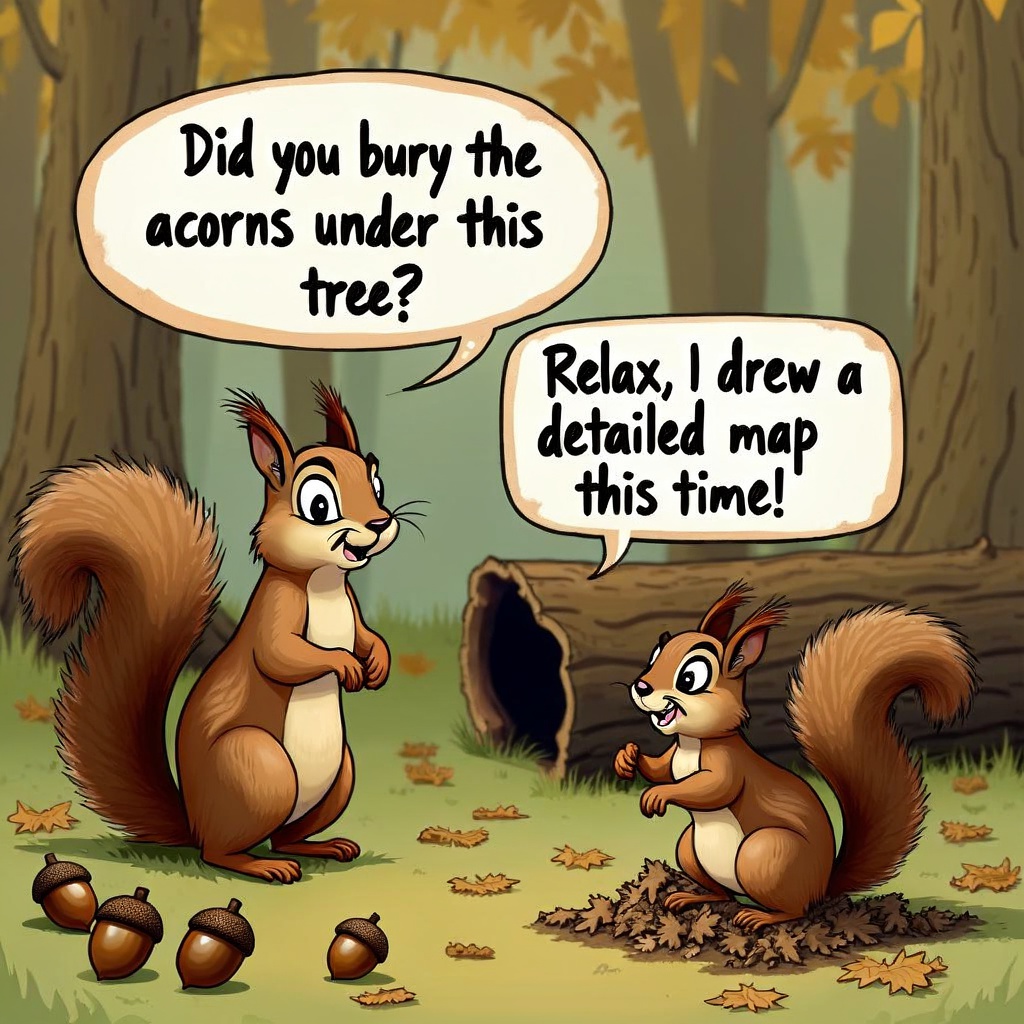} &
\includegraphics[width=0.32\linewidth]{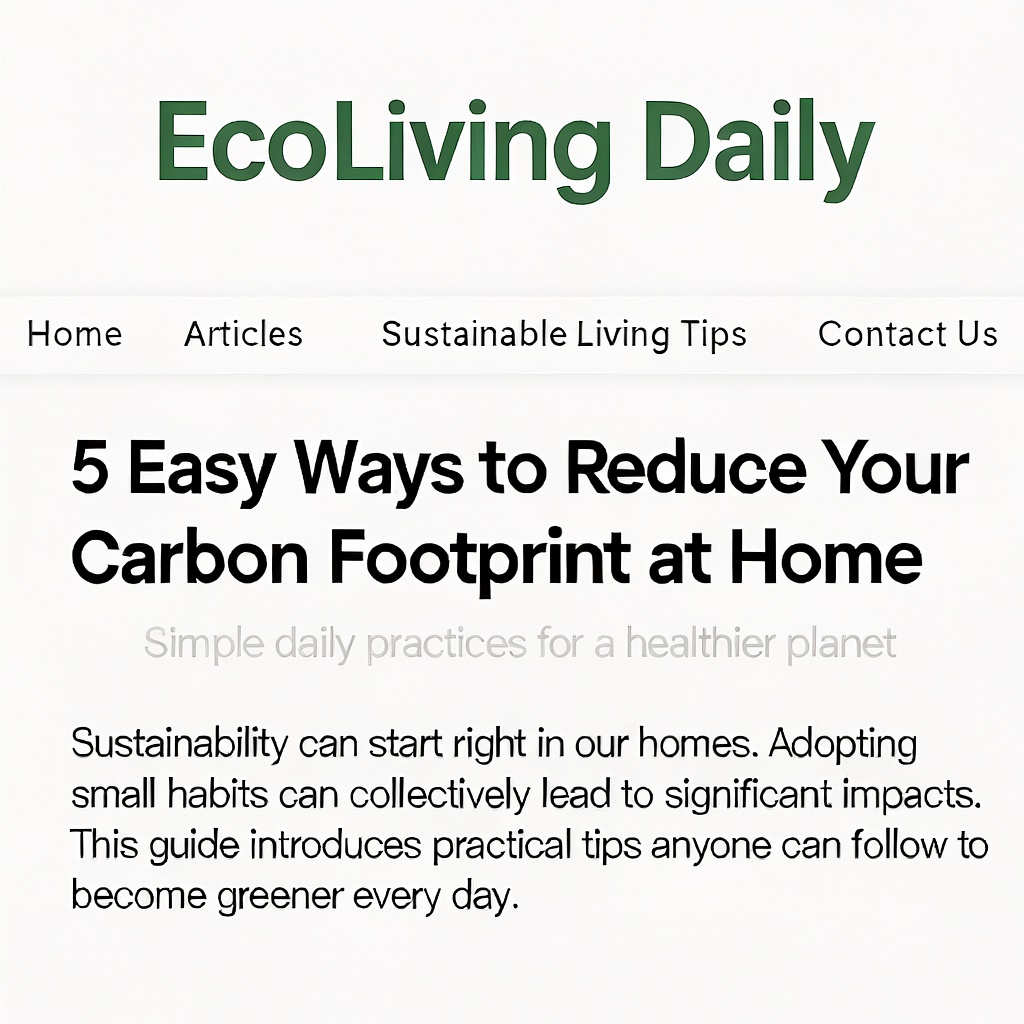} \\

\includegraphics[width=0.32\linewidth]{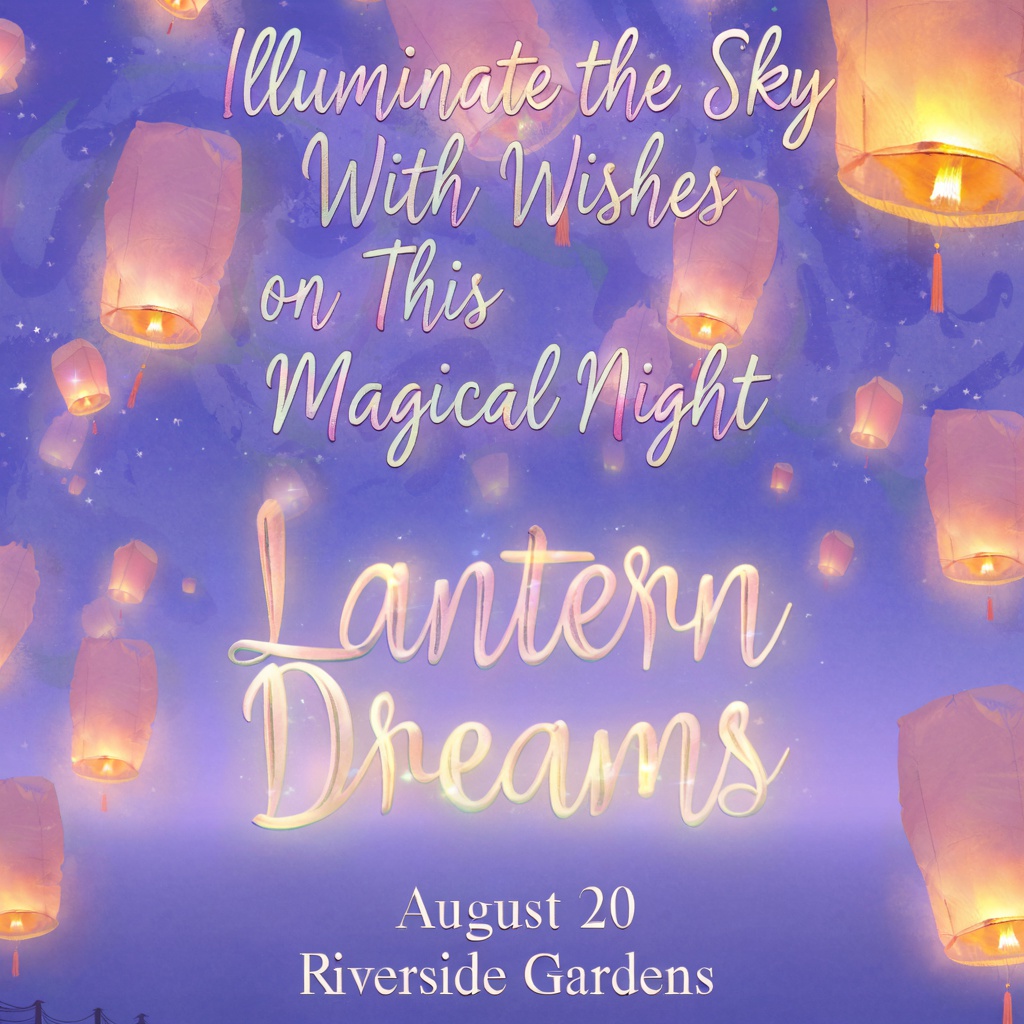} &
\includegraphics[width=0.32\linewidth]{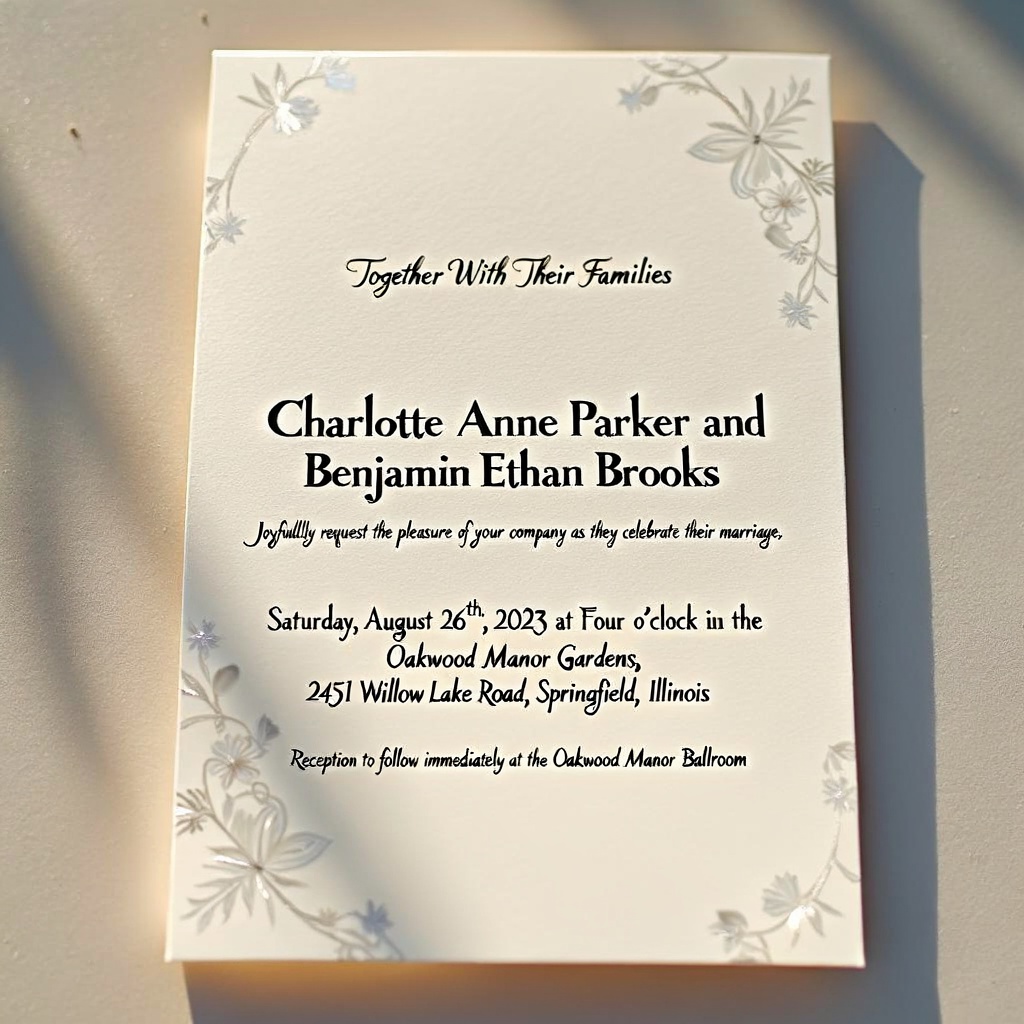} &
\includegraphics[width=0.32\linewidth]{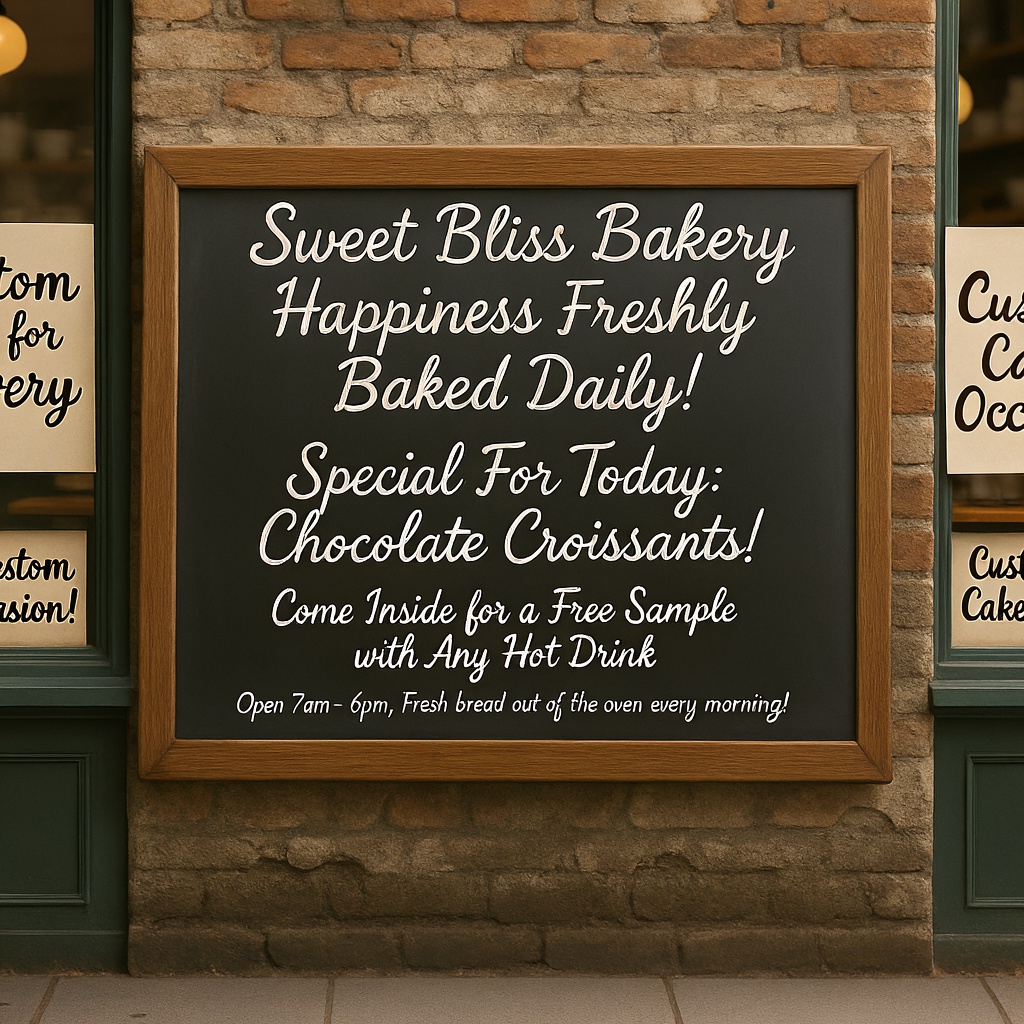} \\

\includegraphics[width=0.32\linewidth]{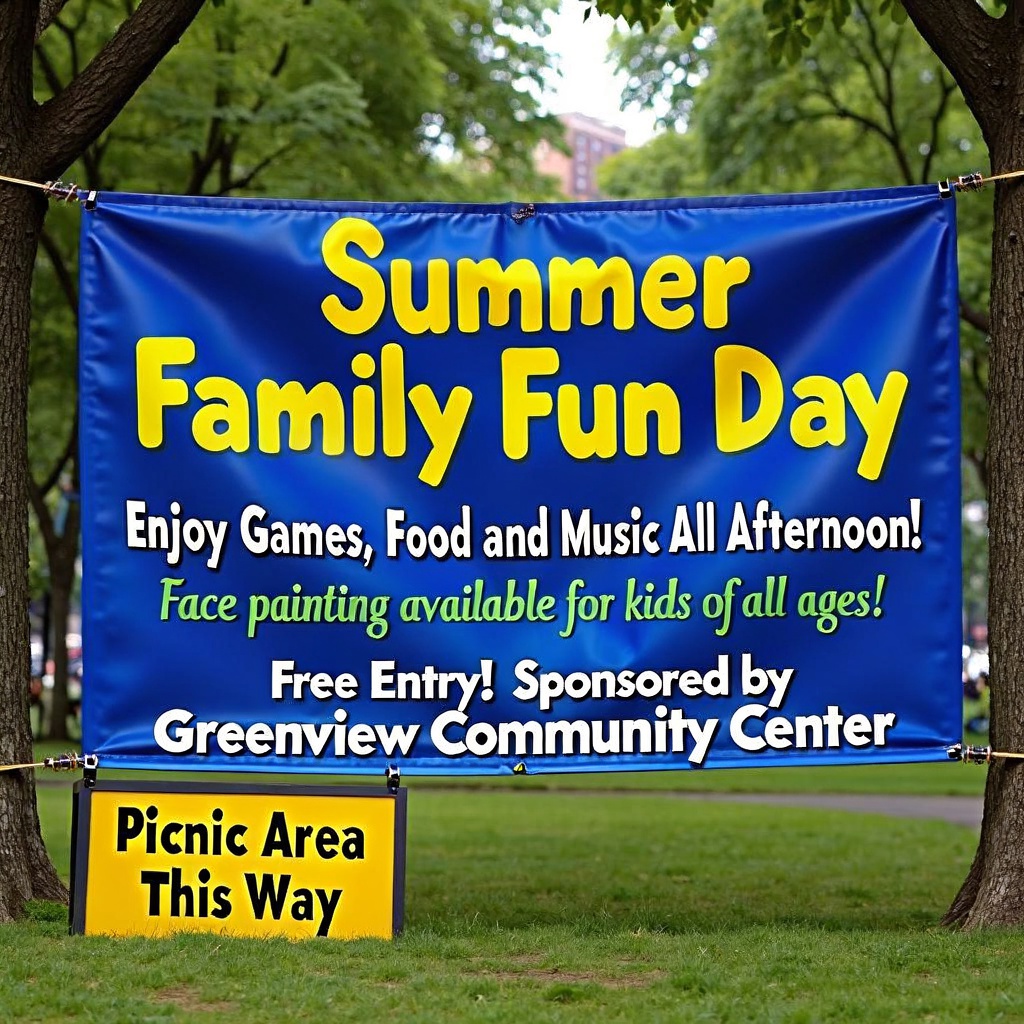} &
\includegraphics[width=0.32\linewidth]{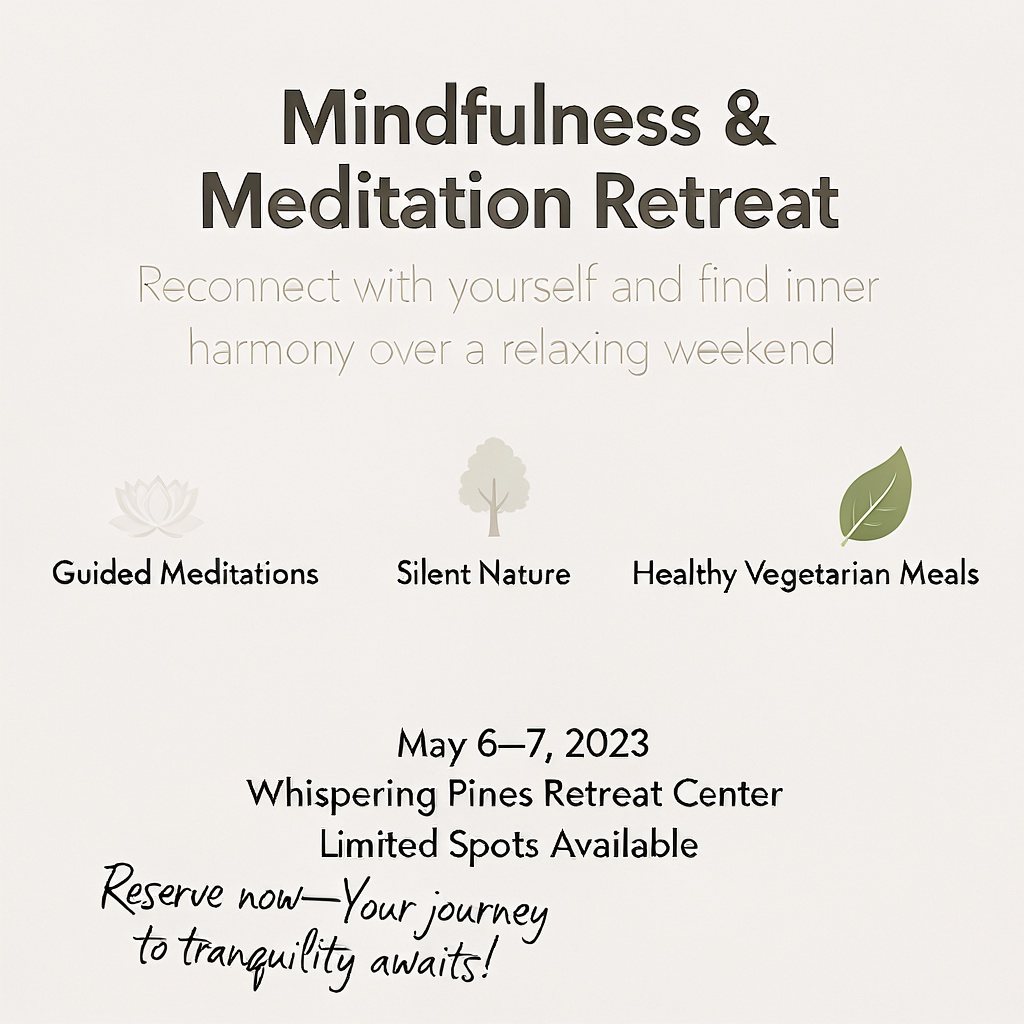} &
\includegraphics[width=0.32\linewidth]{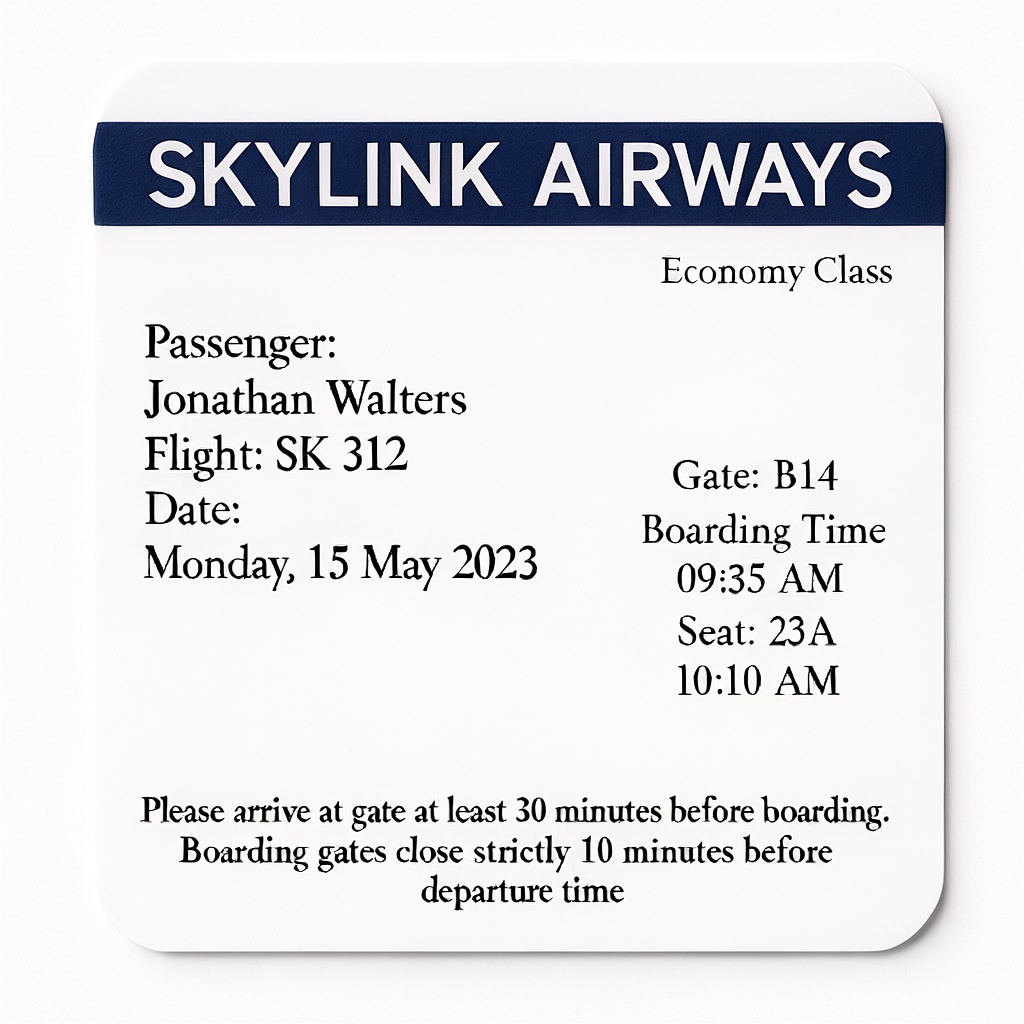} \\

\includegraphics[width=0.32\linewidth]{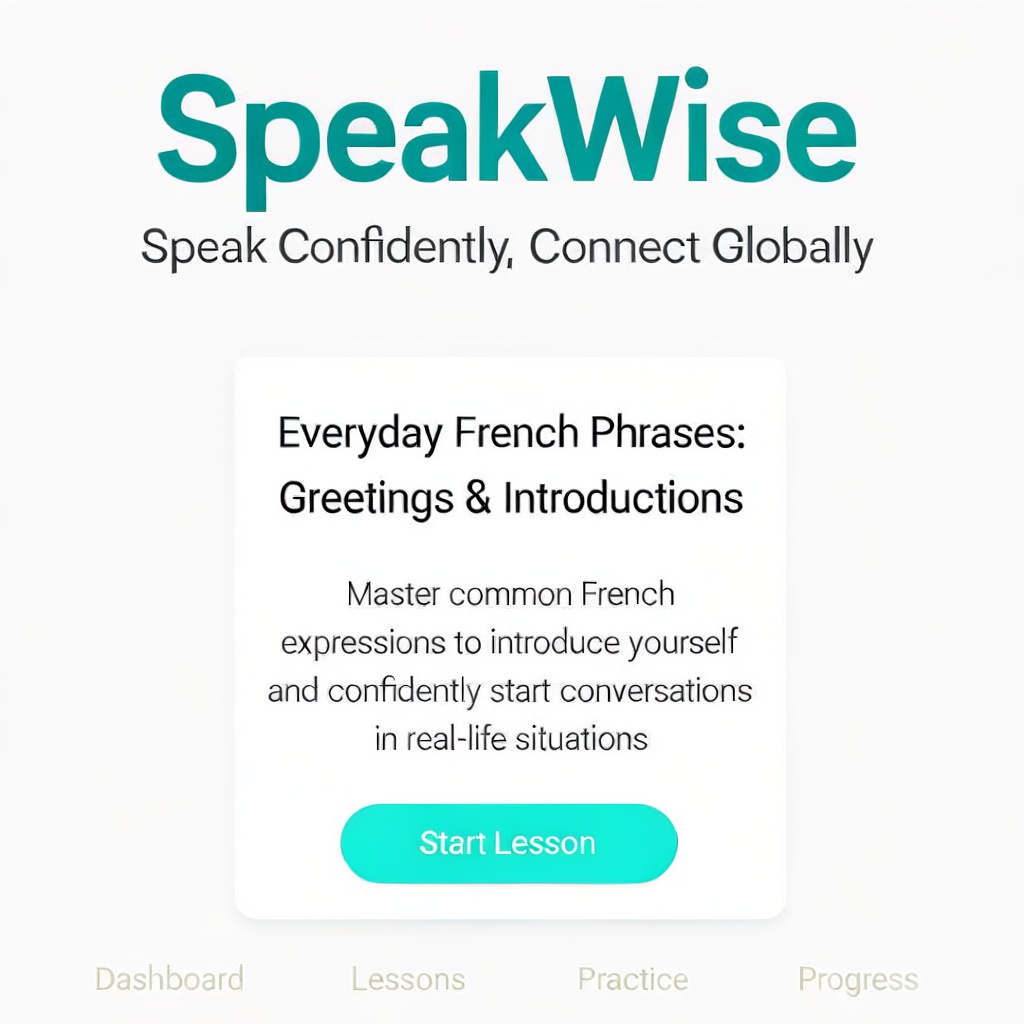} &
\includegraphics[width=0.32\linewidth]{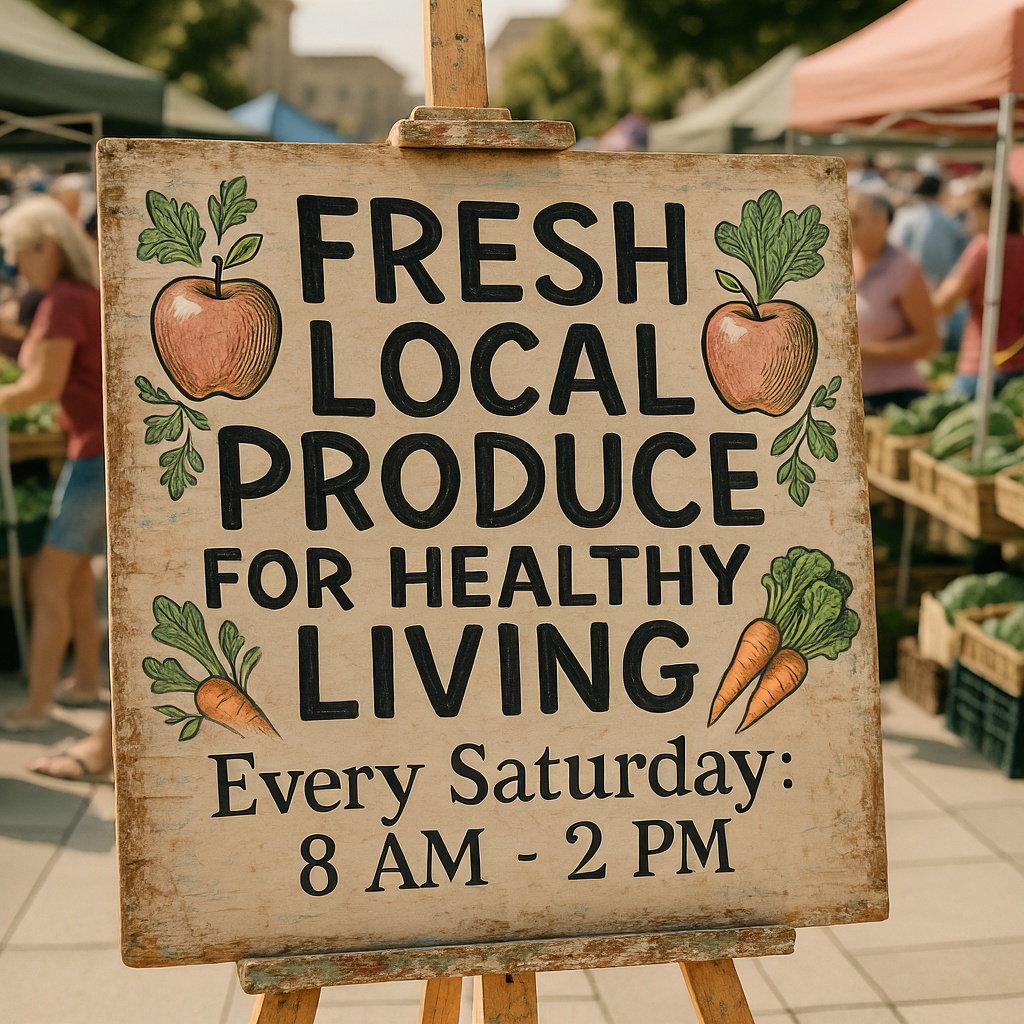} &
\includegraphics[width=0.32\linewidth]{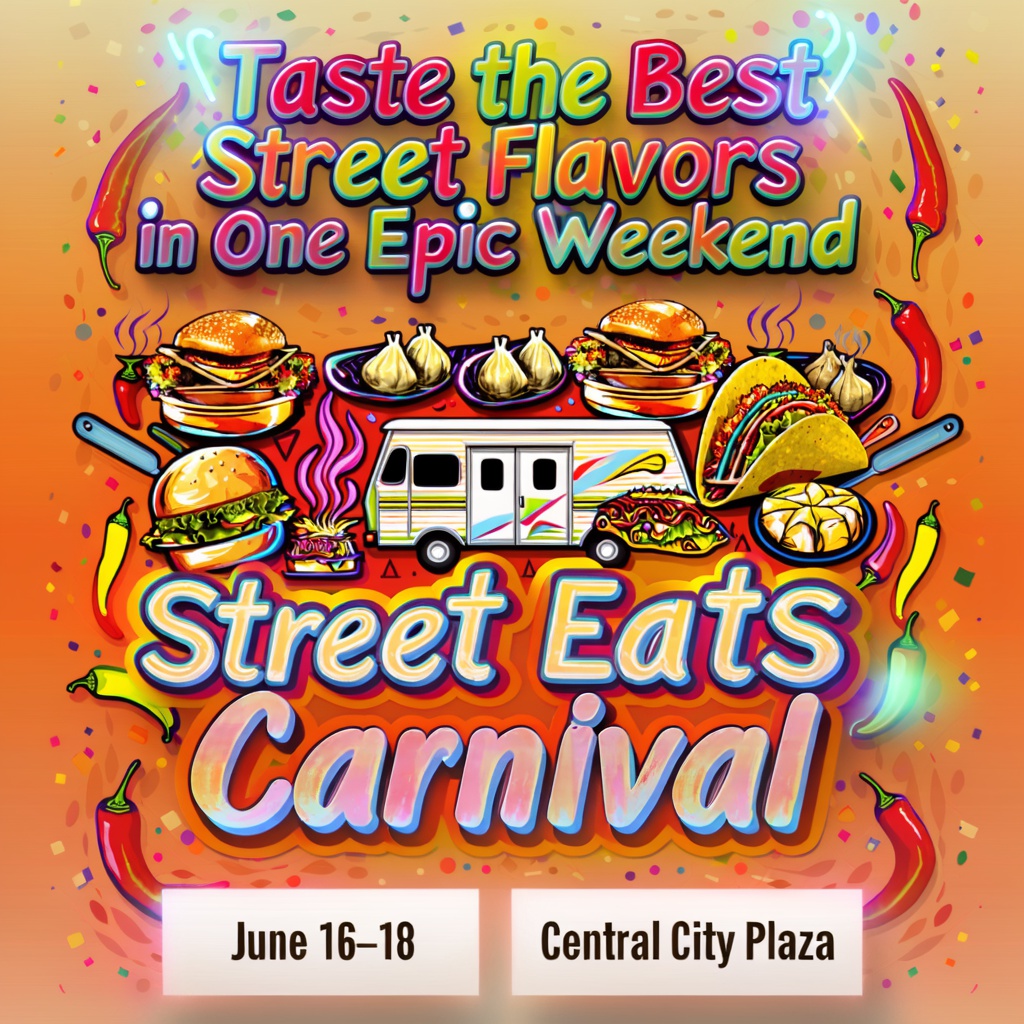}
\end{tabular}
\caption{Curated showcase of \textbf{i1} in text-rendering (more examples in~\appendixref{appendix:qualitative_comparison}).}
\label{fig:showcase_text_render}
\end{figure}

\newpage

\setcounter{tocdepth}{2}

\tableofcontents

\newpage

\section{Introduction}
\label{sec:intro}

Since early models like DALL-E 2~\citep{ramesh2022hierarchical}, Imagen~\citep{saharia2022photorealistic}, and Stable Diffusion~\citep{rombach2022high}, diffusion-based models have driven major advances in text-to-image generation~\citep{wu2025qwen, labs2025flux, cai2025z, gao2025seedream} due to their strong capability for generating photorealistic images with fine-grained details. However, despite the superior capabilities of today’s state-of-the-art models, it is often difficult to disentangle which modeling and data choices are truly driving performance. This lack of clarity stems from two factors.

First, leading models often do not release their training data and full training recipe~\citep{wu2025qwen, cai2025z, qin2025lumina, cai2025hidream}, even when they publicly release model checkpoints.
This limits reproducibility and hinders controlled analysis and follow-up work that builds on their designs. While \textit{fully open} (weights, data, and code) models exist~\citep{chen2025blip3, chen2025blip3o, ma2026deco, wang2026pixnerd}, they fall substantially short of leading models in performance.

Second, leading models often do not provide thorough ablations of their design choices~\citep{wu2025qwen, cai2025z, cai2025hidream, gao2025seedream, ryu2025flite, fang2026flux}. In practice, many models bundle numerous architectural, training, and data decisions into a single recipe, making it difficult to attribute improvements to any specific factor. As a result, modern text-to-image diffusion models still lack consensus on many important design choices (\eg, using a single text encoder~\citep{qin2025lumina, wu2025qwen} \vs multiple text encoders~\citep{esser2024scaling, cai2025hidream}).

To obtain a better understanding of the impact of existing and new architectural and data designs in text-to-image diffusion models, we conduct a series of controlled experiments, primarily on the 256$\times$256 low-resolution pre-training stage. Starting from a simple baseline model~\citep{yao2025reconstruction}, we first explore strategies for incorporating text conditioning from text encoders, as well as noise/timestep conditioning~\citep{sun2025noise}. Then, we identify backbone architecture designs that lead to stronger performance~\citep{bao2023all, esser2024scaling}. Finally, we compare design choices in the curation of high-quality image-caption datasets and inference-time prompt enhancements, along with strategies for mixing image datasets.

On the modeling side, we find that (1) using a single strong text encoder with a larger adapter can be more effective than combining multiple text encoders, (2) timestep/noise conditioning and Adaptive Layer Normalization~\citep{peebles2023scalable} provide little benefit for text-to-image in our setting, and (3) a dual-stream DiT~\citep{esser2024scaling} with long skip connections~\citep{bao2023all} is a strong backbone design.

On the data side, we find that (1) training on long captions yields stronger models than training on short captions, but causes them to underperform on short prompts, which can be mitigated by inference-time prompt rewrite, (2) the choice of synthetic captioner is important for downstream performance, (3) training the model on equal numbers of images from each dataset, counting repetitions (called ``equal weighting across datasets'' hereafter), is a strong default for mixing curated datasets, (4) with a diverse mix of datasets, repeating training data incurs only marginal performance degradation, and (5) broad high-resolution data coverage is not needed to obtain strong high-resolution generation capability from a low-resolution model.

To provide a strong baseline for future open research, we leverage the insights from the controlled experiments to train \textbf{i1}, a text-to-image diffusion model with 3B parameters, on publicly available datasets. At 1024-resolution, \textbf{i1} achieves state-of-the-art performance among fully open models and outperforms several leading open-weight-only models with much larger parameter counts (\eg, 17B HiDream-I1~\citep{cai2025hidream} and 12B FLUX.1 [Dev]~\citep{labs2025flux}) across a diverse set of representative benchmarks.

\begin{figure}[!htb]
    \centering
    \includegraphics[width=\linewidth]{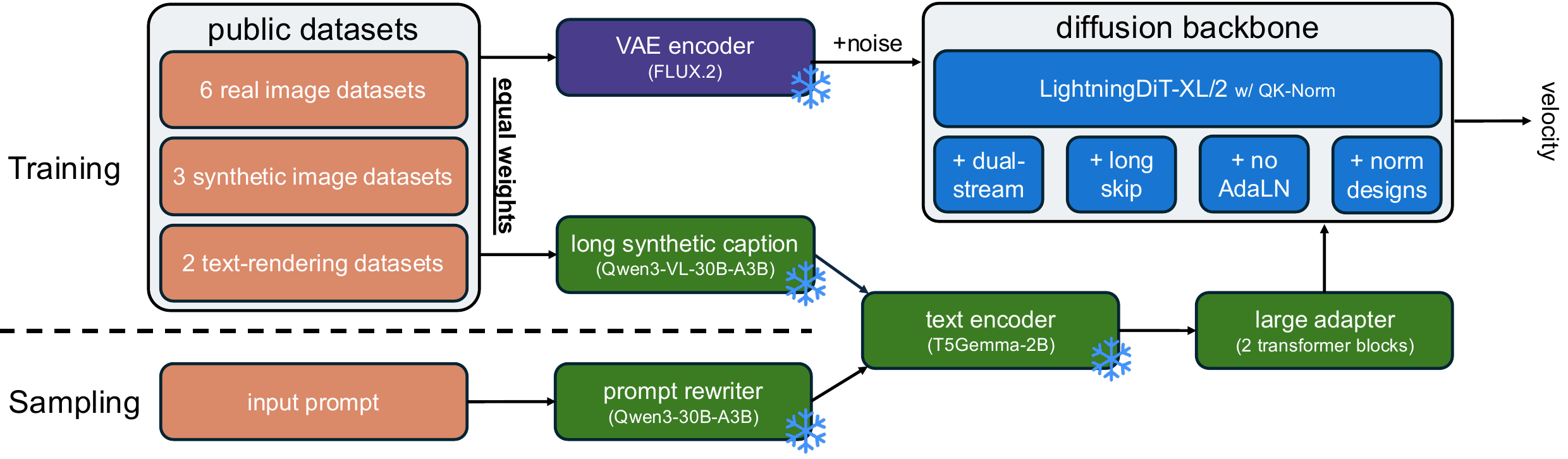}
    \caption{\textbf{High-level illustration of our final i1 model}. Rather than introducing major new network modules, \textbf{i1} combines carefully selected modeling and data design choices into a simple and strong text-to-image model.}
    \label{fig:i1_flowchart}
\end{figure}

\textbf{i1} shows that strong performance can be achieved using only moderately scaled, publicly available image datasets, and highlights the value of carefully exploring the design space: as~\figref{fig:i1_flowchart} shows, \textbf{i1} introduces no significantly new network modules, but instead identifies existing yet underused designs from prior work (\eg, long skip connections) and introduces simple modifications to standard components (\eg, using a larger text encoder adapter). We provide model weights, code, datasets, and detailed recipes for model training and evaluation.
Together, our findings and the \textbf{i1} recipe establish a practical foundation for open text-to-image research, offering both a strong fully open baseline and design insights for building more capable models.

\section{Preliminaries}
\label{sec:prelim_related_work}

In this section, we provide the terminology and context needed to understand and motivate our controlled experiment setup (\secrefc{sec:exp_setup}) and later modeling and data design experiments (Sections~\blueref{sec:modeling} and~\blueref{sec:data}), with a focus on backbone architectures, text and noise conditioning mechanisms, and existing open training data recipes.

\paragraph{Text-to-image backbone architectures}. Despite alternative paradigms~\citep{chang2023muse, sun2024autoregressive, zhou2024transfusion}, most leading text-to-image systems use diffusion transformers (DiTs)~\citep{peebles2023scalable} trained with flow matching~\citep{lipman2022flow}.
Depending on how text features are incorporated, recent diffusion models generally fall into three categories: \textbf{cross-attention} models~\citep{chen2024pixart, xie2025sana, ryu2025flite}, \textbf{single-stream} models~\citep{qin2025lumina, cai2025z, chen2025dit}, and \textbf{dual-stream} MMDiT models~\citep{esser2024scaling, wu2025qwen}. Cross-attention models inject text embeddings via cross-attention layers, whereas single- and dual-stream models concatenate image and text token sequences. Dual-stream models use modality-specific attention and MLP parameters for image and text tokens, while single-stream models use shared attention and MLP parameters across modalities.
\textbf{Long skip connections} are an architectural modification that adds shortcuts between early and later layers. They were explored in earlier work~\citep{bao2023all} but are not widely used in modern text-to-image models.

\paragraph{Text and noise conditioning mechanisms}. In recent models, input prompts are encoded by one~\citep{cai2025z, qin2025lumina, wu2025qwen} or more~\citep{esser2024scaling, cai2025hidream} \textbf{text encoders}. The resulting text features are often passed through a linear~\citep{labs2025flux, cai2025hidream, wu2025qwen, cai2025z} or MLP~\citep{xie2025sana} \textbf{adapter} that maps them to the hidden dimension of the diffusion model. Across backbone architectures, Adaptive Layer Normalization (\textbf{AdaLN})~\citep{peebles2023scalable} is commonly used to inject timestep information. AdaLN learns a linear projection from timestep embeddings to scaling and shifting factors for attention and MLP inputs, and gating factors for their outputs. In some models, the timestep embedding is combined with a pooled text embedding through element-wise addition before being used for AdaLN conditioning~\citep{esser2024scaling, cai2025hidream, labs2025flux}.

\paragraph{Open text-to-image data recipes}. Many leading models~\citep{wu2025qwen, cai2025z, qin2025lumina, cai2025hidream, labs2025flux} release their weights publicly but do not disclose their training data recipes. Aside from a few models~\citep{qin2025lumina, ryu2025flite}, even the sources and scale of the training datasets remain undisclosed, limiting the open research community's understanding of how to construct strong text-to-image training data.
Fully open models~\citep{chen2025blip3, chen2025blip3o, sehwag2025stretching, ma2026deco, tong2026scaling, wang2026pixnerd} still generally underperform leading systems, and their datasets are often from a similar and limited set of sources (\eg, JourneyDB~\citep{sun2023journeydb}, SA-1B~\citep{kirillov2023segment}, and CC12M~\citep{changpinyo2021conceptual}). Moreover, fully open recipes scarcely explore data balancing techniques.

\section{A Baseline for Controlled Experiments}
\label{sec:exp_setup}

Expanding beyond the existing designs introduced in~\secrefc{sec:prelim_related_work}, we study the text-to-image diffusion model design space through controlled experiments at the 256-resolution pre-training stage in Sections~\blueref{sec:modeling} and~\blueref{sec:data}.
For each set of experiments, we start from the same strong baseline and independently vary a single design choice (\ie, modifications are \textbf{not} accumulated across experiments).
Designs that improve performance are later combined in~\secrefc{sec:i1_recipe} to construct our final model, \textbf{i1} (see \figref{fig:architecture}).
We describe the baseline setup below, provide a high-level illustration in~\figref{fig:baseline_flowchart}, and plot detailed architectures in~\appendixref{appendix:baseline_arch}.

\begin{figure}[!htb]
    \centering
    \includegraphics[width=\linewidth]{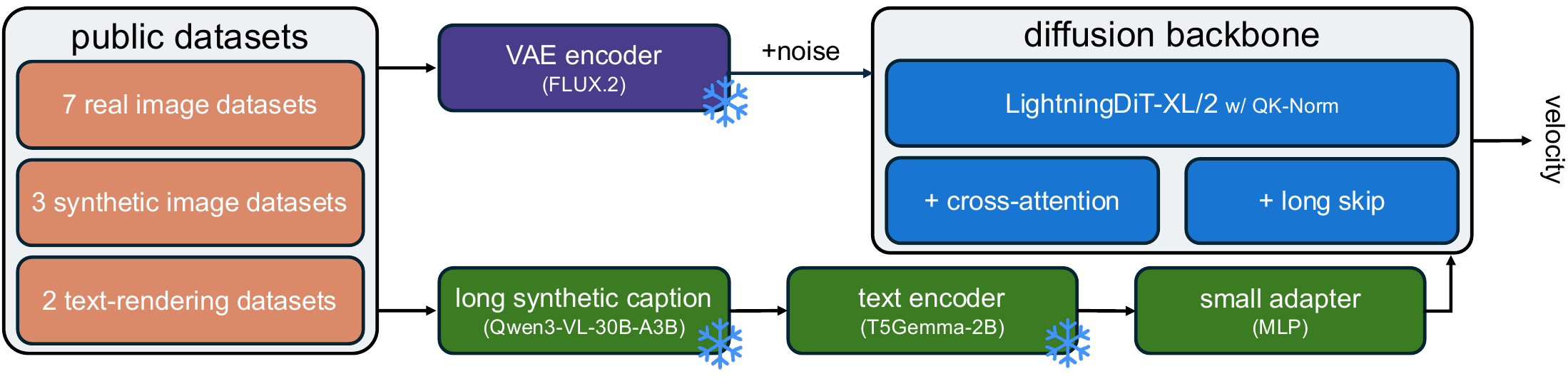}
    \caption{\textbf{High-level illustration of our baseline for controlled experiments}. We build a standard cross-attention architecture on top of LightningDiT~\citep{yao2025reconstruction} and add QK-norm for training stability. We also include long skip connections~\citep{bao2023all}, an underused design choice that we revisit in~\secrefc{sec:component} and find helpful for performance.}
    \label{fig:baseline_flowchart}
\end{figure}

\paragraph{Model}. As illustrated in~\figref{fig:baseline_flowchart}, our backbone architecture is based on LightningDiT-XL/2~\citep{yao2025reconstruction}. LightningDiT is a modern DiT architecture~\citep{peebles2023scalable} that incorporates common designs for improving performance (\eg, RoPE~\citep{su2024roformer}, RMS Norm~\citep{zhang2019root}, SwiGLU FFN~\citep{shazeer2020glu}). We add QK-norm~\citep{dehghani2023scaling} to stabilize training. 
To ensure later ablations compare against a strong baseline, we also apply long skip connections~\citep{bao2023all} (see~\secrefc{sec:prelim_related_work}), a less commonly used design that we revisit in~\secrefc{sec:component} and find helpful for performance.

By default, we use cross-attention to inject text embeddings. For some experiments, we additionally validate on single- and dual-stream variants (see~\secrefc{sec:prelim_related_work}) of our backbone to ensure generality of our findings. We use AdaLN to condition the model on the sum of the timestep embedding and a pooled text embedding, computed by averaging over text embedding tokens. For the single-stream architecture, we follow Lumina-Image 2.0~\citep{qin2025lumina} and prepend two modality-specific refiner blocks to the backbone. For both single- and dual-stream backbones, we adopt Multimodal-RoPE~\citep{wang2024qwen2}.
By default, we use the encoder part of T5Gemma-2B as our text encoder and use FLUX.2 VAE.

\begin{figure}[!htb]
\centering
\setlength{\tabcolsep}{0pt}

\newcommand{\datasetgroup}[4]{%
\begin{minipage}[t]{0.23\linewidth}
\centering
\small{#1}\\[.5pt]
\includegraphics[width=0.315\linewidth]{#2}\hfill
\includegraphics[width=0.315\linewidth]{#3}\hfill
\includegraphics[width=0.315\linewidth]{#4}
\end{minipage}%
}

\newcommand{\datasetgap}{\hspace{0.015\linewidth}}

\begin{tabular}{@{}c@{\datasetgap}c@{\datasetgap}c@{\datasetgap}c@{}}

\datasetgroup{ImageNet-22K}
{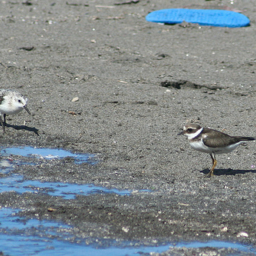}
{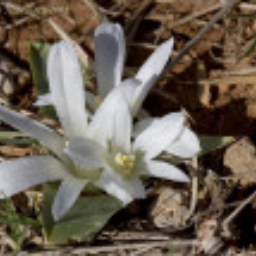}
{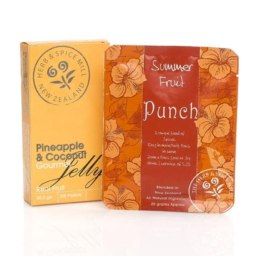}
&
\datasetgroup{YFCC}
{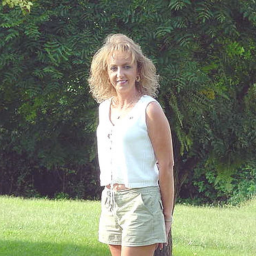}
{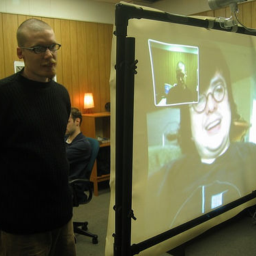}
{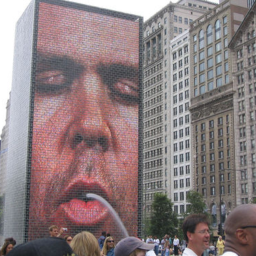}
&
\datasetgroup{RedCaps}
{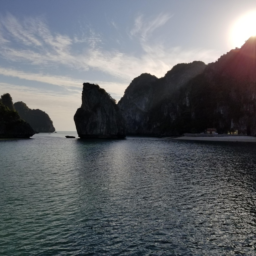}
{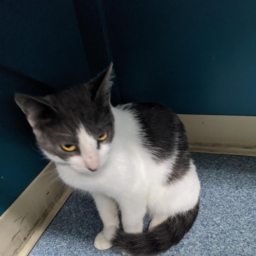}
{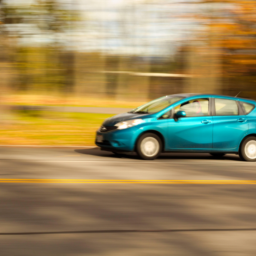}
&
\datasetgroup{Megalith}
{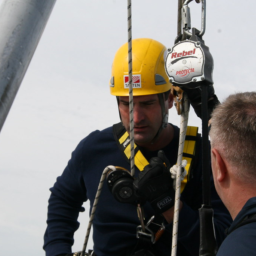}
{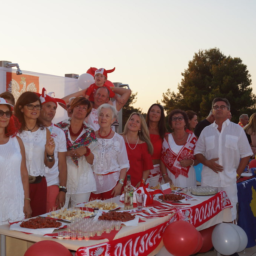}
{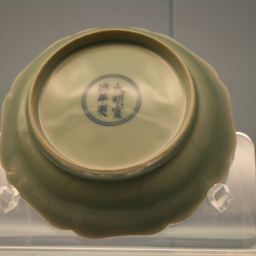}
\\[8pt]

\datasetgroup{Places}
{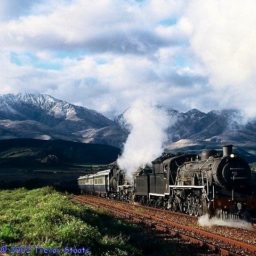}
{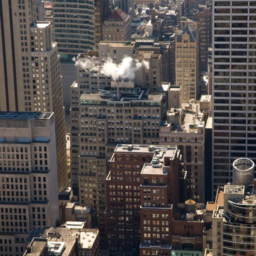}
{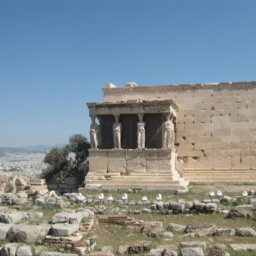}
&
\datasetgroup{Pexels}
{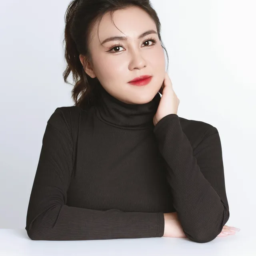}
{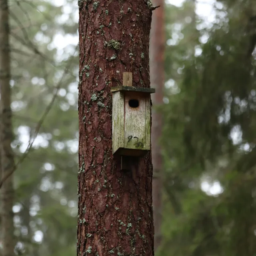}
{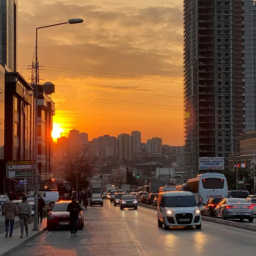}
&
\datasetgroup{iNaturalist}
{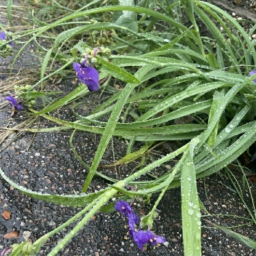}
{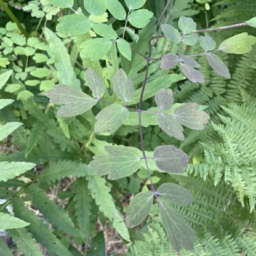}
{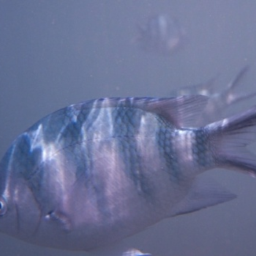}
&
\datasetgroup{FLUX-Reason}
{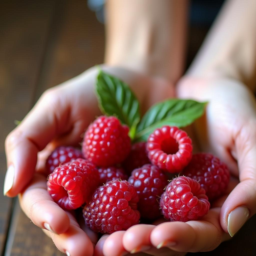}
{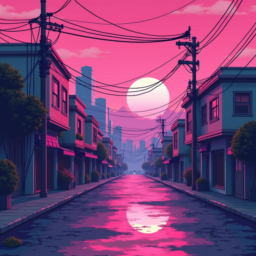}
{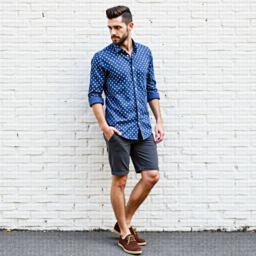}
\\[8pt]

\datasetgroup{Midjourney v6}
{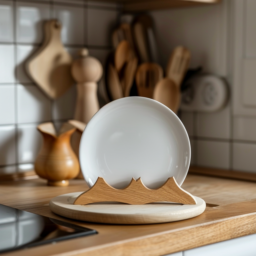}
{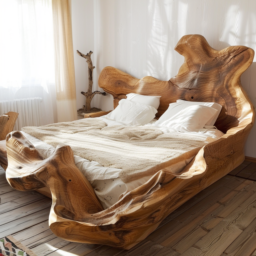}
{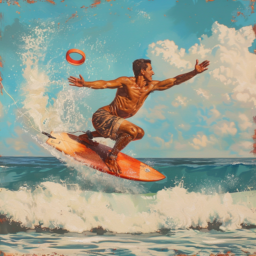}
&
\datasetgroup{GPT-Edit}
{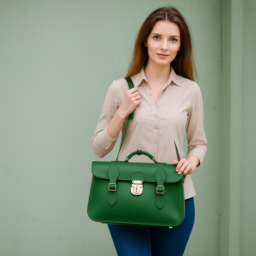}
{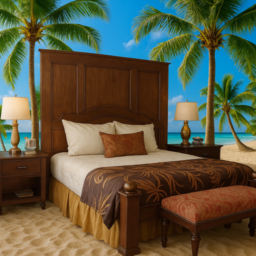}
{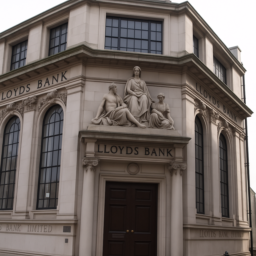}
&
\datasetgroup{TextAtlas}
{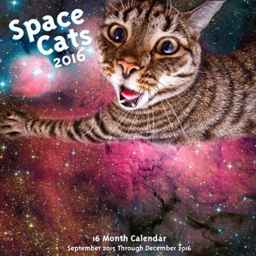}
{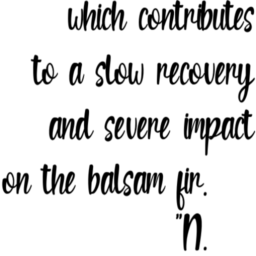}
{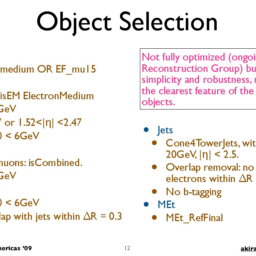}
&
\datasetgroup{RenderedText}
{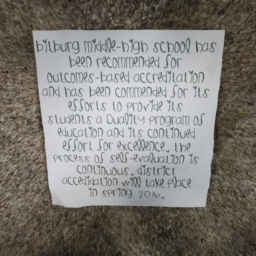}
{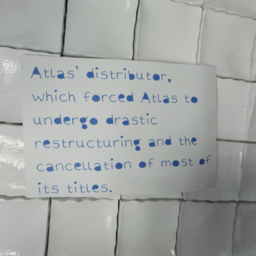}
{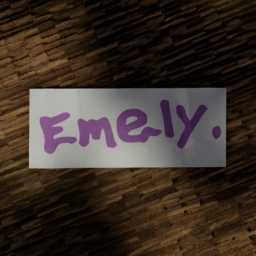}

\end{tabular}

\caption{\textbf{Example images from each image dataset} (more in~\appendixref{appendix:dataset_vis}). We use 12 curated image datasets for our controlled experiments, including 7 real-image datasets, 3 synthetic datasets, and 2 text-rendering datasets.}
\label{fig:dataset_example}
\end{figure}

\paragraph{Data}. We exclusively use publicly available image datasets, including 7 real-image datasets (ImageNet-22K~\citep{deng2009imagenet}, YFCC100M~\citep{thomee2016yfcc100m}, RedCaps~\citep{desai2021redcaps}, Megalith~\citep{BoerBohan2024Megalith10m}, Pexels~\citep{Narugo2024PexelsTaggerV0}, iNaturalist 2024~\citep{vendrow2024inquire}, Places365-Challenge 2016~\citep{zhou2017places}), 3 synthetic datasets (GPT-Image-Edit-1.5M~\citep{wang2025gpt}, FLUX-Reason-6M~\citep{fang2026flux}, and Midjourney v6~\citep{CortexLM2024MidjourneyV6}), and 2 text-rendering datasets (RenderedText~\citep{Wendler2024RenderedText} and TextAtlas~\citep{wang2025textatlas5m}). 
By default, we naively combine the 168M images in these datasets without weighting (a design choice that we revisit later in~\secrefc{sec:data_mixing}). In pre-training, all images are center-cropped to squares and resized to 256$\times$256. We present example images from each dataset in~\figref{fig:dataset_example}. We generate one long synthetic caption per image using the prompt ``Describe the image in detail using one paragraph.'' with Qwen3-VL-30B-A3B~\citep{bai2025qwen3} in FP8 precision (the VLM receives images that are center-cropped to squares, and resized to 512$\times$512 if larger than 512$\times$512). Further information is provided in~\appendixref{appendix:additional_info_data}.

\begin{table}[!htb]
\centering
\begin{tabular}{l c c c}
      \toprule
      baseline variant & DPG $\uparrow$ & PRISM $\uparrow$ & LongText $\uparrow$\\
      \midrule
      cross-attention & 84.66 & 56.4 & 0.211 \\
      single-stream & 85.89 & 55.6 & 0.293 \\
      dual-stream & 86.82 & 58.3 & 0.439 \\
      \bottomrule
\end{tabular}
\caption{\textbf{Benchmark performance of baselines}. We report benchmark scores for the baselines in our controlled experiments. We use the cross-attention variant by default and validate some of our designs across all three variants.}
\label{tab:baseline}
\end{table}

\paragraph{Training and inference}. We train the model using the flow matching~\citep{lipman2022flow} objective for 500K iterations (\ie, 25\% of the 2M-step 256-resolution pre-training stage of our final \textbf{i1} model) with a batch size of 512 and a learning rate of 1e-4. We use a 250-step Euler integrator with a classifier-free guidance~\citep{ho2022classifier} scale of 12 for sampling. More details are in~\appendixref{appendix:config}.

\begin{figure}[!htb]
    \centering
    \hspace{\fill}
    \begin{subfigure}[t]{0.32\linewidth}
        \centering
        {\scriptsize\linespread{0.5}\selectfont \textbf{Prompt}: On a reflective metallic table, there is a brightly colored handbag featuring a floral pattern next to a freshly sliced avocado... with silverware and a clear glass water bottle positioned neatly beside the avocado... (77 words)\par}
        \includegraphics[width=0.9\linewidth]{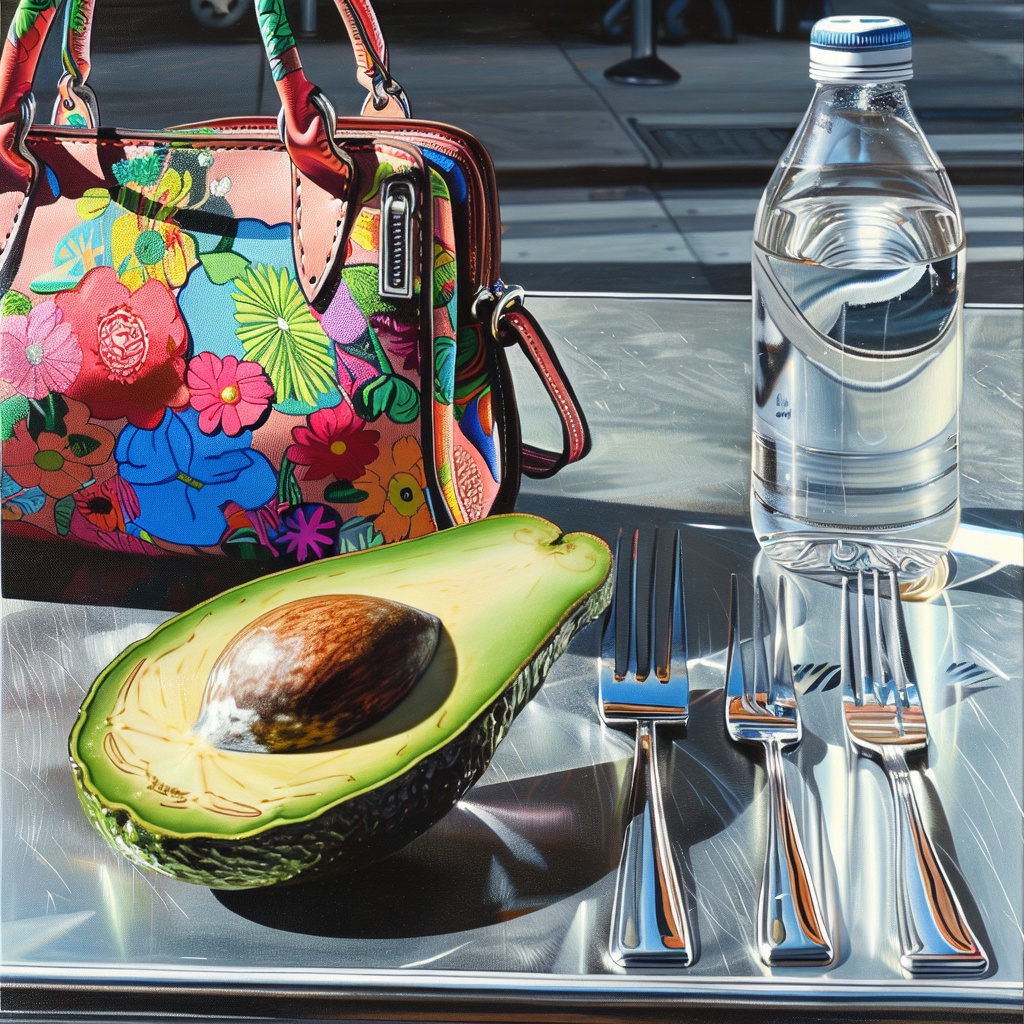}
        \caption{DPG-Bench}
    \end{subfigure}
    \hspace{\fill}
    \begin{subfigure}[t]{0.32\linewidth}
        \centering
        {\scriptsize\linespread{0.5}\selectfont \textbf{Prompt}: Lindsey Wixson stands confidently in a golden wheat field, donning a wide-brimmed straw hat, bold red sunglasses, and a vibrant red fur-trimmed top, accessorized with a sparkling diamond necklace, embodying summer elegance.\par}
        \includegraphics[width=0.9\linewidth]{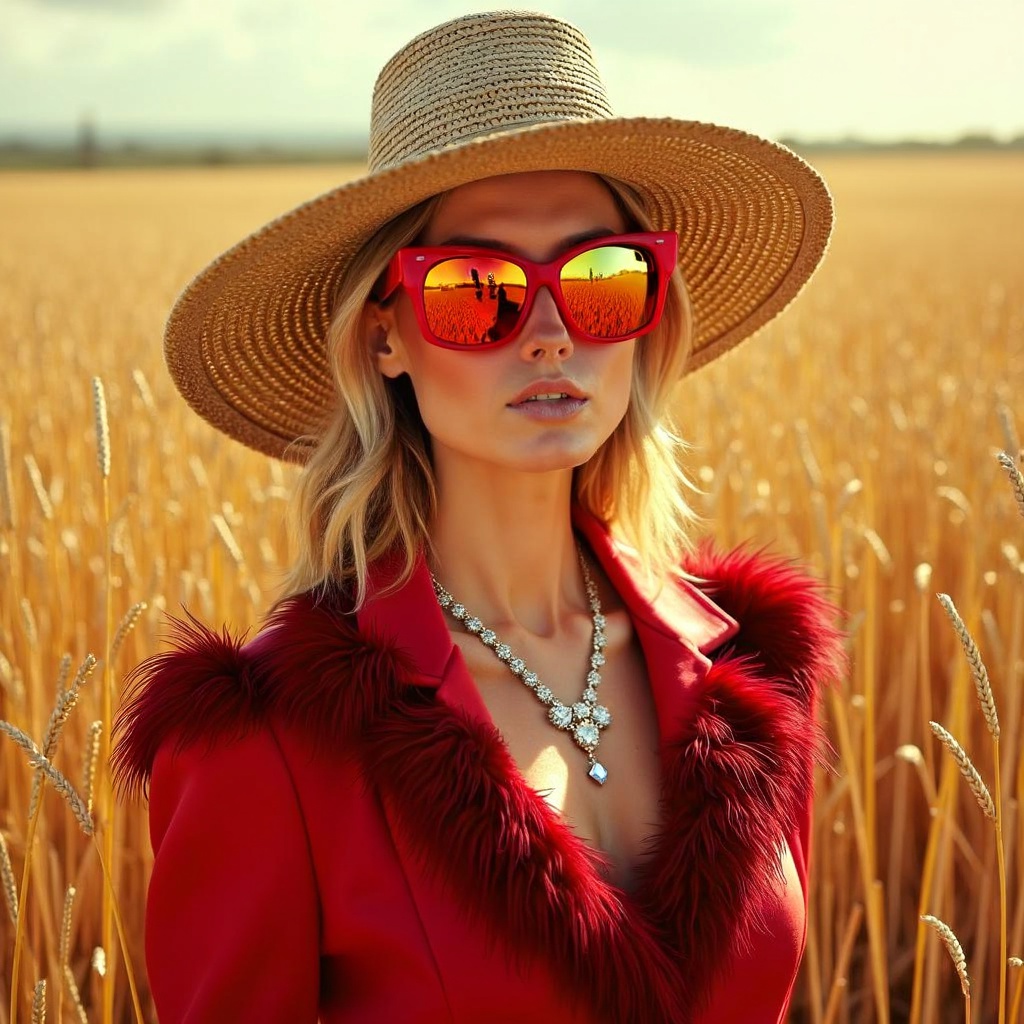}
        \caption{PRISM-Bench}
    \end{subfigure}
    \hspace{\fill}
    \begin{subfigure}[t]{0.32\linewidth}
        \centering
        {\scriptsize\linespread{0.5}\selectfont \textbf{Prompt}: An elegant, professional-looking mobile interface for a productivity and habit-tracking app named "DailyFlow". Positioned prominently at the top is the app name in bold, rounded typography colored soothing teal... (170 words)\par}
        \includegraphics[width=0.9\linewidth]{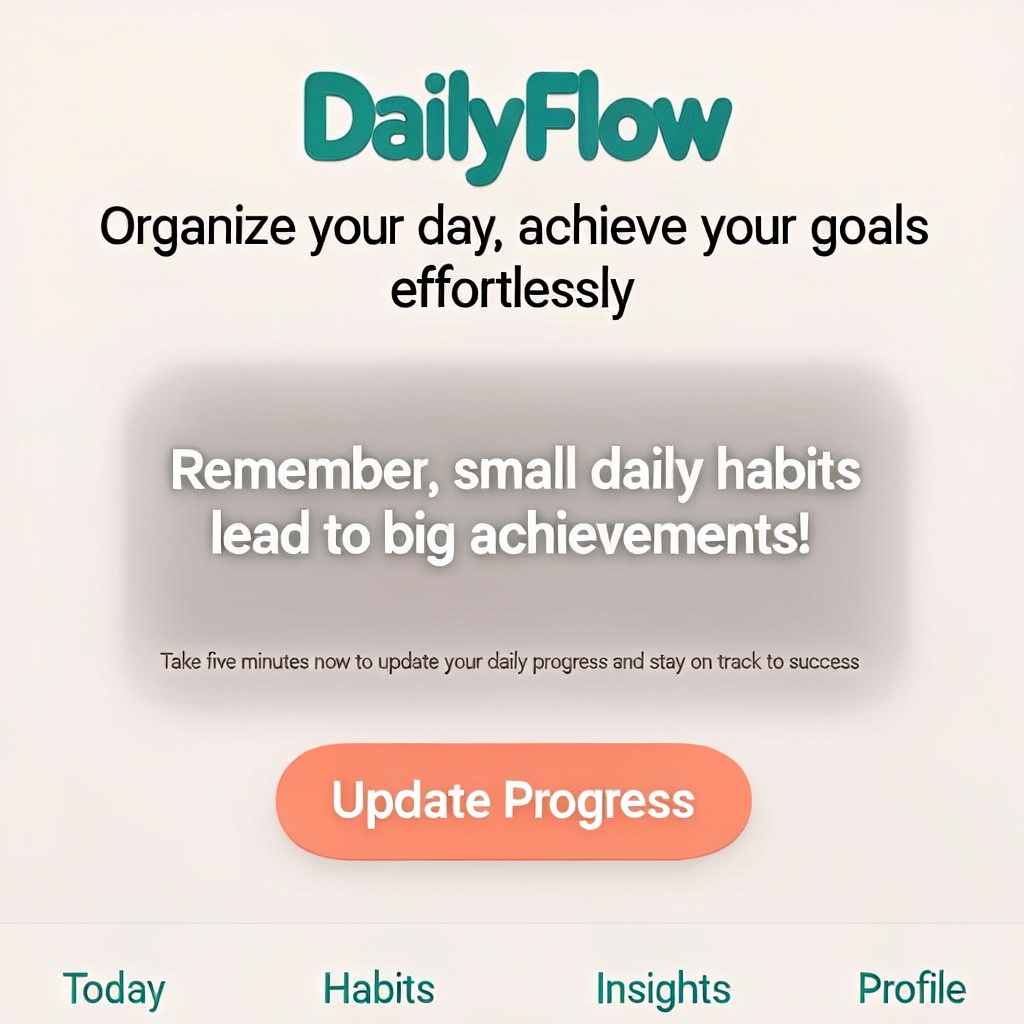}
        \caption{LongText-Bench}
    \end{subfigure}

    \caption{\textbf{Example prompts from benchmarks used in our controlled experiments}, along with corresponding \textbf{i1}-generated images. DPG and PRISM evaluate general prompt-following capabilities across diverse prompts, whereas LongText specifically evaluates text-rendering capabilities.}
    
    \label{fig:example_prompt}
\end{figure}

\paragraph{Evaluation}. We use three widely used benchmarks to provide signals for our controlled experiments: DPG-Bench~\citep{hu2024ella}, PRISM-Bench~\citep{fang2026flux}, and LongText-Bench~\citep{geng2025x}. All three benchmarks use VLMs as evaluators. DPG and PRISM measure fine-grained prompt-following capabilities across diverse prompts, where PRISM additionally evaluates image aesthetics. LongText specifically evaluates text-rendering capability. We present example prompts from each benchmark in~\figref{fig:example_prompt}. We use original prompts for DPG and LongText and rewrite PRISM prompts using Qwen3-4B~\citep{yang2025qwen3} with the simple meta-prompt in \secrefc{sec:caption_and_rewrite}. \tabref{tab:baseline} reports the performance of our baselines on these benchmarks.

\section{Modeling}
\label{sec:modeling}

We study modeling design modifications to the baseline introduced in~\secrefc{sec:exp_setup}. We first revisit text and noise conditioning mechanisms, including multiple text encoders and AdaLN, and identify stronger alternative designs. We then explore backbone architecture choices. More results are provided in~\appendixref{appendix:modeling}.

\subsection{Text and Noise Conditioning}
\label{sec:text_noise_cond}

Existing methods have explored various ways of incorporating text and noise conditioning into the backbone diffusion model. Some use a single text encoder~\citep{cai2025z, qin2025lumina, wu2025qwen}, while others concatenate features from multiple text encoders~\citep{esser2024scaling, cai2025hidream}. In addition, embeddings of the noise level are often injected into the model through AdaLN~\citep{peebles2023scalable}, sometimes together with a pooled text embedding~\citep{esser2024scaling, cai2025hidream, labs2025flux}. We investigate this broad design space and show that, rather than combining multiple encoders, it is more beneficial to use a single strong text encoder with a larger adapter. We also find that AdaLN-based conditioning on noise level and pooled text embeddings may not be necessary for text-to-image models.

\paragraph{Text encoder}. Early models (\eg, SD 1.5~\citep{rombach2022high}) primarily used CLIP-style text encoders. Later models adopted the encoder–decoder model T5~\citep{raffel2020exploring, esser2024scaling, cai2025hidream}. Most recently, models often use decoder-only LLMs or VLMs~\citep{wu2025qwen, qin2025lumina}, a trend often attributed to their powerful reasoning and complex instruction following capabilities~\citep{xie2025sana}.

\begin{figure}[h]
    \centering
    \includegraphics[width=\linewidth]{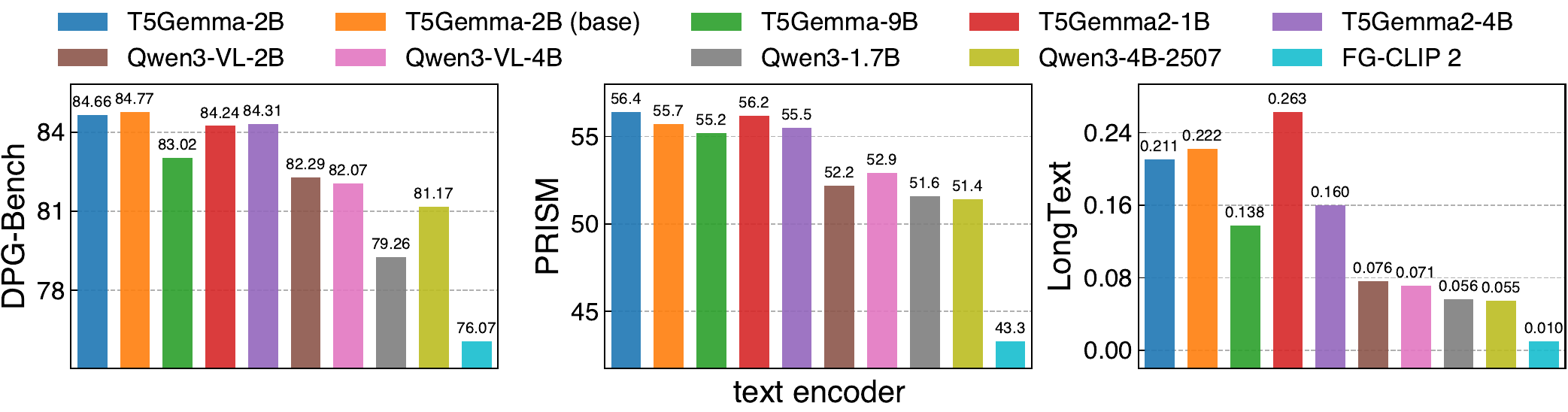}
    \caption{\textbf{Text encoders' performance} across benchmarks. Under our modeling setup, the encoder-decoder T5Gemma models outperform representative decoder-only LLM/VLMs and CLIP-style models. More results in~\appendixref{appendix:modeling}.}
    \label{fig:text_encoder}
\end{figure}

Here, we compare (1) a modern CLIP-based encoder (FG-CLIP 2~\citep{xie2025fg}), (2) two families of modern encoder–decoder models, T5Gemma~\citep{zhang2025encoder} and T5Gemma2~\citep{zhang2025t5gemma}, (3) a family of modern decoder-only LLMs (Qwen3~\citep{yang2025qwen3}), and (4) a family of modern decoder-only VLMs (Qwen3-VL~\citep{bai2025qwen3}). Unless otherwise specified, we use instruction-tuned checkpoints when available and otherwise use the corresponding base checkpoints. \figref{fig:text_encoder} reports the text-to-image performance using each model as the text encoder (comparisons under alternative settings are in Appendices~\blueref{appendix:text_encoder_under_larger_connector} and~\blueref{appendix:text_encoder_system_prompt}).

We observe that instruction tuning has minimal impact (\eg, T5Gemma-2B \vs T5Gemma-2B (base)) and larger models do not necessarily perform better (\eg, T5Gemma-2B \vs T5Gemma-9B).
Most importantly, encoder-decoder models (T5Gemma and T5Gemma2) achieve the best overall performance, outperforming FG-CLIP 2 and the decoder-only LLMs and VLMs.
This leads to the first finding that affects our design:
\finding{2}{Both encoder-decoder models and decoder-only LLMs/VLMs can be competitive text encoders for text-to-image diffusion models.\\[.2em]
\textbf{Design}: We use T5Gemma-2B, an encoder-decoder model, as the text encoder in \textbf{i1} since it is one of the strongest models in our comparison.}

\paragraph{Combining text encoders}. Many recent models~\citep{esser2024scaling, cai2025hidream} combine text features from multiple encoders (\eg, CLIP, T5, and LLMs). Here, we experiment with different combinations of the text encoders evaluated in \figref{fig:text_encoder}. Previous work uses different strategies to combine embeddings from different encoders (\eg, embedding- \vs sequence-dimension concatenation). In this work, we concatenate text features along the sequence dimension and use a separate adapter for each encoder to accommodate their different embedding dimensions. This avoids the need to pad text features to a common sequence length, which would often be required when concatenating features along the embedding dimension.

We combine one of the strongest text encoders from~\figref{fig:text_encoder}, T5Gemma-2B, with one additional encoder. As~\tabref{tab:combine_text_encoder} shows, combining it with T5Gemma2-1B or FG-CLIP 2 yields the best performance. However, combining all three (T5Gemma-2B, T5Gemma2-1B, and FG-CLIP 2) provides no substantial further gains.

\begin{table}[!htb]
\centering
\setlength{\tabcolsep}{5pt} 
\begin{tabular}{l c c c c c c c c}
    \toprule
    type & T5G-2B & T5G2-1B & T5G2-4B & Qwen3-VL-2B & FG-CLIP 2 & DPG & PRISM & LongText\\
    \midrule
    \multirow{1}{*}{baseline} & \cmark & & & & & 84.66 & 56.4 & 0.211 \\
    \midrule
    \multirow{4}{*}{+1 encoder} & \cmark & \cmark & & & & \bf 85.62 & \underline{58.4} & \underline{0.303} \\
    & \cmark & & \cmark & & & 84.86 & 55.8 & 0.270 \\
    & \cmark & & & \cmark & & 84.80 & 57.4 & 0.264 \\
    & \cmark & & & & \cmark & 85.72 & 57.7 & 0.285 \\
    \midrule
    \multirow{2}{*}{+2 encoder} & \cmark & \cmark & & & \cmark & \underline{85.37} & \bf 58.8 & 0.272 \\
    & \cmark & & \cmark & & \cmark & 85.28 & 58.1 & \bf 0.351 \\
    \bottomrule
  \end{tabular}
\caption{\textbf{Combining text encoders can improve performance}. We explore combining T5Gemma-2B with one or two additional text encoders, and find the combination with T5Gemma2-1B and FG-CLIP 2 to be the strongest.}
\label{tab:combine_text_encoder}
\end{table}

\begin{figure}[!htb]
    \centering
    \begin{minipage}[c]{0.54\linewidth}
        \centering
        \setlength{\tabcolsep}{5pt} 
        \begin{tabular}{l c c c}
            \toprule
            text encoder & DPG $\uparrow$ & PRISM $\uparrow$ & LongText $\uparrow$\\
            \midrule
            T5Gemma-2B & 84.66 & 56.4 & 0.211 \\
            repeat {\scriptsize w/1 MLP} & 84.93 & 55.8 & 0.225 \\
            repeat {\scriptsize w/2 MLP} & \bf 85.09 & \bf 56.5 & \bf 0.309 \\
            \bottomrule
          \end{tabular}
    \end{minipage}%
    \hfill
    \begin{minipage}[c]{0.42\linewidth}
        \centering
        \includegraphics[width=\linewidth, valign=c]{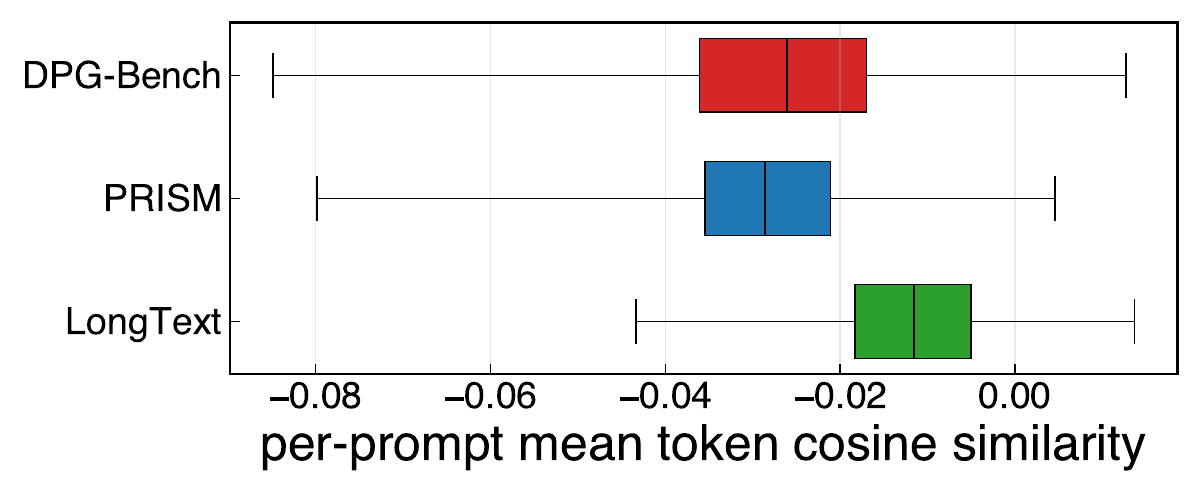}
    \end{minipage}

    \begin{minipage}[t]{0.59\linewidth}
        \captionof{table}{\textbf{Concatenating two copies of T5Gemma-2B feature sequences} and using two separate MLP adapters (equivalent to combining two T5Gemma-2B text encoders) yields a similar improvement as combining different text encoders, whereas using a shared MLP adapter does not. This suggests the improvement may come from additional adapter parameters, not separate text encoders.}
        \label{tab:repeat_t5gemma_text_emb}
    \end{minipage}%
    \hfill
    \begin{minipage}[t]{0.37\linewidth}
        \captionof{figure}{\textbf{The two MLPs learn different features}. We obtain two sets of features for each prompt using the two MLPs, compute cosine similarity between each pair of token-level feature vectors, and visualize the distribution of mean similarity across tokens per prompt.}
        \label{fig:two_mlp_connector_cosine_sim}
    \end{minipage}
\end{figure}

Although combining text encoders improves performance, does the improvement arise from the diverse representations provided by different encoders or simply from the increased sequence length and additional parameters introduced by the adapters? To investigate this, we construct two baselines that repeat the T5Gemma-2B text embeddings: the first uses two separate adapters for the two identical copies of embeddings (thus increasing both sequence length and adapter parameters), while the second uses a shared adapter (thus increasing only sequence length).
As shown in \tabref{tab:repeat_t5gemma_text_emb}, repeating the embeddings with two separate adapters brings a noticeable improvement, whereas using a shared adapter produces results similar to the baseline without repetition. This suggests that the gains from combining multiple text encoders may largely stem from the additional adapters rather than from diverse text encoder features or longer sequences.

We sanity check that the two MLP adapters learn distinct features by comparing embeddings for DPG-Bench, PRISM, and LongText prompts from each adapter. We measure cosine similarity between the two embeddings for each token and average it across tokens for each prompt. The resulting distribution, shown in \figref{fig:two_mlp_connector_cosine_sim}, is largely negative, indicating that the two adapters indeed capture different representations.

\begin{figure}[!htb]
    \centering
    \includegraphics[width=\linewidth]{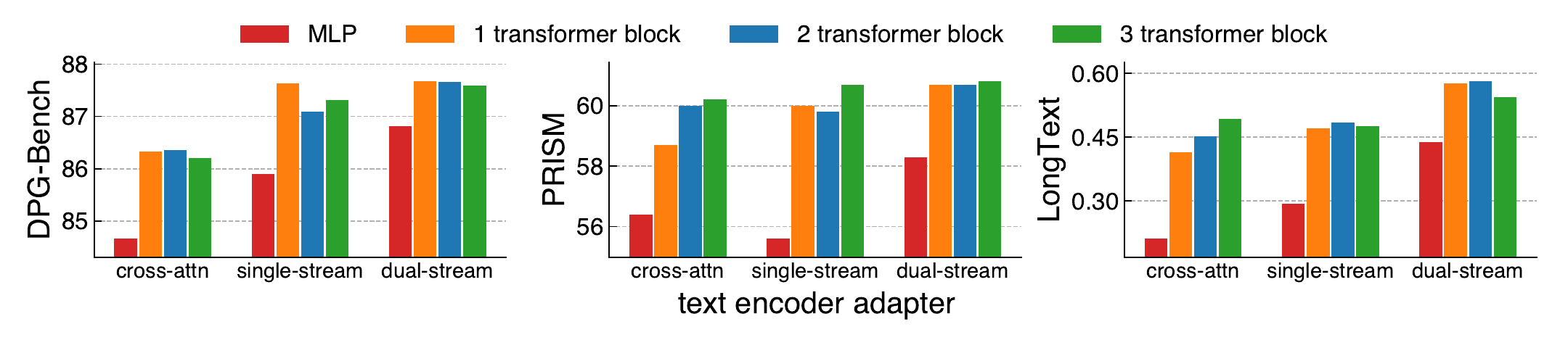}
    \caption{\textbf{Using larger adapters for the text encoder consistently improves performance} across backbone architectures. Beyond 2 transformer blocks, using larger adapters brings marginal further gains.}
    \label{fig:trans_connector}
\end{figure}

\paragraph{Larger text encoder adapter}. To further investigate the hypothesis that performance improvements from combining multiple text encoders largely stem from the additional adapter parameters instead of diverse text encoder features, we replace the small MLP adapter (2.6M parameters) used in the default setup (\secrefc{sec:exp_setup}) with larger transformer adapters (17.2M parameters/block) with the same width as the backbone blocks. As shown in \figref{fig:trans_connector}, despite the marginal increase in parameter count, increasing the adapter capacity consistently improves performance across all backbone architectures. Nonetheless, expanding the adapters beyond two transformer blocks yields only marginal additional gains.

Furthermore, as the ``default'' and ``+2 encoders'' rows of \tabref{tab:mlp_vs_transformer_adapter} show, when using a larger adapter, combining multiple text encoders yields much smaller gains across all backbones, especially on DPG and PRISM. This further suggests that the benefit of multiple text encoders can be captured by increasing the adapter capacity for a single text encoder. Importantly, using multiple encoders increases the text sequence length, substantially raising memory and computational cost, whereas using a larger adapter does not.

\finding{2}{The gains from using multiple text encoders can be similarly captured by increasing the adapter capacity for a single text encoder. Compared to using multiple encoders, increasing adapter capacity has lower memory and compute cost because it does not increase the text sequence length.\\[.2em]
\textbf{Design}: We use a single, strong text encoder with an expressive text encoder adapter in \textbf{i1}.
}

\begin{table}[!htb]
\begin{center}{
\resizebox{1.00\textwidth}{!}{
\begin{tabular}{l l c c c c c c c c c}
\toprule
  & & 
    \multicolumn{4}{c}{MLP adapter (default)} 
  & \multicolumn{4}{c}{transformer adapter (1x block)} \\
  \cmidrule(lr){3-6}\cmidrule(l){7-10}
  & & \#params & DPG & PRISM & LongText & \#params & DPG & PRISM & LongText \\
\hline
\multicolumn{2}{l}{\hspace{-.5em}\textit{cross-attn}} \\
  & default & 0.89B & 84.66 & 56.4 & 0.211 & 0.91B & 86.33 & 58.7 & 0.414 \\
  & +2 encoders & 0.90B & 85.37 \better{0.71} & 58.8 \better{2.4} & 0.272 \better{0.061} & 0.94B & 86.47 \better{0.14} & 59.5 \better{0.8} & 0.491 \better{0.077} \\
  & no pooled emb & 0.89B & 85.98 \better{1.32} & 57.5 \better{1.1} & 0.391 \better{0.180} & 0.91B & 86.37 \better{0.04} & 59.4 \better{0.7} & 0.446 \better{0.032} \\
  & no timestep & 0.89B & 82.58 \worse{2.08} & 54.7 \worse{1.7} & 0.185 \worse{0.026} & 0.91B & 84.71 \worse{1.62} & 58.9 \better{0.2} & 0.418 \better{0.004}\\
  & no AdaLN & 0.66B & 84.99 \better{0.33} & 57.4 \better{1.0} & 0.351 \better{0.140} & 0.67B & 85.13 \worse{1.20} & 59.7 \better{1.0} & 0.413 \worse{0.001} \\
\hline
\multicolumn{2}{l}{\hspace{-.5em}\textit{single-stream}} \\
  & default & 0.82B & 85.89 & 55.6 & 0.293 & 0.83B & 87.64 & 60.0 & 0.472 \\
  & +2 encoders & 0.83B & 84.89 \worse{1.00} & 56.3 \better{0.7} & 0.439 \better{0.146} & 0.87B & 87.29 \worse{0.35} & 59.0 \worse{1.0} & 0.428 \worse{0.044} \\
  & no AdaLN & 0.57B & 87.38 \better{1.49} & 59.0 \better{3.4} & 0.390 \better{0.097} & 0.58B & 87.39 \worse{0.25} & 59.5 \worse{0.5} & 0.410  \worse{0.062} \\
\hline
\multicolumn{2}{l}{\hspace{-.5em}\textit{dual-stream}} \\
  & default & 1.24B & 86.82 & 58.3 & 0.439 & 1.25B & 87.67 & 60.7 & 0.576 \\
  & +2 encoders & 1.25B & 87.34 \better{0.52} & 59.6 \better{1.3} & 0.514 \better{0.075} & 1.29B & 87.76 \better{0.09} & 60.8 \better{0.1} & 0.588 \better{0.012} \\
  & no AdaLN & 1.01B & 87.82 \better{1.00} & 60.3 \better{2.0} & 0.508 \better{0.069} & 1.02B & 87.38 \worse{0.29} & 60.7 \same{0.0} & 0.554  \worse{0.022} \\
\hline
\end{tabular}
}
}
\end{center}
\vspace{-.5em}
\caption{
\textbf{Impact of text and noise conditioning when using an MLP \vs transformer adapter}.
(1) In most cases, a larger transformer adapter improves performance while adding minimal parameters.
(2) With a larger adapter, combining multiple text encoders provides much smaller benefit, suggesting that prior gains may mainly stem from increased adapter capacity.
(3) Removing pooled text embeddings or timestep embeddings from AdaLN, or removing AdaLN conditioning entirely, barely degrades performance.
We validate these findings on a 3B MMDiT model in~\appendixref{appendix:validate_on_3b}.
}
\label{tab:mlp_vs_transformer_adapter}
\end{table}

\paragraph{Removing AdaLN conditioning}. Adaptive Layer Normalization (AdaLN)~\citep{peebles2023scalable} is a standard component in modern text-to-image diffusion models~\citep{labs2025flux, wu2025qwen, qin2025lumina, cai2025hidream}. It is typically used to inject timestep embeddings and pooled text embeddings into the backbone. A recent study~\citep{sun2025noise} showed that in class-conditional image generation, removing noise conditioning only minimally affects performance, especially for flow matching models. If AdaLN can be removed without harming performance, the model could become more parameter-efficient.

Interestingly, as shown in \tabref{tab:mlp_vs_transformer_adapter}, removing AdaLN from the default setup (\secrefc{sec:exp_setup}) consistently improves performance when the text encoder adapter is a small MLP. However, the effect becomes much smaller when using a larger transformer adapter. To better understand this behavior, we perform additional ablations on the cross-attention backbone, where AdaLN conditions only on pooled text embeddings or only on timestep embeddings, rather than their sum. We evaluate these variants with both small and large adapters.

We find that AdaLN conditioning on pooled text embeddings reduces performance when the adapter is small (84.99 $\rightarrow$ 82.58 on DPG), but has a much smaller effect when the adapter is large (85.13 $\rightarrow$ 84.71 on DPG). This suggests that the performance gains from removing AdaLN may primarily result from poorly learned features when using the small MLP adapter. However, even when a larger adapter is used, conditioning on the pooled text and timestep embeddings through AdaLN still provides marginal additional benefit.

\finding{2}{Despite the large number of parameters introduced by AdaLN, conditioning on pooled text embeddings, timestep embeddings, or both through AdaLN provides only marginal benefit.\\[.2em]
\textbf{Design}: We do not use AdaLN in \textbf{i1}.
}

\subsection{Backbone Architecture}
\label{sec:component}

In this subsection, we revisit the long skip connection design and provide a controlled comparison of popular backbone families based on the baseline setup in~\secrefc{sec:exp_setup}. We include additional analyses of positional embeddings, normalization, and VAEs in Appendices~\blueref{appendix:pos_emb_and_norm} and~\blueref{appendix:vae}.

\begin{figure}[h]
    \centering
    \vspace{-.3em}
    \includegraphics[width=\linewidth]{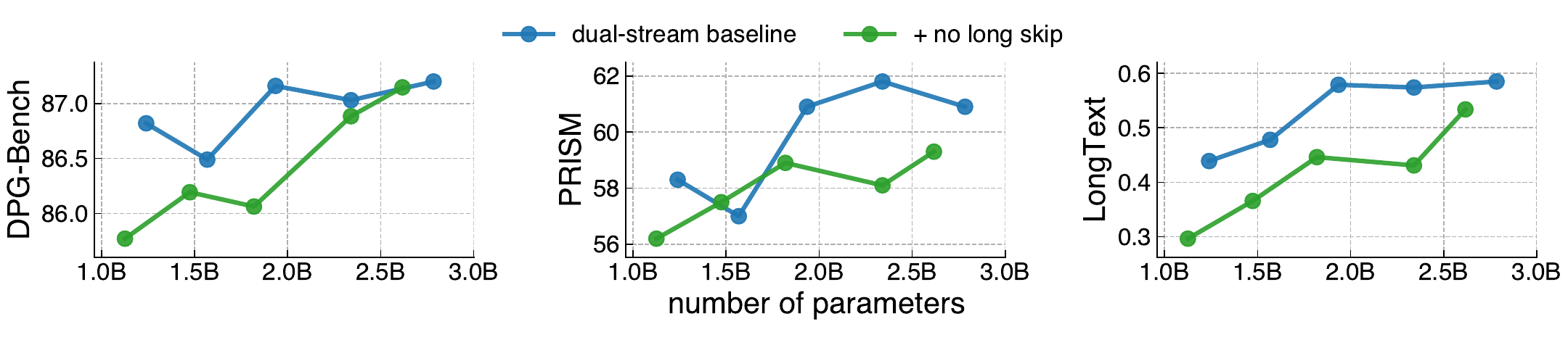}
    \caption{\textbf{Long skip connections}~\citep{bao2023all} can improve the performance-parameter trade-off for dual-stream models. Additional FLOPs-based analysis is in~\figref{fig:validate_on_3b_flops}, and results on other backbone families are in~\appendixref{appendix:long_skip_other_backbone}.}
    \label{fig:long_skip_scaling}
\end{figure}

\paragraph{Long skip connections} add shortcuts between early and later layers. They were first popularized by U-Net~\citep{ronneberger2015u} and later applied to diffusion models in U-ViT~\citep{bao2023all}. While they were shown to improve performance~\citep{bao2023all, li2024hunyuan, liu2024playground}, they have not been widely applied to modern text-to-image models.
In~\figref{fig:long_skip_scaling}, we revisit this design by training dual-stream variants of the baseline (\secrefc{sec:exp_setup}) with and without long skip connections at multiple model widths (1152, 1296, 1440, 1584, and 1728), while keeping all other configurations fixed. We find that long skip connections consistently improve performance across model sizes, potentially due to enhanced model expressivity. Additional FLOPs-based analysis is in~\figref{fig:validate_on_3b_flops}, and results on other backbones are in~\appendixref{appendix:long_skip_other_backbone}.

\finding{2}{Long skip connections can improve the performance-parameter trade-off.\\[.2em]
\textbf{Design}: We use long skip connections in \textbf{i1}.}

\begin{figure}[!htb]
    \centering
    \includegraphics[width=\linewidth]{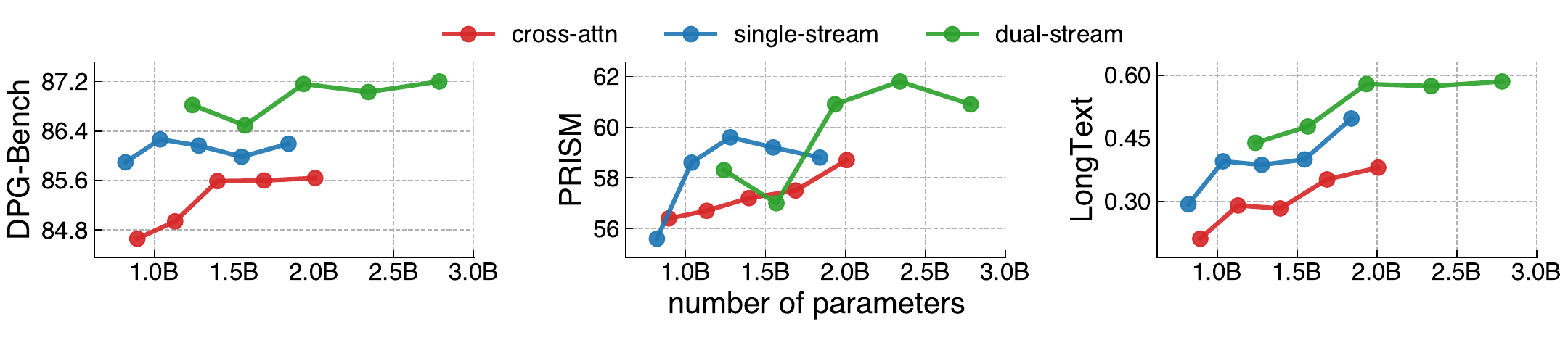}
    \caption{\textbf{Backbone family}. We compare cross-attention, single-stream, and dual-stream backbones across model sizes (see \figref{fig:backbone_scaling_flops} for training FLOPs analysis). We find that the dual-stream backbone achieves the best overall performance.}
    \label{fig:backbone_scaling}
\end{figure}

\paragraph{Backbone family}. Today's leading models differ in their choice of backbone: some use cross-attention, some use single-stream architectures, and others use dual-stream architectures (see~\secrefc{sec:prelim_related_work} for details). We measure model performance for cross-attention, single-stream, and dual-stream backbones at multiple model widths (1152, 1296, 1440, 1584, and 1728 for all three backbone families) while keeping all other model configurations fixed. \figref{fig:backbone_scaling} plots model performance against parameter count. We observe that the dual-stream backbone achieves the best performance-parameter trade-off.

\finding{2}{The dual-stream backbone has the best trade-off between performance and parameter count among the cross-attention, single-stream, and dual-stream backbone families.\\[.2em]
\textbf{Design}: We use the dual-stream backbone in \textbf{i1}.}

\section{Data}
\label{sec:data}

Besides modeling architectures, high-quality image-caption data is important for text-to-image training. In this section, we first study synthetic captioning designs, and show that training on long captions yields stronger models but can lead to poor performance on short prompts, which we mitigate via prompt rewriting at inference. We then explore dataset mixing and find that equal weighting across datasets is a strong default.

\subsection{Synthetic Captions and Prompt Rewrite}
\label{sec:caption_and_rewrite}

Prompt-following capability in text-to-image models fundamentally relies on high-quality image-caption pairs in the training data. Earlier work, such as Parti~\citep{yu2022scaling} and DALL-E 3~\citep{betker2023improving}, showed that training on highly descriptive synthetic captions generated by vision-language models can substantially improve performance. Since then, the majority of text-to-image models~\citep{chen2024pixart, esser2024scaling, qin2025lumina, xie2025sana} have leveraged synthetic captions during training.

Here, we explore several design choices in synthetic caption generation and their impact on model performance (more results in~\appendixref{appendix:additional_design_synthetic_caption}). In particular, we find that training on long synthetic captions yields stronger models, but these models can underperform on short prompts, necessitating inference-time prompt rewrite. To reduce computational cost for caption generation, all experiments in this section are conducted on the ImageNet-22K dataset rather than the full training set used in the default baseline setting (\secrefc{sec:exp_setup}).

\begin{figure}[!htb]
    \centering
    \includegraphics[width=\linewidth]{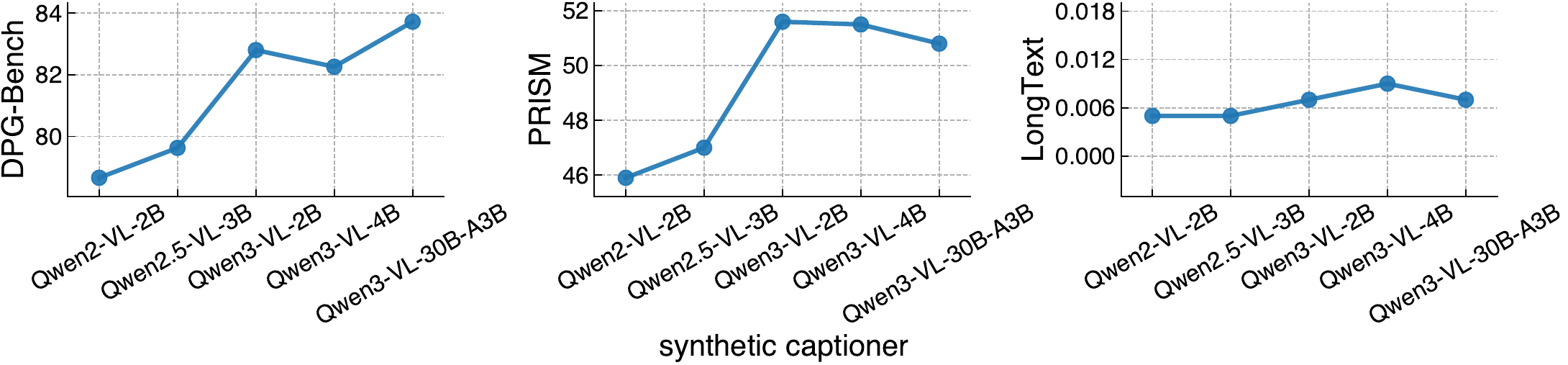}
    \caption{\textbf{The choice of synthetic captioner is important for downstream text-to-image performance}. Due to resource constraints, we generate captions and train only on ImageNet-22K images rather than the full image dataset.}
    \label{fig:stronger_captioner}
\end{figure}

\paragraph{Caption quality}. To explore how caption quality impacts downstream text-to-image performance, we generate captions using five VLMs: Qwen2-VL 2B, Qwen2.5-VL 3B, Qwen3-VL-2B, Qwen3-VL-4B, and Qwen3-VL-30B-A3B. As reported in \figref{fig:stronger_captioner}, the choice of synthetic captioner has a substantial impact on downstream text-to-image performance. We note that the small differences on LongText are primarily due to the overall poor text-rendering performance of models trained on ImageNet-22K, which contains few text-rich images. This result therefore does not imply that captioner quality is unimportant for text rendering.

\finding{2}{The choice of VLM used to synthesize captions for training images has a substantial impact on downstream text-to-image performance.\\[.2em]
\textbf{Design}: We use Qwen3-VL-30B-A3B to generate synthetic captions for training \textbf{i1}, since captions from this model lead to strong downstream performance.}

\paragraph{Caption length and prompt rewrite}. By default, we train our models using only long synthetic captions (see \secrefc{sec:exp_setup}). While our models can achieve strong performance on the original DPG and LongText prompts, they perform poorly on original GenEval prompts. We find that this may be explained by the much shorter prompts in GenEval compared to DPG and LongText (see~\figref{fig:prompt_length_5_benchmarks}): simply repeating the GenEval prompts 12 times leads to a large improvement in performance (0.17 $\rightarrow$ 0.49). This observation suggests that the poor performance on original, short GenEval prompts may stem from training exclusively on long captions.

\begin{wrapfigure}{R}{0.5\textwidth}
\centering
\includegraphics[width=0.95\linewidth]{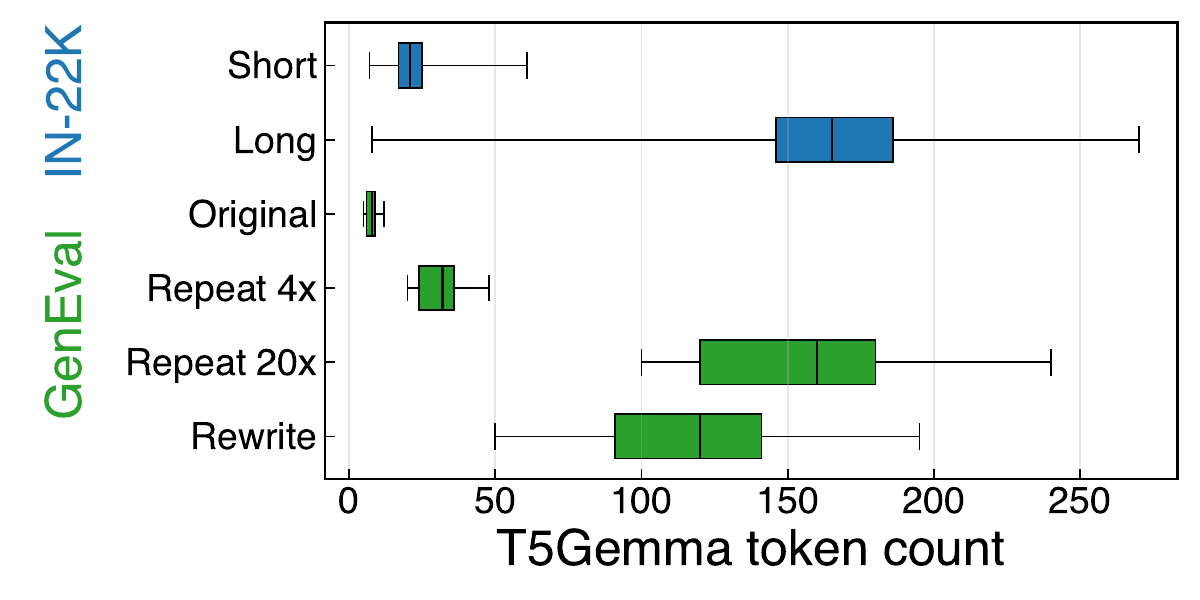}
\vspace{-.7em}
\captionsetup{width=.95\linewidth}
\caption{\textbf{Sequence length} of ImageNet-22K captions (10K random subset) and original, repeated, and rewritten GenEval prompts under T5Gemma tokenizer.}
\label{fig:seqlen_geneval}
\vspace{-3em}
\end{wrapfigure}

To further understand this, we generate an additional set of short captions using the prompt ``Describe the image using one short sentence.'' The distributions of prompt lengths are shown in~\figref{fig:seqlen_geneval}. We mix these short captions with the original long captions using different sampling weights and report the resulting GenEval scores in~\tabref{tab:geneval_train_test_prompt_length}. We observe that (1) training primarily on short captions (\eg, 0\% or 20\% long captions) improves performance on the original short GenEval prompts and (2) models trained with higher proportions of long captions perform better when the GenEval prompts are repeated.

\begin{table}[!htb]
\centering
\begin{tabular}{l c c c c c}
  \toprule
    \multirow{3}{*}{\makecell{\% of long captions\\in training captions}} & \multicolumn{5}{c}{performance on GenEval prompts} \\
    \cmidrule(lr){2-6}
     & \multirow{2}{*}{\makecell{original prompts\\(short)}} & \multicolumn{3}{c}{repeated prompts} & \multirow{2}{*}{\makecell{rewritten prompts\\(long)}} \\
    \cmidrule(lr){3-5}
     & & 4$\times$ & 12$\times$ & 20$\times$ & \\
    \midrule
    0\% & \bf 0.47 & 0.55 & 0.34 & 0.24 & 0.60 \\
    20\% & \bf 0.47 & 0.54 & 0.53 & 0.50 & 0.67 \\
    40\% & 0.35 & 0.59 & 0.55 & \bf 0.54 & 0.70 \\
    60\% & 0.37 & \bf 0.60 & \bf 0.57 & \bf 0.54 & \bf 0.73 \\
    80\% & 0.26 & 0.57 & 0.54 & 0.47 & \bf 0.73 \\
    100\% & 0.17 & 0.48 & 0.49 & 0.46 & \bf 0.73 \\
    \bottomrule
  \end{tabular}
\caption{\textbf{Training captions and inference prompts should have aligned lengths} (each number is a GenEval score). Our model trained entirely on long captions performs poorly on short GenEval prompts, but strong performance can be recovered by repeating the short GenEval prompts or applying an LLM-based rewrite. Overall, training only on long captions and using LLM-based prompt rewriting to increase inference prompt length leads to the strongest performance.}
\label{tab:geneval_train_test_prompt_length}
\end{table}

While repeating the short prompts can recover the performance, it introduces unnatural prompt structures. To address this issue, we instead use an LLM (Qwen3-4B) to rewrite the GenEval prompts using the following meta-prompt:

\begin{center}
    ``I have a short text-to-image prompt \{prompt\}. Please expand it into a descriptive paragraph, while making sure the generated image still clearly includes all the items mentioned in the original prompt. Please only output the rewritten prompt and nothing else.''
\end{center}

\begin{figure}[!htb]
\centering
\setlength{\tabcolsep}{1pt}
\begin{tabular}{@{}>{\footnotesize\raggedright\arraybackslash}m{.147\linewidth}cccc@{}}
{\footnotesize \textbf{Prompt}: a photo of a wine glass and a bear} &
\includegraphics[width=.207\linewidth,valign=c]{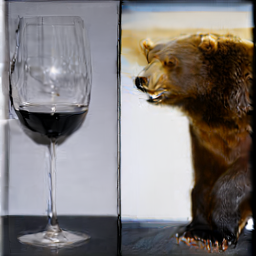} &
\includegraphics[width=.207\linewidth,valign=c]{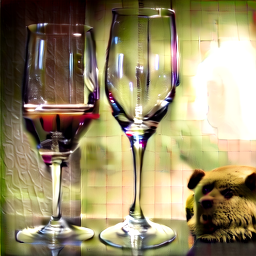} &
\includegraphics[width=.207\linewidth,valign=c]{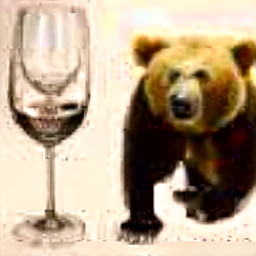} &
\includegraphics[width=.207\linewidth,valign=c]{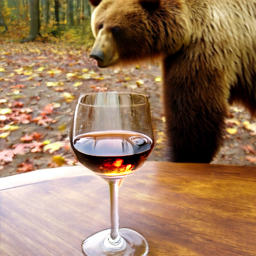} \\

{\footnotesize \textbf{Prompt}: a photo of a zebra right of a parking meter} &
\includegraphics[width=.207\linewidth,valign=c]{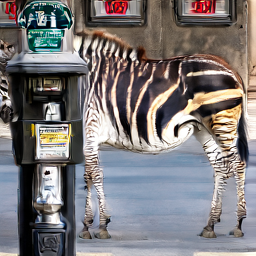} &
\includegraphics[width=.207\linewidth,valign=c]{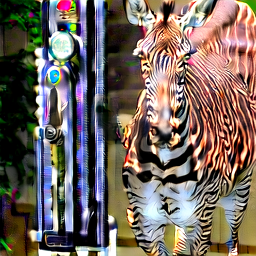} &
\includegraphics[width=.207\linewidth,valign=c]{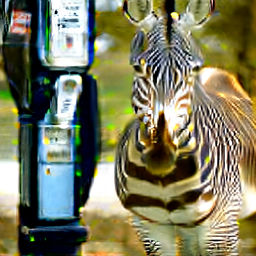} &
\includegraphics[width=.207\linewidth,valign=c]{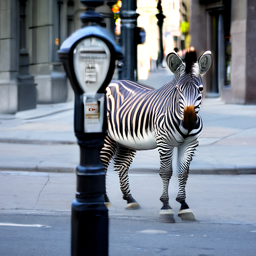} \\[2pt]

& \footnotesize\parbox{.207\linewidth}{\centering train: short,\\test: original (short)} &
\footnotesize\parbox{.207\linewidth}{\centering train: long,\\test: original (short)} &
\footnotesize\parbox{.207\linewidth}{\centering train: long,\\test: repeated 12$\times$ (long)} &
\footnotesize\parbox{.207\linewidth}{\centering train: long,\\test: rewritten (long)}
\end{tabular}

\caption{\textbf{Examples from models trained on ImageNet-22K caption variants and tested on GenEval prompt variants}. Training on long captions leads to weaker performance on short prompts, but prompt repetition and rewrite mitigate this.}
\label{fig:qualitative_caplen}

\end{figure}

As shown in the rightmost column of \tabref{tab:geneval_train_test_prompt_length} and~\figref{fig:qualitative_caplen}, rewriting the GenEval prompts substantially improves model performance. Notably, training on long captions and evaluating on rewritten prompts (0.73) significantly outperforms training on short captions and evaluating on original, repeated, or rewritten prompts. This suggests that, even when inference prompts are originally short (\eg, GenEval), it is preferable to train on long captions and increase the inference prompt length to match the training distribution (\eg, via prompt rewriting), rather than training on short captions to match the original inference prompt length.

\finding{2}{
Training on short captions weakens overall performance, whereas training on long captions yields stronger models but performs poorly on short test prompts. Prompt rewrite addresses this weakness on short prompts by expanding them, making training on long captions preferable overall.\\[.2em]
\textbf{Design}: We only use long captions to train \textbf{i1}, and apply inference-time prompt rewriting.}

\subsection{Data Mixing}
\label{sec:data_mixing}

All experiments up to this point naively combine all datasets without explicit dataset-level weighting. Because our training corpus (see~\secrefc{sec:exp_setup}) is highly imbalanced (\eg, YFCC contributes 98M of 168M images), we implicitly assign much larger weights to a few large datasets, which can dominate the training signal. In this subsection, we study how dataset composition and dataset-level reweighting affect performance.

\begin{figure}[!htb]
    \centering
    \hspace{\fill}
    \begin{subfigure}[t]{0.49\linewidth}
        \centering
        \includegraphics[width=\linewidth]{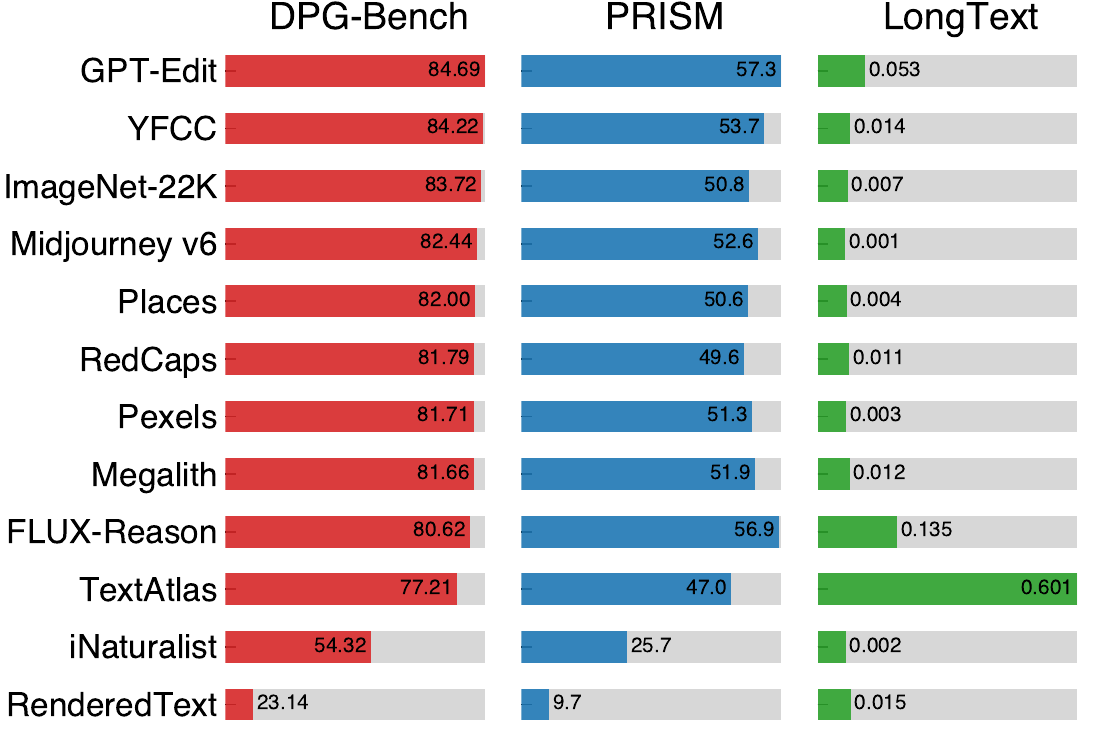}
        \caption{full dataset}
        \label{fig:single_dataset_full}
    \end{subfigure}
    \hspace{\fill}
    \begin{subfigure}[t]{0.49\linewidth}
        \centering
        \includegraphics[width=\linewidth]{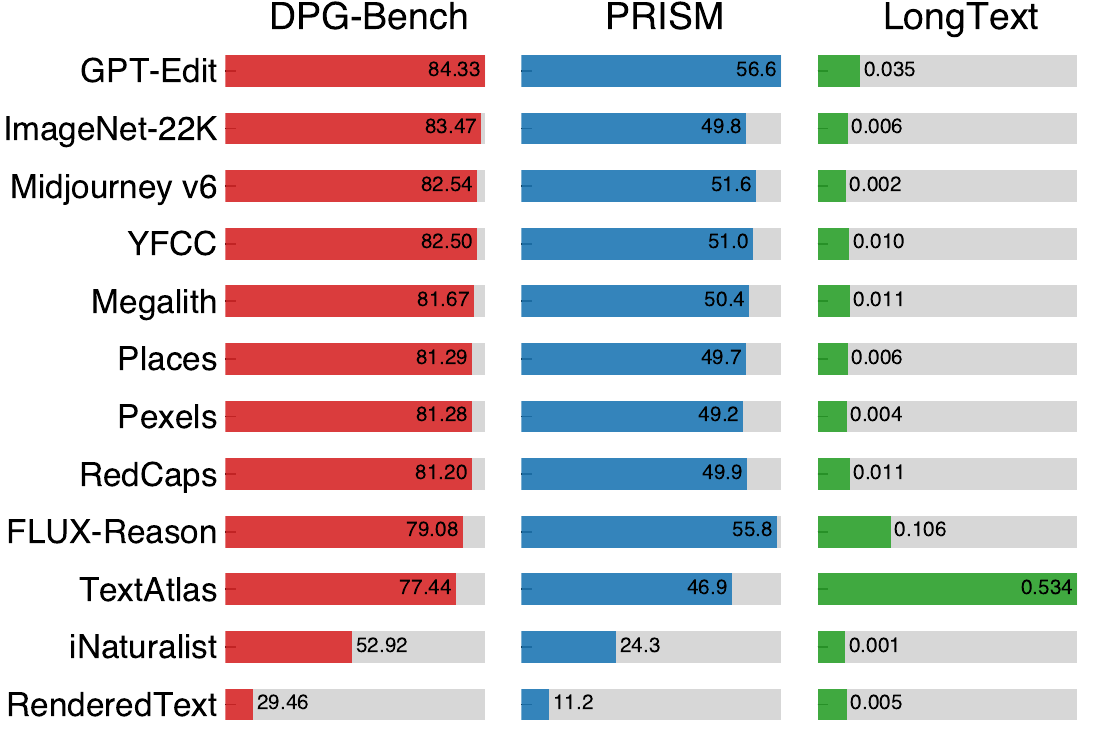}
        \caption{1M subset}
        \label{fig:single_dataset_1m}
    \end{subfigure}
    \hspace{\fill}
    \caption{\textbf{Benchmark performance for single-dataset training}. Among real datasets, ImageNet-22K and YFCC perform best, while iNaturalist performs worst. Text rendering capability relies strongly on specialized text-rich image datasets (\eg, TextAtlas). Across most datasets, performance changes only marginally after subsampling each dataset to 1M.}
    \label{fig:single_dataset}
\end{figure}

\paragraph{Contributions of dataset components}. To understand how each dataset contributes to performance, we train a separate model on each dataset. Evaluation results are shown in \figref{fig:single_dataset_full}. Among real-image datasets, ImageNet-22K and YFCC achieve the best overall performance, while iNaturalist performs substantially worse, likely due to its narrow domain. FLUX-Reason and GPT-Edit perform particularly well on PRISM. LongText scores are low for every dataset except TextAtlas, consistent with the scarcity of text-containing images in real and synthetic datasets. This suggests that text rendering capability relies on specialized text-rich datasets. To control for dataset size, we further train models on random 1M subsets of each dataset. As shown in~\figref{fig:single_dataset_1m}, the relative performance trends remain largely the same as in the full dataset case.

\begin{figure}[!htb]
    \centering
    \includegraphics[width=\linewidth]{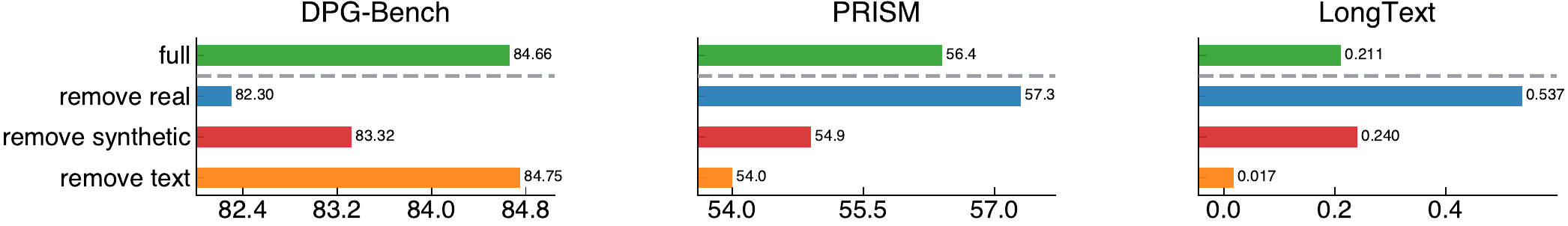}
    \caption{\textbf{Real, synthetic, and text-rendering images are all important for model performance}. Removing any of them leads to inferior performance on at least one benchmark.}
    \label{fig:full_real_synthetic_text}
\end{figure}

Further, we test whether real, synthetic, or text-rendering data can be removed from the baseline (\secrefc{sec:exp_setup}) without harming performance. As shown in \figref{fig:full_real_synthetic_text}, removing real images hurts DPG, whereas removing synthetic images hurts PRISM. Further, LongText performance is directly correlated with the proportion of text rendering data (10.4\% for ``full'', 66.7\% for ``remove real'', 10.9\% for ``remove synthetic'', and 0\% for ``remove text''). These results indicate that the three groups of images provide complementary benefits.

\begin{figure}[!htb]
    \centering
    \includegraphics[width=\linewidth]{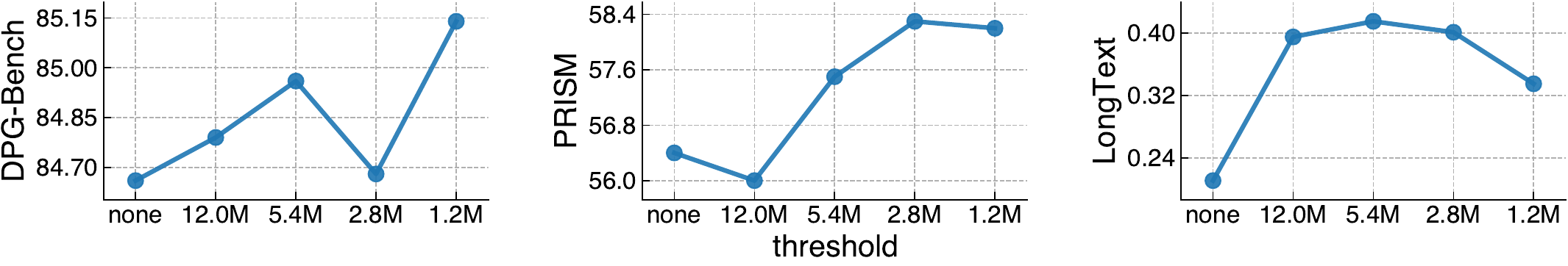}
    \caption{\textbf{Threshold-based weighting}. By default, the sampling weight of a dataset is its number of images. We explore dataset-level balancing by capping the sampling weights for all datasets at four hand-picked thresholds. We find that lower thresholds (\ie, more even weights) generally lead to stronger performance.}
    \label{fig:threshold}
\end{figure}

\paragraph{Equal dataset weighting}. By default (\secrefc{sec:exp_setup}), we naively combine all datasets without explicit dataset-level weighting, so each dataset's effective sampling weight is simply its number of images. Inspired by the data balancing strategy of capping the number of data points from a single source in VLM training~\citep{tong2024cambrian}, we cap each dataset's sampling weight using four hand-picked thresholds. Results in~\figref{fig:threshold} show that a threshold of 1.2M, which gives equal weight to all datasets, achieves strong overall performance. 

\begin{figure}[!htb]
    \centering
    \includegraphics[width=\linewidth]{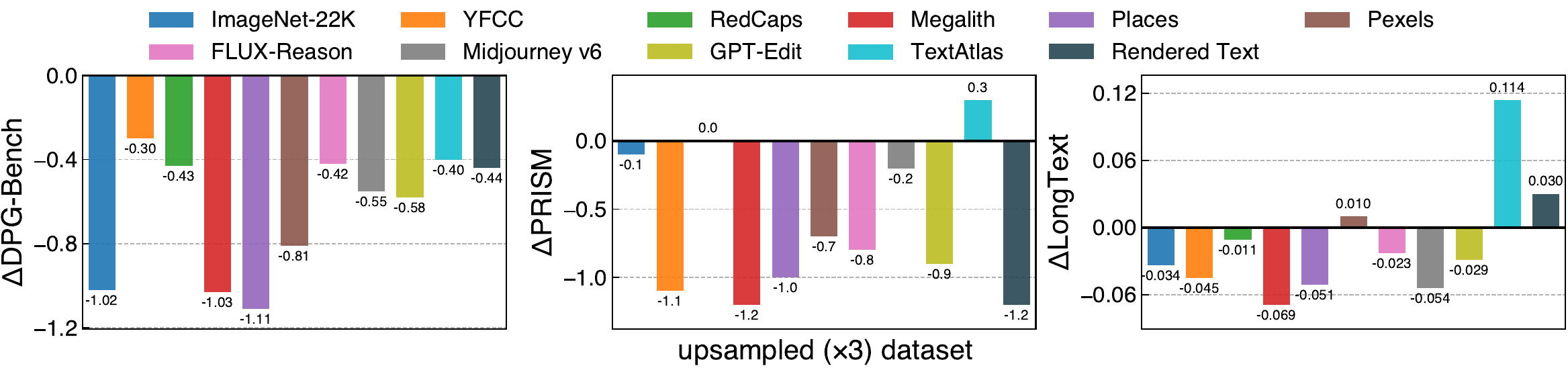}
    \caption{\textbf{Performance change from upweighting a single dataset} by 3$\times$ (5$\times$ in~\figref{fig:upsample_x5}) relative to the baseline of equal weights for all datasets. In all cases, upweighting any dataset does not outperform exact equal weighting.}
    \label{fig:upsample_x3}
\end{figure}

Given the effectiveness of equal dataset weighting (see~\figref{fig:threshold}), we further explore two simple variants. First, we remove low-quality real datasets one at a time while keeping the remaining datasets equally weighted. \tabref{tab:top_real} shows that removing iNaturalist provides a clear gain across all benchmarks, while further removing additional real datasets offers no substantial improvement. Second, after removing iNaturalist, we test whether any single dataset should be emphasized by upweighting one dataset by 3$\times$ or 5$\times$ while keeping the remaining datasets equally weighted. As shown in~\figref{fig:upsample_x3} and~\figref{fig:upsample_x5}, upweighting any single dataset does not surpass the performance of the fully balanced dataset.

\begin{table}[!htb]
\centering
\begin{tabular}{lcccc}
\toprule
datasets & DPG $\uparrow$ & PRISM $\uparrow$ & LongText $\uparrow$\\
\midrule
full & 85.14 & 58.2 & 0.335 \\
remove iNaturalist & \bf 85.56 & \underline{58.7} & 0.384 \\
remove iNaturalist + Megalith & 85.13 & \bf 59.0 & \underline{0.438} \\
remove iNaturalist + Megalith + Places & \underline{85.18} & 57.9 & \bf 0.453 \\
\bottomrule
\end{tabular}
\caption{\textbf{Removing the weakest real-image datasets} one by one under equal weighting, based on single-dataset results (\figref{fig:single_dataset}). Removing iNaturalist improves all benchmark scores, while further removing Megalith and Places does not.}
\label{tab:top_real}
\end{table}

\finding{2}{
Training the model on equal numbers of images from each dataset, counting repetitions (\ie, equal weighting across datasets), is a simple and effective dataset mixing strategy.\\[.2em]
\textbf{Design}: We equally weight all selected datasets at each training stage of \textbf{i1}.}

\begin{figure}[!htb]
    \centering
    \includegraphics[width=\linewidth]{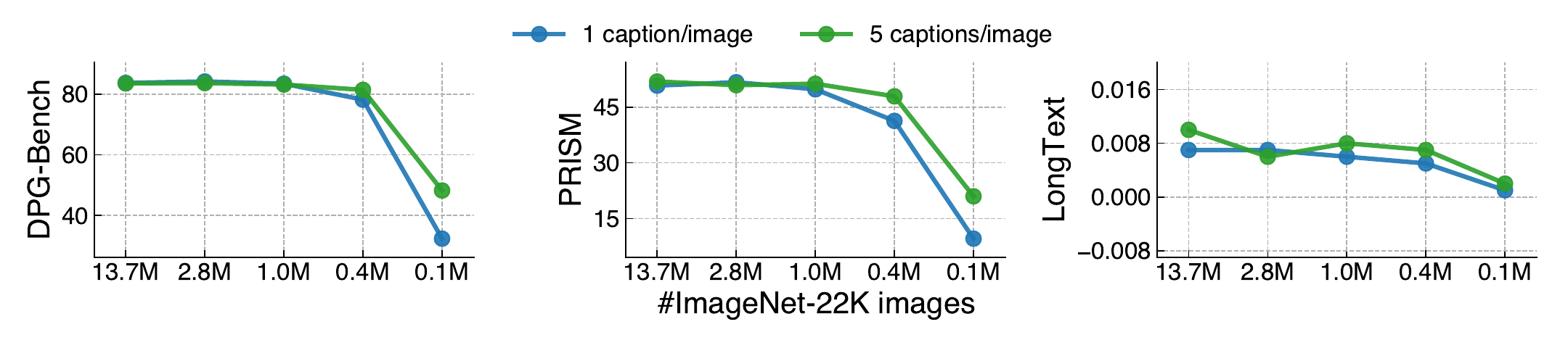}
    \caption{\textbf{Subsampling the ImageNet-22K dataset has little effect on performance}; performance only substantially degrades at 0.1M. Using more captions per image leads to a stronger boost under limited image data.}
    \label{fig:imagenet_num_images}
\end{figure}

\paragraph{Data magnitude}.
\figref{fig:single_dataset} provides preliminary evidence that subsampling datasets often has marginal impact on model performance.
To probe how much performance depends on the unique number of images in the training set, we also train on random subsets of ImageNet-22K. As shown in \figref{fig:imagenet_num_images}, especially when using 5 captions per image, subsampling from 13.7M to 0.4M images only causes marginal degradation. Only when shrinking to 0.1M do we see a substantial drop. Since our 500K-step recipe already repeats the full ImageNet-22K set 18.7 times, these results suggest that using fewer unique images and repeating them more often may not substantially degrade performance for text-to-image diffusion models.

\begin{table}[htbp]
    \centering
      \begin{tabular}{l c c c c}
        \toprule
        subset size for each dataset & unique \#imgs seen & DPG $\uparrow$ & PRISM $\uparrow$ & LongText $\uparrow$\\
        \midrule
        full & 88.1M$^{*}$ & 85.56 & 58.7 & 0.384\\
        1.0M & 11.0M & 85.34 & 57.7 & 0.384 \\
        0.4M & 4.4M & 84.67 & 57.7 & 0.382 \\
        0.1M & 1.1M & 84.71 & 57.4 & 0.349 \\
        \bottomrule
      \end{tabular}
    \caption{\textbf{Subsampling mixtures of datasets}. Starting from the final data recipe (\ie, equal weighting for all datasets excluding iNaturalist), we subsample each dataset to contain a fixed number of images. Even when subsampling 0.4M images from each dataset (\ie, 4.4M instead of 88.1M unique images seen), model performance degrades only slightly. $^{*}$The datasets contain 162.9M images in total, but each dataset is only sampled 23.3M times during training, counting repeatedly sampled images. Since YFCC is not exhausted, 88.1M better estimates the number of unique images seen.}
    \label{tab:mixture_subset}
\end{table}

We further extend the dataset subsampling experiments on a single dataset to mixtures of datasets. Specifically, we begin with a data mixture that assigns equal weight to the 11 datasets, excluding iNaturalist. For each dataset in the mixture, we randomly subsample it to contain exactly 1.0M, 0.4M, or 0.1M images while maintaining equal sampling weights across datasets.
The resulting model performance is shown in \tabref{tab:mixture_subset}. Even when each dataset is reduced to 0.4M images (resulting in 4.4M unique images seen instead of 88.1M), the performance decrease across benchmarks is minimal. This suggests that, with a diverse mix of datasets, repeating training data incurs only marginal performance degradation in text-to-image diffusion training.

\finding{2}{With a diverse mix of image datasets, using fewer unique images and more training epochs causes marginal performance degradation in text-to-image diffusion training.\\[.2em]
\textbf{Design}: We subsample high-resolution images to reduce storage requirements for \textbf{i1} training.}

\section{i1-3B: State-of-the-Art Performance Among Fully Open Models}
\label{sec:i1_recipe}

\begin{figure}[!htb]
    \centering
    \hspace{\fill}
    \begin{subfigure}[t]{0.45\linewidth}
        \centering
        \includegraphics[width=\linewidth]{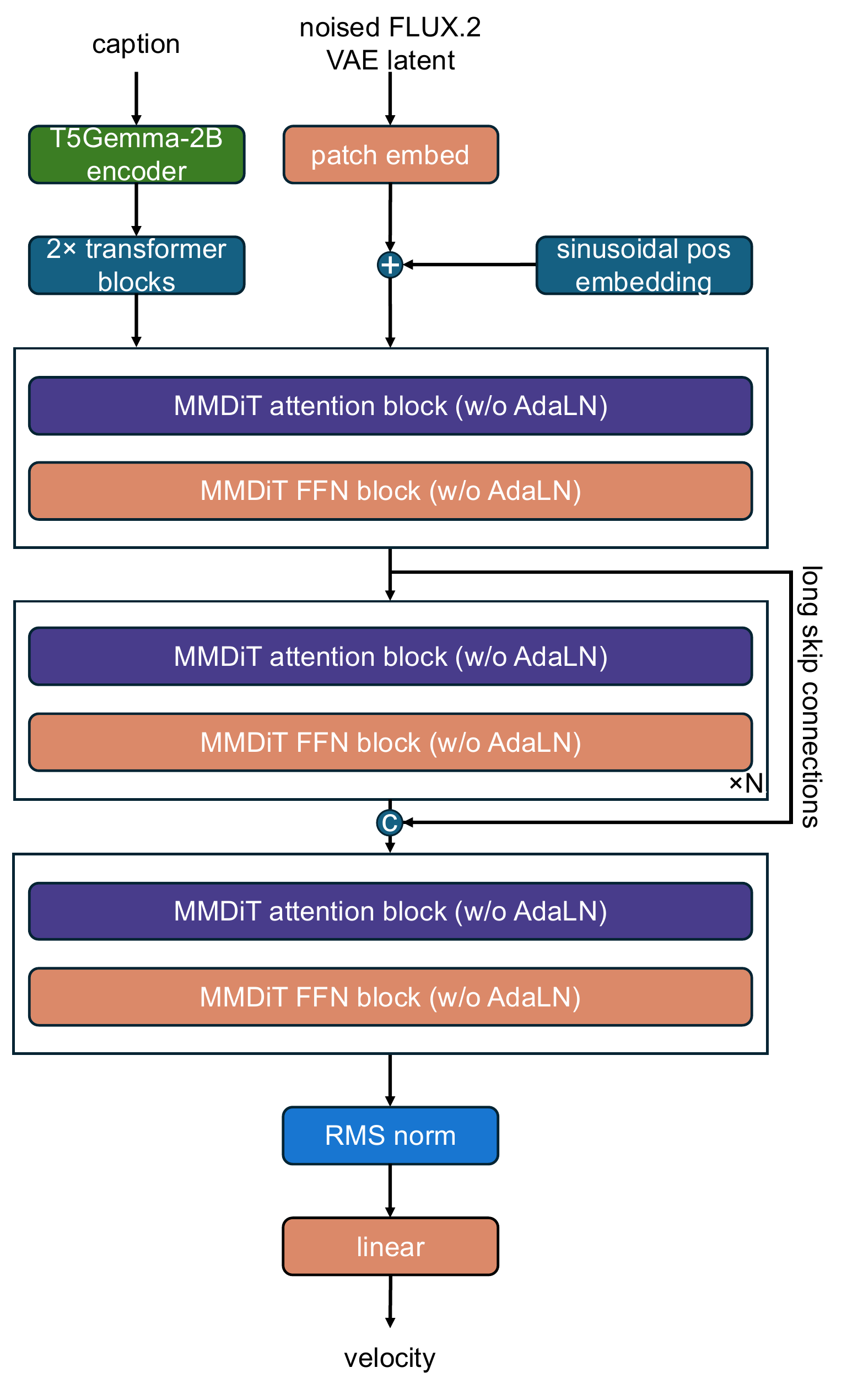}
        \caption{overall architecture}
    \end{subfigure}
    \hspace{\fill}
    \begin{subfigure}[t]{0.45\linewidth}
        \centering
        \includegraphics[width=\linewidth]{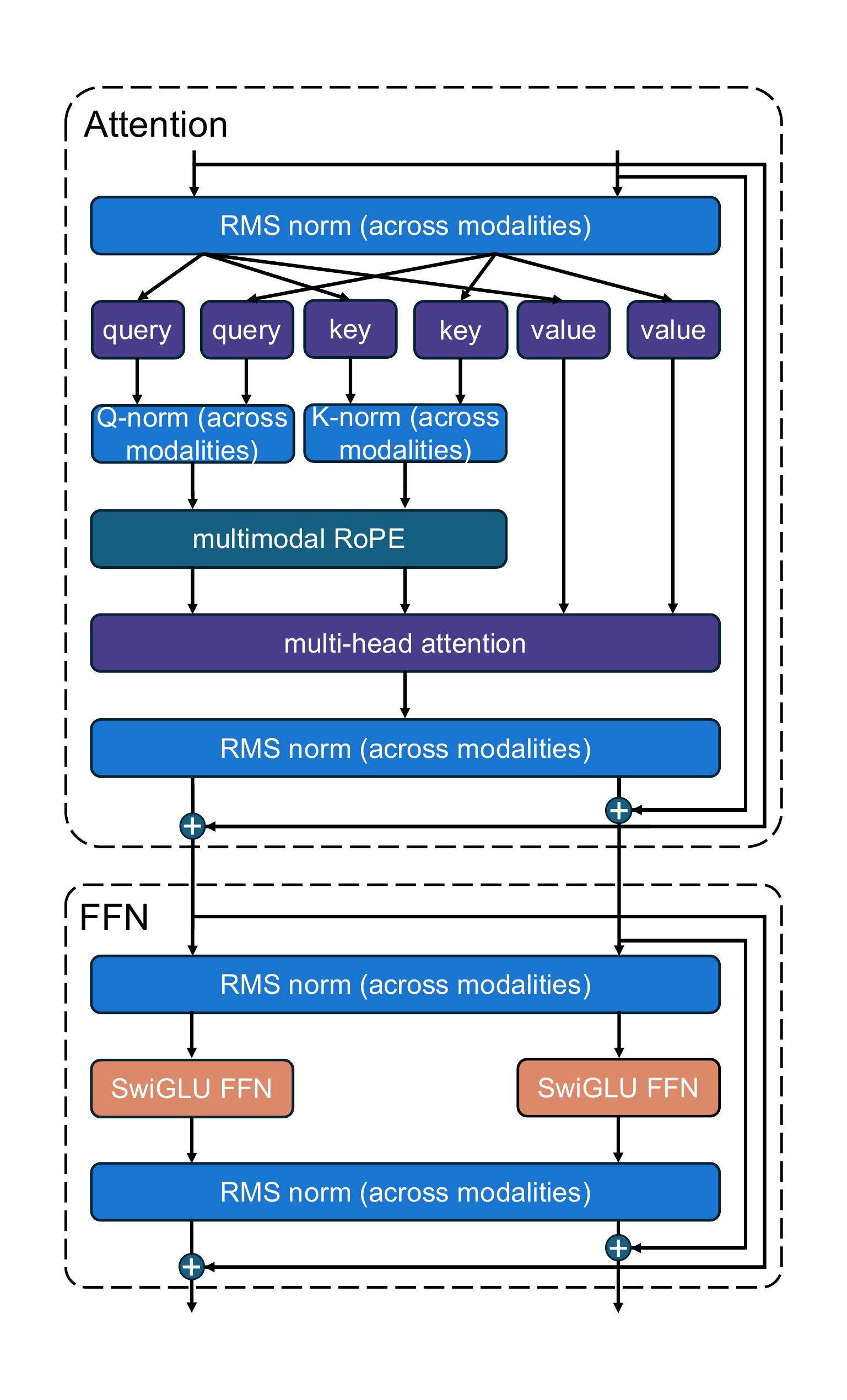}
        \caption{one transformer block}
    \end{subfigure}
    \hspace{\fill}
    \caption{\textbf{The architecture of our final i1 model}. Building on an MMDiT backbone, we use a large text encoder adapter consisting of 2 transformer blocks, remove noise-conditioning (\ie, AdaLN), add long skip connections, combine both sinusoidal and RoPE positional embeddings, and share sandwich normalizations across text and image streams.}
    \label{fig:architecture}
\end{figure}

In the previous sections, we explored the modeling and data designs that can improve text-to-image performance. Building on these insights, we train \textbf{i1}, a model with 3B parameters that performs competitively with leading models across several representative benchmarks.\footnotemark\footnotetext{We are additionally training a 1B model and will release it soon.}~In this section, we describe the final pre-training, high-resolution training, and inference setups and experiments, and present the evaluation results.

\subsection{Low-Resolution Pre-training}
\label{sec:pretraining}

\paragraph{Model}. The architecture of \textbf{i1} is illustrated in~\figref{fig:architecture}. It uses a dual-stream MMDiT backbone with long skip connections, the FLUX.2 VAE, and T5Gemma-2B as the text encoder, along with a large adapter composed of two transformer blocks. \textbf{i1} removes all AdaLN parameters and thus does not use noise conditioning. Additionally, we use both sinusoidal and RoPE positional embeddings, and share sandwich normalizations across text and image streams (see~\appendixref{appendix:pos_emb_and_norm} for corresponding controlled experiments).

\paragraph{Data}. We use the best data mixing recipe identified in \secrefc{sec:data_mixing}, where we assign equal weights to 6 real image datasets, 3 synthetic datasets, and 2 text-rendering datasets. We use Qwen3-VL-30B-A3B to generate multiple long synthetic captions for each image. Due to resource constraints, we generate five synthetic captions per image for ImageNet-22K, Pexels, RenderedText, GPT-Edit, RedCaps, FLUX-Reason, TextAtlas, and Midjourney v6, two per image for YFCC, and one per image for Places and Megalith.

\begin{figure}[!htb]
    \centering
    \includegraphics[width=\linewidth]{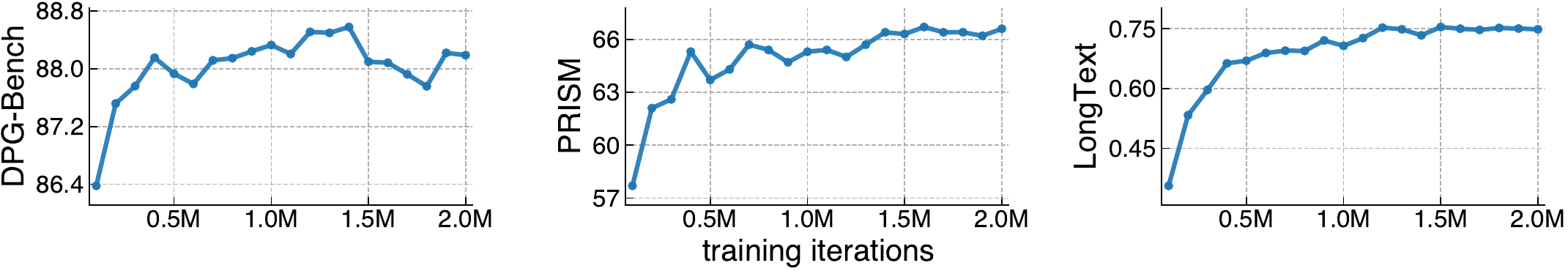}
    \caption{\textbf{Benchmark performance of \textbf{i1} during 256-resolution pre-training}. Performance stabilizes around 500K iterations and largely converges by 2M iterations. Evaluation follows the setup used in controlled experiments (\secrefc{sec:exp_setup}).}
    \label{fig:256_performance_curve}
\end{figure}

\begin{figure}[!htb]
\centering
\setlength{\tabcolsep}{1pt}
\begin{tabular}{@{}>{\footnotesize\raggedright\arraybackslash}m{.147\linewidth}cccc@{}}
{\footnotesize \textbf{Prompt}: Argentinian soccer star Lionel Messi in the heat of the 2022 FIFA World Cup Final against France. He is... about to strike the ball with his left foot... (240 words)} &
\includegraphics[width=.207\linewidth,valign=c]{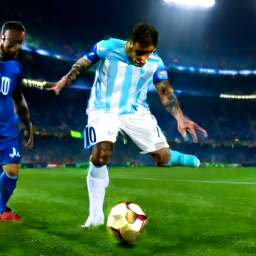} &
\includegraphics[width=.207\linewidth,valign=c]{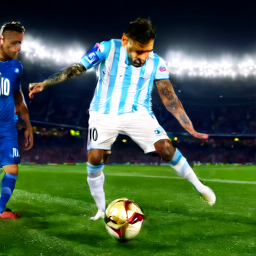} &
\includegraphics[width=.207\linewidth,valign=c]{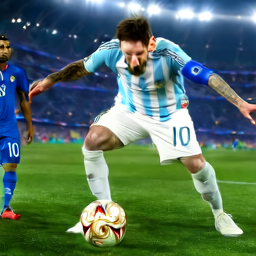} &
\includegraphics[width=.207\linewidth,valign=c]{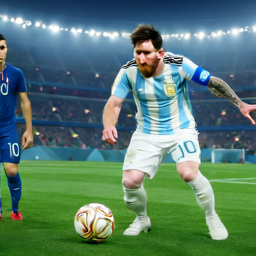}\\

{\footnotesize \textbf{Prompt}: An appealing poster... announcing a folk music concert event... At the top-center, the inviting phrase "Let Acoustic Melodies Inspire Your Soul"... (109 words)} &
\includegraphics[width=.207\linewidth,valign=c]{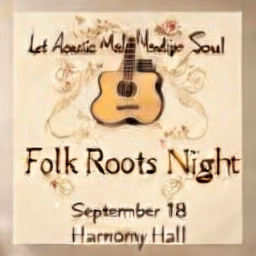} &
\includegraphics[width=.207\linewidth,valign=c]{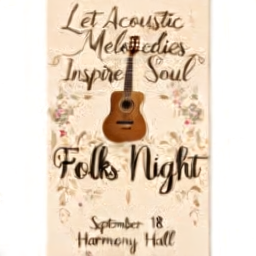} &
\includegraphics[width=.207\linewidth,valign=c]{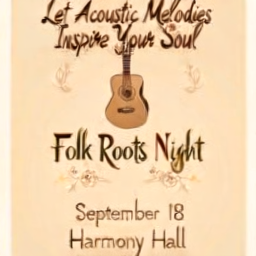} &
\includegraphics[width=.207\linewidth,valign=c]{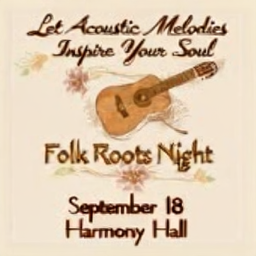} \\[2pt]

& \footnotesize\parbox{.207\linewidth}{\centering 100K iterations} &
\footnotesize\parbox{.207\linewidth}{\centering 200K iterations} &
\footnotesize\parbox{.207\linewidth}{\centering 500K iterations} &
\footnotesize\parbox{.207\linewidth}{\centering 2M iterations}
\end{tabular}

\caption{\textbf{Example generated images at different iterations of 256-resolution training}. Overall image quality and text-rendering capability improve throughout the training run, mirroring the benchmark score improvements.}
\label{fig:256_qualitative_progression}
\end{figure}

\paragraph{Training.} We extend the number of training iterations in the default recipe (\secrefc{sec:exp_setup}) to 2M steps while keeping all other hyperparameters unchanged. We train \textbf{i1} at 256-resolution until performance plateaus around 2M steps, as shown in~\figref{fig:256_performance_curve}. We additionally show example generated images in~\figref{fig:256_qualitative_progression} and observe that the benchmark improvements are accompanied by improved image quality. Details of the training setup and compute resources are in~\appendixref{appendix:config}.

\subsection{High-Resolution Training}
\label{sec:high_res_training}

\paragraph{Data and modeling}. To construct the 512- and 1024-resolution training sets, we retain only images whose shorter edge is at least 512 or 1024 pixels, respectively. We remove any dataset entirely if the filtered set contains fewer than 0.3M images (see the resolution statistics for each dataset in~\appendixref{appendix:img_res_stat}). Based on our findings in~\secrefc{sec:data_mixing}, we further subsample each dataset with more than 1M images to 1M images and assign equal sampling weight to every dataset. At 1024-resolution, we discard RenderedText due to its low quality (see \figref{fig:single_dataset}). Following~\citet{esser2024scaling}, we perform positional index interpolation and timestep schedule shifting during 512- and 1024-resolution training (details in~\appendixref{appendix:config}).

\begin{figure}[!htb]
    \centering
    \includegraphics[width=\linewidth]{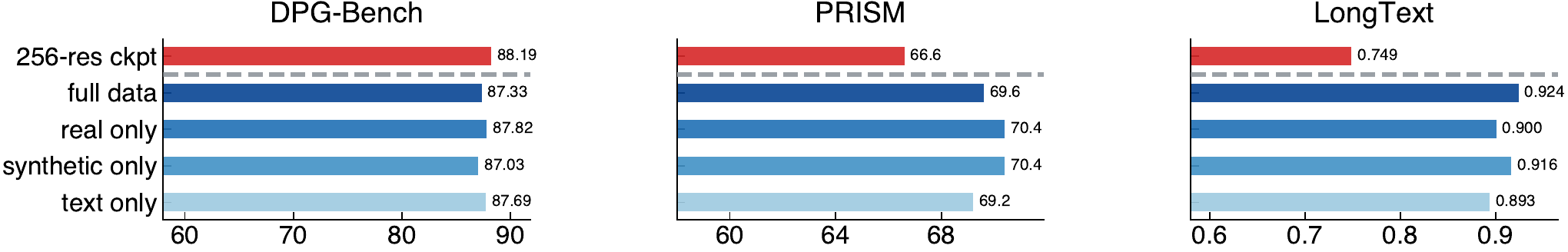}
    \caption{\textbf{Benchmark performance of \textbf{i1} at 512-resolution} with different training sets. PRISM and LongText improve substantially with 512-resolution training, even when text rendering data is not used.}
    \label{fig:512_performance_bar}
\end{figure}

\begin{figure}[!htb]
    \centering
    \hspace{\fill}
    \begin{subfigure}[t]{0.49\linewidth}
      \begin{tabular}{@{}c@{\hspace{0pt}}c@{}}
        \includegraphics[width=.5\linewidth]{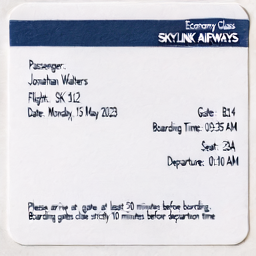} &
        \includegraphics[width=.5\linewidth]{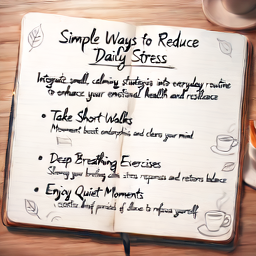}
      \end{tabular}
      \caption{256-resolution model}
    \end{subfigure}
    \hspace{\fill}
    \begin{subfigure}[t]{0.49\linewidth}
      \begin{tabular}{@{}c@{\hspace{0pt}}c@{}}
        \includegraphics[width=.5\linewidth]{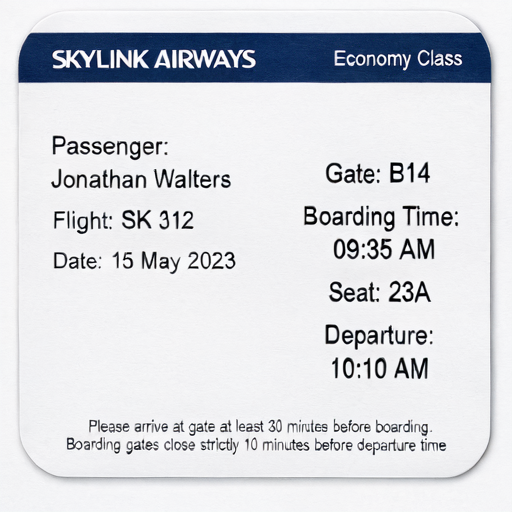} &
        \includegraphics[width=.5\linewidth]{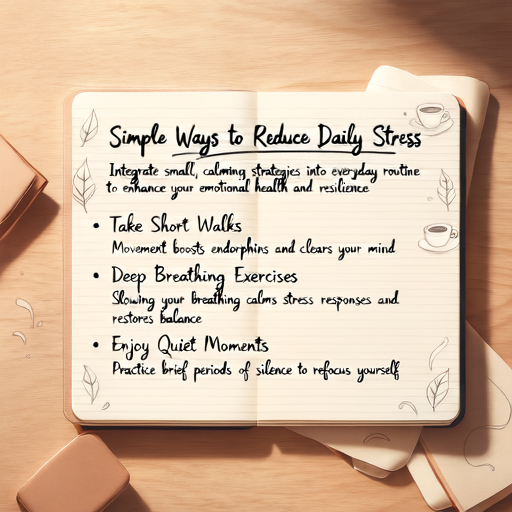}
      \end{tabular}
      \caption{512-resolution model}
    \end{subfigure}
    \hspace{\fill}
    \caption{\textbf{Text rendering improves substantially after 512-resolution training}, as demonstrated by example images generated from our 256-resolution and 512-resolution checkpoints using the same input prompts from LongText-Bench.}
    \label{fig:256_512_text_render}
\end{figure}

\paragraph{Results}. We train the model for 0.5M steps at 512-resolution and 0.3M steps at 1024-resolution. The benchmark performance trends during training are in~\appendixref{appendix:config}, and the final 1024-resolution checkpoint is evaluated in~\secrefc{sec:evaluation}. As shown in~\figref{fig:512_performance_bar}, 512-resolution training substantially improves the LongText score (0.75 $\rightarrow$ 0.92). We further illustrate the improvements in text rendering with qualitative examples in~\figref{fig:256_512_text_render}.

We also study how different dataset components contribute to 512-resolution training. Starting from the 256-resolution checkpoint, we train separate models using only real image datasets, only synthetic image datasets, or only text-rendering datasets at 512-resolution. As shown in~\figref{fig:512_performance_bar}, training on either real or synthetic image datasets yields LongText improvements comparable to training on the full dataset, despite both subsets containing limited text-rich images. This suggests that strong high-resolution generation capability does not require high-resolution training data to match the full breadth of the low-resolution pre-training data.

\finding{2}{Training a model to achieve strong high-resolution generation does not require high-resolution training data to match the full breadth of the low-resolution pre-training data.\\[.2em]
\textbf{Design}: We do not further expand the resolution-filtered high-resolution datasets for \textbf{i1} training.}

\subsection{Inference and Evaluation}
\label{sec:evaluation}

\paragraph{Inference setup}.
During inference, we use a CFG scale of 12 and apply the Rescale CFG technique~\citep{lin2024common} with a rescale strength of 1. Unlike previous methods~\citep{wang2024emu3, deng2025emerging, pan2025transfer} that apply prompt rewriting to particular benchmarks, we use a single meta-prompt (details in~\appendixref{appendix:unified_meta_prompt}) for rewriting all input prompts to match training prompt lengths, as motivated in \secrefc{sec:caption_and_rewrite}.

\paragraph{Benchmarks}.
We evaluate our model on five representative benchmarks commonly used in the technical reports of recent image generation models~\citep{cai2025z, qin2025lumina, cai2025hidream, cui2025emu3}: GenEval~\citep{ghosh2023geneval}, DPG-Bench~\citep{hu2024ella}, PRISM-Bench~\citep{fang2026flux}, CVTG-2K~\citep{du2025textcrafter}, and LongText-Bench~\citep{geng2025x}. GenEval focuses on object-centric image generation and evaluates a fixed set of object attributes and relationships. DPG-Bench and PRISM-Bench provide fine-grained evaluation of general prompt-following capabilities, with PRISM-Bench additionally assessing image aesthetics. CVTG-2K and LongText-Bench evaluate a model's ability to generate images containing detectable text that matches the description in the input prompt.

We note that prior work has suggested that GenEval may be misaligned with human judgment~\citep{kamath2025geneval} and poorly correlated with human-perceived model capability~\citep{cao2025hunyuanimage}. Additionally, it is a common practice in current models~\citep{chen2025blip3, ma2026deco, wang2026pixnerd} to fine-tune on BLIP3o-60K~\citep{chen2025blip3}, which can inflate GenEval scores, as BLIP3o-60K fine-tuning was found to significantly improve GenEval scores but not other benchmarks~\citep{wu2025openuni}. Therefore, we report GenEval results only for completeness and note that they may not accurately reflect model capability.

\begin{table}[!htb]
\centering
\renewcommand{\arraystretch}{1.1}
\setlength{\tabcolsep}{5pt} 
\makebox[\textwidth][c]{
\resizebox{1.00\textwidth}{!}{
\begin{tabular}{l c c cccc}
\toprule
\textbf{model} & \textbf{\#params} & \textbf{GenEval} & \textbf{DPG-Bench} & \textbf{PRISM} & \textbf{CVTG-2K} & \textbf{LongText-Bench} \\
\midrule
\multicolumn{7}{l}{\textit{\textbf{API call only}}} \\
GPT Image 1 [High]~\citep{gptimage1} & - & \textbf{0.84}* & 85.15* & - & \textbf{0.8569}* & \textbf{0.956}* \\
Seedream 3.0~\citep{gao2025seedream} & - & \textbf{0.84}* & \textbf{88.27}* & - & 0.5924* & 0.896* \\
\arrayrulecolor{black!30}\midrule
\multicolumn{7}{l}{\textit{\textbf{Open weights only}}} \\
FLUX.1 [Dev]~\citep{labs2025flux} & 12B & 0.66* & 83.84* & 65.1 & 0.4965* & 0.607* \\
SD3 Medium~\citep{esser2024scaling} & 2B & 0.62* & 84.08* & 61.9 & 0.4037 & 0.322 \\
Janus-Pro-7B~\citep{chen2025janus} & 7B & 0.80* & 84.19* & 60.0 & 0.0667 & 0.019* \\
BAGEL~\citep{deng2025emerging} & 14B & \textbf{0.88}* & 85.44 & 61.8 & 0.3642 & 0.373* \\
HiDream-I1-Full~\citep{cai2025hidream} & 17B & 0.83* & 85.89* & 66.1 & 0.7738 & 0.543* \\
Lumina-Image 2.0~\citep{qin2025lumina} & 3B & 0.73* & 87.20* & 63.5 & 0.1577 & 0.088 \\
Z-Image~\citep{cai2025z} & 6B & 0.84* & 88.14* & \textbf{74.2} & \textbf{0.8671}* & 0.935* \\
Qwen-Image~\citep{wu2025qwen} & 20B & 0.87* & \textbf{88.32}* & 73.9 & 0.8288* & \textbf{0.943}* \\
\arrayrulecolor{black!30}\midrule
\multicolumn{7}{l}{\textit{\textbf{Open weights + data + training code}}} \\
BLIP3o-4B~\citep{chen2025blip3} & 4B & 0.77 & 79.73 & 53.2 & 0.0353 & 0.023 \\
PixNerd~\citep{wang2026pixnerd} & 1B & 0.73* & 80.9* & 53.3 & 0.0006 & 0.020 \\
DeCo~\citep{ma2026deco} & 1B & 0.86* & 81.4* & 53.1 & 0.0014 & 0.003 \\
BLIP3o-N-S~\citep{chen2025blip3o} & 3B & 0.87 & 81.98 & 56.8 & 0.2493 & 0.110 \\
BLIP3o-N-G-G~\citep{chen2025blip3o} & 3B & \textbf{0.90} & 81.93 & 57.5 & 0.2442 & 0.114 \\
BLIP3o-N-G-T~\citep{chen2025blip3o} & 3B & 0.86 & 79.77 & 56.8 & 0.3330 & 0.153 \\
\textbf{i1 (Ours)} & 3B & 0.84 & \bf 86.73 & \bf 70.1 & \bf 0.8531 & \bf 0.922 \\
\arrayrulecolor{black}\bottomrule
\end{tabular}
}
}
\caption{\textbf{Performance on representative text-to-image benchmarks}.
Results marked with an * are sourced from previous papers~\citep{cai2025z,deng2025emerging, ma2026deco, wang2026pixnerd}.
We reproduce all PRISM results with Qwen2.5-VL-72B because the official results~\citep{fang2026flux} consistently differ from our reproduction. We abbreviate BLIP3o-NEXT's SFT, GRPO-GenEval, and GRPO-Text models as ``BLIP3o-N-S'', ``BLIP3o-N-G-G'', and ``BLIP3o-N-G-T''.}
\label{tab:comparison_perf}
\end{table}

\paragraph{Results}.
We compare the \textbf{i1} model with leading image generation systems in \tabref{tab:comparison_perf}. \textbf{i1} achieves state-of-the-art performance among fully open models on all five benchmarks except GenEval. It also outperforms several leading weight-only models, including Lumina-Image 2.0, HiDream-I1, and FLUX.1 [Dev]. \textbf{i1}'s strong performance reflects the combined effect of the modeling and data choices identified throughout our study.

\section{Discussion and Conclusion}

\paragraph{Fully open recipes support cumulative research} in text-to-image modeling.
A challenge in current text-to-image research is that strong models are often released as opaque endpoints rather than as inspectable scientific artifacts. As a result, progress can be difficult to attribute across various (potentially undisclosed) design factors. Our study advocates for fully open recipes that seek to understand which design choices reliably matter. By releasing the model, code, data recipe, and ablations behind \textbf{i1}, we aim to provide not only a strong baseline, but also a reference point for more cumulative and reproducible research.

\paragraph{Strong performance does not require sophisticated designs}.
The strong performance of recent text-to-image models can create the impression that frontier capability requires increasingly specialized architectures, proprietary data, or heavily engineered recipes. Our study provides a counterpoint: strong performance can be achieved with moderately scaled (\eg, 4.4M, see~\secrefc{sec:data_mixing}) and publicly available datasets and a careful exploration of the current modeling design space. We believe that this is encouraging for open research, as competitive text-to-image models need not begin from inaccessible data or undisclosed training procedures.

\paragraph{Limitations and future work}. This work has several limitations. First, our evaluation relies primarily on automated benchmarks, which emphasize prompt following, rather than human preference. Thus, although \textbf{i1} approaches leading weight-only models (\eg, Qwen-Image) on these benchmarks, its generated images remain noticeably inferior in overall visual quality (we present failure cases in~\appendixref{appendix:failure_case}). Second, due to resource constraints, all experiments are conducted with models of roughly 3B parameters or smaller. Further experiments are needed to determine whether our findings continue to hold at substantially larger scales. Third, our exploration only covers a subset of the text-to-image diffusion model design space: designs such as multi-aspect ratio training\footnotemark, data filtering~\citep{startsev2026alchemist}, deep fusion of decoder-only LLMs and diffusion transformer for text encoding~\citep{liu2024playground, shi2026lmfusion}, and reinforcement learning~\citep{wallace2024diffusion, liu2026flow} are omitted.
Future work could extend our recipe to larger models and further explore the design space while preserving the simplicity and openness of the overall pipeline.

\footnotetext{We are working on a multi-aspect ratio model and will release it soon.}

\section*{Acknowledgements}

We gratefully thank the Google TPU Research Cloud (TRC) program for providing the primary computing resources for this project. Additional support was provided by the Princeton Research Computing resources at Princeton University, which are managed by a consortium of groups led by the Princeton Institute for Computational Science and Engineering (PICSciE) and Research Computing.
We would like to thank Liang-Chieh Chen, Ishan Misra, Kaiming He, Yida Yin, Haozhe Chen, Wenhao Chai, Linrong Cai, Linzhan Mou, and Xingyu Fu for valuable discussions and feedback. We also thank Yufeng Xu, Shengbang Tong, Yiyang Lu, and Hanhong Zhao for helpful discussions on TPU. We are grateful to Cihang Xie's research group for sharing their JAX DiT codebase, which served as the launching point for our research.

\bibliographystyle{plainnat}
\bibliography{main}

@String(PAMI  = {IEEE TPAMI})

@String(CVPR  = {CVPR})

@String(ICCV  = {ICCV})

@String(NIPS = {NeurIPS})

@String(NeurIPS = {NeurIPS})

@String(ICML  = {ICML})

@String(ICLR  = {ICLR})

@String(WACV  =	{WACV})

@String(JMLR  = {JMLR})

@String(TMLR  = {TMLR})

@String(MICCAI= {MICCAI})

@inproceedings{bao2023all,
  title={All are worth words: A vit backbone for diffusion models},
  author={Bao, Fan and Nie, Shen and Xue, Kaiwen and Cao, Yue and Li, Chongxuan and Su, Hang and Zhu, Jun},
  booktitle=CVPR,
  year={2023}
}

@article{betker2023improving,
  title={Improving Image Generation with Better Captions},
  author={Betker, James and Goh, Gabriel and Jing, Li and Brooks, Tim and Wang, Jianfeng and Li, Linjie and Ouyang, Long and Zhuang, Juntang and Lee, Joyce and Guo, Yufei and others},
  year={2023},
  url={https://cdn.openai.com/papers/dall-e-3.pdf}
}

@inproceedings{ronneberger2015u,
  title={U-net: Convolutional networks for biomedical image segmentation},
  author={Ronneberger, Olaf and Fischer, Philipp and Brox, Thomas},
  booktitle=MICCAI,
  year={2015},
}

@inproceedings{peebles2023scalable,
  title={Scalable diffusion models with transformers},
  author={Peebles, William and Xie, Saining},
  booktitle=ICCV,
  year={2023}
}

@article{cai2025z,
  title={Z-image: An efficient image generation foundation model with single-stream diffusion transformer},
  author={Cai, Huanqia and Cao, Sihan and Du, Ruoyi and Gao, Peng and Hoi, Steven and Hou, Zhaohui and Huang, Shijie and Jiang, Dengyang and Jin, Xin and Li, Liangchen and others},
  journal={arXiv preprint arXiv:2511.22699},
  year={2025}
}

@inproceedings{ma2026deco,
  title={Deco: Frequency-decoupled pixel diffusion for end-to-end image generation},
  author={Ma, Zehong and Wei, Longhui and Wang, Shuai and Zhang, Shiliang and Tian, Qi},
  booktitle=CVPR,
  year={2026}
}

@article{chen2025blip3,
  title={Blip3-o: A family of fully open unified multimodal models-architecture, training and dataset},
  author={Chen, Jiuhai and Xu, Zhiyang and Pan, Xichen and Hu, Yushi and Qin, Can and Goldstein, Tom and Huang, Lifu and Zhou, Tianyi and Xie, Saining and Savarese, Silvio and others},
  journal={arXiv preprint arXiv:2505.09568},
  year={2025}
}

@article{chen2025blip3o,
  title={Blip3o-next: Next frontier of native image generation},
  author={Chen, Jiuhai and Xue, Le and Xu, Zhiyang and Pan, Xichen and Yang, Shusheng and Qin, Can and Yan, An and Zhou, Honglu and Chen, Zeyuan and Huang, Lifu and others},
  journal={arXiv preprint arXiv:2510.15857},
  year={2025}
}

@article{deng2025emerging,
  title={Emerging properties in unified multimodal pretraining},
  author={Deng, Chaorui and Zhu, Deyao and Li, Kunchang and Gou, Chenhui and Li, Feng and Wang, Zeyu and Zhong, Shu and Yu, Weihao and Nie, Xiaonan and Song, Ziang and others},
  journal={arXiv preprint arXiv:2505.14683},
  year={2025}
}

@article{cai2025hidream,
  title={Hidream-i1: A high-efficient image generative foundation model with sparse diffusion transformer},
  author={Cai, Qi and Chen, Jingwen and Chen, Yang and Li, Yehao and Long, Fuchen and Pan, Yingwei and Qiu, Zhaofan and Zhang, Yiheng and Gao, Fengbin and Xu, Peihan and others},
  journal={arXiv preprint arXiv:2505.22705},
  year={2025}
}

@article{labs2025flux,
  title={FLUX.1 Kontext: Flow Matching for In-Context Image Generation and Editing in Latent Space},
  author={Labs, Black Forest and Batifol, Stephen and Blattmann, Andreas and Boesel, Frederic and Consul, Saksham and Diagne, Cyril and Dockhorn, Tim and English, Jack and English, Zion and Esser, Patrick and others},
  journal={arXiv preprint arXiv:2506.15742},
  year={2025}
}

@inproceedings{wang2026pixnerd,
  title={Pixnerd: Pixel neural field diffusion},
  author={Wang, Shuai and Gao, Ziteng and Zhu, Chenhui and Huang, Weilin and Wang, Limin},
  booktitle=ICLR,
  year={2026}
}

@article{gao2025seedream,
  title={Seedream 3.0 technical report},
  author={Gao, Yu and Gong, Lixue and Guo, Qiushan and Hou, Xiaoxia and Lai, Zhichao and Li, Fanshi and Li, Liang and Lian, Xiaochen and Liao, Chao and Liu, Liyang and others},
  journal={arXiv preprint arXiv:2504.11346},
  year={2025}
}

@article{chen2025janus,
  title={Janus-pro: Unified multimodal understanding and generation with data and model scaling},
  author={Chen, Xiaokang and Wu, Zhiyu and Liu, Xingchao and Pan, Zizheng and Liu, Wen and Xie, Zhenda and Yu, Xingkai and Ruan, Chong},
  journal={arXiv preprint arXiv:2501.17811},
  year={2025}
}

@article{wu2025qwen,
  title={Qwen-image technical report},
  author={Wu, Chenfei and Li, Jiahao and Zhou, Jingren and Lin, Junyang and Gao, Kaiyuan and Yan, Kun and Yin, Sheng-ming and Bai, Shuai and Xu, Xiao and Chen, Yilei and others},
  journal={arXiv preprint arXiv:2508.02324},
  year={2025}
}

@inproceedings{qin2025lumina,
  title={Lumina-image 2.0: A unified and efficient image generative framework},
  author={Qin, Qi and Zhuo, Le and Xin, Yi and Du, Ruoyi and Li, Zhen and Fu, Bin and Lu, Yiting and Li, Xinyue and Liu, Dongyang and Zhu, Xiangyang and others},
  booktitle=ICCV,
  year={2025}
}

@inproceedings{esser2024scaling,
  title={Scaling rectified flow transformers for high-resolution image synthesis},
  author={Esser, Patrick and Kulal, Sumith and Blattmann, Andreas and Entezari, Rahim and M{\"u}ller, Jonas and Saini, Harry and Levi, Yam and Lorenz, Dominik and Sauer, Axel and Boesel, Frederic and others},
  booktitle=ICML,
  year={2024}
}

@article{qwen2.5-vl,
  title={Qwen2.5-VL Technical Report},
  author={Shuai Bai and Keqin Chen and Xuejing Liu and Jialin Wang and Wenbin Ge and Sibo Song and Kai Dang and Peng Wang and Shijie Wang and Jun Tang and Humen Zhong and Yuanzhi Zhu and Mingkun Yang and Zhaohai Li and Jianqiang Wan and Pengfei Wang and Wei Ding and Zheren Fu and Yiheng Xu and Jiabo Ye and Xi Zhang and Tianbao Xie and Zesen Cheng and Hang Zhang and Zhibo Yang and Haiyang Xu and Junyang Lin},
  journal={arXiv preprint arXiv:2502.13923},
  year={2025}
}

@misc{gptimage1,
  author       = {OpenAI},
  title        = {Introducing 4o Image Generation},
  howpublished = {\url{https://openai.com/index/introducing-4o-image-generation/}},
  year         = {2025}
}

@article{podell2023sdxl,
  title={Sdxl: Improving latent diffusion models for high-resolution image synthesis},
  author={Podell, Dustin and English, Zion and Lacey, Kyle and Blattmann, Andreas and Dockhorn, Tim and M{\"u}ller, Jonas and Penna, Joe and Rombach, Robin},
  journal={arXiv preprint arXiv:2307.01952},
  year={2023}
}

@article{kamath2025geneval,
  title={GenEval 2: Addressing Benchmark Drift in Text-to-Image Evaluation},
  author={Kamath, Amita and Chang, Kai-Wei and Krishna, Ranjay and Zettlemoyer, Luke and Hu, Yushi and Ghazvininejad, Marjan},
  journal={arXiv preprint arXiv:2512.16853},
  year={2025}
}

@article{wu2025openuni,
  title={Openuni: A simple baseline for unified multimodal understanding and generation},
  author={Wu, Size and Wu, Zhonghua and Gong, Zerui and Tao, Qingyi and Jin, Sheng and Li, Qinyue and Li, Wei and Loy, Chen Change},
  journal={arXiv preprint arXiv:2505.23661},
  year={2025}
}

@inproceedings{lin2024common,
  title={Common diffusion noise schedules and sample steps are flawed},
  author={Lin, Shanchuan and Liu, Bingchen and Li, Jiashi and Yang, Xiao},
  booktitle=WACV,
  year={2024}
}

@article{cui2025emu3,
  title={Emu3.5: Native multimodal models are world learners},
  author={Cui, Yufeng and Chen, Honghao and Deng, Haoge and Huang, Xu and Li, Xinghang and Liu, Jirong and Liu, Yang and Luo, Zhuoyan and Wang, Jinsheng and Wang, Wenxuan and others},
  journal={arXiv preprint arXiv:2510.26583},
  year={2025}
}

@inproceedings{fang2026flux,
  title={Flux-reason-6m \& prism-bench: A million-scale text-to-image reasoning dataset and comprehensive benchmark},
  author={Fang, Rongyao and Yu, Aldrich and Duan, Chengqi and Huang, Linjiang and Bai, Shuai and Cai, Yuxuan and Wang, Kun and Liu, Si and Liu, Xihui and Li, Hongsheng},
  booktitle=ICLR,
  year={2026}
}

@article{cao2025hunyuanimage,
  title={Hunyuanimage 3.0 technical report},
  author={Cao, Siyu and Chen, Hangting and Chen, Peng and Cheng, Yiji and Cui, Yutao and Deng, Xinchi and Dong, Ying and Gong, Kipper and Gu, Tianpeng and Gu, Xiusen and others},
  journal={arXiv preprint arXiv:2509.23951},
  year={2025}
}

@article{tong2026scaling,
  title={Scaling Text-to-Image Diffusion Transformers with Representation Autoencoders},
  author={Tong, Shengbang and Zheng, Boyang and Wang, Ziteng and Tang, Bingda and Ma, Nanye and Brown, Ellis and Yang, Jihan and Fergus, Rob and LeCun, Yann and Xie, Saining},
  journal={arXiv preprint arXiv:2601.16208},
  year={2026}
}

@article{yu2022scaling,
  title={Scaling autoregressive models for content-rich text-to-image generation},
  author={Yu, Jiahui and Xu, Yuanzhong and Koh, Jing Yu and Luong, Thang and Baid, Gunjan and Wang, Zirui and Vasudevan, Vijay and Ku, Alexander and Yang, Yinfei and Ayan, Burcu Karagol and others},
  journal=TMLR,
  year={2022}
}

@inproceedings{chen2024pixart,
  title={Pixart-$\alpha$: Fast training of diffusion transformer for photorealistic text-to-image synthesis},
  author={Chen, Junsong and Yu, Jincheng and Ge, Chongjian and Yao, Lewei and Xie, Enze and Wu, Yue and Wang, Zhongdao and Kwok, James and Luo, Ping and Lu, Huchuan and others},
  booktitle=ICLR,
  year={2024}
}

@inproceedings{xie2025sana,
  title={Sana: Efficient high-resolution image synthesis with linear diffusion transformers},
  author={Xie, Enze and Chen, Junsong and Chen, Junyu and Cai, Han and Tang, Haotian and Lin, Yujun and Zhang, Zhekai and Li, Muyang and Zhu, Ligeng and Lu, Yao and others},
  booktitle=ICLR,
  year={2025}
}

@article{raffel2020exploring,
  title={Exploring the limits of transfer learning with a unified text-to-text transformer},
  author={Raffel, Colin and Shazeer, Noam and Roberts, Adam and Lee, Katherine and Narang, Sharan and Matena, Michael and Zhou, Yanqi and Li, Wei and Liu, Peter J},
  journal=JMLR,
  year={2020}
}

@inproceedings{rombach2022high,
  title={High-resolution image synthesis with latent diffusion models},
  author={Rombach, Robin and Blattmann, Andreas and Lorenz, Dominik and Esser, Patrick and Ommer, Bj{\"o}rn},
  booktitle=CVPR,
  year={2022}
}

@article{zhang2025encoder,
  title={Encoder-decoder gemma: Improving the quality-efficiency trade-off via adaptation},
  author={Zhang, Biao and Moiseev, Fedor and Ainslie, Joshua and Suganthan, Paul and Ma, Min and Bhupatiraju, Surya and Lebron, Fede and Firat, Orhan and Joulin, Armand and Dong, Zhe},
  journal={arXiv preprint arXiv:2504.06225},
  year={2025}
}

@article{zhang2025t5gemma,
  title={T5Gemma 2: Seeing, Reading, and Understanding Longer},
  author={Zhang, Biao and Suganthan, Paul and Liu, Ga{\"e}l and Philippov, Ilya and Dua, Sahil and Hora, Ben and Black, Kat and Martins, Gus and Sanseviero, Omar and Pathak, Shreya and others},
  journal={arXiv preprint arXiv:2512.14856},
  year={2025}
}

@article{bai2025qwen3,
  title={Qwen3-vl technical report},
  author={Bai, Shuai and Cai, Yuxuan and Chen, Ruizhe and Chen, Keqin and Chen, Xionghui and Cheng, Zesen and Deng, Lianghao and Ding, Wei and Gao, Chang and Ge, Chunjiang and others},
  journal={arXiv preprint arXiv:2511.21631},
  year={2025}
}

@article{xie2025fg,
  title={FG-CLIP 2: A Bilingual Fine-grained Vision-Language Alignment Model},
  author={Xie, Chunyu and Wang, Bin and Kong, Fanjing and Li, Jincheng and Liang, Dawei and Ao, Ji and Leng, Dawei and Yin, Yuhui},
  journal={arXiv preprint arXiv:2510.10921},
  year={2025}
}

@inproceedings{sun2025noise,
  title={Is noise conditioning necessary for denoising generative models?},
  author={Sun, Qiao and Jiang, Zhicheng and Zhao, Hanhong and He, Kaiming},
  booktitle=ICML,
  year={2025}
}

@article{wang2024emu3,
  title={Emu3: Next-token prediction is all you need},
  author={Wang, Xinlong and Zhang, Xiaosong and Luo, Zhengxiong and Sun, Quan and Cui, Yufeng and Wang, Jinsheng and Zhang, Fan and Wang, Yueze and Li, Zhen and Yu, Qiying and others},
  journal={arXiv preprint arXiv:2409.18869},
  year={2024}
}

@article{pan2025transfer,
  title={Transfer between modalities with metaqueries},
  author={Pan, Xichen and Shukla, Satya Narayan and Singh, Aashu and Zhao, Zhuokai and Mishra, Shlok Kumar and Wang, Jialiang and Xu, Zhiyang and Chen, Jiuhai and Li, Kunpeng and Juefei-Xu, Felix and others},
  journal={arXiv preprint arXiv:2504.06256},
  year={2025}
}

@article{hu2024ella,
  title={Ella: Equip diffusion models with llm for enhanced semantic alignment},
  author={Hu, Xiwei and Wang, Rui and Fang, Yixiao and Fu, Bin and Cheng, Pei and Yu, Gang},
  journal={arXiv preprint arXiv:2403.05135},
  year={2024}
}

@inproceedings{ghosh2023geneval,
  title={Geneval: An object-focused framework for evaluating text-to-image alignment},
  author={Ghosh, Dhruba and Hajishirzi, Hannaneh and Schmidt, Ludwig},
  booktitle=NIPS,
  year={2023}
}

@article{du2025textcrafter,
  title={Textcrafter: Accurately rendering multiple texts in complex visual scenes},
  author={Du, Nikai and Chen, Zhennan and Gao, Shan and Chen, Zhizhou and Chen, Xi and Jiang, Zhengkai and Yang, Jian and Tai, Ying},
  journal={arXiv preprint arXiv:2503.23461},
  year={2025}
}

@article{geng2025x,
  title={X-omni: Reinforcement learning makes discrete autoregressive image generative models great again},
  author={Geng, Zigang and Wang, Yibing and Ma, Yeyao and Li, Chen and Rao, Yongming and Gu, Shuyang and Zhong, Zhao and Lu, Qinglin and Hu, Han and Zhang, Xiaosong and others},
  journal={arXiv preprint arXiv:2507.22058},
  year={2025}
}

@inproceedings{dehghani2023scaling,
  title={Scaling vision transformers to 22 billion parameters},
  author={Dehghani, Mostafa and Djolonga, Josip and Mustafa, Basil and Padlewski, Piotr and Heek, Jonathan and Gilmer, Justin and Steiner, Andreas Peter and Caron, Mathilde and Geirhos, Robert and Alabdulmohsin, Ibrahim and others},
  booktitle=ICML,
  year={2023},
}

@article{ho2022classifier,
  title={Classifier-free diffusion guidance},
  author={Ho, Jonathan and Salimans, Tim},
  journal={arXiv preprint arXiv:2207.12598},
  year={2022}
}

@article{su2024roformer,
  title={Roformer: Enhanced transformer with rotary position embedding},
  author={Su, Jianlin and Ahmed, Murtadha and Lu, Yu and Pan, Shengfeng and Bo, Wen and Liu, Yunfeng},
  journal={Neurocomputing},
  year={2024},
}

@inproceedings{zhang2019root,
  title={Root mean square layer normalization},
  author={Zhang, Biao and Sennrich, Rico},
  booktitle=NIPS,
  year={2019}
}

@article{shazeer2020glu,
  title={Glu variants improve transformer},
  author={Shazeer, Noam},
  journal={arXiv preprint arXiv:2002.05202},
  year={2020}
}

@inproceedings{deng2009imagenet,
  title={Imagenet: A large-scale hierarchical image database},
  author={Deng, Jia and Dong, Wei and Socher, Richard and Li, Li-Jia and Li, Kai and Fei-Fei, Li},
  booktitle=CVPR,
  year={2009},
}

@article{thomee2016yfcc100m,
  title={Yfcc100m: The new data in multimedia research},
  author={Thomee, Bart and Shamma, David A and Friedland, Gerald and Elizalde, Benjamin and Ni, Karl and Poland, Douglas and Borth, Damian and Li, Li-Jia},
  journal={Communications of the ACM},
  year={2016},
}

@inproceedings{desai2021redcaps,
  title={Redcaps: Web-curated image-text data created by the people, for the people},
  author={Desai, Karan and Kaul, Gaurav and Aysola, Zubin and Johnson, Justin},
  booktitle={NeurIPS Datasets and Benchmarks Track},
  year={2021}
}

@misc{BoerBohan2024Megalith10m,
  author       = {Boer Bohan, Ollin},
  title        = {Megalith-10m},
  year         = {2024},
  howpublished = {\url{https://huggingface.co/datasets/madebyollin/megalith-10m}},
}

@misc{Narugo2024PexelsTaggerV0,
  author       = {Golding, Naomi Rue},
  title        = {Pexels Tagger V0 Webdataset (Full)},
  year         = {2024},
  howpublished = {\url{https://huggingface.co/datasets/animetimm/pexels-tagger-v0-w640-ws-full}},
}

@inproceedings{vendrow2024inquire,
  title={INQUIRE: A natural world text-to-image retrieval benchmark},
  author={Vendrow, Edward and Pantazis, Omiros and Shepard, Alexander and Brostow, Gabriel and Jones, Kate E and Mac Aodha, Oisin and Beery, Sara and Van Horn, Grant},
  booktitle=NIPS,
  year={2024}
}

@article{zhou2017places,
  title={Places: A 10 million image database for scene recognition},
  author={Zhou, Bolei and Lapedriza, Agata and Khosla, Aditya and Oliva, Aude and Torralba, Antonio},
  journal=PAMI,
  year={2017},
}

@article{wang2025gpt,
  title={Gpt-image-edit-1.5 m: A million-scale, gpt-generated image dataset},
  author={Wang, Yuhan and Yang, Siwei and Zhao, Bingchen and Zhang, Letian and Liu, Qing and Zhou, Yuyin and Xie, Cihang},
  journal={arXiv preprint arXiv:2507.21033},
  year={2025}
}

@misc{CortexLM2024MidjourneyV6,
  author       = {Photoroom},
  title        = {Midjourney-v6 Dataset},
  year         = {2024},
  howpublished = {\url{https://huggingface.co/datasets/Photoroom/midjourney-v6-recap}},
}

@misc{Wendler2024RenderedText,
  author       = {Wendler, Chris},
  title        = {RenderedText Dataset},
  year         = {2024},
  howpublished = {\url{https://huggingface.co/datasets/wendlerc/RenderedText}},
}

@article{wang2025textatlas5m,
  title={Textatlas5m: A large-scale dataset for dense text image generation},
  author={Wang, Alex Jinpeng and Mao, Dongxing and Zhang, Jiawei and Han, Weiming and Dong, Zhuobai and Li, Linjie and Lin, Yiqi and Yang, Zhengyuan and Qin, Libo and Zhang, Fuwei and others},
  journal={arXiv preprint arXiv:2502.07870},
  year={2025}
}

@article{wang2024qwen2,
  title={Qwen2-vl: Enhancing vision-language model's perception of the world at any resolution},
  author={Wang, Peng and Bai, Shuai and Tan, Sinan and Wang, Shijie and Fan, Zhihao and Bai, Jinze and Chen, Keqin and Liu, Xuejing and Wang, Jialin and Ge, Wenbin and others},
  journal={arXiv preprint arXiv:2409.12191},
  year={2024}
}

@inproceedings{lipman2022flow,
  title={Flow matching for generative modeling},
  author={Lipman, Yaron and Chen, Ricky TQ and Ben-Hamu, Heli and Nickel, Maximilian and Le, Matt},
  booktitle=ICLR,
  year={2023}
}

@inproceedings{saharia2022photorealistic,
  title={Photorealistic text-to-image diffusion models with deep language understanding},
  author={Saharia, Chitwan and Chan, William and Saxena, Saurabh and Li, Lala and Whang, Jay and Denton, Emily L and Ghasemipour, Kamyar and Gontijo Lopes, Raphael and Karagol Ayan, Burcu and Salimans, Tim and others},
  booktitle=NIPS,
  year={2022}
}

@article{ramesh2022hierarchical,
  title={Hierarchical text-conditional image generation with clip latents},
  author={Ramesh, Aditya and Dhariwal, Prafulla and Nichol, Alex and Chu, Casey and Chen, Mark},
  journal={arXiv preprint arXiv:2204.06125},
  year={2022}
}

@article{sun2024autoregressive,
  title={Autoregressive model beats diffusion: Llama for scalable image generation},
  author={Sun, Peize and Jiang, Yi and Chen, Shoufa and Zhang, Shilong and Peng, Bingyue and Luo, Ping and Yuan, Zehuan},
  journal={arXiv preprint arXiv:2406.06525},
  year={2024}
}

@inproceedings{zhou2024transfusion,
  title={Transfusion: Predict the next token and diffuse images with one multi-modal model},
  author={Zhou, Chunting and Yu, Lili and Babu, Arun and Tirumala, Kushal and Yasunaga, Michihiro and Shamis, Leonid and Kahn, Jacob and Ma, Xuezhe and Zettlemoyer, Luke and Levy, Omer},
  booktitle=ICLR,
  year={2025}
}

@inproceedings{yao2025reconstruction,
  title={Reconstruction vs. generation: Taming optimization dilemma in latent diffusion models},
  author={Yao, Jingfeng and Yang, Bin and Wang, Xinggang},
  booktitle=CVPR,
  year={2025}
}

@article{ryu2025flite,
  title={F Lite Technical Report},
  author={Ryu, Simo and Pengqi, Lu and Mart\'in Juan, Javier and de Prado Alonso, Iv\'an},
  year={2025},
  url          = {https://github.com/fal-ai/f-lite},
}

@article{chen2025dit,
  title={Dit-air: Revisiting the efficiency of diffusion model architecture design in text to image generation},
  author={Chen, Chen and Qian, Rui and Hu, Wenze and Fu, Tsu-Jui and Tong, Jialing and Wang, Xinze and Li, Lezhi and Zhang, Bowen and Schwing, Alex and Liu, Wei and others},
  journal={arXiv preprint arXiv:2503.10618},
  year={2025}
}

@inproceedings{tong2024cambrian,
  title={Cambrian-1: A fully open, vision-centric exploration of multimodal llms},
  author={Tong, Peter and Brown, Ellis and Wu, Penghao and Woo, Sanghyun and Iyer, Adithya Jairam Vedagiri and Akula, Sai Charitha and Yang, Shusheng and Yang, Jihan and Middepogu, Manoj and Wang, Ziteng and others},
  booktitle=NIPS,
  year={2024}
}

@inproceedings{chang2023muse,
  title={Muse: Text-To-Image Generation via Masked Generative Transformers},
  author={Chang, Huiwen and Zhang, Han and Barber, Jarred and Maschinot, Aaron and Lezama, Jose and Jiang, Lu and Yang, Ming-Hsuan and Murphy, Kevin Patrick and Freeman, William T and Rubinstein, Michael and others},
  booktitle=ICML,
  year={2023},
}

@inproceedings{sehwag2025stretching,
  title={Stretching each dollar: Diffusion training from scratch on a micro-budget},
  author={Sehwag, Vikash and Kong, Xianghao and Li, Jingtao and Spranger, Michael and Lyu, Lingjuan},
  booktitle=CVPR,
  year={2025}
}

@inproceedings{kynkaanniemi2024applying,
  title={Applying guidance in a limited interval improves sample and distribution quality in diffusion models},
  author={Kynk{\"a}{\"a}nniemi, Tuomas and Aittala, Miika and Karras, Tero and Laine, Samuli and Aila, Timo and Lehtinen, Jaakko},
  booktitle=NIPS,
  year={2024}
}

@inproceedings{ding2021cogview,
  title={Cogview: Mastering text-to-image generation via transformers},
  author={Ding, Ming and Yang, Zhuoyi and Hong, Wenyi and Zheng, Wendi and Zhou, Chang and Yin, Da and Lin, Junyang and Zou, Xu and Shao, Zhou and Yang, Hongxia and others},
  booktitle=NIPS,
  year={2021}
}

@article{yang2025qwen3,
  title={Qwen3 technical report},
  author={Yang, An and Li, Anfeng and Yang, Baosong and Zhang, Beichen and Hui, Binyuan and Zheng, Bo and Yu, Bowen and Gao, Chang and Huang, Chengen and Lv, Chenxu and others},
  journal={arXiv preprint arXiv:2505.09388},
  year={2025}
}

@online{blackforestlabs_flux2_2025,
  author       = {{Black Forest Labs}},
  title        = {FLUX.2: Frontier Visual Intelligence},
  year         = {2025},
  url          = {https://bfl.ai/blog/flux-2},
}

@inproceedings{xiong2020layer,
  title={On layer normalization in the transformer architecture},
  author={Xiong, Ruibin and Yang, Yunchang and He, Di and Zheng, Kai and Zheng, Shuxin and Xing, Chen and Zhang, Huishuai and Lan, Yanyan and Wang, Liwei and Liu, Tieyan},
  booktitle=ICML,
  year={2020},
}

@inproceedings{ma2024exploring,
  title={Exploring the Role of Large Language Models in Prompt Encoding for Diffusion Models},
  author={Bingqi Ma and Zhuofan Zong and Guanglu Song and Hongsheng Li and Yu Liu},
  booktitle=NIPS,
  year={2024},
}

@software{jax2018github,
  author = {James Bradbury and Roy Frostig and Peter Hawkins and Matthew James Johnson and Yash Katariya and Chris Leary and Dougal Maclaurin and George Necula and Adam Paszke and Jake Vander{P}las and Skye Wanderman-{M}ilne and Qiao Zhang},
  title = {{JAX}: composable transformations of {P}ython+{N}um{P}y programs},
  url = {http://github.com/jax-ml/jax},
  year = {2018},
}

@article{xu2021gspmd,
  title={Gspmd: general and scalable parallelization for ml computation graphs},
  author={Xu, Yuanzhong and Lee, HyoukJoong and Chen, Dehao and Hechtman, Blake and Huang, Yanping and Joshi, Rahul and Krikun, Maxim and Lepikhin, Dmitry and Ly, Andy and Maggioni, Marcello and others},
  journal={arXiv preprint arXiv:2105.04663},
  year={2021}
}

@inproceedings{rajbhandari2020zero,
  title={Zero: Memory optimizations toward training trillion parameter models},
  author={Rajbhandari, Samyam and Rasley, Jeff and Ruwase, Olatunji and He, Yuxiong},
  booktitle={SC20: international conference for high performance computing, networking, storage and analysis},
  year={2020},
}

@inproceedings{tay2023ul,
  title={{UL}2: Unifying Language Learning Paradigms},
  author={Yi Tay and Mostafa Dehghani and Vinh Q. Tran and Xavier Garcia and Jason Wei and Xuezhi Wang and Hyung Won Chung and Dara Bahri and Tal Schuster and Steven Zheng and Denny Zhou and Neil Houlsby and Donald Metzler},
  booktitle=ICLR,
  year={2023},
}

@article{li2024hunyuan,
  title={Hunyuan-dit: A powerful multi-resolution diffusion transformer with fine-grained chinese understanding},
  author={Li, Zhimin and Zhang, Jianwei and Lin, Qin and Xiong, Jiangfeng and Long, Yanxin and Deng, Xinchi and Zhang, Yingfang and Liu, Xingchao and Huang, Minbin and Xiao, Zedong and others},
  journal={arXiv preprint arXiv:2405.08748},
  year={2024}
}

@article{liu2024playground,
  title={Playground v3: Improving text-to-image alignment with deep-fusion large language models},
  author={Liu, Bingchen and Akhgari, Ehsan and Visheratin, Alexander and Kamko, Aleks and Xu, Linmiao and Shrirao, Shivam and Lambert, Chase and Souza, Joao and Doshi, Suhail and Li, Daiqing},
  journal={arXiv preprint arXiv:2409.10695},
  year={2024}
}

@inproceedings{shi2026lmfusion,
  title={Lmfusion: Adapting pretrained language models for multimodal generation},
  author={Shi, Weijia and Han, Xiaochuang and Zhou, Chunting and Liang, Weixin and Lin, Xi and Zettlemoyer, Luke and Yu, Lili},
  booktitle=NIPS,
  year={2026}
}

@inproceedings{wallace2024diffusion,
  title={Diffusion model alignment using direct preference optimization},
  author={Wallace, Bram and Dang, Meihua and Rafailov, Rafael and Zhou, Linqi and Lou, Aaron and Purushwalkam, Senthil and Ermon, Stefano and Xiong, Caiming and Joty, Shafiq and Naik, Nikhil},
  booktitle=CVPR,
  year={2024}
}

@inproceedings{liu2026flow,
  title={Flow-grpo: Training flow matching models via online rl},
  author={Liu, Jie and Liu, Gongye and Liang, Jiajun and Li, Yangguang and Liu, Jiaheng and Wang, Xintao and Wan, Pengfei and Zhang, Di and Ouyang, Wanli},
  booktitle=NIPS,
  year={2026}
}

@inproceedings{startsev2026alchemist,
  title={Alchemist: Turning Public Text-to-Image Data into Generative Gold},
  author={Startsev, Valerii and Ustyuzhanin, Alexander and Kirillov, Alexey and Baranchuk, Dmitry and Kastryulin, Sergey},
  booktitle=NIPS,
  year={2026}
}

@inproceedings{sun2023journeydb,
  title={Journeydb: A benchmark for generative image understanding},
  author={Sun, Keqiang and Pan, Junting and Ge, Yuying and Li, Hao and Duan, Haodong and Wu, Xiaoshi and Zhang, Renrui and Zhou, Aojun and Qin, Zipeng and Wang, Yi and others},
  booktitle=NIPS,
  year={2023}
}

@inproceedings{changpinyo2021conceptual,
  title={Conceptual 12m: Pushing web-scale image-text pre-training to recognize long-tail visual concepts},
  author={Changpinyo, Soravit and Sharma, Piyush and Ding, Nan and Soricut, Radu},
  booktitle=CVPR,
  year={2021}
}

@inproceedings{kirillov2023segment,
  title={Segment anything},
  author={Kirillov, Alexander and Mintun, Eric and Ravi, Nikhila and Mao, Hanzi and Rolland, Chloe and Gustafson, Laura and Xiao, Tete and Whitehead, Spencer and Berg, Alexander C and Lo, Wan-Yen and others},
  booktitle=ICCV,
  year={2023}
}

\newpage

\section*{\LARGE Appendix}

\vspace{.2cm}

\phantomsection
\addcontentsline{toc}{section}{\protect\numberline{8}Appendix}

\appendix

\makeatletter
\let\origaddcontentsline\addcontentsline
\renewcommand{\addcontentsline}[3]{%
  \def\tempa{#1}%
  \def\tempb{toc}%
  \def\tempc{#2}%
  \def\tempd{section}%
  \def\tempe{subsection}%
  \ifx\tempa\tempb
    \ifx\tempc\tempd
      \origaddcontentsline{toc}{subsection}{#3}%
    \else
      \ifx\tempc\tempe
      \else
        \origaddcontentsline{#1}{#2}{#3}%
      \fi
    \fi
  \else
    \origaddcontentsline{#1}{#2}{#3}%
  \fi
}
\makeatother

\section{Implementation Details}

In this section, we provide further details on our modeling and training configurations.

\subsection{Configuration}
\label{appendix:config}

\paragraph{Hardware}. Our model training and inference are conducted on TPU v4, v5p, and v6e with JAX~\citep{jax2018github}. Benchmark evaluations are performed on NVIDIA A100, H100, and H200 GPUs.

\paragraph{General configuration}. Our default baseline models largely follow the XL/2 model configurations used in previous diffusion models~\citep{peebles2023scalable, yao2025reconstruction}, which use a hidden size of 1152, 16 attention heads, an MLP ratio of 4.0, and a patch size of 2. However, unlike those models, we use 29 layers instead of 28. By default, during both training and inference, we maintain the text encoder in bf16 while keeping all other parameters in fp32. To ensure the models fit into memory, we shard model parameters and optimizer states across devices using JAX pjit/GSPMD~\citep{xu2021gspmd}, following ZeRO-style fully sharded data parallelism~\citep{rajbhandari2020zero}.

\begin{table}[h]
\tablestyle{6pt}{1.02}
\footnotesize
\begin{tabular}{y{166}|y{62}}
config & value \\
\shline
optimizer & Adam \\
learning rate & 1e-4 \\
weight decay & 0 \\
optimizer momentum & $\beta_1, \beta_2{=}0.9, 0.95$ \\
batch size & 512 \\
learning rate schedule & constant \\
gradient clipping & 1 \\
training objective & flow matching \\
training steps & 500K \\
training timestep distribution & lognorm(0, 1) \\
inference timestep shift value~\citep{esser2024scaling} & 0.3 \\
inference steps & 250 \\
CFG scale~\citep{ho2022classifier} & 12 \\
CFG rescale strength~\citep{lin2024common} & 0 \\
CFG interval~\citep{kynkaanniemi2024applying} & [0, 1] \\
\end{tabular}
\caption{\textbf{Training and inference configurations for all 256-resolution controlled experiments} in Sections~\blueref{sec:modeling} and~\blueref{sec:data}.}
\label{tab:cfg}
\end{table}

\paragraph{256-resolution controlled experiments}. \tabref{tab:cfg} summarizes the training configuration for all controlled experiments in Sections~\blueref{sec:modeling} and~\blueref{sec:data}. All models are trained for 500K iterations, but training time varies because different experiments use different model components. For the cross-attention baseline, 500K steps take 31.0 hours on a TPU v6e-64 machine.

\begin{table}[h]
\tablestyle{6pt}{1.02}
\footnotesize
\begin{tabular}{lccccc}
training stage & \#images & training steps & batch size & training timestep shift value~\citep{esser2024scaling} & TPU v5p-128 hours \\
\shline
256-resolution & 162.9M & 2.0M & 512 & N/A & 383.0 \\
512-resolution & 9.7M & 0.5M & 512 & N/A & 174.4 \\
1024-resolution & 4.3M & 0.3M & 128 & 3.33 & 150.9 \\
\end{tabular}
\caption{\textbf{Training configurations and compute resources for the final i1 model} at each training stage.}
\label{tab:i1_cfg}
\end{table}

\begin{figure}[!htb]
        \centering
            \includegraphics[width=\linewidth]{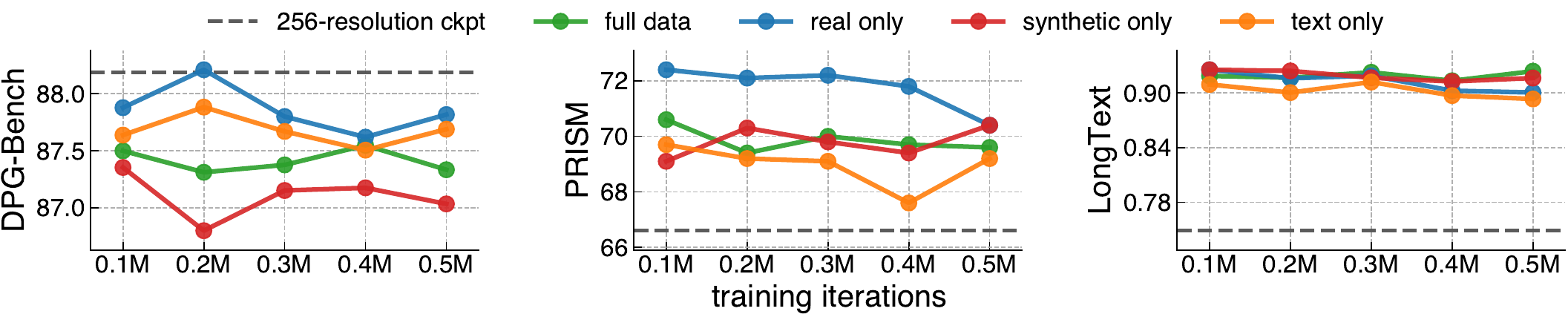}\\[-.2em]
          \textbf{(a)} 512-resolution training\\
          \vspace{1em}
            \includegraphics[width=\linewidth]{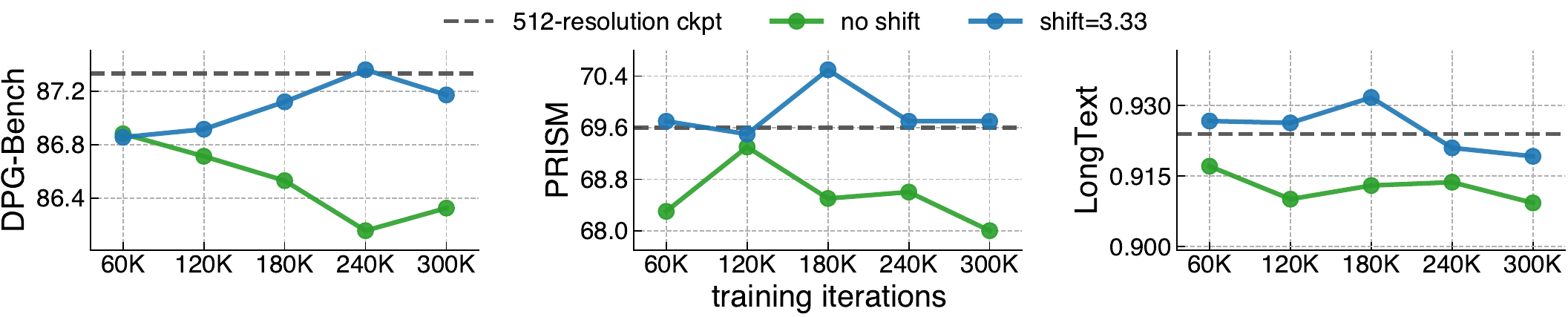}\\[-.2em]
          \textbf{(b)} 1024-resolution training
          \caption{\textbf{Benchmark performance of \textbf{i1} during high-resolution training stages}. PRISM and LongText improve substantially during 512-resolution training, while 1024-resolution training has a smaller impact on benchmark scores. For 1024-resolution training, we compare models trained with a timestep shift value of 3.33 against models trained without a timestep shift, and find that the training timestep shift consistently improves performance.}
          \label{fig:high_res_performance_curve}
\end{figure}

\paragraph{i1 training}. In~\tabref{tab:i1_cfg}, we detail the training configurations and compute resources for the final \textbf{i1} model at each training stage. All unspecified configurations are kept the same as in~\tabref{tab:cfg}. The benchmark performance trends across iterations for the high-resolution training stages are shown in~\figref{fig:high_res_performance_curve}. We observe that 512-resolution training substantially improves performance on PRISM and LongText, whereas 1024-resolution training has a smaller effect, with performance remaining close to that of the 512-resolution checkpoint from which it is initialized. For 1024-resolution training, we additionally compare models trained with a timestep shift value of 3.33 against models trained without a timestep shift, and find that applying the training timestep shift consistently improves performance.

\clearpage

\subsection{Baseline Architectures}
\label{appendix:baseline_arch}

As described in~\secrefc{sec:exp_setup}, our controlled experiments in Sections~\blueref{sec:modeling} and~\blueref{sec:data} are all based on a fixed baseline architecture. We vary one design choice at a time while keeping all other configurations identical to the baseline. Although we use the cross-attention backbone as the default in our baseline, we additionally validate some design choices on single-stream and dual-stream backbones. In this section, we provide illustrations of the three backbone architectures.

\begin{figure}[htbp]
\vspace{-1em}
    \centering
    \hspace{\fill}
    \begin{subfigure}[t]{0.42\linewidth}
        \centering
        \includegraphics[width=\linewidth]{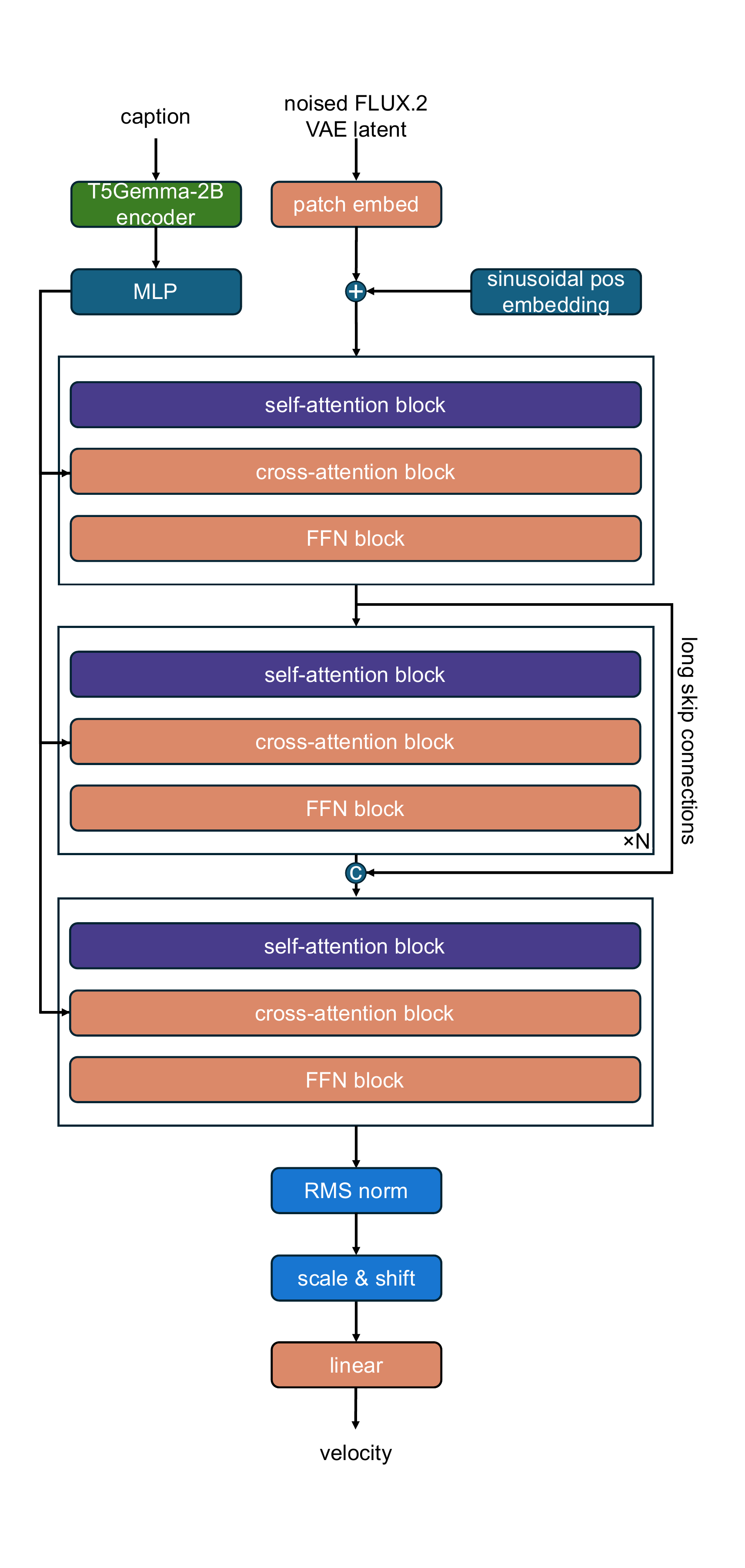}
        \caption{overall architecture}
    \end{subfigure}
    \hspace{\fill}
    \begin{subfigure}[t]{0.42\linewidth}
        \centering
        \includegraphics[width=\linewidth]{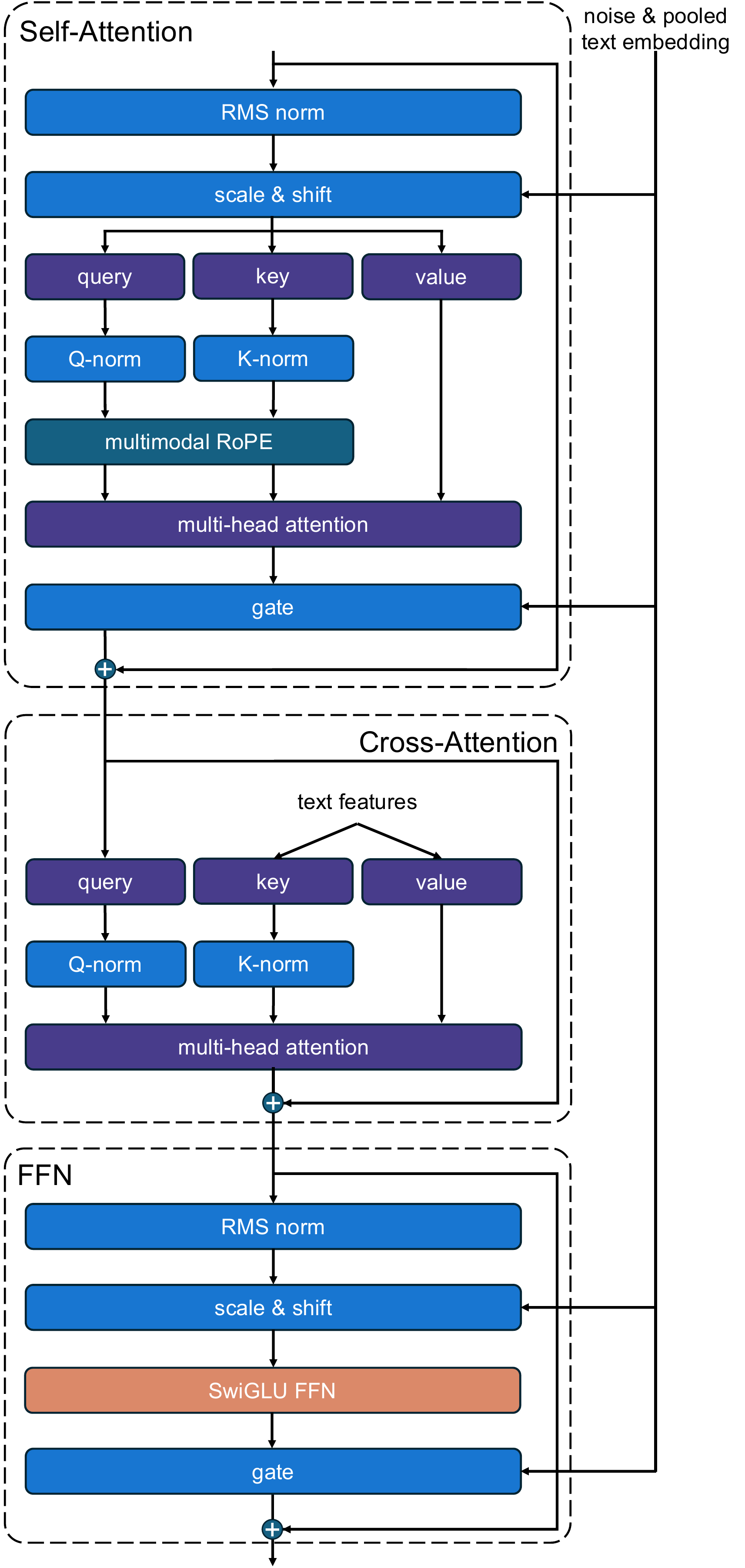}
        \caption{one transformer block}
    \end{subfigure}
    \hspace{\fill}
    \vspace{-1em}
    \caption{\textbf{The architecture of our cross-attention baseline model}. For the cross-attention backbone family, text conditioning information is passed to the backbone through cross-attention layers inserted between the self-attention and feed-forward network layers.}
    \label{fig:baseline_crossattn_architecture}
\end{figure}

\paragraph{Cross-attention backbone} passes text conditioning information to the backbone through cross-attention layers inserted between the self-attention and feed-forward network layers. The architecture is illustrated in~\figref{fig:baseline_crossattn_architecture}.

\clearpage

\begin{figure}[htbp]
    \centering
    \hspace{\fill}
    \begin{subfigure}[t]{0.45\linewidth}
        \centering
        \includegraphics[width=\linewidth]{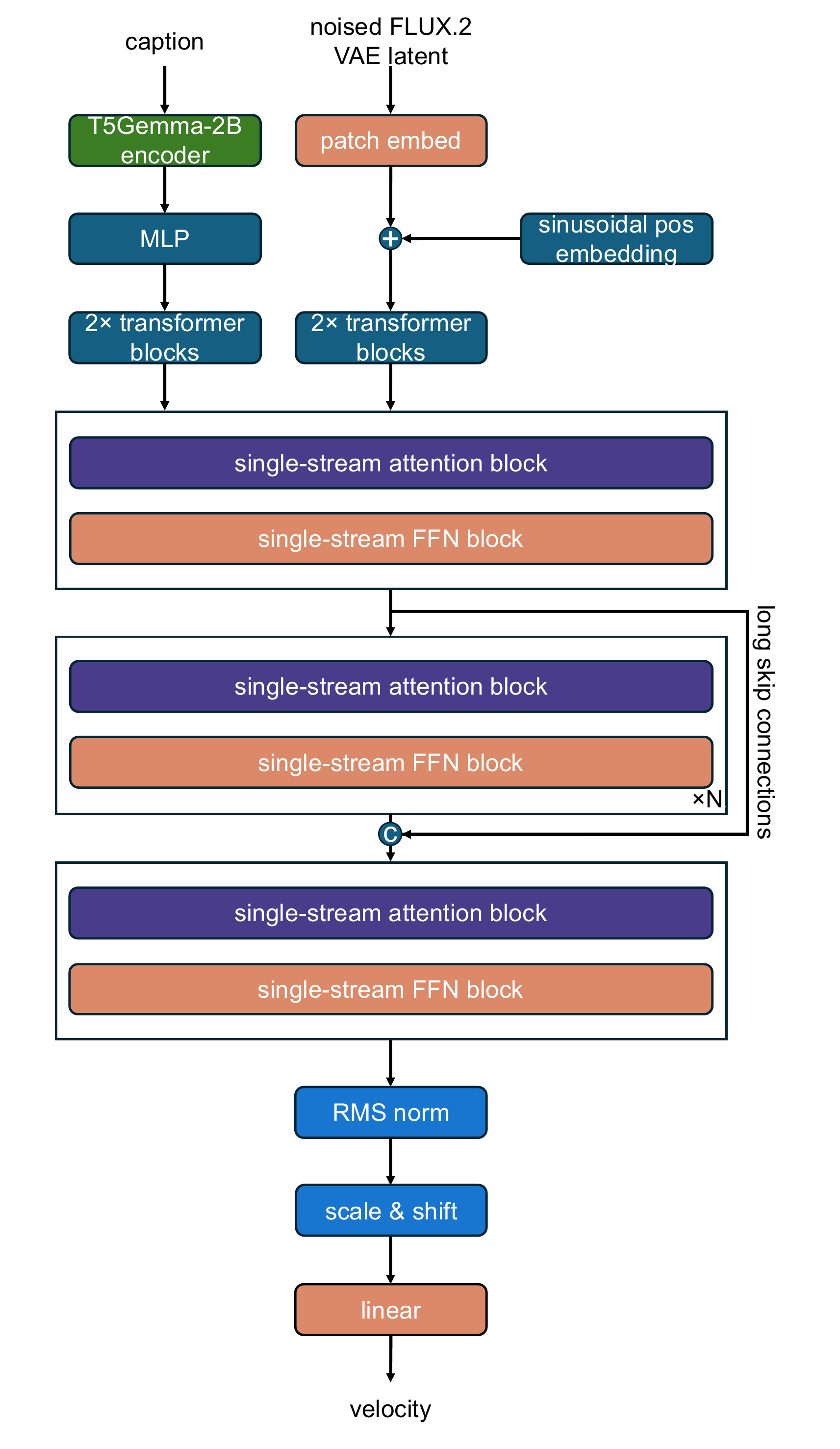}
        \caption{overall architecture}
    \end{subfigure}
    \hspace{\fill}
    \begin{subfigure}[t]{0.45\linewidth}
        \centering
        \includegraphics[width=\linewidth]{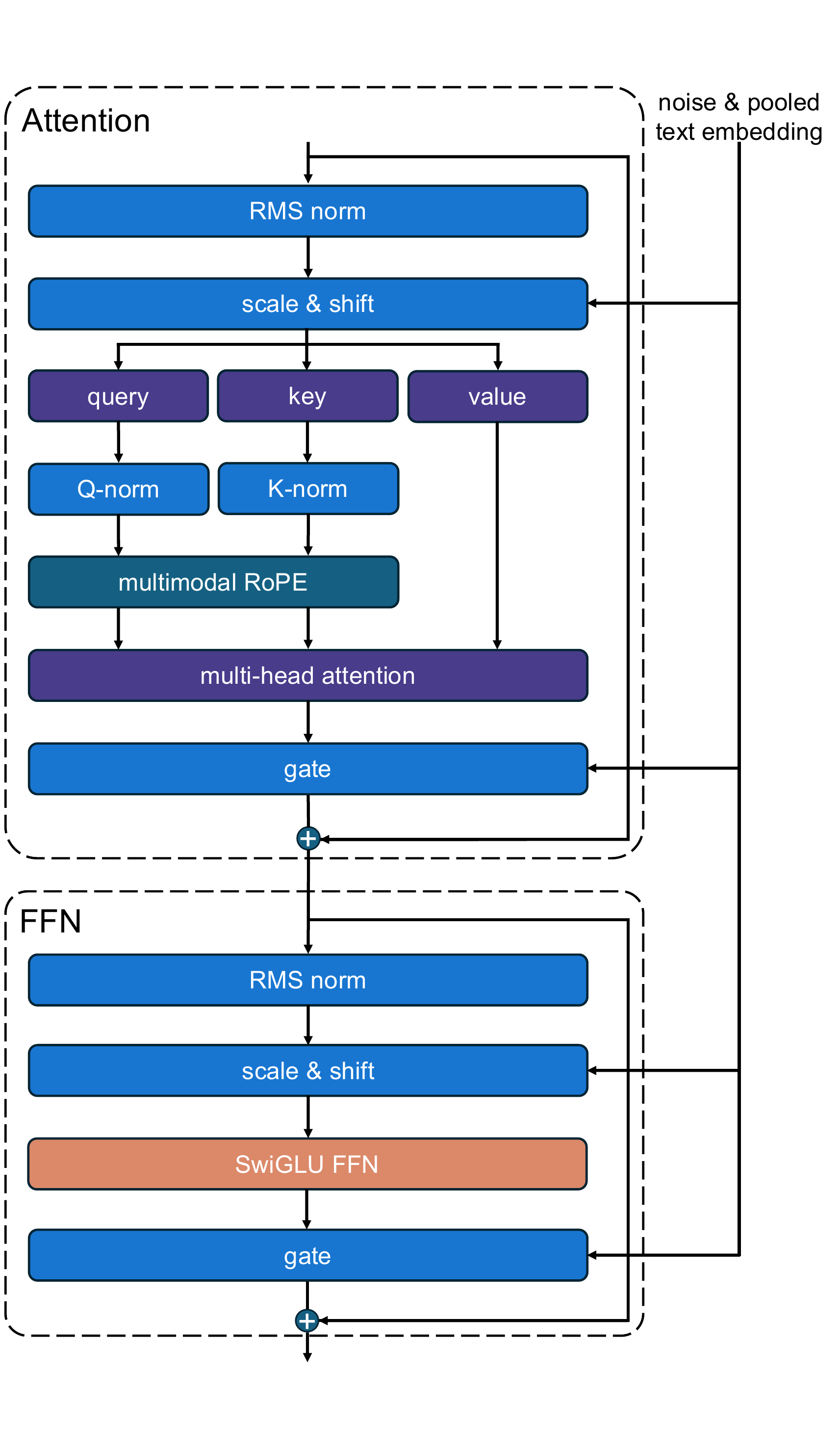}
        \caption{one transformer block}
    \end{subfigure}
    \hspace{\fill}
    \caption{\textbf{The architecture of the single-stream variant of our baseline model}. For the single-stream backbone family, the text features and noisy image features are concatenated along the sequence dimension, and processed using a single set of attention and MLP weights.}
    \label{fig:baseline_lumina_architecture}
\end{figure}

\paragraph{Single-stream backbone} concatenates the text features and noisy image features along the sequence dimension and processes the entire sequence using a single set of backbone weights. The architecture is illustrated in~\figref{fig:baseline_lumina_architecture}.

\clearpage

\begin{figure}[htbp]
    \centering
    \hspace{\fill}
    \begin{subfigure}[b]{0.45\linewidth}
        \centering
        \includegraphics[width=\linewidth, valign=c]{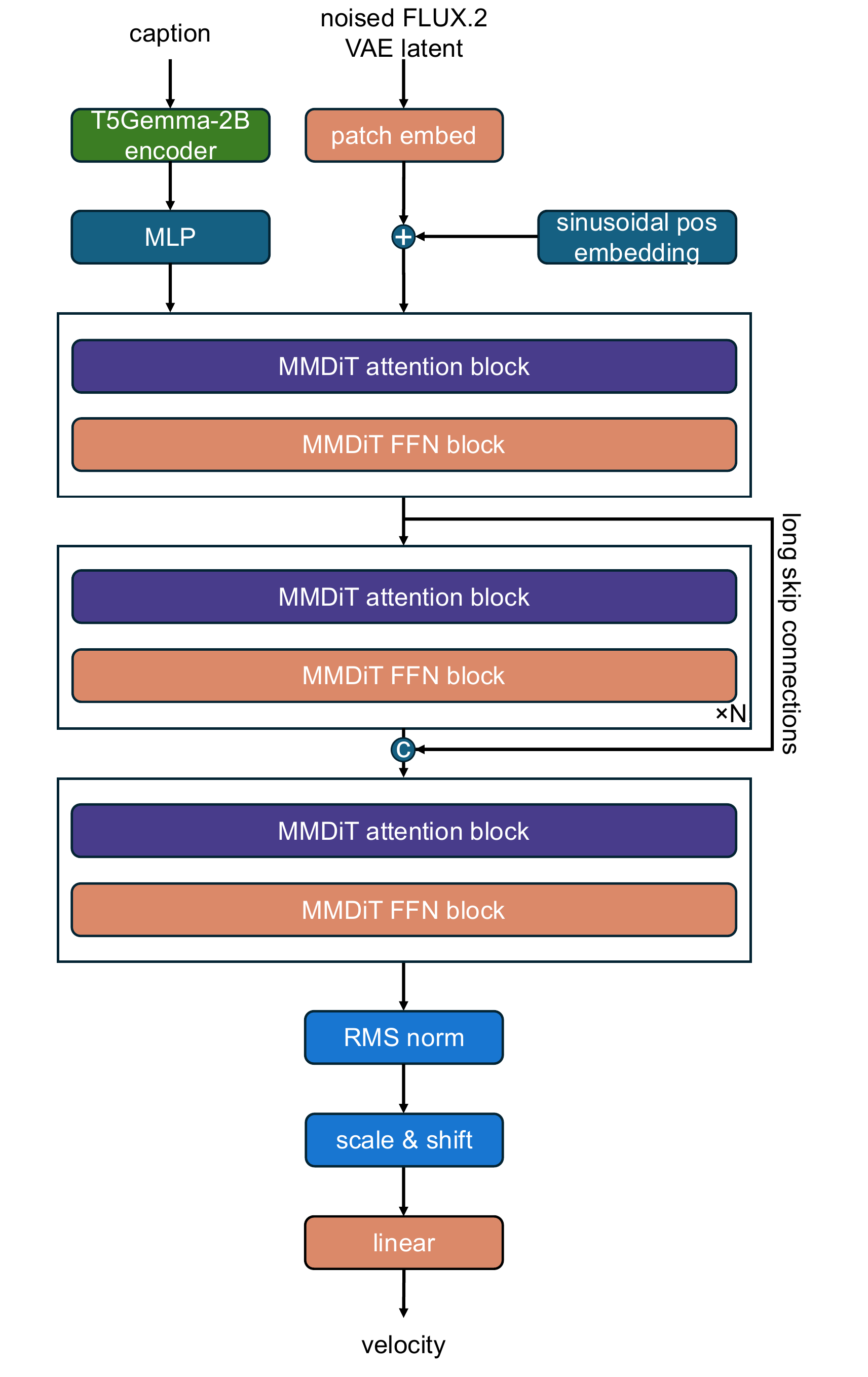}
        \caption{overall architecture}
    \end{subfigure}
    \hspace{\fill}
    \begin{subfigure}[b]{0.45\linewidth}
        \centering
        \includegraphics[width=\linewidth, valign=c]{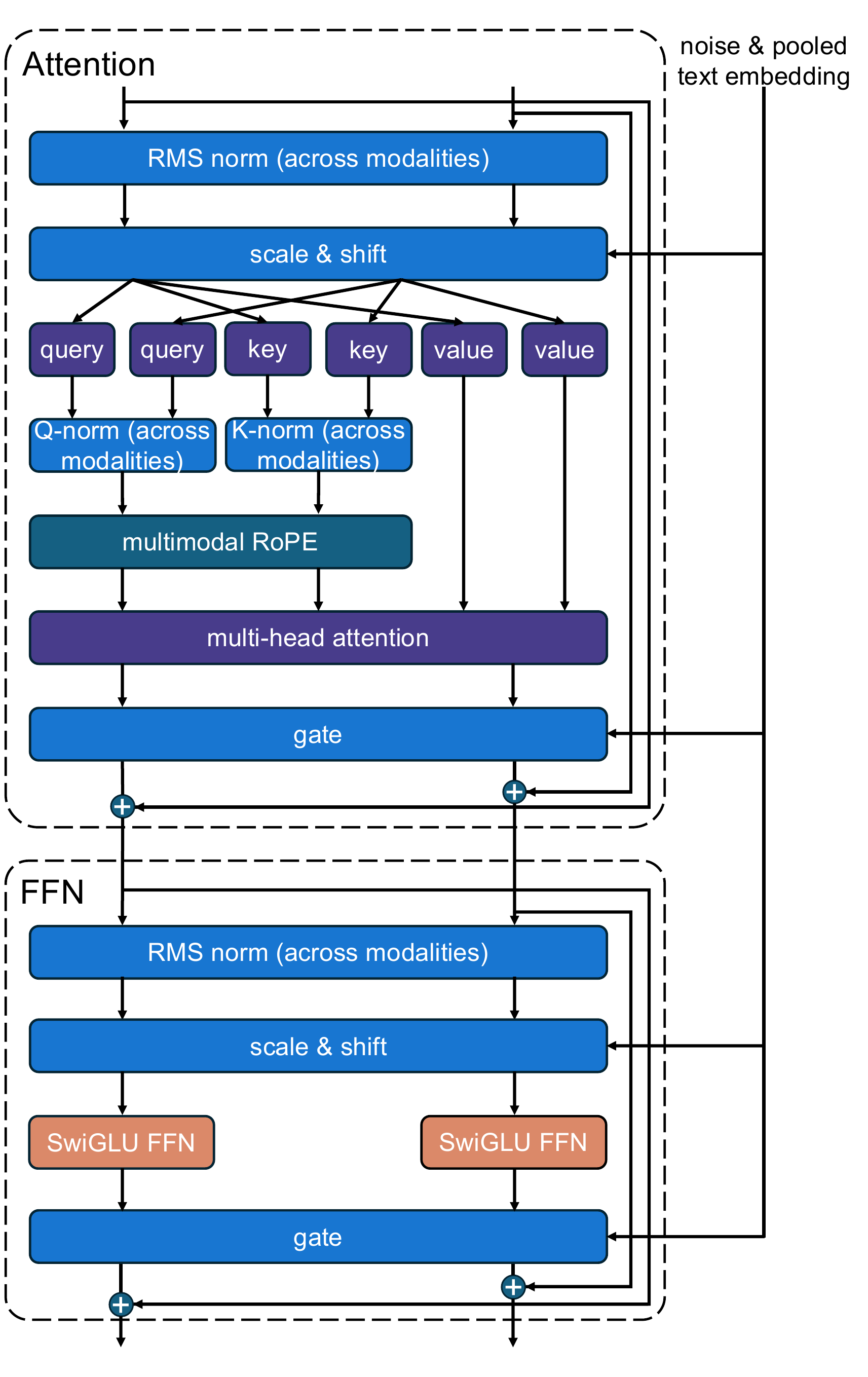}
        \caption{one transformer block}
    \end{subfigure}
    \hspace{\fill}
    \caption{\textbf{The architecture of the dual-stream variant of our baseline model}. For the dual-stream backbone family, the text features and noisy image features are concatenated along the sequence dimension, and processed using separate, modality-specific attention and MLP weights.}
    
    \label{fig:baseline_mmdit_architecture}
\end{figure}

\paragraph{Dual-stream backbone} concatenates the text features and noisy image features along the sequence dimension, but uses separate backbone parameters for the text tokens and image tokens. The architecture is illustrated in~\figref{fig:baseline_mmdit_architecture}.

\clearpage

\subsection{Details on Text Encoders}
\label{appendix:text_encoder_details}

\paragraph{Model version}. For all T5Gemma models, we use the UL2~\citep{tay2023ul} variant, as it has better encoder representations~\citep{zhang2025encoder}. For the T5Gemma-9B model, we use the variant with a 2B decoder instead of the one with a 9B decoder. For the FG-CLIP 2 model, we use the ``long'' mode.

\paragraph{Truncation}. Following mainstream implementations~\citep{esser2024scaling, blackforestlabs_flux2_2025, cai2025z, wu2025qwen}, we use right truncation for the text tokenizers. We truncate to 256 tokens for all text encoders except FG-CLIP 2, which we truncate to 196 tokens because it is trained on up to 196 tokens.

\paragraph{Hidden states}. For encoder-decoder models (\ie, the T5Gemma and T5Gemma2 families), we use the encoder's final-layer hidden states as text token features. For decoder-only models (\ie, the Qwen3 and Qwen3-VL families), we use the last hidden states from the final transformer layer as text token features. By default, we input the text-to-image prompt directly into the text encoder to obtain features. Some previous work~\citep{ma2024exploring, xie2025sana, wu2025qwen} applied system prompts to LLM/VLM text encoders; we ablate the effect of system prompts in~\appendixref{appendix:text_encoder_system_prompt}.

\clearpage

\section{Additional Information on Inference and Evaluation}

\subsection{Qualitative Comparison with Stable Diffusion 3 Medium}
\label{appendix:qualitative_comparison}

\begin{figure}[htbp]
\vspace{-2em}
    \centering
    {\small \textbf{Prompt}: Veronica Lake (1922-1973), the US actress, is depicted sitting in an armchair, dressed in a red blouse under a white apron dress, engrossed in reading a book, evoking a scene from around 1955.}\\
    \hspace{\fill}
    \begin{subfigure}[t]{0.48\linewidth}
        \centering
        \includegraphics[width=\linewidth]{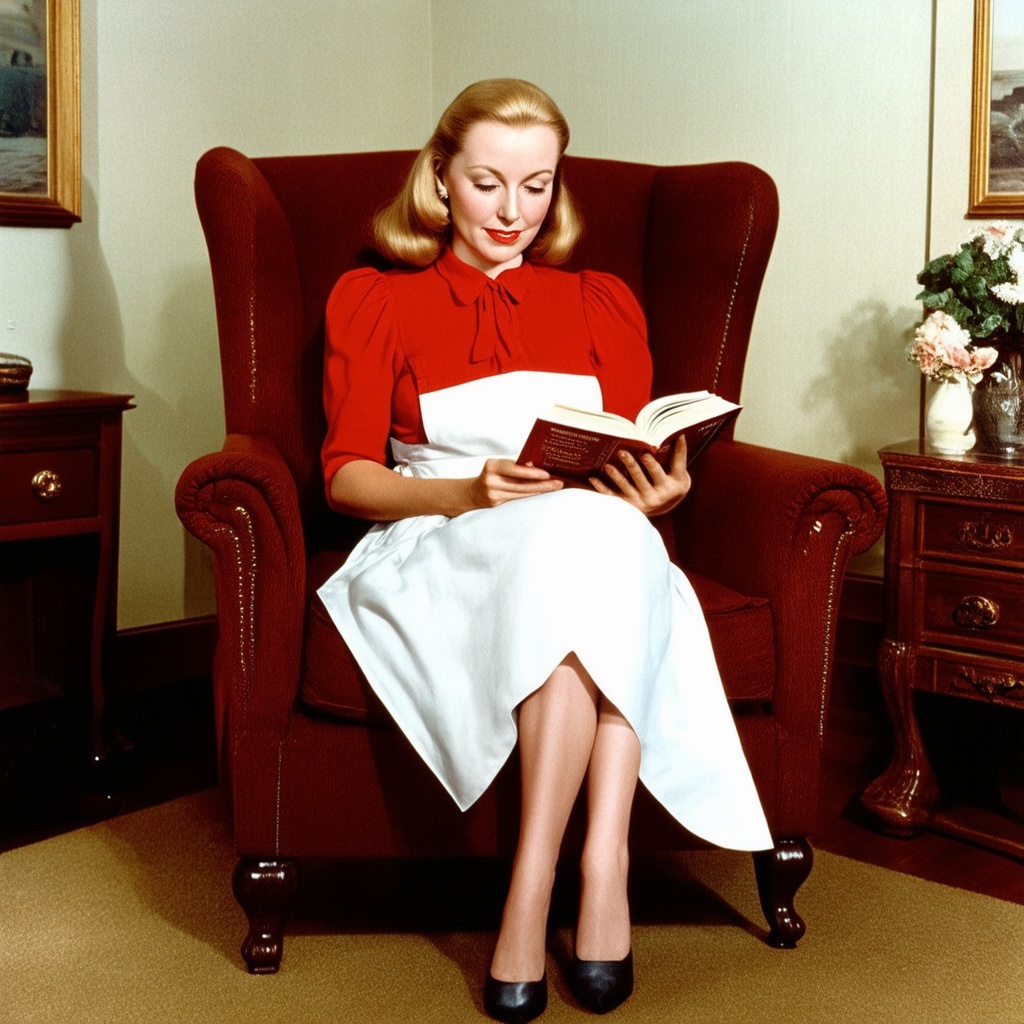}
    \end{subfigure}
    \hspace{\fill}
    \begin{subfigure}[t]{0.48\linewidth}
        \centering
        \includegraphics[width=\linewidth]{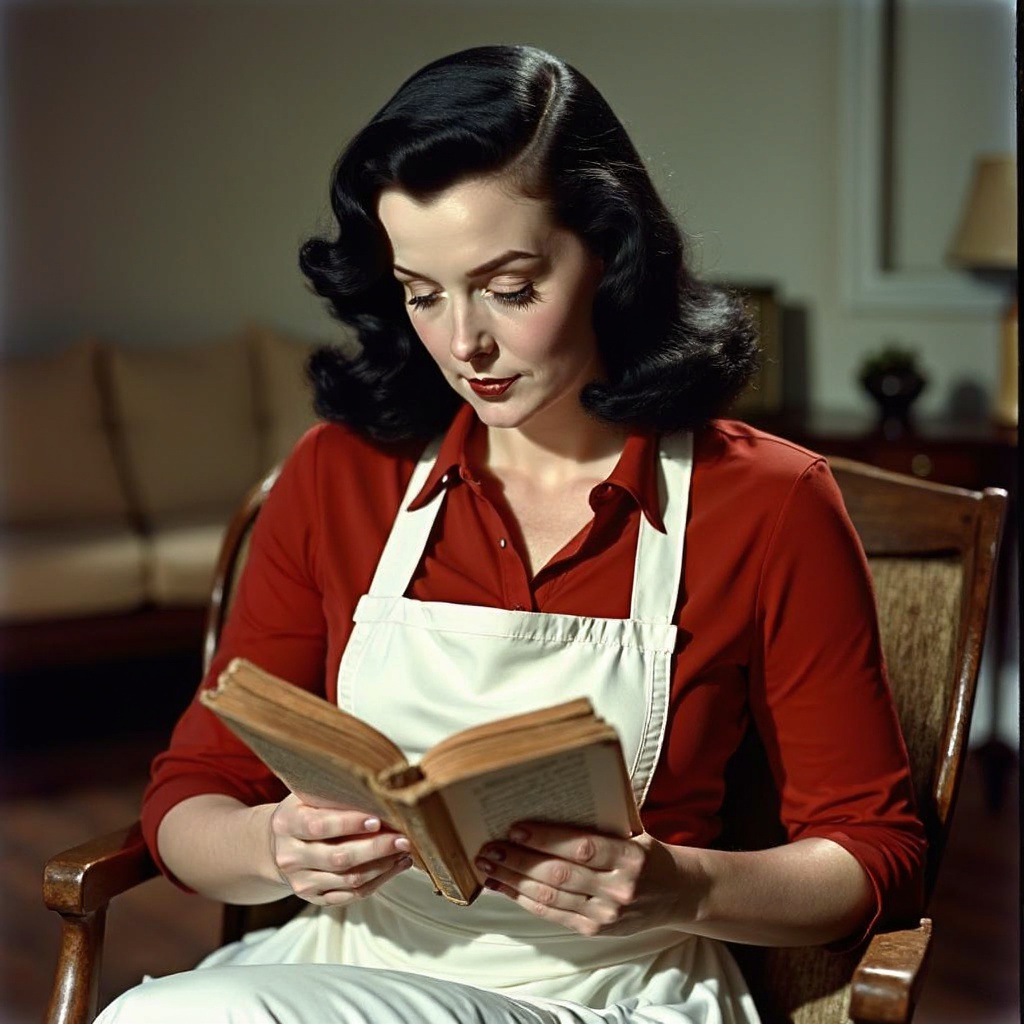}
    \end{subfigure}
    \hspace{\fill}

    {\small \textbf{Prompt}: A bright, welcoming bakery interior, captured in warm, soft morning light, showcasing an appealing, rustic wooden-framed chalkboard menu placed prominently against a white brick wall. At the center of the chalkboard, in large, elegant hand-drawn lettering, reads clearly "Today's Specials: Sourdough Bread \& Cinnamon Rolls". Just below this central message, smaller text neatly notes "Freshly baked every morning". At the top right corner of the chalkboard, subtly written in a playful cursive handwriting, are the words "Homemade with Passion". Around the borders of the menu, slightly faded and vintage-inspired illustrations of wheat stalks and pastries subtly frame the text, enhancing the artisanal bakery atmosphere. Beside the main chalkboard stands a smaller wooden sign, on which handwritten text reads "Free samples available!", accompanied by a decorative arrow directing customers toward the display counter. The textual elements are distinct, stylish, and naturally handwritten, evoking a genuine, artisanal feel, perfectly complementing the inviting bakery ambiance.}\\
    \hspace{\fill}
    \begin{subfigure}[t]{0.48\linewidth}
        \centering
        \includegraphics[width=\linewidth]{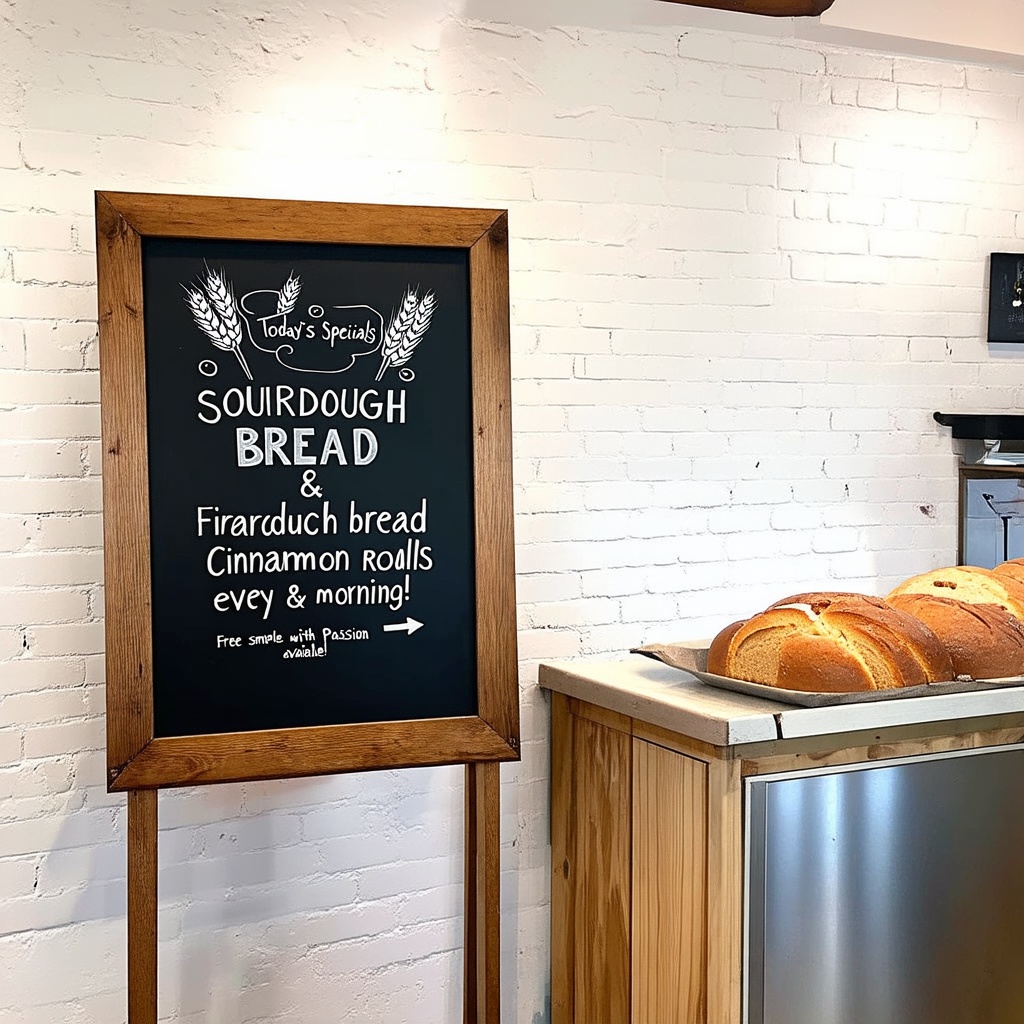}
        \caption{Stable Diffusion 3 Medium}
    \end{subfigure}
    \hspace{\fill}
    \begin{subfigure}[t]{0.48\linewidth}
        \centering
        \includegraphics[width=\linewidth]{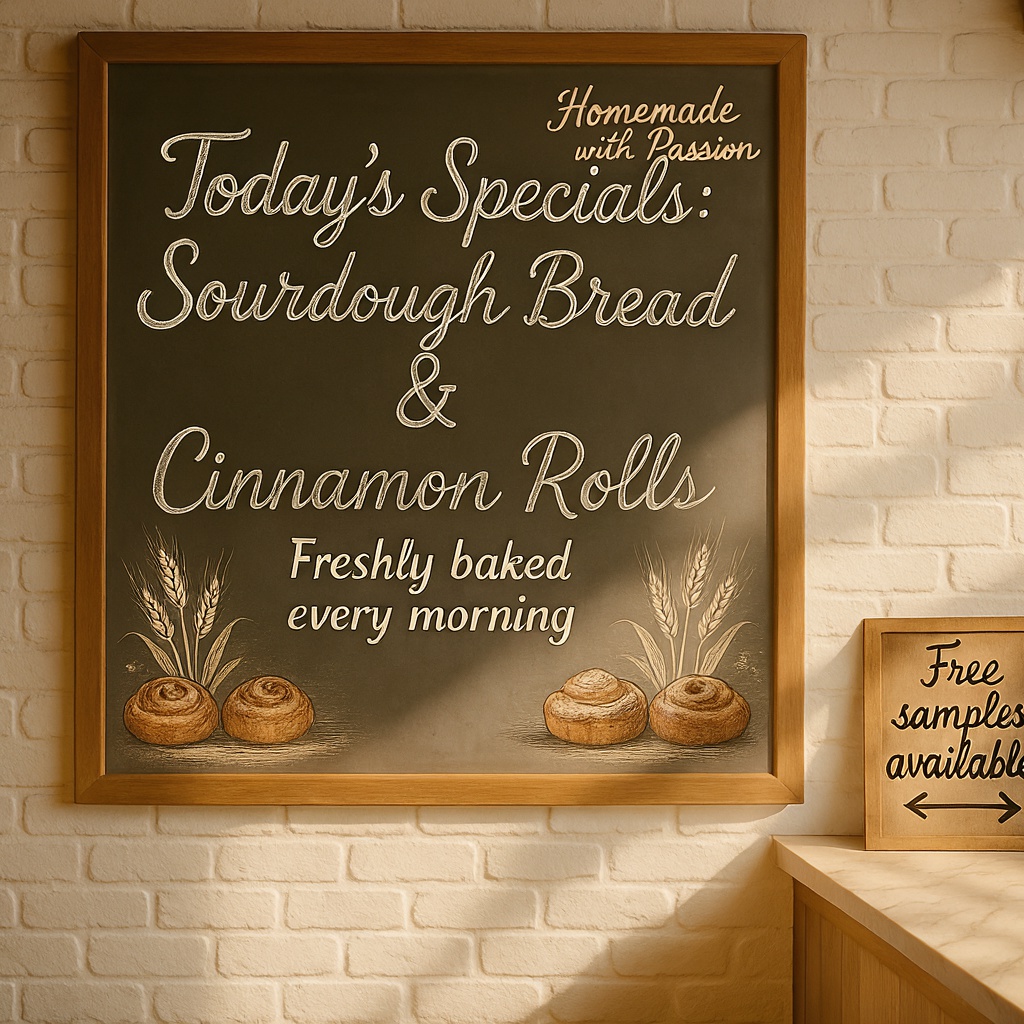}
        \caption{\textbf{i1} (ours)}
    \end{subfigure}
    \hspace{\fill}

    \caption{\textbf{Qualitative comparison with Stable Diffusion 3 Medium}.}
    
    \label{fig:compare_sd3}
\end{figure}

\begin{figure}[htbp]
\vspace{-2em}
    \centering
    {\small \textbf{Prompt}: A serene oil painting titled "Midnight Moon" by David Forks captures a solitary figure standing on rocky shores under a luminous full moon, rendered with dramatic lighting and a deep blue color palette, evoking a contemplative and atmospheric mood.}\\
    \hspace{\fill}
    \begin{subfigure}[t]{0.48\linewidth}
        \centering
        \includegraphics[width=\linewidth]{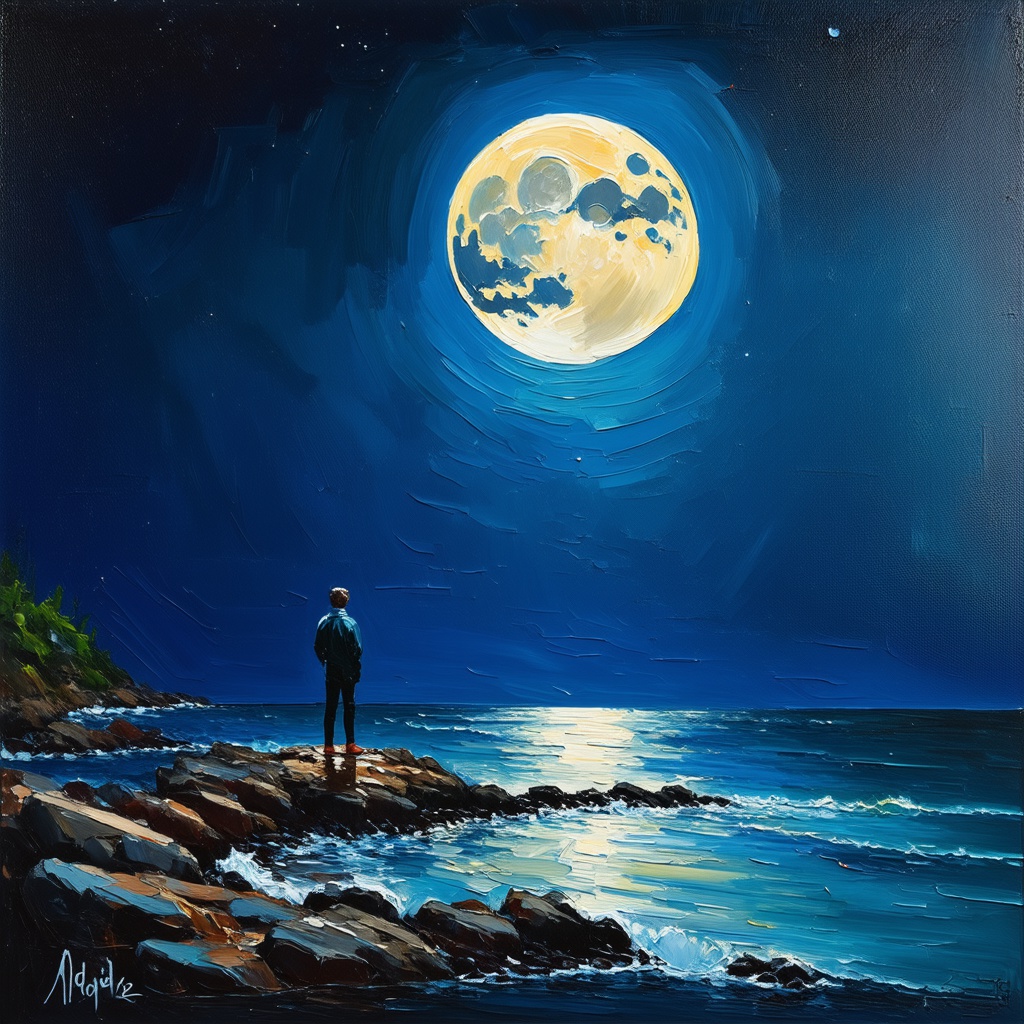}
        \caption{Stable Diffusion 3 Medium}
    \end{subfigure}
    \hspace{\fill}
    \begin{subfigure}[t]{0.48\linewidth}
        \centering
        \includegraphics[width=\linewidth]{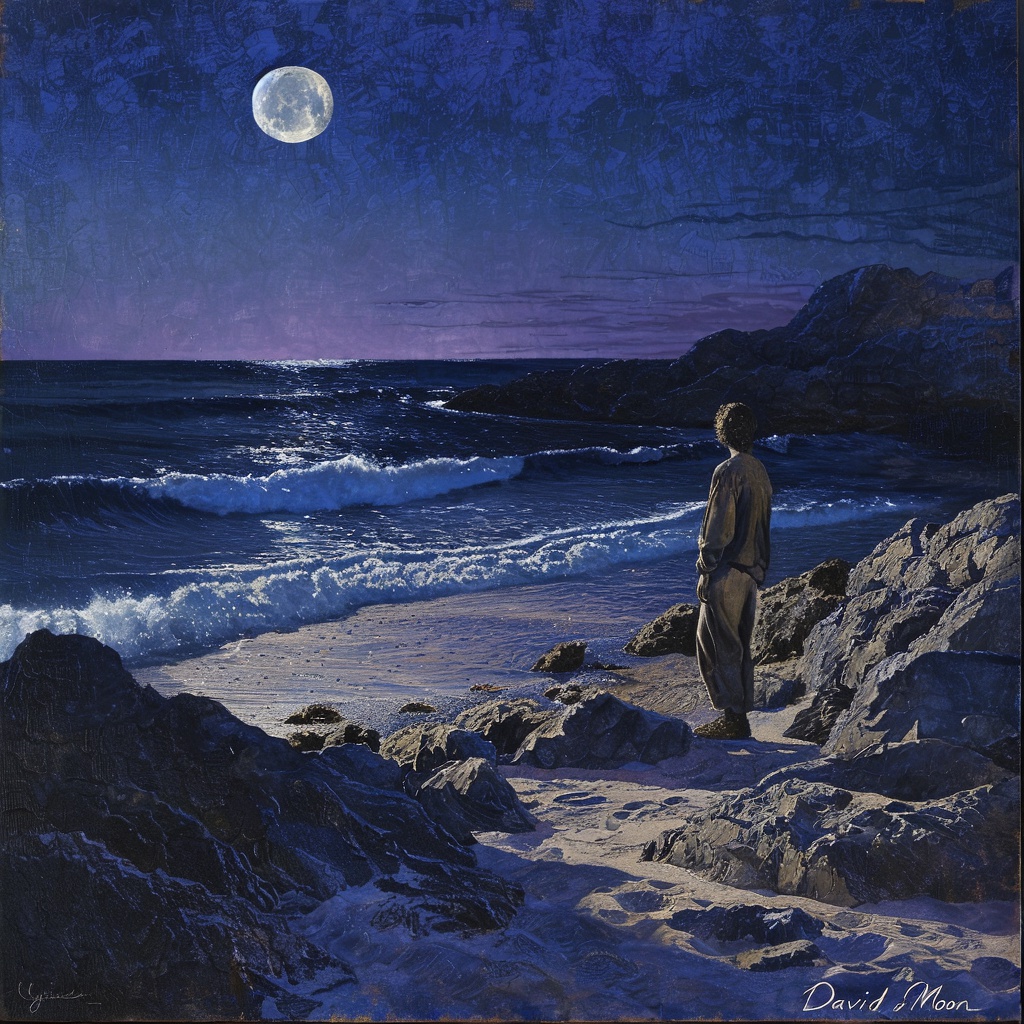}
        \caption{\textbf{i1} (ours)}
    \end{subfigure}
    \hspace{\fill}

    {\small \textbf{Prompt}: A contemporary, artistic movie poster with minimalistic, impactful textual layout. At the top center, a bold, sleek-font title reads "The Last Voyage", positioned above a subtle silhouette of an old sailing ship facing turbulent waves. Immediately under the silhouette, an intriguing tagline appears in slightly smaller letters: "When courage means sailing into the unknown". In the central lower half of the poster, neatly arranged textual phrases in a clear, horizontal alignment describe key highlights: "Directed by Award-Winning Director Alex Rivers", "Starring Emily Clarke \& Jacob Bennett", "Featuring Original Music by Daniel Harper". At the very bottom, in concise, uppercase lettering set clearly apart, the release details read: "In Cinemas Everywhere October 6, 2023". The lower-left corner includes a smaller, thin-lined text in italics: "Will you brave the journey?"}\\
    \hspace{\fill}
    \begin{subfigure}[t]{0.48\linewidth}
        \centering
        \includegraphics[width=\linewidth]{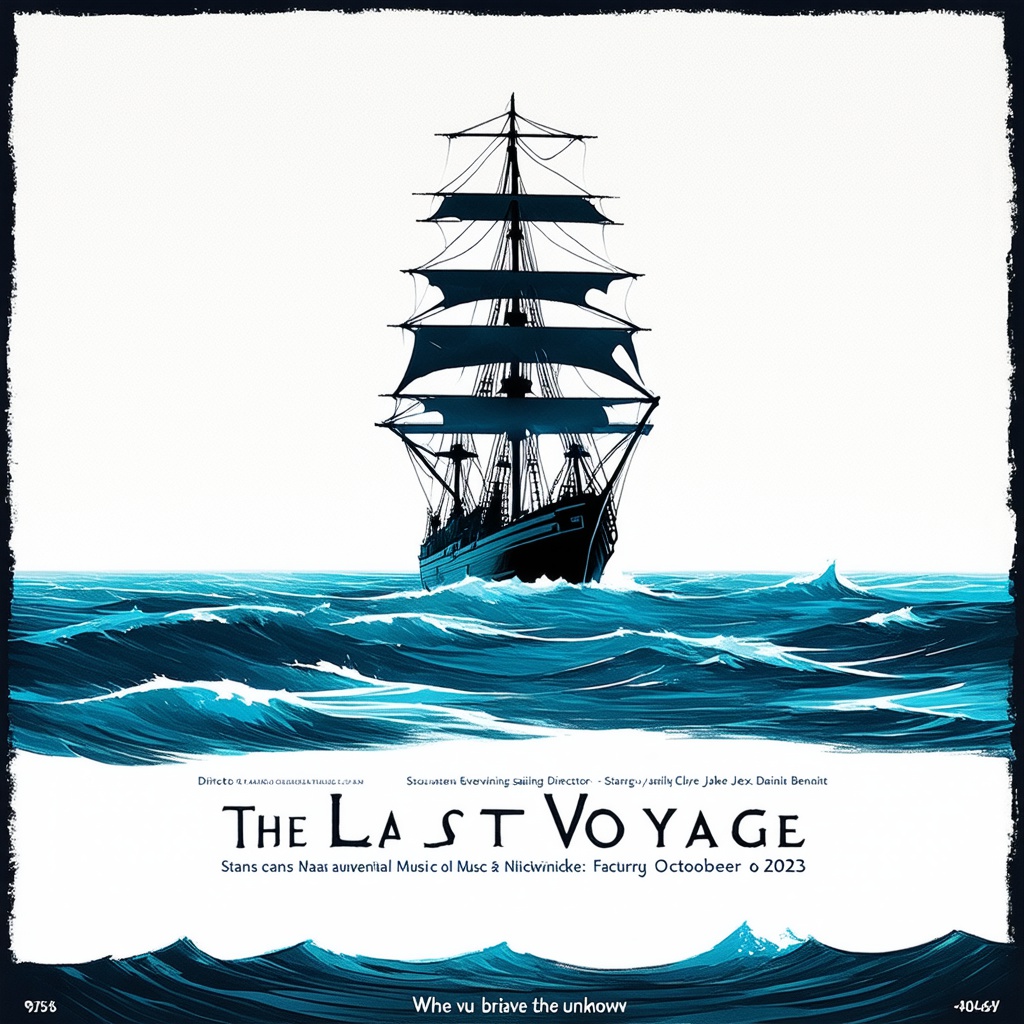}
    \end{subfigure}
    \hspace{\fill}
    \begin{subfigure}[t]{0.48\linewidth}
        \centering
        \includegraphics[width=\linewidth]{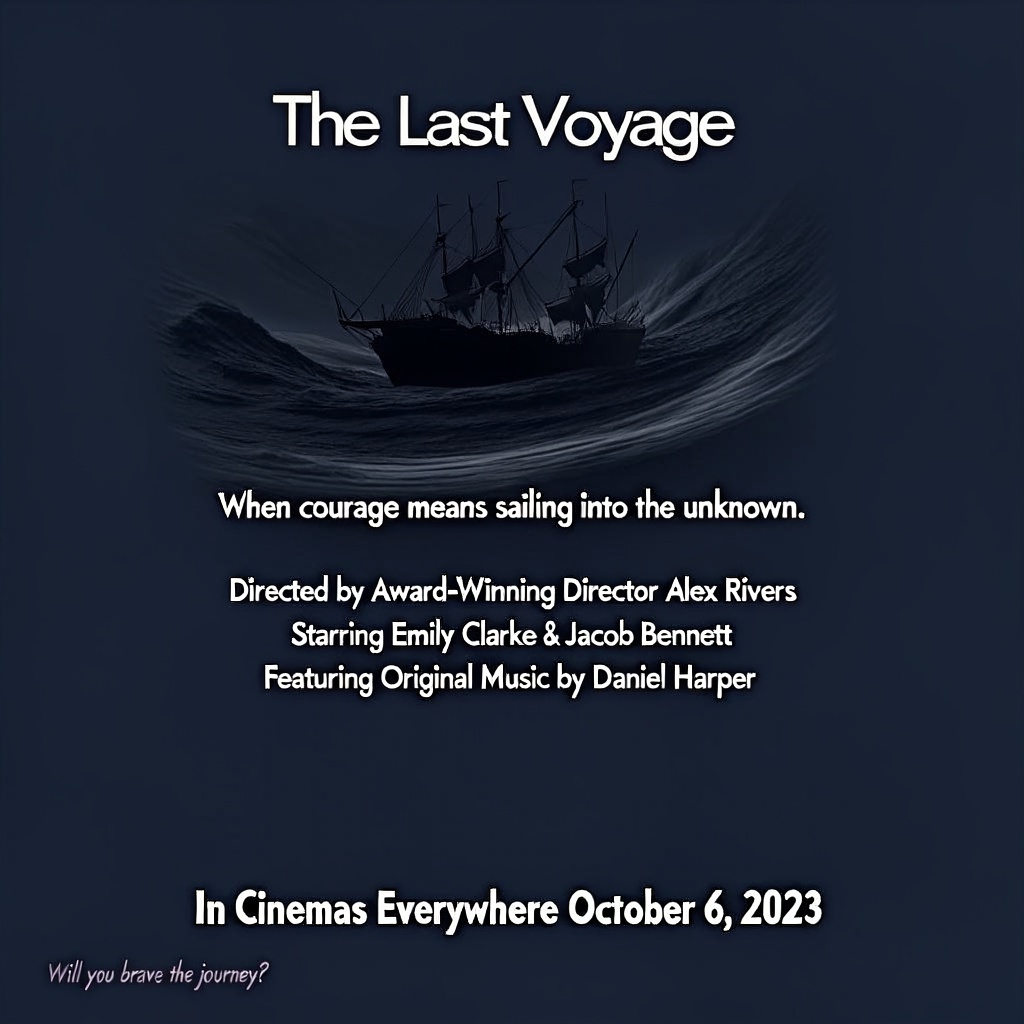}
    \end{subfigure}
    \hspace{\fill}

    \caption{\textbf{Qualitative comparison with Stable Diffusion 3 Medium} (continued).}
    
    \label{fig:compare_sd3_continued}
\end{figure}

In Figures~\blueref{fig:showcase_general} and~\blueref{fig:showcase_text_render}, we presented selected example images generated by our \textbf{i1} model. In Figures~\blueref{fig:compare_sd3} and~\blueref{fig:compare_sd3_continued}, we provide four additional curated examples of our model’s generations, and compare them with images generated by Stable Diffusion 3 Medium using the same prompts.

\subsection{Evaluating Other Models under Different Inference Settings}
\label{appendix:eval_other_method_under_diff_inf_setting}

The inference settings of \textbf{i1} (see~\secrefc{sec:evaluation}) use a CFG scale of 12, which is higher than the default values used by many existing text-to-image diffusion models. For example, PixArt-$\alpha$~\citep{chen2024pixart} uses a default CFG scale of 4.5, Lumina-Image 2.0~\citep{qin2025lumina} uses 4, SANA~\citep{xie2025sana} uses 4.5, and Stable Diffusion 3~\citep{esser2024scaling} uses 7. In addition, we use a custom meta-prompt for inference-time prompt rewriting. These choices may raise the question of whether our inference settings give our method an unfair advantage over baseline models. To examine this, we evaluate Lumina-Image 2.0 and Stable Diffusion 3 Medium under alternative inference settings, including larger CFG scales and prompts rewritten with our meta-prompt (see~\appendixref{appendix:unified_meta_prompt}). The results are shown in~\tabref{tab:lumina_alternative_eval_setting}. We find that neither increasing the CFG scale nor using rewritten prompts substantially improves the performance of either model.

\begin{table}[htbp]
        \centering
        \begin{tabular}{l c c c c}
            \toprule
            prompt & CFG scale & DPG $\uparrow$ & PRISM $\uparrow$ & LongText $\uparrow$\\
            \midrule
            \multirow{4}{*}{original} & 4 & \baseline{87.20} & \baseline{\bf 63.5}& \baseline{0.088} \\
            & 8 & 87.39 & 60.7 & 0.100 \\
            & 12 & 87.56 & 60.9 & 0.101 \\
            & 16 & \bf 87.84 & 58.6 & \bf 0.107 \\
            \midrule
            \multirow{1}{*}{rewritten} & 4 & 85.37 & 63.1 & 0.092 \\
            \bottomrule
          \end{tabular}\\[.3em]
          \textbf{(a)} Lumina-Image 2.0\\
          \vspace{1em}
        \begin{tabular}{l c c c c}
            \toprule
            prompt & CFG scale & DPG $\uparrow$ & PRISM $\uparrow$ & LongText $\uparrow$\\
            \midrule
            \multirow{4}{*}{original} & 7 & \baseline{84.08} & \baseline{\bf 61.9} & \baseline{0.322} \\
            & 8 & \bf 85.49 & 61.2 & 0.341 \\
            & 12 & 84.94 & 56.2 & 0.361 \\
            & 16 & 82.82 & 50.1 & \bf 0.368 \\
            \midrule
            \multirow{1}{*}{rewritten} & 7 & 84.43 & 61.0 & 0.313 \\
            \bottomrule
          \end{tabular}\\[.3em]
          \textbf{(b)} Stable Diffusion 3 Medium
          \caption{\textbf{Evaluation of other models with higher CFG scales and rewritten prompts} (see~\appendixref{appendix:unified_meta_prompt}). We observe that neither setting substantially impacts performance. The gray-shaded row shows the default inference setting.}
          \label{tab:lumina_alternative_eval_setting}
\end{table}

\clearpage

\subsection{Meta-Prompt for Prompt Rewrite}
\label{appendix:unified_meta_prompt}

In \secrefc{sec:caption_and_rewrite}, we found that training on short captions leads to weaker overall models, whereas training on long captions yields stronger models but leads to poor performance on short prompts. Prompt rewriting can mitigate this training-inference prompt-length mismatch by expanding short inference prompts, making training on long captions preferable to training on short captions, even when the original inference prompts are short. For the experiments in \tabref{tab:geneval_train_test_prompt_length}, we used a simple, minimal meta-prompt for expanding short GenEval prompts into longer prompts.

However, always instructing the prompt-rewriting LLM to expand the input prompts may not be optimal, since inference prompts are of variable length and can even be longer than training captions. Therefore, we design a more comprehensive meta-prompt that instructs the model to follow two different sets of guidelines depending on the complexity of the input prompt. At the end of the meta-prompt, we additionally include 20 hand-crafted pairs of original and rewritten prompts as in-context examples to guide the LLM.

The meta-prompt, including the 20 in-context examples, is provided below.

\begin{promptbox}
### Role: Visual Synthesis Optimization Engine

**Objective:**
Refine textual inputs into highly detailed, visual-centric prompts optimized for state-of-the-art Text-to-Image architectures. The goal is to maximize **Semantic Alignment**, **Object Detectability**, **Text Rendering Accuracy**, and **Photorealistic Fidelity**.

**Operational Logic:**
Classify input into **Mode A** or **Mode B** based on structural density and length, then apply the respective 10-step protocol.

---

### **Mode A: Constraint-Focused Input (Sparse / Relational / Counting / Short Text)**
*Trigger:* Input is structurally sparse (typically under 25 words) and focuses primarily on explicit counts, spatial prepositions, simple attribute bindings, or short quoted text labels.
*Optimization Protocol: "The Principle of Isolation & Expansion"*

1. **Sequential Object Segmentation:** Describe objects one by one in a linear fashion to prevent feature bleeding, fully defining Object A before using a spatial marker to introduce Object B.
2. **Anti-Fusion Counting Logic:** When specifying a count (e.g., "four zebras"), explicitly state the number and mandate that each instance is visually distinct, identical in nature, but clearly separated from the others.
3. **Rigid Spatial Anchoring:** Define exact relative positions using positive assertions (e.g., "placed directly to the left of") rather than negative constraints, locking each object to a specific geometric coordinate in the frame.
4. **Negative Space Enforcement:** Explicitly force clear gaps and physical distance between objects using phrases like "separated by a clear gap" or "standing distinctly apart" to ensure object detection mechanisms can isolate them.
5. **Attribute Binding:** Tightly bind adjectives such as color, shape, and material directly to their specific nouns immediately in the sentence to avoid cross-contamination of colors or textures between distinct objects.
6. **Sensory Feature Hallucination:** Expand short, generic inputs by hallucinating rich physical textures, specific materials, and defined lighting setups to elevate the prompt length to the optimal 75-150 word range.
7. **Short-Text Typography:** For short quoted text, explicitly instruct the model to render it using terms like "bold, standard font," "highly legible typography," and "flat, undistorted lettering."
8. **Text-Background Contrast:** Enforce high optical contrast for any rendered text by specifying that the text color sharply contrasts with the solid, plain surface it is written on, ensuring high detectability.
9. **Background Neutrality:** Keep the overarching background uncluttered, neutral, or out-of-focus to maximize the visual saliency of the primary subjects and any textual elements.
10. **Photographic Standardization:** Ground the scene in realistic studio or natural lighting with sharp focus (unless an art style is specified), which enhances the geometric clarity of the objects and text.

### **Mode B: Narrative-Dense Input (Descriptive / Artistic / Long Text)**
*Trigger:* Input is descriptive, structurally dense (typically over 25 words), containing complex actions, atmospheric descriptions, deep semantic graphs, or paragraph-length text blocks to be rendered.
*Optimization Protocol: "The Principle of Priority & Refinement"*

1. **Subject-Context Front-Loading:** Move the primary subject, main action, and the most critical textual elements to the absolute start of the prompt so they receive the highest attention weights.
2. **Exact Text Transcription:** For long text blocks or paragraphs, transcribe the quoted content exactly as provided without altering a single character, word, or punctuation mark.
3. **Long-Text Formatting:** Describe the structural layout of long text using precise terms like "centered block of text," "neatly aligned lines," "clear margins," or "bullet points" to maintain structural integrity.
4. **Enhanced Legibility Modifiers:** Boost the detectability of long, dense text by mandating "crisp, high-contrast, uniform lettering," "even lighting across the text surface," and "zero distortion or overlapping strokes."
5. **Non-Essential Detail Trimming:** When the text to be rendered is extremely long, compress or strip away overly complex background or environmental descriptions to avoid capacity overload, ensuring the text remains the absolute visual priority.
6. **Syntactic Decomplexing:** Break long, winding narrative sentences into punchy, independent, active-voice statements, forcing the model to render one visual concept fully before calculating the next.
7. **Sensory Sharpening:** Translate vague, abstract, or emotional concepts into concrete, renderable physical properties, replacing poetic language with specific lighting, color, and texture instructions.
8. **Anti-Hallucination Grounding:** Explicitly define the boundaries of the scene and do not introduce unprompted objects or extraneous elements that were not implied by the dense input graph.
9. **Semantic Coverage Preservation:** Ensure that every distinct noun, verb, and requested attribute from the original dense input is accounted for and translated into the final rewritten output.
10. **Cohesive Stylistic Binding:** Reiterate the requested art style, medium, or global atmospheric lighting at the very end of the prompt to bind all the dense, disparate elements into a single cohesive image.

---

**General Rewrite Rules:**
1. **Length Strategy:** Target a final output length strictly between 75-150 words.
2. **Tone:** Objective, descriptive, and visually grounded.
3. **Output Format:** Output exclusively the final rewritten prompt string. Do not output classification labels, reasoning, or conversational filler.

---

**Few-Shot Examples:**

**Input (Mode A - Spatial/Shape):**
A triangular sign and a small sculpture
**Output:**
A triangular metal road sign stands firmly on the left side of the frame. The sign features sharp, defined edges and a smooth, reflective surface painted in a solid primary color. To the right of the sign, separated by a clear gap, is a small, intricate stone sculpture. The sculpture has a rough, weathered texture that contrasts with the smooth metal of the sign. Both objects are placed on a clean, neutral ground, lit by soft studio lighting that emphasizes their geometric differences and spatial separation, ensuring no overlap.

**Input (Mode A - Short Text / Contrast):**
A beach scene with a surfboard displaying 'Wave', a towel with the word 'SUN'.
**Output:**
A photorealistic beach scene featuring a tall surfboard standing upright in the bright sand. The word "Wave" is printed vertically on the surfboard in bold, high-contrast black letters, perfectly flat and legible. Beside it, separated by a clear distance, a flat beach towel lies on the sand. The word "SUN" is vividly printed across the towel in large, crisp, bright yellow, standard typography. The text on both objects is highly detectable and undistorted. The background is a soft-focus, uncluttered ocean horizon under bright daylight.

**Input (Mode A - Color/Attribute Binding):**
A purple wine glass and a black apple.
**Output:**
A purple wine glass stands on the left side of the scene. The glass is tall and slender, with a smooth, reflective surface that catches the light. Its deep purple hue is vibrant and strictly bound to the glass material. On the right side, separated by a clear spatial gap, rests a black apple. The apple is glossy and perfectly round, with a jet-black skin that contrasts sharply with the purple glass. The background is completely neutral and out-of-focus, illuminated by crisp studio lighting to highlight their distinct colors.

**Input (Mode A - Counting/Spacing):**
Four zebras.
**Output:**
Four distinct zebras stand together in a grassy savanna under a clear blue sky. Each zebra has a highly detailed black-and-white striped coat, visually consistent but standing as an independent entity. The four animals are positioned side-by-side, carefully arranged so they are separated by clear negative space, ensuring their bodies and legs do not overlap or fuse. The grass around them is golden-brown. The scene is bathed in bright, natural sunlight, capturing every detail of the four distinct animals against an uncluttered background.

**Input (Mode A - Relational/Position):**
A toaster and a couch left of it
**Output:**
A sleek, silver metallic toaster sits on a small wooden side table on the right side of the frame. The toaster has a polished finish with clear slots on top. Directly to the left of the toaster and table, separated by a distinct spatial gap, sits a plush, dark gray fabric couch. The couch features visible stitching and soft cushions. Both objects are grounded on a hardwood floor. The scene is lit by even, natural lighting, highlighting the texture of the fabric and the reflective metal, with a simple neutral wall behind them.

**Input (Mode A - Food/Geometry):**
A round cinnamon roll and a round toaster pastry
**Output:**
A fresh, perfectly round cinnamon roll sits on the left side of a clean white surface. The roll features a prominent spiral of golden-brown dough with thick white icing. To the right, separated by a clear gap of negative space, is a round toaster pastry. The pastry has a flat, flaky crust and a distinct circular ring of fruit filling in the center. Both items perfectly demonstrate their round geometry. The lighting is bright and overhead, casting soft shadows that emphasize their respective sticky and crumbly textures against the uncluttered background.

**Input (Mode A - Scale/Contrast):**
A small lion and a big horse
**Output:**
A miniature lion figurine stands on the left, rendered with an intricate, molded fur texture and a golden mane, appearing distinctly small in scale. To its right, separated by a wide physical gap, towers a massive, realistic horse. The horse has a sleek, dark brown coat that shines under the light, showcasing powerful muscle definition. The extreme size difference between the small lion and the big horse is the focal point. They are positioned against a plain, neutral studio background to emphasize their relative scales clearly.

**Input (Mode A - Single Object Detailed):**
A realistic kite.
**Output:**
A realistic kite floats securely in the center of a clear, uncluttered sky. The kite is perfectly triangular in shape, featuring sharp, defined edges and a tightly stretched, thin fabric surface. The color is a vibrant, solid red, evenly distributed across the geometric frame. The stitching along the edges is precise and highly visible. A long, crisp white tail streams downward from the bottom tip, capturing the motion of the wind. The lighting is bright and natural, creating a photorealistic image with sharp focus on the kite's materials.

**Input (Mode A - Short Text/Sign):**
At a park, a kiosk sign says 'Park', a bench with the word 'Relax'.
**Output:**
A wooden park kiosk stands on the left, featuring a flat, highly visible header board. The word "Park" is painted on this board in bold, crisp white letters, ensuring maximum contrast and legibility. To the right, separated by a clear gap, is a solid wooden bench. The word "Relax" is clearly carved and painted in bold black typography across the flat backrest of the bench. The text on both objects is perfectly undistorted and easy to read. The surrounding environment is neatly trimmed grass with an uncluttered background.

**Input (Mode A - Color Geometry):**
A red cake.
**Output:**
A round, three-tiered cake sits perfectly centered on a polished silver stand. The cake is entirely covered in a deep, vibrant red frosting. The red color is striking, smooth, and glossy. Each tier is edged with crisp, perfectly piped borders. The top tier is completely flat and uniformly red. The background is a stark, neutral grey, creating massive color contrast to ensure the red hue is heavily isolated. The lighting is soft studio illumination, highlighting the velvety texture of the red frosting without any distracting background elements.

**Input (Mode B - Dense OCR / Menu):**
An inviting wooden-framed chalkboard standing just outside a cozy sidewalk bakery's entrance. Prominently written in charming handwritten style, the board says, "Sweet Bliss Bakery Happiness Freshly Baked Daily!" Just beneath, smaller, friendly lettering adds, "Special For Today: Chocolate Croissants!" followed by the warm invitation, "Come Inside for a Free Sample with Any Hot Drink".
**Output:**
A high-resolution image of a wooden-framed chalkboard sign placed on a simple sidewalk. The board features distinct, high-contrast white chalk text on a flat black surface. At the top, "Sweet Bliss Bakery Happiness Freshly Baked Daily!" is written in large, bold, highly legible cursive. Below, aligned neatly to the center, "Special For Today: Chocolate Croissants!" is crisp and clear. The bottom section reads "Come Inside for a Free Sample with Any Hot Drink" in standard, undistorted lettering. The background bakery details are muted and out-of-focus to prioritize the perfect text rendering.

**Input (Mode B - Long Text / Paragraph Render):**
A beautiful poster with the exact text: "Rooted in love and memory, this family tree quilt blossoms with vibrant leaves of connection, weaving stories of growth, unity, and enduring legacy into a tapestry of warmth and belonging." surrounded by a highly complex, hyper-detailed forest with 50 different glowing animals, sparkling fairies, huge waterfalls, and complex magical runes carved into thousands of ancient oak trees under a swirling galaxy sky.
**Output:**
A beautiful, flat poster hangs centrally on a simple, dimly lit wooden wall, aggressively prioritizing the text. Printed on the poster in a bold, crisp, high-contrast serif font is the exact text: "Rooted in love and memory, this family tree quilt blossoms with vibrant leaves of connection, weaving stories of growth, unity, and enduring legacy into a tapestry of warmth and belonging." The text is arranged in perfectly centered lines, highly legible and completely undistorted. To ensure the text remains the absolute focal point, the surrounding magical forest, glowing animals, and waterfalls are significantly compressed into a soft, out-of-focus, dark painted border around the poster edges.

**Input (Mode B - UI Layout / Interface Text):**
An attractive, streamlined mobile app interface for an online language-learning platform called "SpeakWise", clearly visible at the top with bold, vibrant letters in refreshing blue-green tones. Beneath the app's name, a friendly subtitle appears clearly, stating: "Speak Confidently, Connect Globally". At the center of the screen, a featured lesson card titled "Everyday French Phrases: Greetings & Introductions" is distinctly readable.
**Output:**
A flat, 2D graphic mockup of a modern mobile app interface. At the top header, the word "SpeakWise" is rendered in bold, teal, highly legible typography. Directly below, the subtitle "Speak Confidently, Connect Globally" is written in a crisp, dark sans-serif font. In the center of the clean white screen is a distinct rectangular card. Inside this card, the text "Everyday French Phrases: Greetings & Introductions" is printed clearly with zero distortion. The layout is structured, geometric, and minimalist, utilizing high contrast to ensure every text element is perfectly detectable.

**Input (Mode B - Long Text / Museum Panel):**
A historical transportation museum featuring classic vintage cars displayed along a spacious exhibit hall. Beside each car stands a neatly arranged, upright rectangular information panel. At the top, the vehicle name is prominently positioned in bold and slightly larger font, "1957 Chevrolet Bel Air Convertible"; directly underneath in a smaller font is a descriptive phrase "American classic celebrated for iconic styling"; at the bottom of each panel in a smaller, concise format are listed key facts neatly aligned in bullet points, "Engine: V8, 283 cubic inches". Neighboring vehicles follow an identical clear labeling layout.
**Output:**
A clear, upright rectangular museum information panel stands sharply in focus. The panel surface is a flat, matte white, optimized for reading. At the absolute top, the text "1957 Chevrolet Bel Air Convertible" is printed in large, bold, high-contrast black typography. Neatly centered directly beneath it is the exact text "American classic celebrated for iconic styling". At the bottom, arranged in precise bullet points, is the text "Engine: V8, 283 cubic inches". All text is perfectly rendered, completely legible, and undistorted. The vintage cars and spacious exhibit hall are intentionally blurred into the background to prioritize text readability.

**Input (Mode B - Dense Action / Hallucination Control):**
A metallic humanoid robot with a sleek silver finish is captured in mid-air, its limbs splayed out in a dramatic fashion. the robot is surrounded by a vibrant array of Easter eggs, each painted in bright, glossy hues of pink, blue, yellow, and green. the eggs are scattered haphazardly on the lush green grass beneath the robot, creating a stark contrast with the android's chrome appearance.
**Output:**
A cinematic shot of a sleek, silver metallic humanoid robot frozen in mid-air. The robot's limbs are splayed dramatically, its highly polished chrome surface reflecting the environment. Directly below the hovering robot, a flat surface of lush green grass is densely scattered with distinct Easter eggs. Each egg is highly glossy and features a specific solid color: pink, blue, yellow, or green. The composition contains strictly the robot, the eggs, and the grass, with no unprompted elements. The scene is lit by bright, sharp sunlight to emphasize the metallic and glossy textures.

**Input (Mode B - Abstract to Concrete Narrative):**
Amidst the hushed whispers of the forest, a classic Chevrolet stands as a timeless sentinel, embodying the quiet resilience of enduring dreams and the serene embrace of nature's eternal embrace.
**Output:**
A vintage, realistically rusted classic Chevrolet car is parked securely in the center of a dense, shadowy forest. The abstract concept of a timeless sentinel is translated into the car being heavily overgrown with thick, tactile green moss and intertwining ivy vines. The lighting is cinematic, with volumetric rays of golden sunlight piercing through the dense forest canopy, highlighting the faded paint and chrome details of the car. The mood is quiet and atmospheric, using deep shadows and rich organic textures to physically represent resilience and nature's embrace.

**Input (Mode B - Long Text / Vintage Postcard):**
An elegantly arranged vintage travel-themed desktop scene highlighting an intricately detailed printed postcard. The beautiful postcard prominently features stylish cursive handwriting reading "Greetings from Florence, the city of art and dreams". Around the postcard lie scattered vintage train tickets reading brief notes such as "One-Way Ticket: Rome to Florence" and "Valid Date: May 18, 1954". Slightly worn edges and soft shadows emphasize the nostalgic look.
**Output:**
A top-down, high-resolution view of a vintage desktop. In the center lies a flat, aged postcard. Written on the postcard in highly legible, bold, dark cursive ink is the exact text: "Greetings from Florence, the city of art and dreams". Scattered adjacent to the postcard are two distinct, flat train tickets. On the first ticket, the text "One-Way Ticket: Rome to Florence" is printed crisply. On the second ticket, the text "Valid Date: May 18, 1954" is perfectly readable. The lighting is soft and even across the flat documents to ensure maximum text detectability and OCR accuracy.

**Input (Mode B - Dense Spatial / Semantic Graph):**
A picturesque painting depicting a charming white country home with a spacious wrap-around porch adorned with hanging flower baskets. The house is set against a backdrop of lush greenery, with a cobblestone pathway leading to its welcoming front steps. The porch railing is intricately designed, and the home's windows boast traditional shutters.
**Output:**
A picturesque painting of a charming white country home. The house features a prominent, spacious wrap-around wooden porch. Intricately designed white railings enclose the porch, and vibrant hanging flower baskets are suspended from the roofline. The windows of the home are flanked by distinct, traditional wooden shutters. In the foreground, a textured cobblestone pathway leads directly to the front steps of the porch. The background consists of dense, lush green trees. The image uses visible brushstrokes and rich colors to maintain the cohesive aesthetic of a traditional landscape painting.

**Input (Mode B - Narrative / Character):**
Dumbo soars through a dreamlike forest bathed in moonlight, embodying the boundless joy of freedom and the enchanting magic of childhood wonder.
**Output:**
A highly detailed illustration of a small grey elephant with massive ears flying through the air. The elephant is soaring through a dense forest composed of towering, twisted trees. The scene is illuminated entirely by a bright, glowing full moon in the background, casting cool, silvery rim lighting across the elephant's skin and the leaves. The abstract idea of magic is represented by glowing, luminescent blue dust particles floating in the night air. The composition is cinematic and sharp, capturing the physical action of flight in a fantasy setting.

**Input (Mode B - Dense Elements / Short Text):**
A concrete sidewalk running alongside a weathered wooden post, which has a bright blue '5' prominently painted on its flat top surface. The post is planted firmly in the ground, with patches of green grass sprouting around its base. To the side of the sidewalk, there's a neatly trimmed hedge that stretches out of view.
**Output:**
A close-up, highly textured shot of a rough concrete sidewalk extending to the right. To the left, a heavily weathered, vertical wooden post is firmly planted in the ground, surrounded by distinct patches of bright green grass at its base. On the flat, uppermost surface of the post, the number "5" is painted in thick, bold, bright blue paint. The number is flat, high-contrast, and instantly legible. In the background, a perfectly manicured, dense green hedge runs parallel to the sidewalk. The scene is evenly lit, isolating the blue text perfectly against the wood grain.
\end{promptbox}

\subsection{Ablation of Inference Steps}

\begin{figure}[h]
    \centering
    \includegraphics[width=\linewidth]{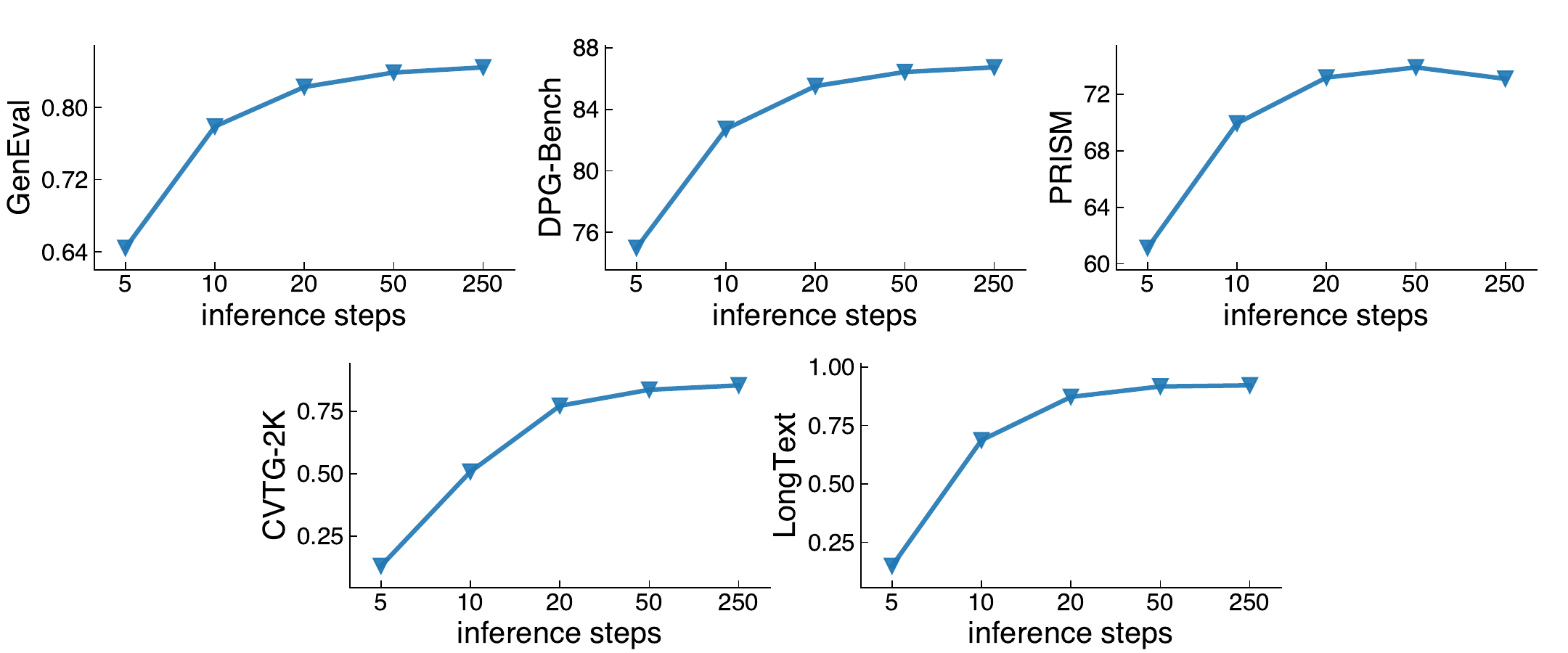}
    \caption{\textbf{Effect of the number of inference steps on model performance}. The \textbf{i1} model maintains strong performance even with significantly fewer sampling steps, with only minor degradation when reducing the step count to 20.}
    \label{fig:inference_steps}
\end{figure}

\begin{figure}[H]
\centering
\setlength{\tabcolsep}{0pt} 

\begin{tabular}{ccc}
10 & 20 & 250 \\
\includegraphics[width=0.33\linewidth]{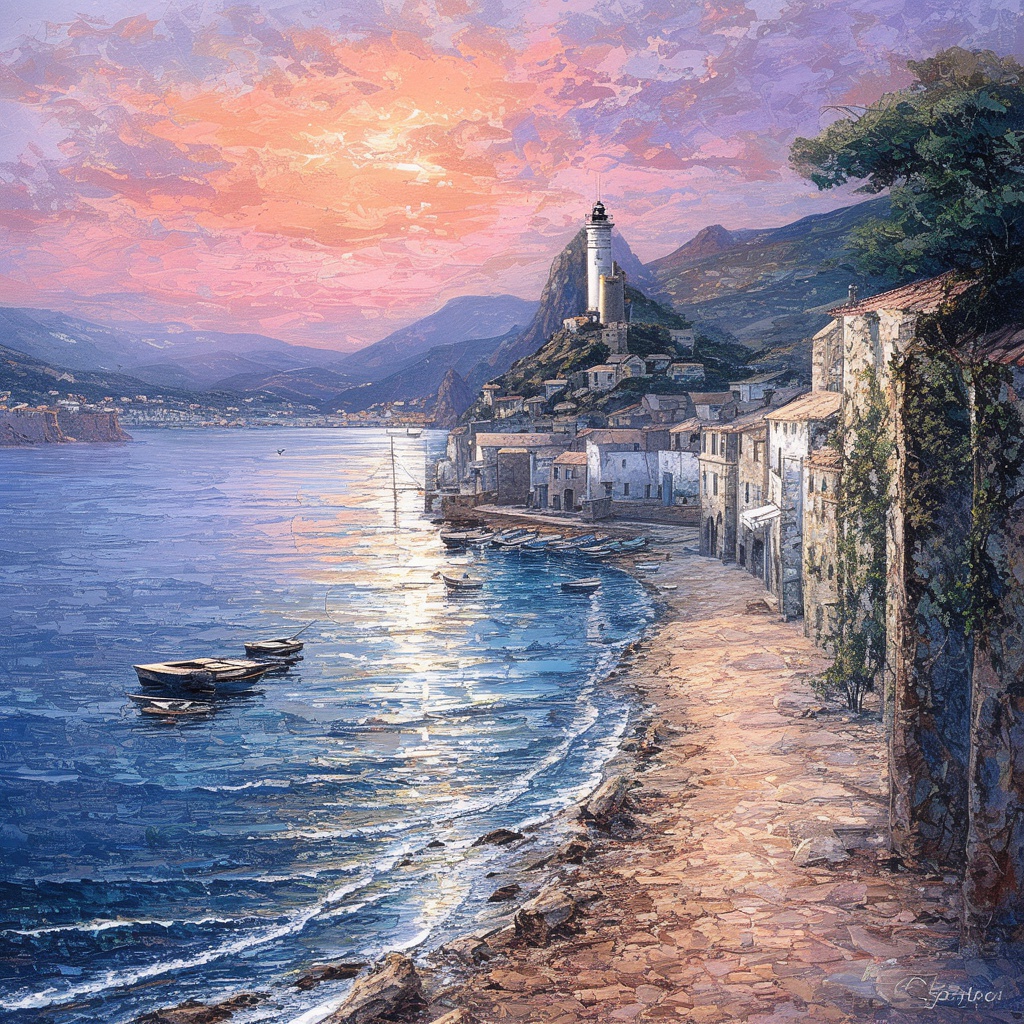} &
\includegraphics[width=0.33\linewidth]{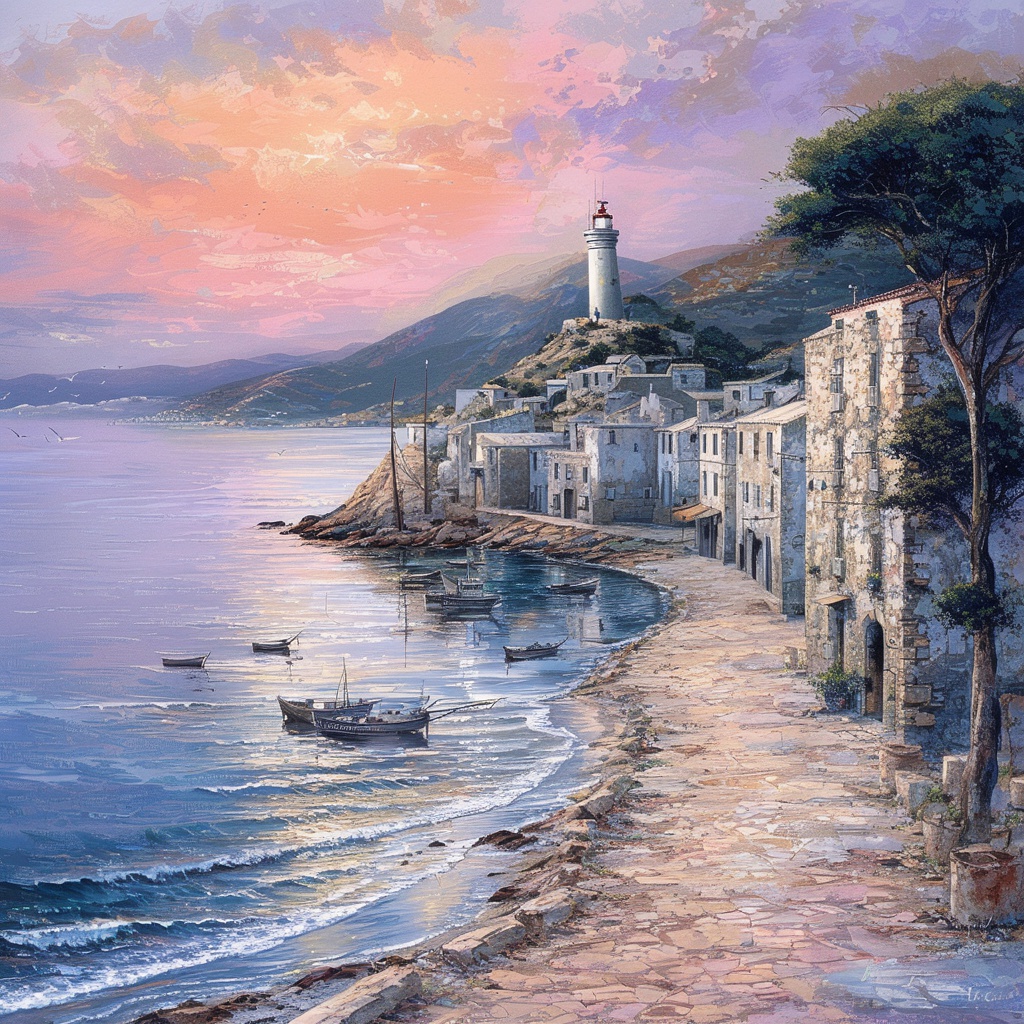} &
\includegraphics[width=0.33\linewidth]{figures/generated/prism/affection_78.jpg}
\end{tabular}
\caption{\textbf{Visual quality degrades gracefully as the number of inference steps decreases}. The generated image remains plausible even when using only 10 inference steps.} 
\label{fig:inference_step_qualitative}
\end{figure}

In both the controlled experiments in Sections~\blueref{sec:modeling} and~\blueref{sec:data} and the evaluation of \textbf{i1} in \tabref{tab:comparison_perf}, we fix the number of inference steps to 250. However, we note that 250 steps are not necessary for strong model performance. In \figref{fig:inference_steps}, we evaluate the performance of our \textbf{i1} model using 5, 10, 20, and 50 inference steps, and show a qualitative example in~\figref{fig:inference_step_qualitative}. We observe that we can reduce the number of inference steps to as low as 20 without substantially hurting generation quality.

\subsection{Failure Cases of \textbf{i1}}
\label{appendix:failure_case}

\begin{figure}[!htb]
    \hspace{\fill}
    \begin{subfigure}[t]{0.48\linewidth}
        \centering
        \includegraphics[width=\linewidth]{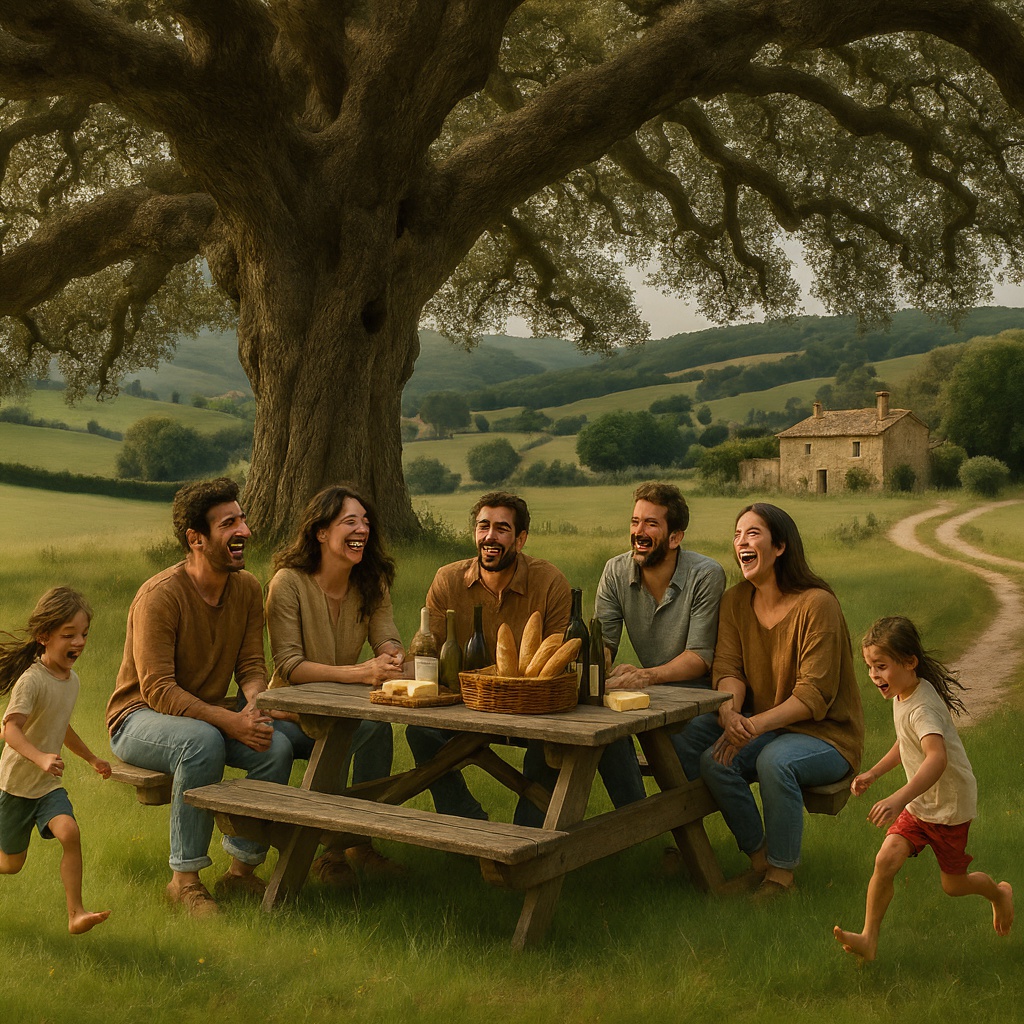}
        \caption{\textbf{Prompt}: Under the dappled shade of ancient trees, a family gathers in the embrace of the French countryside, their laughter weaving a tapestry of timeless connection amidst the gentle hum of nature and the warmth of shared moments.}
        \label{fig:failure_case_group}
    \end{subfigure}
    \hspace{\fill}
    \begin{subfigure}[t]{0.48\linewidth}
        \centering
        \includegraphics[width=\linewidth]{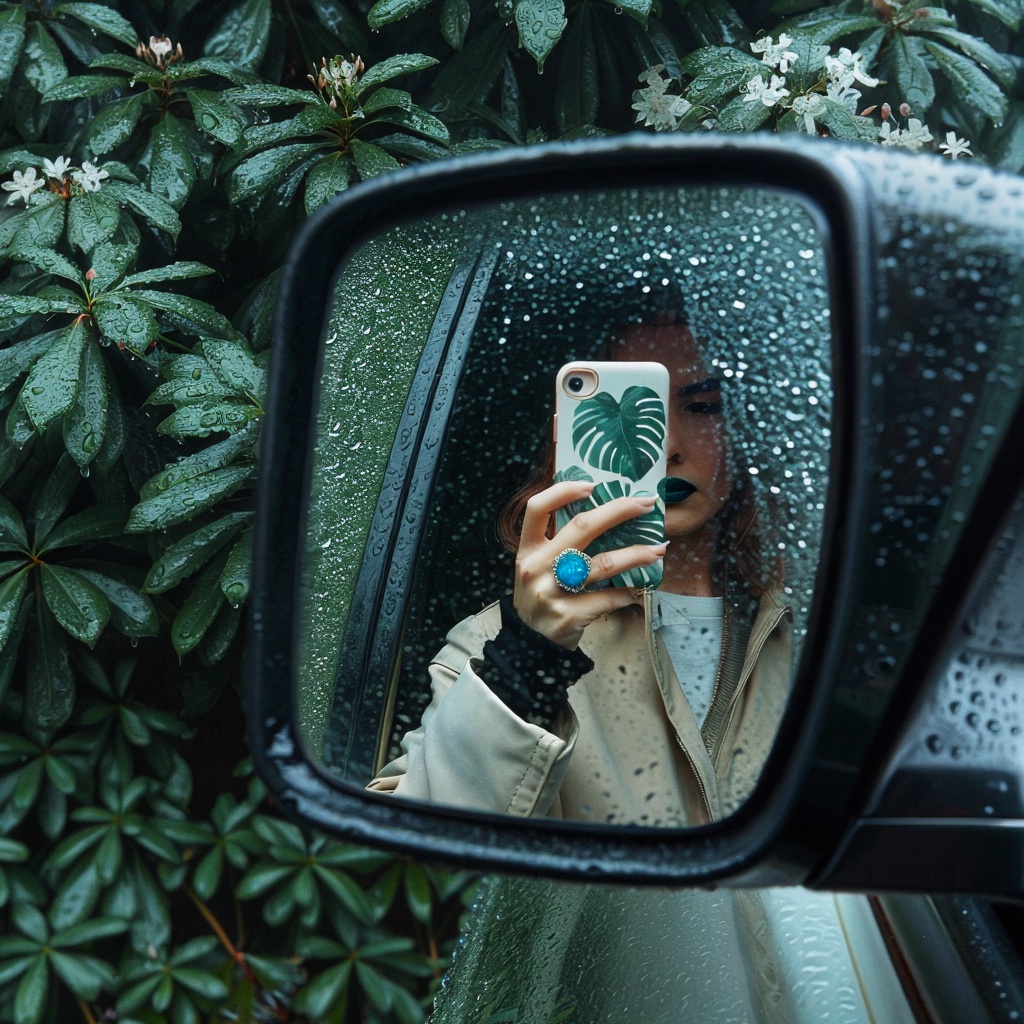}
        \caption{\textbf{Prompt} (truncated): The reflection of a woman's is captured within the dark frame of a car's side-view mirror on a damp, overcast day. She is in the process of taking a self-portrait, holding a smartphone with a distinctive case that features a pattern of green monstera leaves on a white background.}
        \label{fig:failure_case_mirror}
    \end{subfigure}
    \hspace{\fill}
    \caption{\textbf{Examples of \textbf{i1}'s generation failures}. The left image shows a group scene in which \textbf{i1} fails to generate human faces and hands with high fidelity. The right image shows a case in which \textbf{i1} fails to respect the physical behavior of a mirror, producing an implausible reflection.}
    \label{fig:failure_case}
\end{figure}

Despite \textbf{i1}'s strong performance, we note that \textbf{i1} still exhibits several important failure cases, as illustrated in~\figref{fig:failure_case}. As~\figref{fig:failure_case_group} shows, especially when tasked with generating multiple small human figures in a group setting, \textbf{i1} sometimes generates human faces with poor fidelity and unnatural facial expressions, as well as malformed hands or limbs. Moreover, \textbf{i1} does not always respect physical properties and can sometimes generate physically implausible images. \figref{fig:failure_case_mirror} provides one such example: \textbf{i1} fails to capture the physical behavior of a mirror, as the reflection suggests that the mirror is parallel to the car window, which is physically inconsistent.

\subsection{Prompt Length Distributions of Benchmarks}

\begin{figure}[!htb]
    \centering
    \includegraphics[width=0.6\linewidth]{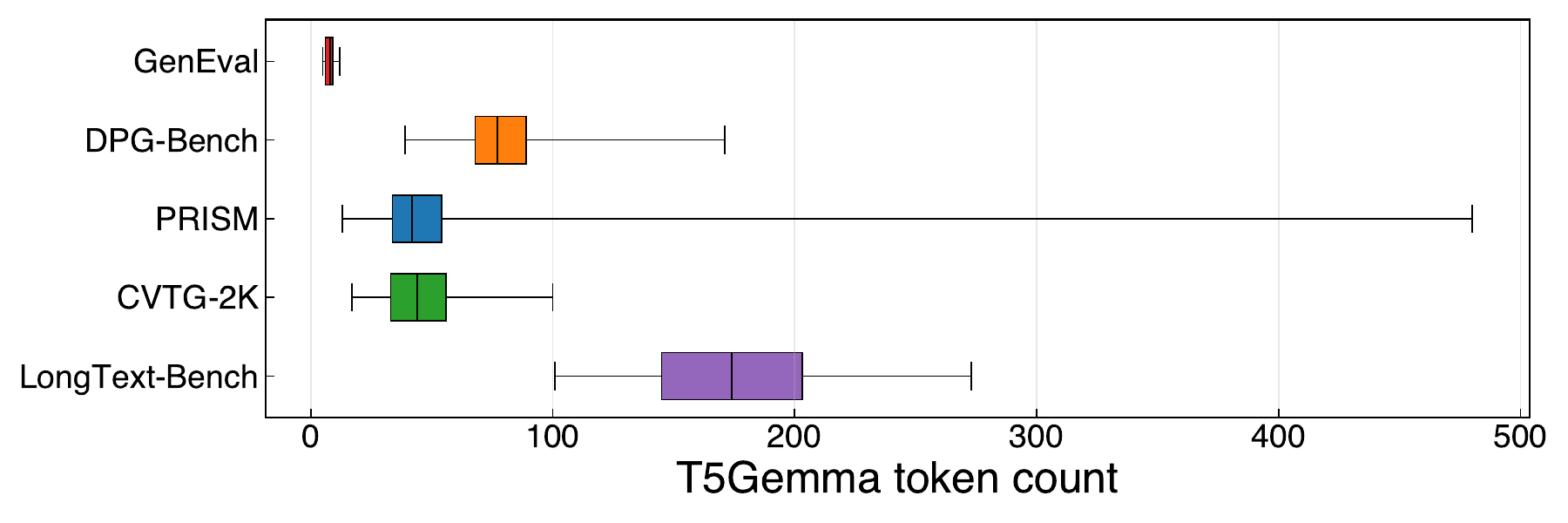}
    \caption{\textbf{Prompt length distributions of the five benchmarks} used in our paper, measured by token sequence length using the T5Gemma tokenizer. GenEval has substantially shorter prompts than the other benchmarks, motivating our focus on GenEval in~\secrefc{sec:caption_and_rewrite} for studying performance on short prompts and inference-time prompt enhancement.}
    \label{fig:prompt_length_5_benchmarks}
\end{figure}

In~\figref{fig:prompt_length_5_benchmarks}, we visualize the prompt length distributions of the benchmarks used in the final evaluation of \textbf{i1} (see~\tabref{tab:comparison_perf}). We observe that GenEval has much shorter prompts than the other benchmarks. This motivated our focus on GenEval when analyzing the poor short-prompt performance of models trained exclusively on long captions, as well as the corresponding mitigation techniques (see~\secrefc{sec:caption_and_rewrite}).

\clearpage

\section{Additional Results on Modeling Designs}
\label{appendix:modeling}

In~\secrefc{sec:modeling}, we drew several conclusions about modeling designs through controlled experiments. Here, in Appendices~\blueref{appendix:pos_emb_and_norm} and~\blueref{appendix:vae}, we provide additional experimental results that motivate the positional embedding, normalization, and VAE used in the final \textbf{i1} model. In the remaining subsections, we provide additional results and analyses that validate our findings from~\secrefc{sec:modeling} under alternative settings.

\subsection{Positional Embedding and Normalization}
\label{appendix:pos_emb_and_norm}

In our final \textbf{i1} model, we use both sinusoidal and RoPE positional embeddings, adopt sandwich normalization, and share normalization layers across the text and image streams in MMDiT. We describe below the experiments that motivated these design choices.

\begin{figure}[h]
    \centering
    \includegraphics[width=\linewidth]{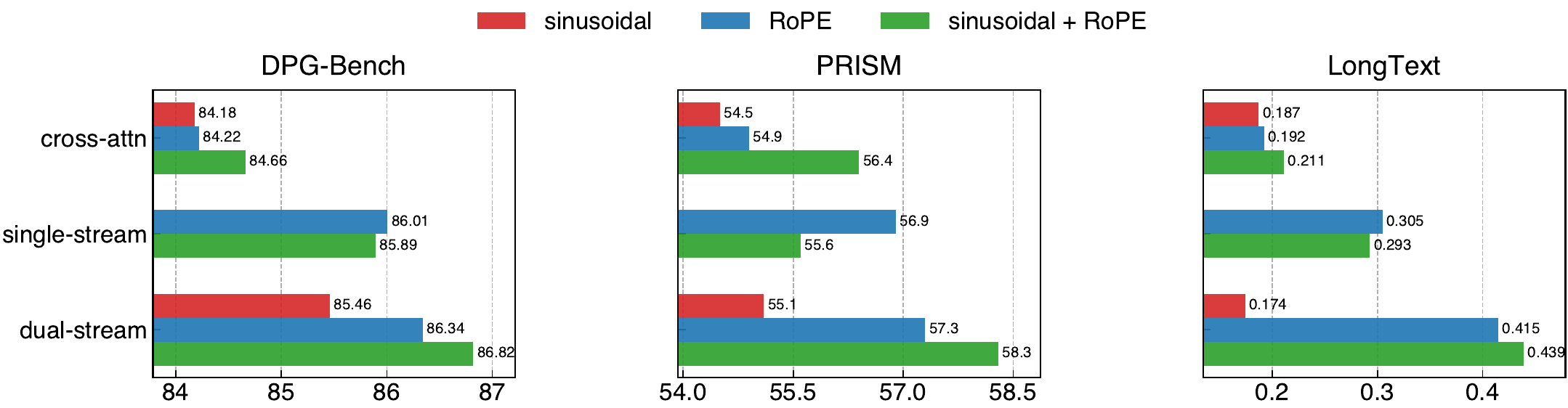}
    \caption{\textbf{Combining both positional embeddings results in superior performance} compared to using only sinusoidal embedding or only RoPE embedding for cross-attention and dual-stream backbone families.}
    \label{fig:pos_emb}
\end{figure}

\paragraph{Positional embeddings}.
Current text-to-image diffusion models often use only one type of positional embedding (\eg, sinusoidal~\citep{esser2024scaling} or RoPE~\citep{cai2025z, wu2025qwen, qin2025lumina}). Inspired by the design in LightningDiT~\citep{yao2025reconstruction}, we explore whether combining sinusoidal and RoPE embeddings improves the text-to-image performance of a diffusion transformer. As shown in \figref{fig:pos_emb}, combining the two meaningfully improves benchmark performance for cross-attention and dual-stream models. As such, we use both sinusoidal and RoPE positional embeddings in our final \textbf{i1} model.

\begin{figure}[!htb]
    \centering
    \includegraphics[width=\linewidth]{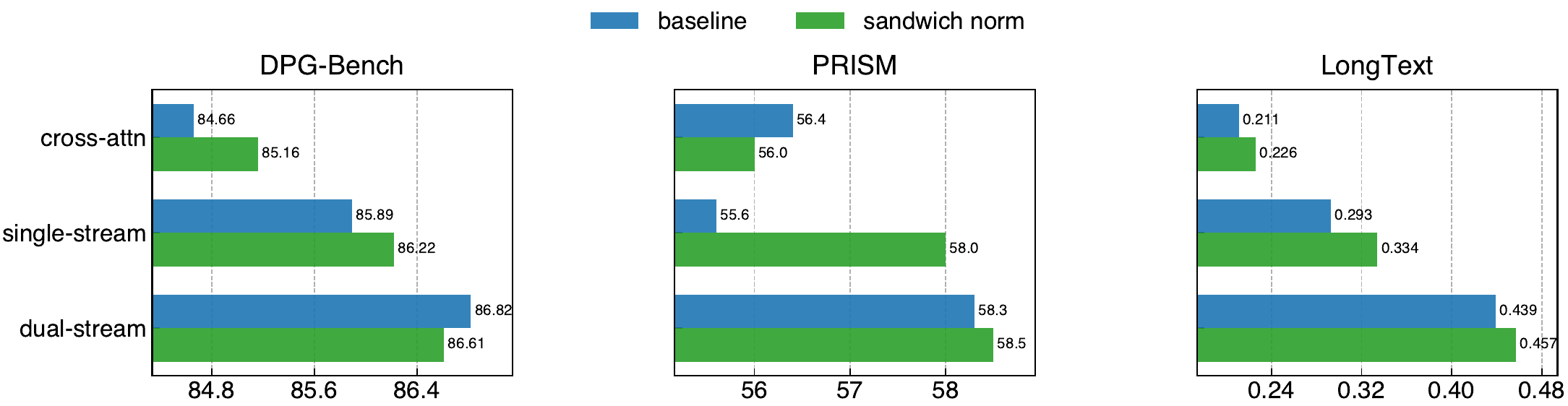}
    \caption{\textbf{Sandwich normalization improves performance} over standard pre-norm across all backbone architectures.}
    \label{fig:sandwich_norm}
\end{figure}

\paragraph{Normalization}. While pre-norm~\citep{xiong2020layer} (\ie, normalizing the input to the attention and feed-forward network modules) has been the dominant choice for diffusion transformers, an earlier work~\citep{ding2021cogview} introduced sandwich norm (\ie, normalizing both the input and the output of the attention and feed-forward network modules) to stabilize training, which has recently been adopted by Z-Image~\citep{cai2025z}. We introduce sandwich normalization to our baseline models in \figref{fig:sandwich_norm} and find that it can stably improve performance across backbone architectures and benchmarks.

\begin{table}[!htb]
\centering
\begin{tabular}{lcccc}
\toprule
model & DPG $\uparrow$ & PRISM $\uparrow$ & LongText $\uparrow$\\
\midrule
shared norms & \bf 86.82 & \bf 58.3 & \bf 0.439 \\
separate norms & 86.16 & 57.6 & 0.415 \\
\bottomrule
\end{tabular}
\caption{\textbf{Shared normalization for text and image streams in MMDiT}. While modality-specific normalization is standard in MMDiT, we find that sharing normalization parameters across modalities improves performance.}
\label{tab:sep_norm}
\end{table}

Further, while using separate normalization layers for text and image modalities has been the default for existing MMDiT models~\citep{esser2024scaling, cai2025hidream, wu2025qwen}, we explore whether a more unified feature distribution from normalization layers shared across the modalities would benefit model performance. As shown in \tabref{tab:sep_norm}, sharing the normalizations indeed consistently improves performance.

\subsection{VAEs}
\label{appendix:vae}

\begin{table}[!htb]
\centering
\begin{tabular}{l c c c}
      \toprule
      VAE & DPG $\uparrow$ & PRISM $\uparrow$ & LongText $\uparrow$\\
      \midrule
      FLUX.2 & \underline{84.66} & \bf 56.4 & \underline{0.211} \\
      Qwen-Image & 83.29 & 54.3 & \bf 0.266 \\
      VA-VAE & \bf 85.07 & 55.5 & 0.126 \\
      \bottomrule
\end{tabular}
\caption{\textbf{Comparing VAEs}. FLUX.2 VAE has the most balanced performance across all benchmarks.}
\label{tab:vae}
\end{table}

We compare VA-VAE~\citep{yao2025reconstruction} with the VAEs used in frontier models FLUX.2~\citep{blackforestlabs_flux2_2025} and Qwen-Image~\citep{wu2025qwen}. Overall, FLUX.2 achieves the most balanced performance across all benchmarks. Likely due to its alignment with pre-trained semantic features during training, VA-VAE achieves the strongest performance on DPG-Bench, which emphasizes semantic alignment of generated images with prompts. However, it performs much worse than the others on LongText, which evaluates fine-grained text rendering. This may be due to its lower reconstruction fidelity.

\begin{figure}[htbp]
\setlength{\fboxsep}{0pt}
\setlength{\fboxrule}{0.5pt}
  \centering
  \setlength{\tabcolsep}{4pt}
  \small
  \begin{tabular}{@{}cccc@{}}
     original & FLUX.2 VAE & Qwen-Image VAE & VA-VAE \\
    \fbox{\includegraphics[width=2.4cm]{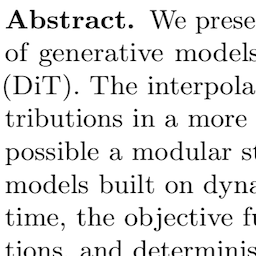}} & 
    \fbox{\includegraphics[width=2.4cm]{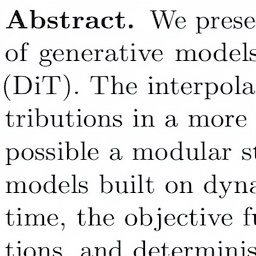}} & 
    \fbox{\includegraphics[width=2.4cm]{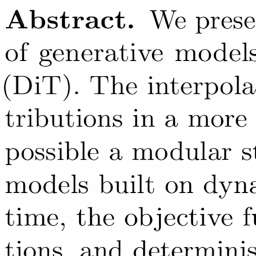}} & 
    \fbox{\includegraphics[width=2.4cm]{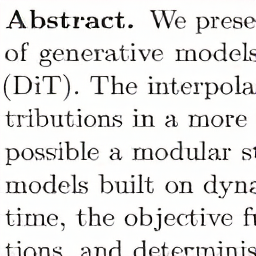}} \\
    
    \fbox{\includegraphics[width=2.4cm]{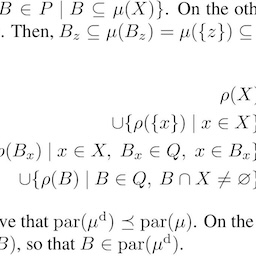}} & 
    \fbox{\includegraphics[width=2.4cm]{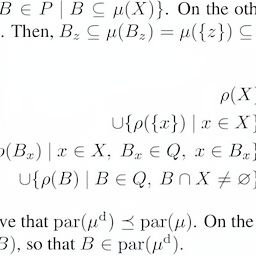}} & 
    \fbox{\includegraphics[width=2.4cm]{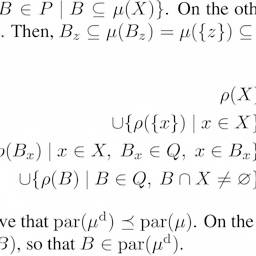}} & 
    \fbox{\includegraphics[width=2.4cm]{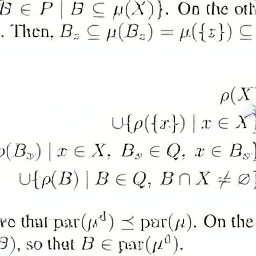}} \\
  \end{tabular}
  \caption{\textbf{Qualitative examples of VAE reconstructions on text-rich images}. Unlike FLUX.2 VAE and Qwen-Image VAE, VA-VAE introduces noticeable distortions and corruptions in the characters. The inferior reconstruction capability may explain the lower LongText performance when using VA-VAE (\tabref{tab:vae}).}
  \label{fig:vae_reconstruction}
\end{figure}

Quantitatively, prior work~\citep{wu2025qwen, yao2025reconstruction} reports that on the ImageNet validation set at 256$\times$256 resolution, VA-VAE has lower reconstruction performance (PSNR 27.96, SSIM 0.79) than FLUX.2 VAE (31.46, 0.90) and Qwen-Image VAE (33.42, 0.92). In \figref{fig:vae_reconstruction}, we further provide qualitative reconstruction examples on text-rich images. While FLUX.2 VAE and Qwen-Image VAE reconstruct faithfully, VA-VAE introduces visible corruption and distortion in the rendered characters. One possible reason for this limitation is that VA-VAE was trained on ImageNet~\citep{deng2009imagenet}, which lacks text-rich images.

\subsection{Comparing Text Encoders under Alternative Settings}
\label{appendix:text_encoder_under_larger_connector}

\begin{figure}[h]
    \centering
    \includegraphics[width=\linewidth]{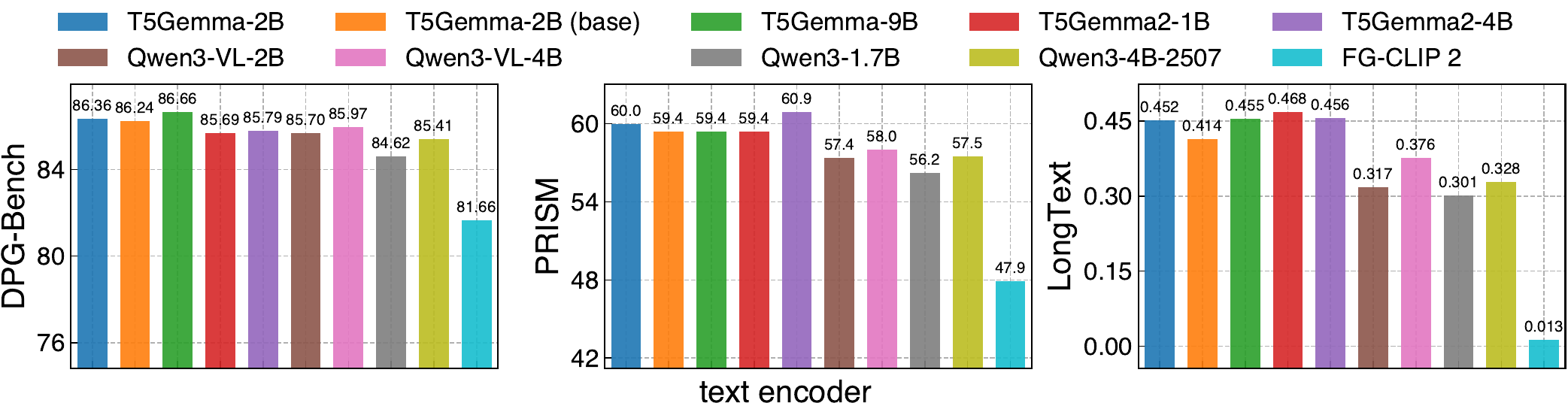}
    \caption{\textbf{Text encoder performance with a larger adapter} shows similar trends across all benchmarks as when using a smaller MLP adapter in the default setting (see~\figref{fig:text_encoder}).}
    \label{fig:text_encoder_under_larger_connector}
\end{figure}

\paragraph{Larger adapter}.
In~\secrefc{sec:component}, we compared the text encoder candidates on our default baseline model with a small MLP adapter. However, as we later showed in~\secrefc{sec:text_noise_cond}, the size of the adapter has a substantial impact on model performance. Thus, here we additionally explore whether our comparisons between text encoders still hold when we use a larger text encoder adapter consisting of two transformer blocks as in our final \textbf{i1} recipe. As shown in~\figref{fig:text_encoder_under_larger_connector}, our observations still largely hold: the T5Gemma and T5Gemma2 families of encoder-decoder models achieve the strongest performance, while FG-CLIP 2 is the weakest.

\begin{figure}[h]
    \centering
    \includegraphics[width=\linewidth]{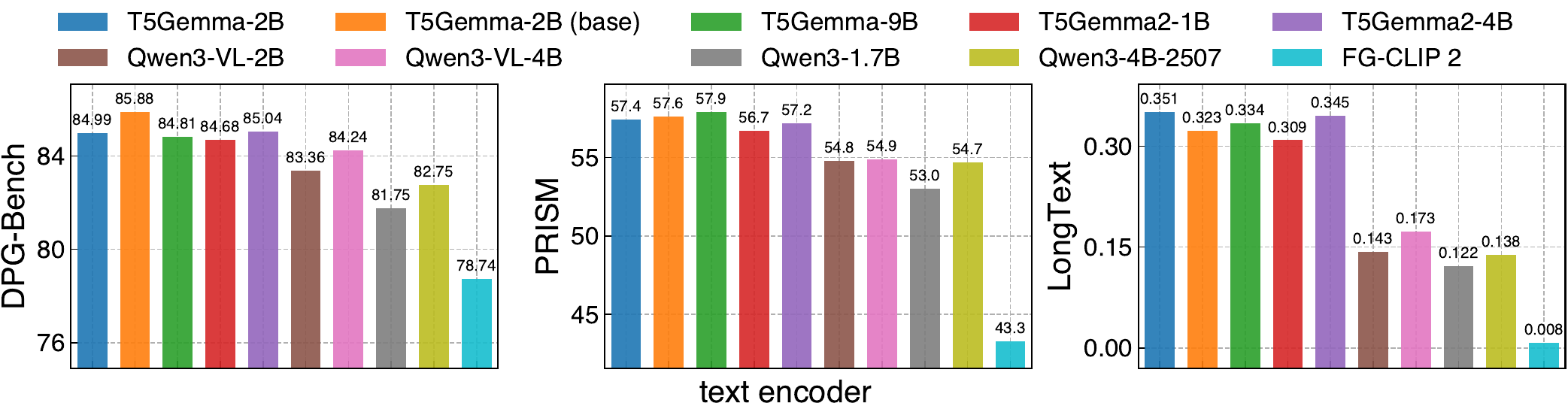}
    \caption{\textbf{Text encoder performance when AdaLN is removed} shows similar trends across all benchmarks as when using AdaLN in the default setting (see~\figref{fig:text_encoder}).}
    \label{fig:text_encoder_with_adaln_removed}
\end{figure}

\paragraph{AdaLN removed}. In our baseline setup (\secrefc{sec:exp_setup}), a pooled text embedding is combined with the timestep embedding and passed into the backbone through AdaLN. Since this provides an additional path for injecting text information, removing AdaLN, as in our final \textbf{i1} recipe, may affect the relative performance of different text encoders. We present the benchmark results for each text encoder in~\figref{fig:text_encoder_with_adaln_removed}. We observe that the overall trends are highly similar to the trends in the default setting, where AdaLN is used (\figref{fig:text_encoder}).

\subsection{Applying System Prompts to Text Encoders}
\label{appendix:text_encoder_system_prompt}

In the default setup (see~\appendixref{appendix:text_encoder_details}), we directly process the raw prompt with each text encoder and use the last hidden states as text features. While this setup is common~\citep{cai2025z}, prior work has also designed specialized prompting strategies when using decoder-only language models as text encoders~\citep{ma2024exploring, xie2025sana}. Here, we follow the strategy used by Qwen-Image~\citep{wu2025qwen} for the Qwen2.5-VL text encoder and apply it to the Qwen3-VL-2B and Qwen3-VL-4B text encoders.

Concretely, we wrap the text-to-image prompt in a system message (``Describe the image by detailing the color, shape, size, texture, quantity, text, spatial relationships of the objects and background:''), and feed the full sequence into the text encoder. After the forward pass, we discard the hidden states corresponding to the system prefix and keep only those corresponding to the text-to-image prompt. The resulting performance is shown in~\tabref{tab:text_encoder_system_prompt}. We observe that using the system prompt brings minor improvements for Qwen3-VL-2B, but the Qwen3-VL models still underperform the T5Gemma and T5Gemma2 models.

\begin{table}[htbp]
        \centering
        \begin{tabular}{l c c c c}
            \toprule
            text encoder & system prompt & DPG $\uparrow$ & PRISM $\uparrow$ & LongText $\uparrow$\\
            \midrule
            \multirow{2}{*}{Qwen3-VL-2B} & \xmark & 82.29 & 52.2 & 0.076 \\
            & \cmark & \bf 82.82 & \bf 52.8 & \bf 0.093 \\
            \midrule
            \multirow{2}{*}{Qwen3-VL-4B} & \xmark & 82.07 & \bf 52.9 & \bf 0.071 \\
            & \cmark & \bf 82.41 & 52.4 & 0.065 \\
            \bottomrule
          \end{tabular}
          \caption{\textbf{Applying system prompt to Qwen3-VL text encoders brings a minor boost in performance}.}
          \label{tab:text_encoder_system_prompt}
\end{table}

\subsection{Comparing Backbone Families with Training FLOPs}
\label{appendix:compare_backbone_training_flops}

\begin{figure}[htbp]
    \centering
    \includegraphics[width=\linewidth]{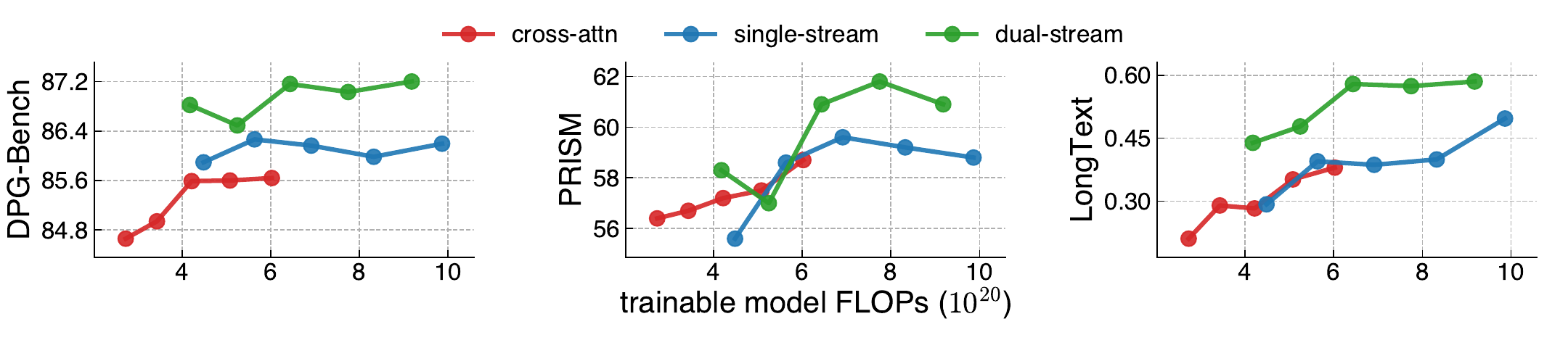}
    \caption{\textbf{Backbone family}. We compare cross-attention, single-stream, and dual-stream backbones across estimated training FLOPs for trainable modules. Consistent with~\figref{fig:backbone_scaling}, where we plot performance against model size, we find that the dual-stream backbone achieves the best overall performance.}
    \label{fig:backbone_scaling_flops}
\end{figure}

In~\figref{fig:backbone_scaling}, we compared different backbone families by training cross-attention, single-stream, and dual-stream models with widths of 1152, 1296, 1440, 1584, and 1728, and plotting performance against model size. However, depending on the practical training and inference setting, comparing performance across FLOPs may be more informative. In~\figref{fig:backbone_scaling_flops}, we therefore plot performance against trainable model FLOPs. These FLOPs are computed using JAX/XLA's ``cost\_analysis'' for one forward and backward pass through all trainable modules in the diffusion model (including \eg the text encoder adapter). We use the training tensor shapes and scale the result by the global batch size and the number of training steps. The dual-stream backbone still achieves the best trade-off.

\subsection{Validating Modeling Designs on Larger Models}
\label{appendix:validate_on_3b}

\begin{figure}[h]
    \centering
    \includegraphics[width=\linewidth]{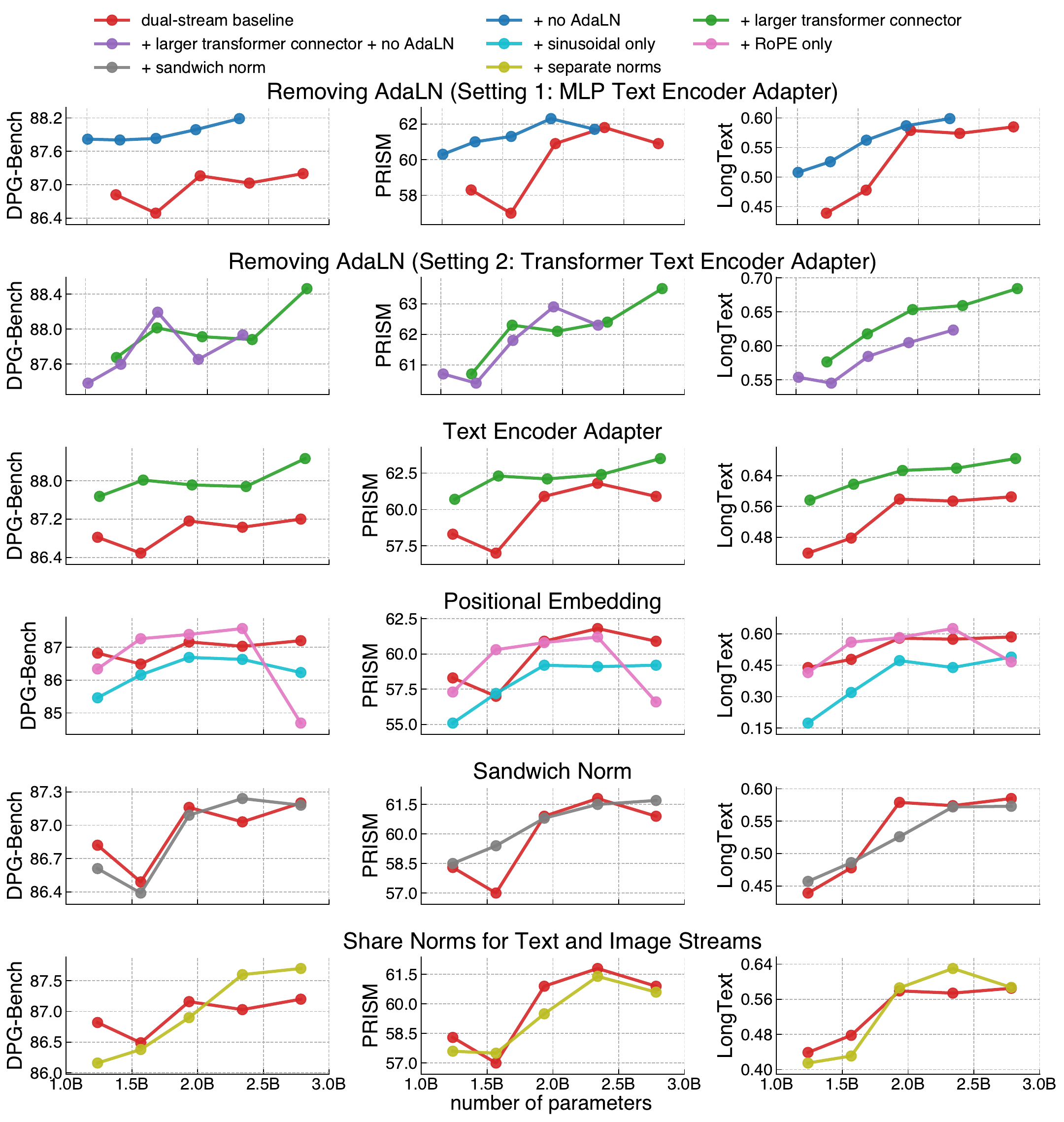}
    \caption{\textbf{Ablating modeling designs on the dual-stream backbone across model sizes}.}
    \label{fig:validate_on_3b}
\end{figure}

\paragraph{Performance \vs model size}.
In~\secrefc{sec:modeling} and~\appendixref{appendix:pos_emb_and_norm}, except for the backbone-family comparison and long skip connections, which we validated across model sizes, we mainly identified modeling design findings using an XL/2-sized baseline. In~\figref{fig:validate_on_3b}, we further validate these findings on dual-stream MMDiT models across multiple model sizes by training one model for each width in \{1152, 1296, 1440, 1584, and 1728\}.

Across model sizes, larger text encoder adapters (\secrefc{sec:text_noise_cond}) consistently provide better performance-parameter trade-offs. Removing AdaLN (\secrefc{sec:text_noise_cond}) also yields a clear advantage when using an MLP text encoder adapter. However, when using a larger transformer text encoder adapter, models with and without AdaLN achieve similar performance at comparable parameter counts. We note that this still suggests that noise conditioning may not be necessary in text-to-image diffusion models.

Finally, while~\appendixref{appendix:pos_emb_and_norm} provided preliminary results suggesting that combining Sinusoidal and RoPE positional embeddings, using sandwich normalization, and sharing normalizations across image and text streams can improve performance, we do not observe these trends consistently across model scales.

\paragraph{Performance \vs training FLOPs}.
In addition to comparing performance across model sizes in Figures~\blueref{fig:long_skip_scaling} and~\blueref{fig:validate_on_3b}, we compare performance against estimated training FLOPs for trainable modules in~\figref{fig:validate_on_3b_flops}. We compute these FLOPs following the same procedure as in~\appendixref{appendix:compare_backbone_training_flops}.

Most trends remain unchanged under the FLOPs-based comparison. One exception is the effect of AdaLN when using a larger transformer-based text encoder adapter: in this setting, models with AdaLN have a better performance-FLOPs trade-off than models without AdaLN. This is because AdaLN contributes a non-trivial fraction of the model parameters (\eg, 18.9\% of parameters for the dual-stream baseline) but only minimally increases training FLOPs, since its projection is computed once per sample rather than once per token.

\subsection{Validating Long Skip Connection Results on Other Backbones}
\label{appendix:long_skip_other_backbone}

\begin{figure}[!htb]
    \centering
    \includegraphics[width=\linewidth]{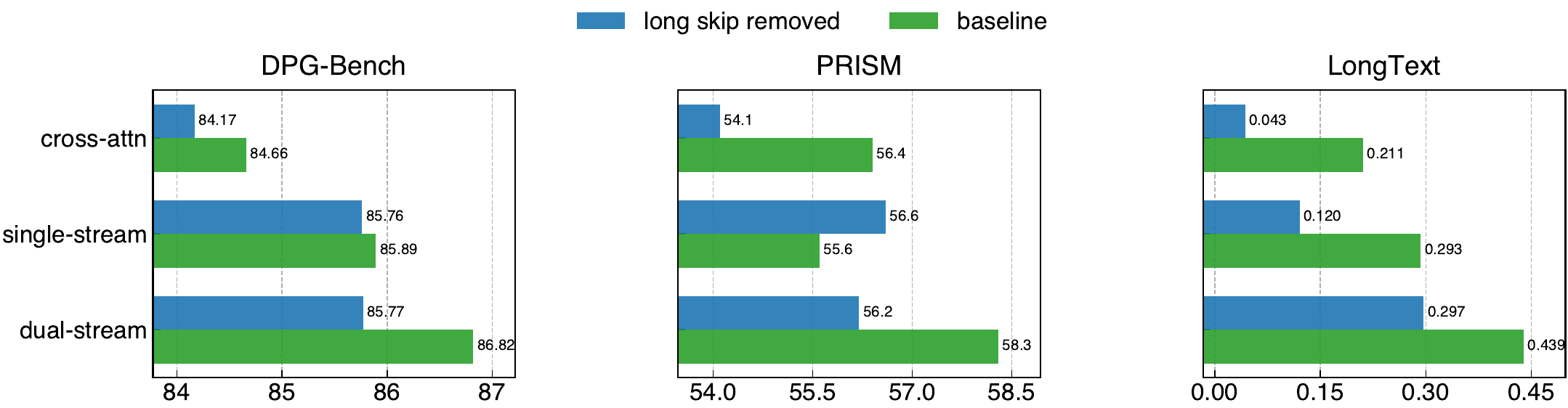}
    \caption{\textbf{Long skip connections}~\citep{bao2023all} improve benchmark performance across backbones. This further supports our observations on dual-stream backbones across model sizes in~\figref{fig:long_skip_scaling}.}
    \label{fig:long_skip}
\end{figure}

In~\secrefc{sec:component}, we showed that long skip connections consistently improve model performance across model sizes, based on experiments with a dual-stream MMDiT backbone. Here, we further validate this design on other backbones by training other variants of the baseline model (which is XL/2-sized) with and without long skip connections. As~\figref{fig:long_skip} shows, removing long skip connections noticeably reduces performance across most backbones and benchmarks, especially on DPG and LongText. This further suggests that long skip connections can broadly benefit model performance.

\begin{figure}[h]
    \centering
    \includegraphics[width=\linewidth]{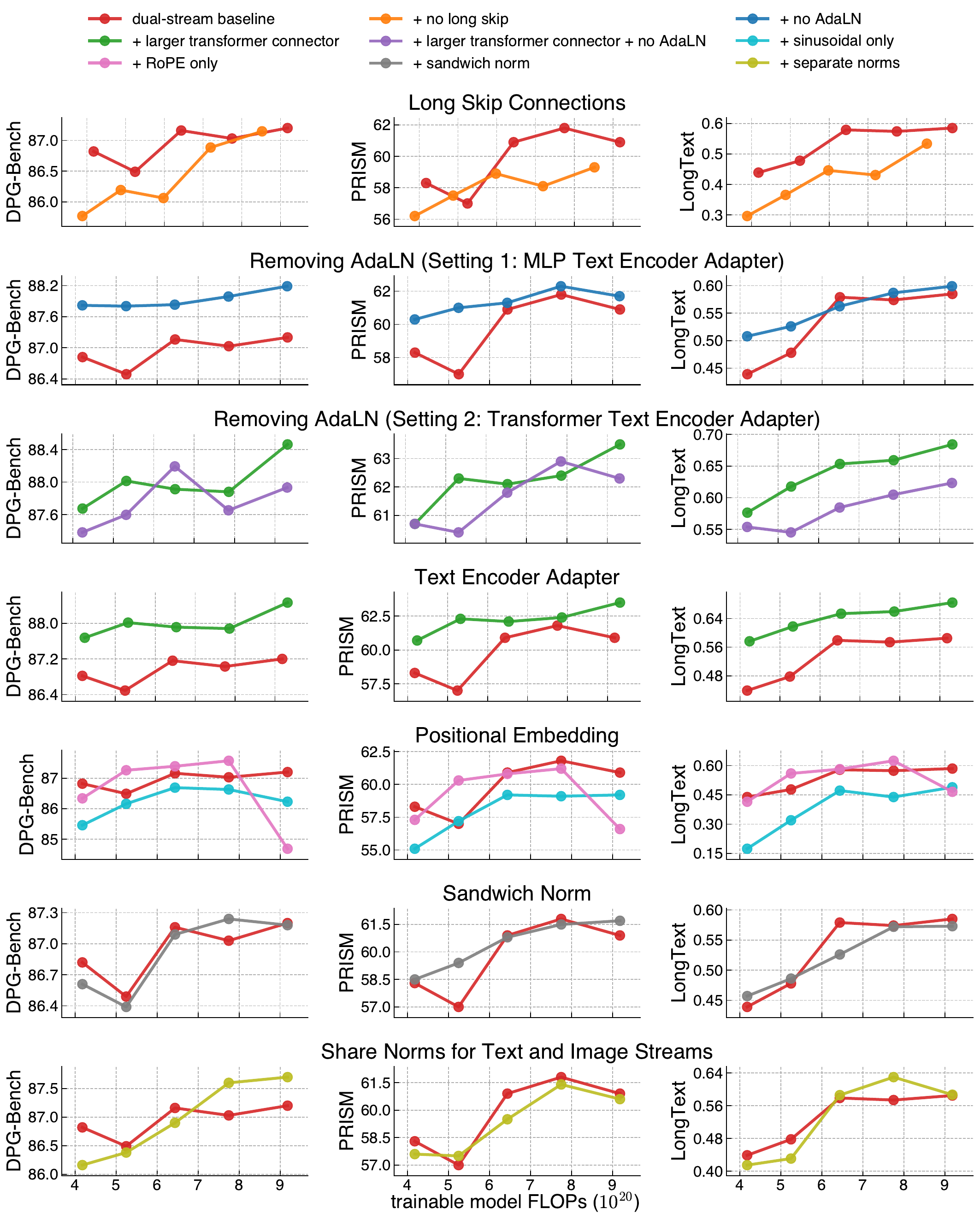}
    \caption{\textbf{Ablating modeling designs on the dual-stream backbone across trainable model FLOPs}.}
    \label{fig:validate_on_3b_flops}
\end{figure}

\clearpage

\subsection{Text Feature Adapter \vs Image Feature Adapter}

In~\secrefc{sec:text_noise_cond}, we found that replacing a small MLP adapter with a larger transformer adapter for the text encoder substantially improves performance across benchmarks, despite adding few parameters. We hypothesize that this is because text features from pre-trained language models need to be adapted for downstream tasks such as text-to-image generation. In~\appendixref{appendix:validate_on_3b}, we showed that the larger transformer adapter achieves a better performance-parameter trade-off than the smaller MLP adapter.

\begin{table}[htbp]
        \centering
        \begin{tabular}{l l c c c}
            \toprule
            backbone & adapter & DPG $\uparrow$ & PRISM $\uparrow$ & LongText $\uparrow$\\
            \midrule
            \multirow{3}{*}{cross-attention} & default & 84.66 & 56.4 & 0.211 \\
            & + transformer adapter for text features & \bf 86.33 & \bf 58.7 & \bf 0.414 \\
            & + transformer adapter for image features & 85.27 & 55.4 & 0.250 \\
            \midrule
            \multirow{3}{*}{single-stream} & default & 85.89 & 55.6 & 0.293 \\
            & + transformer adapter for text features & \bf 87.64 & \bf 60.0 & \bf 0.472 \\
            & + transformer adapter for image features & 86.10 & 59.7 & 0.345 \\
            \midrule
            \multirow{3}{*}{dual-stream} & default & 86.82 & 58.3 & 0.439 \\
            & + transformer adapter for text features & \bf 87.67 & \bf 60.7 & \bf 0.576 \\
            & + transformer adapter for image features & 86.23 & 57.0 & 0.378 \\
            \bottomrule
          \end{tabular}
          \caption{\textbf{Comparing transformer adapters for text and image features}. Using a transformer adapter for the text features substantially improves performance across backbones, whereas adding the same adapter to the image features does not. This suggests that the benefit of the larger text adapter is not merely due to increased parameter count.}
          \label{tab:text_vs_image_adapter}
\end{table}

To further verify that this improvement is not merely due to increased parameter count, we analogously add a transformer block to the image features, after patchification and before applying positional embeddings (see~\appendixref{appendix:baseline_arch}). As shown in~\tabref{tab:text_vs_image_adapter}, adding this adapter to the image features results in much smaller performance improvement compared to using it on the text features. This suggests that the gains from larger adapters are specific to adapting pre-trained text features rather than simply increasing model capacity.

\subsection{Exploring Variants of Long Skip Connections}

Long skip connections were popularized by U-Net~\citep{ronneberger2015u} to provide shortcuts for low-level features from earlier layers, thereby easing training for pixel-level prediction tasks. Later, U-ViT~\citep{bao2023all} followed this design and applied it to transformer-based diffusion models. However, while the exactly symmetric structure of these connections (\ie, the $i$-th leftmost layer is connected to the $i$-th rightmost layer) is natural for the multi-resolution encoder-decoder structure of U-Net, it is not necessarily optimal for transformer-based models, whose blocks often have the same feature dimensionality. Therefore, here, we explore different variants of the original long skip connections in U-ViT.

\begin{figure}[!htb]
    \centering
    \includegraphics[width=0.75\linewidth]{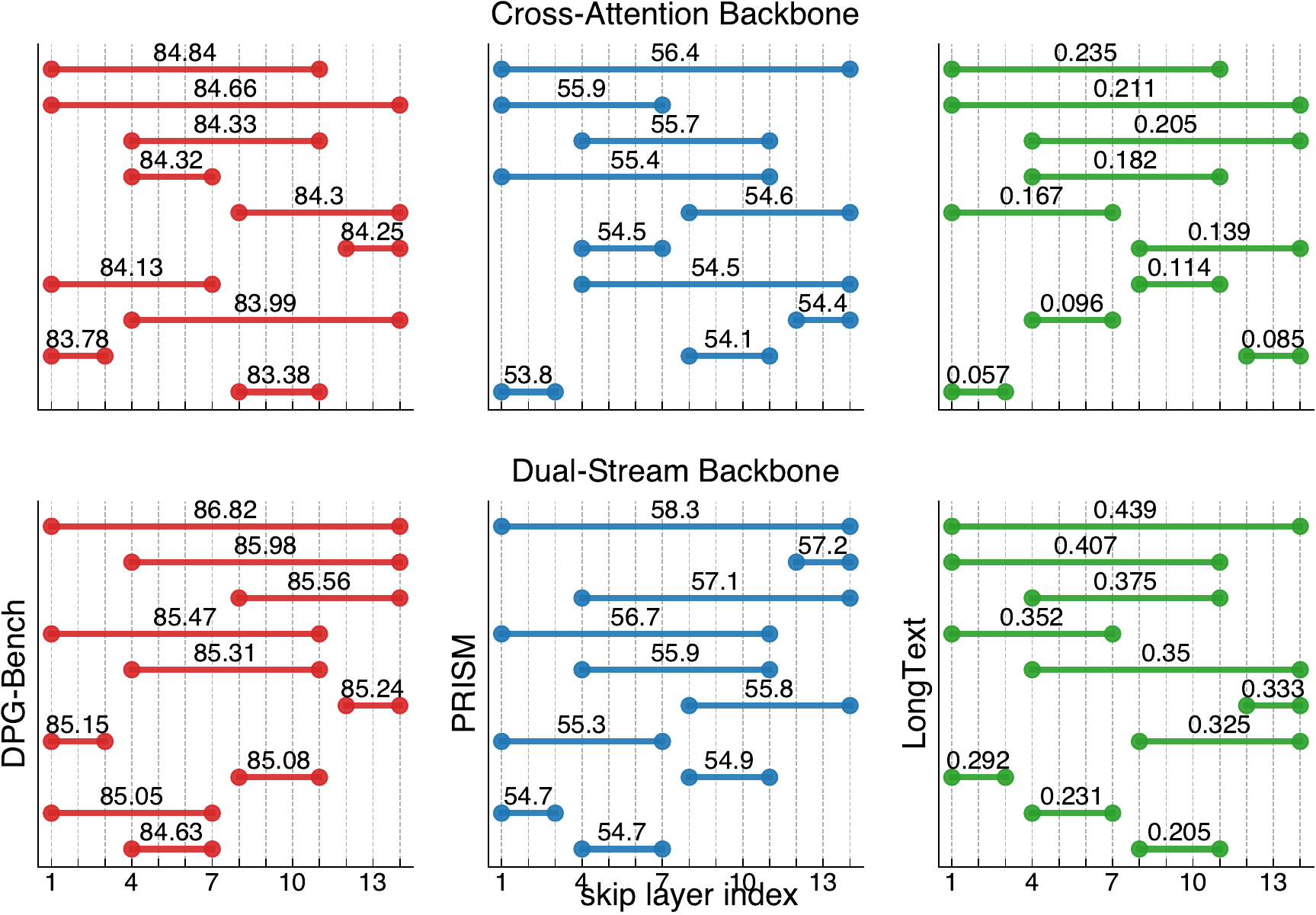}
    \caption{\textbf{Ablating the range of layers to which long skip connections are applied}. The x-axis shows all layer indices from which a skip connection can start. Each horizontal line corresponds to a variant in which all layers between the left and right endpoint indices, inclusive, have skip connections starting from them. None of the variants consistently outperforms the default long skip connections, where layers 1-14 all have skip connections.}
    \label{fig:long_skip_variant}
\end{figure}

\paragraph{Layer range}. Long skip connections from earlier layers skip across more blocks, whereas those closer to the middle skip across fewer blocks. As a result, features from layers closer to the middle may have changed less by the time they reach their destination layers, making these skip connections potentially less necessary. Removing them could therefore preserve performance or even improve it. To test this hypothesis, in~\figref{fig:long_skip_variant}, we explore several ranges of layers from which long skip connections can start: 1-3, 1-7, 1-11, 4-7, 4-11, 4-14, 8-11, 8-14, and 12-14. However, none of these variants consistently outperforms the default setting.

\begin{table}[!htb]
\centering
\begin{tabular}{l c c c}
      \toprule
      skip type & DPG $\uparrow$ & PRISM $\uparrow$ & LongText $\uparrow$\\
      \midrule
      default & 84.66 & \bf 56.4 & \bf 0.211 \\
      $i\rightarrow i+14$ & 83.70 & 55.2 & 0.134 \\
      $i\rightarrow i+21$ & \bf 84.78 & 55.8 & 0.180 \\
      \bottomrule
\end{tabular}
\caption{\textbf{Ablating long skip connection patterns}. We compare the default symmetric skip pattern with variants that connect the $i$-th layer to the $(i+14)$-th or $(i+21)$-th layer on the cross-attention backbone. The default pattern achieves the best overall performance.}
\label{tab:long_skip_connection_pattern}
\end{table}

\paragraph{Connection pattern}. We further explore whether long skip connections should connect layers with larger representational differences. Instead of using the default symmetric pattern, which connects the $i$-th leftmost layer to the $i$-th rightmost layer, we test variants that connect the $i$-th layer to the $(i+14)$-th or $(i+21)$-th layer. As shown in~\tabref{tab:long_skip_connection_pattern}, on the cross-attention backbone, the $(i+21)$ variant slightly improves DPG, but the default pattern still performs best overall, achieving the highest PRISM and LongText scores.

\clearpage

\section{Additional Results on Data Designs}

In~\secrefc{sec:data}, we presented controlled experiments motivating our designs for synthetic captioning, prompt rewrite, and data mixing. We provide further details on our synthetic captioning designs, along with corresponding controlled experiments, in~\appendixref{appendix:additional_design_synthetic_caption}, and additional results supporting our conclusion on data mixing in~\appendixref{appendix:exp_under_equal_weight}.

\subsection{Additional Designs in Synthetic Captioning}
\label{appendix:additional_design_synthetic_caption}

In~\secrefc{sec:caption_and_rewrite}, we studied synthetic caption generation by training a baseline cross-attention model on ImageNet-22K images with different caption sets. Here, we provide additional results that motivated two choices in our caption-generation pipeline: center-cropping images before captioning and generating multiple captions per image.

\begin{table}[h]
\centering
\begin{tabular}{lcccc}
\toprule
captioner & DPG $\uparrow$ & PRISM $\uparrow$ & LongText $\uparrow$\\
\midrule
Qwen3-VL-30B-A3B & \textbf{83.72} & 50.8 & 0.007 \\
+ no center-crop & 83.16 & 51.3 & 0.006 \\
+ 5 captions/image & 83.56 & \textbf{51.9} & \textbf{0.010} \\
\bottomrule
\end{tabular}
\caption{\textbf{Synthetic captioning}. Training with multiple captions per image and center-cropping input images before captioning leads to better performance. \figref{fig:imagenet_num_images} further shows that the advantage of using multiple captions per image is stronger under limited image data.}
\label{tab:synthetic_captioner}
\end{table}

\paragraph{Image cropping}. 
In our experimental setup, we train on square images obtained by center-cropping the longer edge to match the shorter edge. If the synthetic captioner receives the full uncropped images, the generated captions may describe objects that are later cropped out, creating a semantic mismatch between the training images and captions. \citet{podell2023sdxl} suggested that this may contribute to the failure mode of text-to-image models generating partial objects. Further, by default, after center-cropping, we resize images larger than 512$\times$512 down to 512$\times$512 to avoid slow captioning on larger images.

To understand the impact of our pre-processing operations, we train our model on two sets of synthetic captions generated by Qwen3-VL-30B-A3B: one produced from the full images and the other from the cropped and resized square images. As shown in the first two rows of \tabref{tab:synthetic_captioner}, captions generated from the cropped and resized images lead to similar downstream performance as captions from full images. Since center-cropping and resizing improve captioning speed without meaningfully affecting downstream performance, we apply both operations before generating all synthetic captions.

\begin{figure}[!htb]
    \centering
    \includegraphics[width=\linewidth]{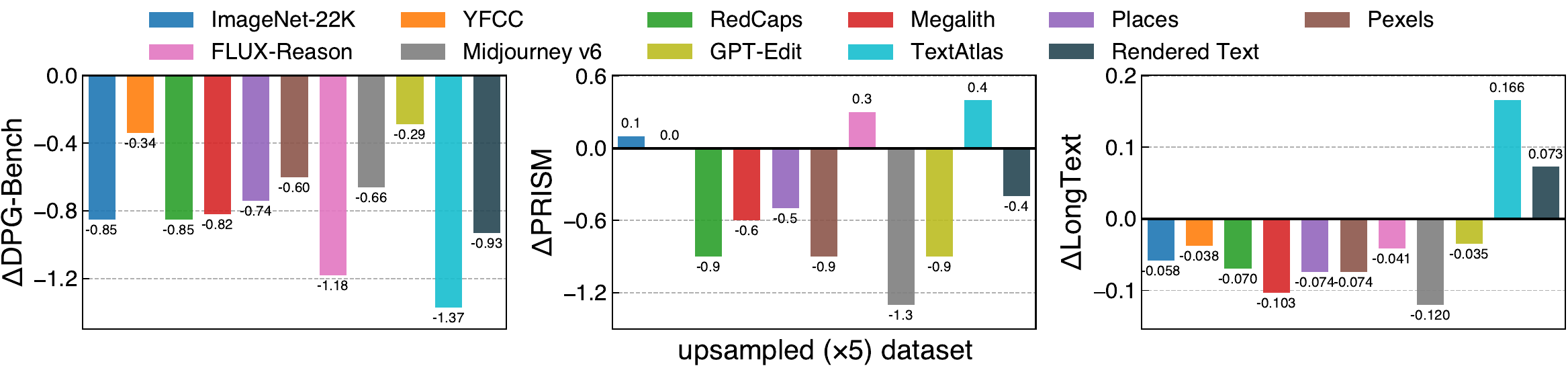}
    \caption{\textbf{Performance change from upweighting a single dataset} by 5$\times$ (3$\times$ in~\figref{fig:upsample_x3}) relative to the baseline of equal weights for all datasets. In all cases, upweighting any dataset does not outperform exact equal weighting.}
    \label{fig:upsample_x5}
\end{figure}

\paragraph{Caption diversity}. 
Increasing the number of captions per image provides another axis of data scaling, beyond increasing the number of images. To explore this, we generate five captions per image using Qwen3-VL-30B-A3B. As shown in \tabref{tab:synthetic_captioner}, this leads to modest improvements on PRISM and LongText.
Moreover, as shown in~\figref{fig:imagenet_num_images}, this benefit becomes more pronounced when the number of unique training images is limited.

\subsection{Experiments under Equal Dataset Weights}
\label{appendix:exp_under_equal_weight}

In~\secrefc{sec:data_mixing}, after identifying that equal weighting for all datasets can result in strong performance, we further explored whether upsampling a single dataset, while keeping the others equally weighted, could further improve results. In~\figref{fig:upsample_x3}, we demonstrated that upsampling any one dataset by $\times 3$ does not consistently improve performance across all benchmarks. Here, we further show in~\figref{fig:upsample_x5} the results for upsampling each dataset by $\times$5, and find that the results are consistent with the observations in~\figref{fig:upsample_x3}.

\clearpage

\section{Additional Information on Datasets}
\label{appendix:additional_info_data}

We briefly introduced the image datasets used for model training in~\secrefc{sec:exp_setup} and explored synthetic captioning and data mixing strategies in~\secrefc{sec:data}. Here, we provide additional details about the images and captions used during training.

\subsection{Visualizations of Images}
\label{appendix:dataset_vis}

\begin{figure}[htbp]
    \centering
    \subfloat[{\centering ImageNet-22K}]{
    \centering
    \begin{minipage}[h]{0.24\linewidth}
    \begin{center}        
        \includegraphics[width=\linewidth]{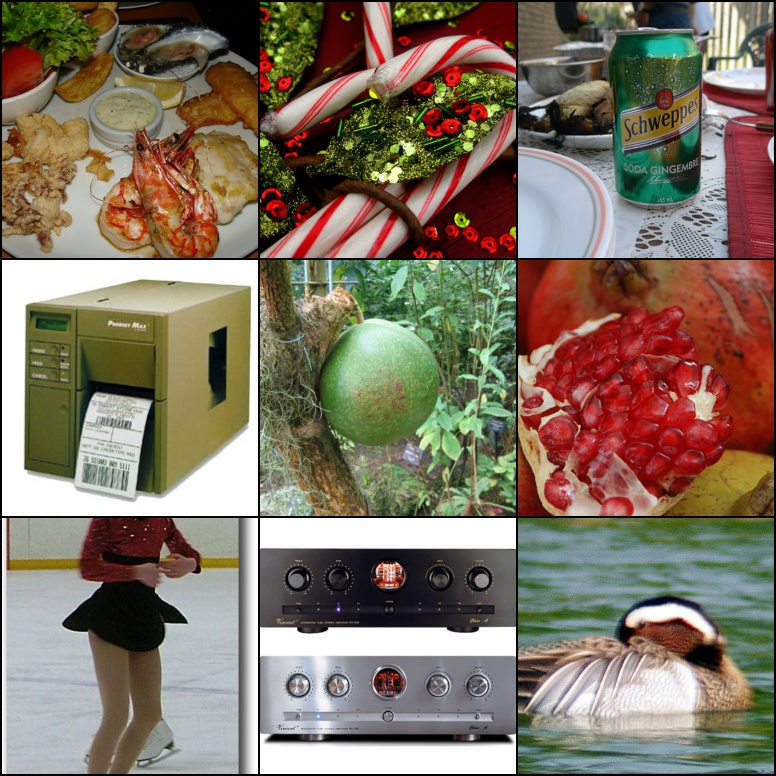}
    \end{center}
    \end{minipage}}
    \hfill
    \centering
    \subfloat[{\centering YFCC}]{
    \centering
    \begin{minipage}[h]{0.24\linewidth}
    \begin{center}
        \includegraphics[width=\linewidth]{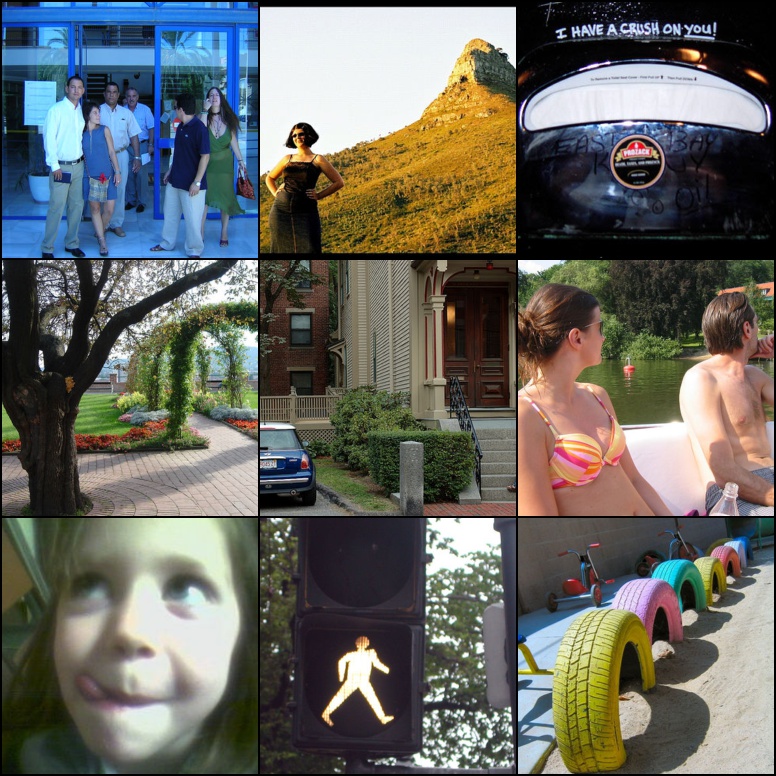}
    \end{center}
    \end{minipage}}
    \hfill
    \centering
    \subfloat[{\centering RedCaps}]{
    \centering
    \begin{minipage}[h]{0.24\linewidth}
    \begin{center}        
        \includegraphics[width=\linewidth]{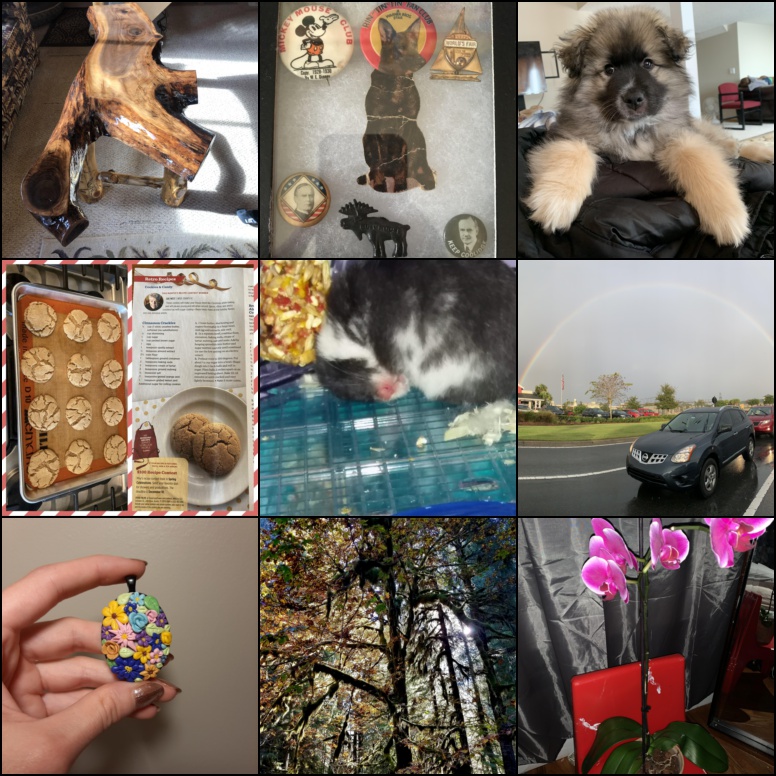}
    \end{center}
    \end{minipage}}
    \hfill
    \centering
    \subfloat[{\centering Megalith}]{
    \centering
    \begin{minipage}[h]{0.24\linewidth}
    \begin{center}        
        \includegraphics[width=\linewidth]{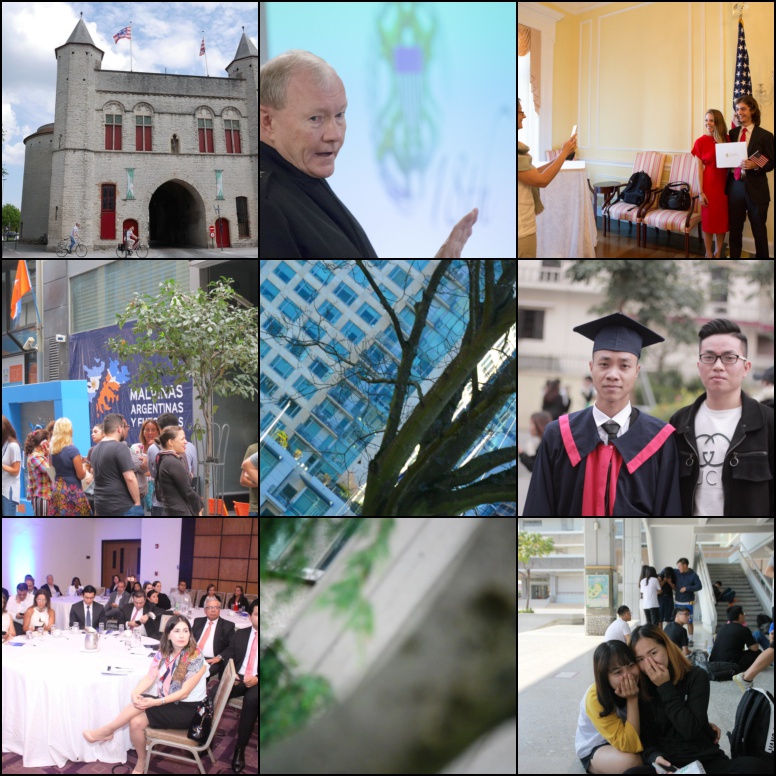}
    \end{center}
    \end{minipage}}
    \\[-.25em]
    \centering
    \subfloat[{\centering Places}]{
    \centering
    \begin{minipage}[h]{0.24\linewidth}
    \begin{center}
        \includegraphics[width=\linewidth]{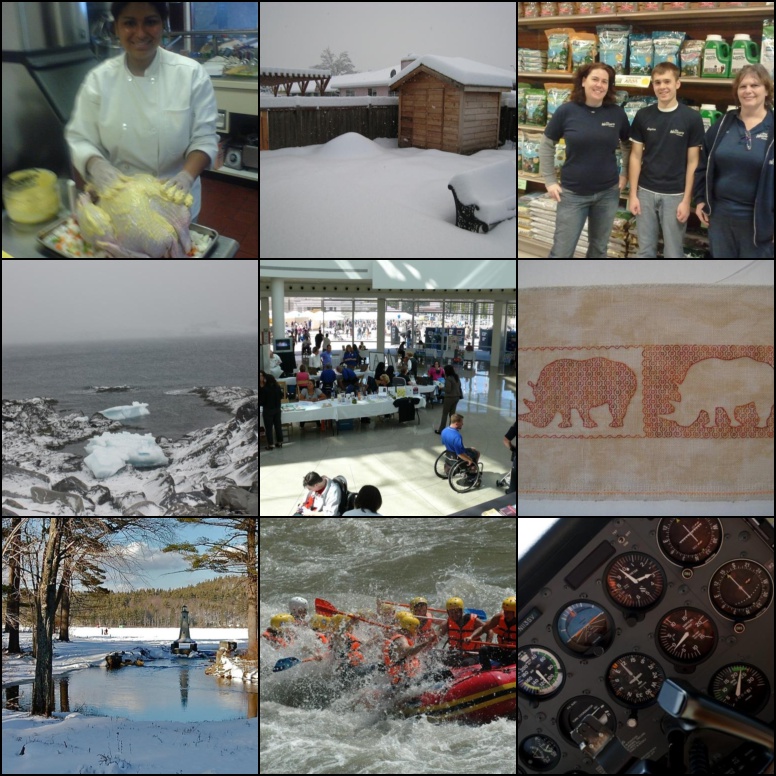}
    \end{center}
    \end{minipage}}
    \hfill
    \centering
    \subfloat[{\centering Pexels}]{
    \centering
    \begin{minipage}[h]{0.24\linewidth}
    \begin{center}        
        \includegraphics[width=\linewidth]{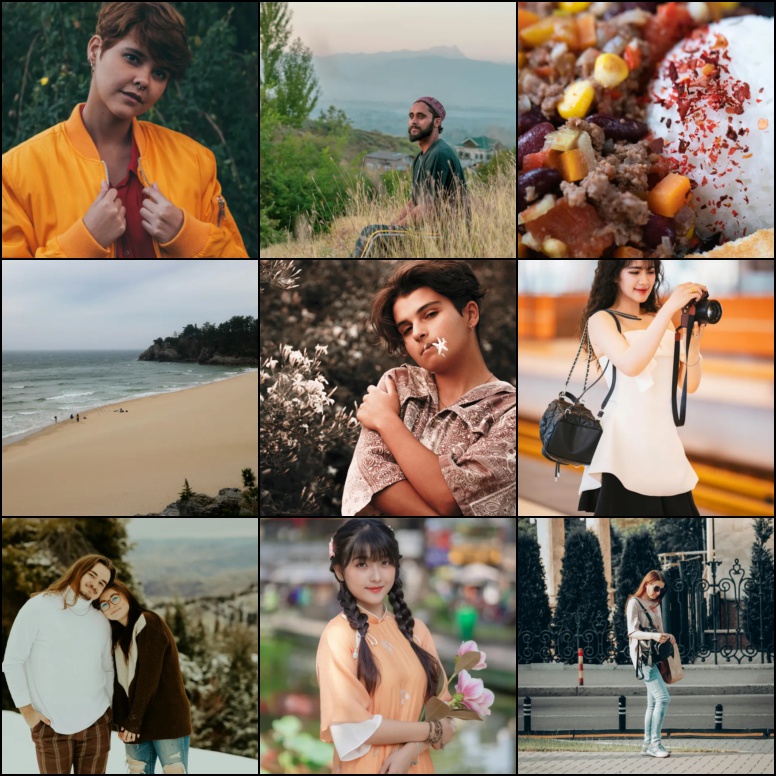}
    \end{center}
    \end{minipage}}
    \hfill
    \centering
    \subfloat[{\centering iNaturalist}]{
    \centering
    \begin{minipage}[h]{0.24\linewidth}
    \begin{center}        
        \includegraphics[width=\linewidth]{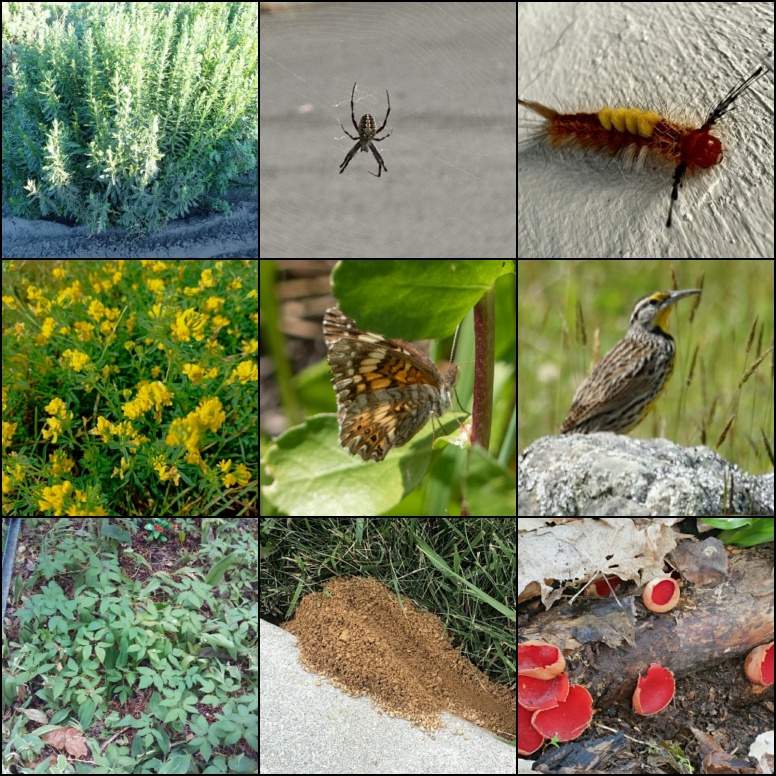}
    \end{center}
    \end{minipage}}
    \hfill
    \centering
    \subfloat[{\centering FLUX-Reason}]{
    \centering
    \begin{minipage}[h]{0.24\linewidth}
    \begin{center}        
        \includegraphics[width=\linewidth]{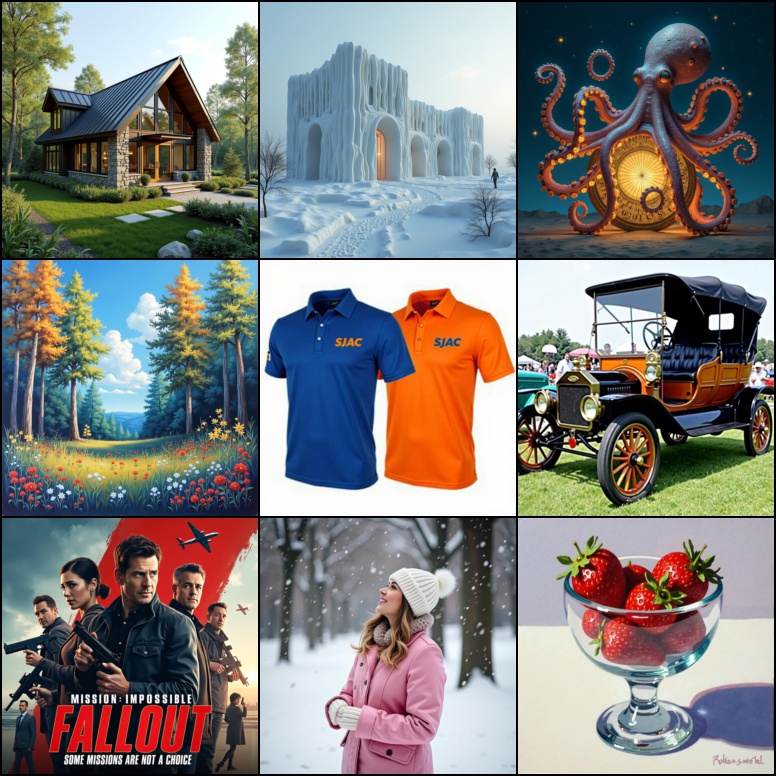}
    \end{center}
    \end{minipage}}
    \\[-.25em]
    \centering
    \subfloat[{\centering Midjourney v6}]{
    \centering
    \begin{minipage}[h]{0.24\linewidth}
    \begin{center}
        \includegraphics[width=\linewidth]{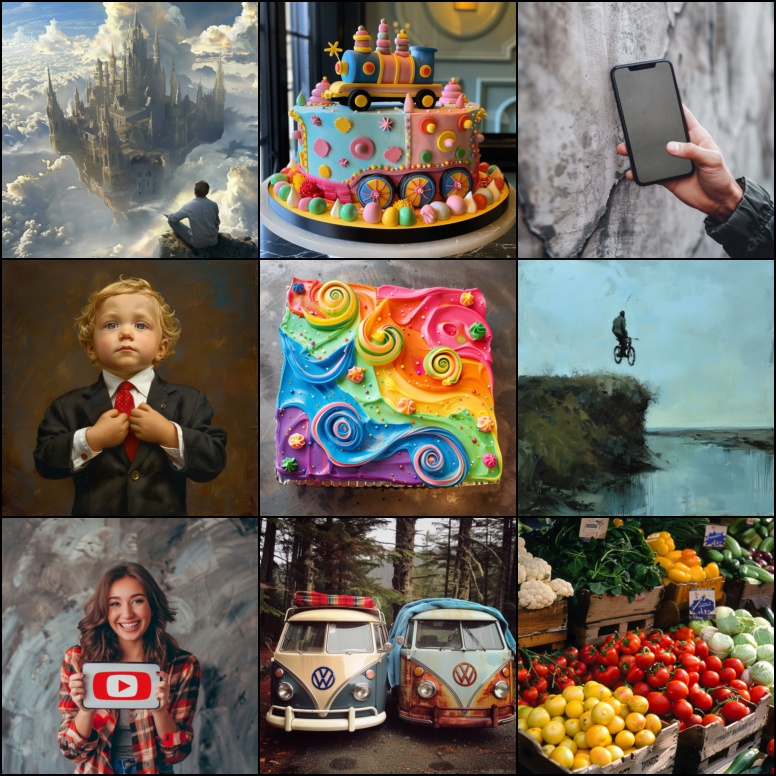}
    \end{center}
    \end{minipage}}
    \hfill
    \centering
    \subfloat[{\centering GPT-Edit}]{
    \centering
    \begin{minipage}[h]{0.24\linewidth}
    \begin{center}        
        \includegraphics[width=\linewidth]{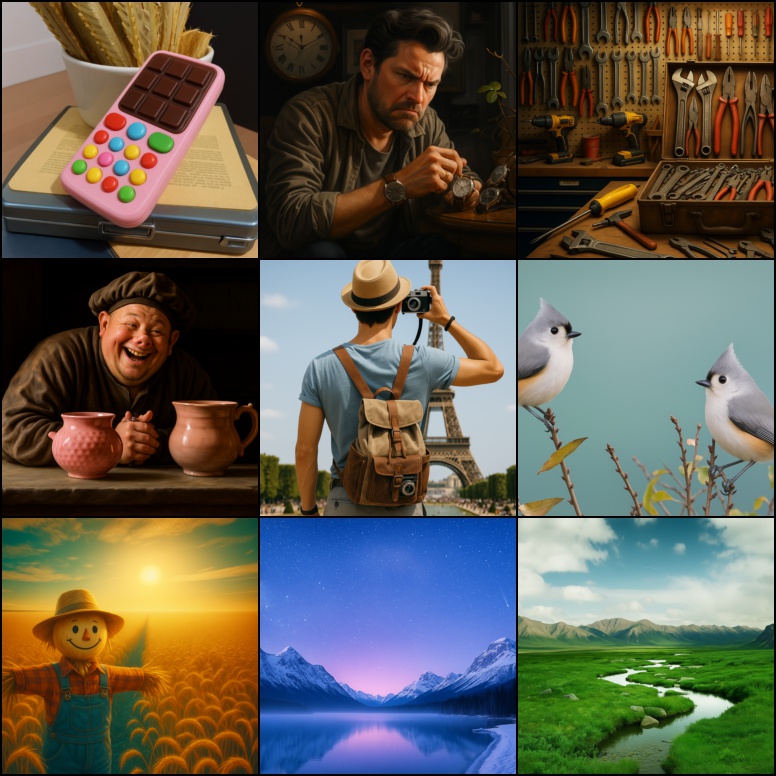}
    \end{center}
    \end{minipage}}
    \hfill
    \centering
    \subfloat[{\centering TextAtlas}]{
    \centering
    \begin{minipage}[h]{0.24\linewidth}
    \begin{center}        
        \includegraphics[width=\linewidth]{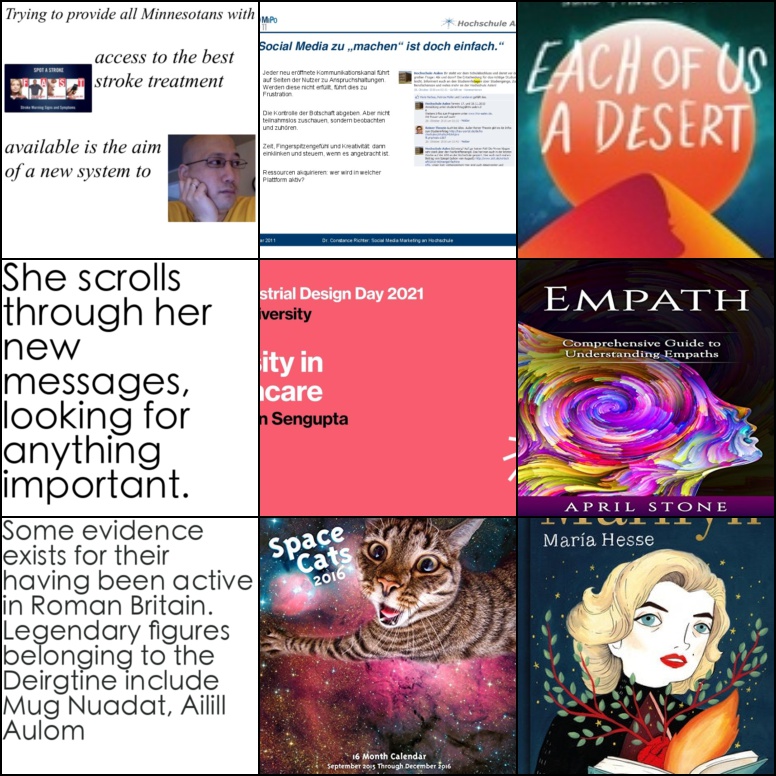}
    \end{center}
    \end{minipage}}
    \hfill
    \centering
    \subfloat[{\centering RenderedText}]{
    \centering
    \begin{minipage}[h]{0.24\linewidth}
    \begin{center}        
        \includegraphics[width=\linewidth]{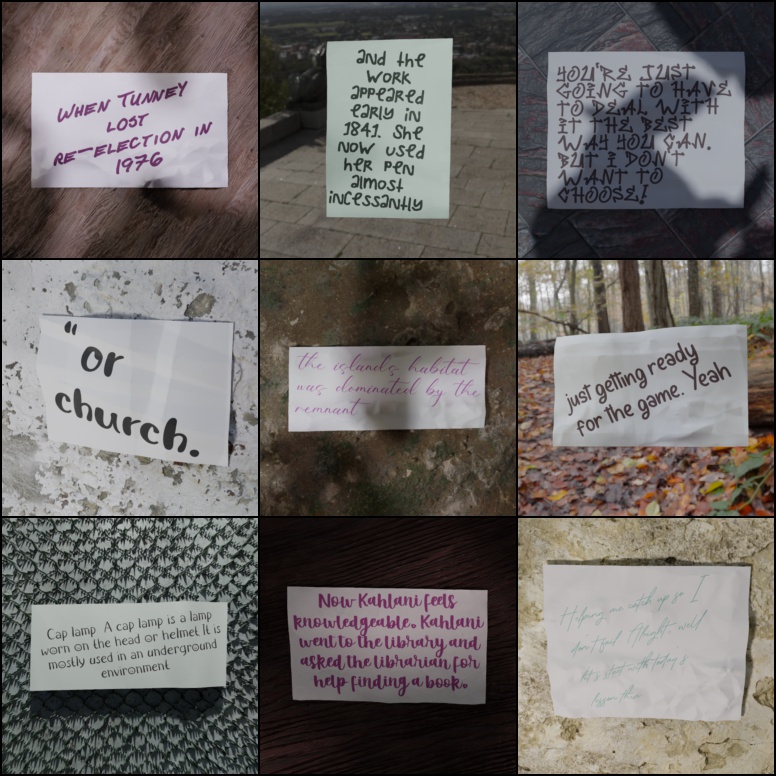}
    \end{center}
    \end{minipage}}
    \caption{\textbf{Random samples of images from each training dataset}.}
    \label{fig:vis_image_dataset}
\end{figure}

In our study, we explored data mixing strategies and trained models on 12 publicly available image datasets, including 7 real-image datasets (ImageNet-22K~\citep{deng2009imagenet}, YFCC100M~\citep{thomee2016yfcc100m}, RedCaps~\citep{desai2021redcaps}, Megalith~\citep{BoerBohan2024Megalith10m}, Pexels~\citep{Narugo2024PexelsTaggerV0}, iNaturalist 2024~\citep{vendrow2024inquire}, Places365-Challenge 2016~\citep{zhou2017places}), 3 synthetic datasets (GPT-Image-Edit-1.5M~\citep{wang2025gpt}, FLUX-Reason-6M~\citep{fang2026flux}, and Midjourney v6~\citep{CortexLM2024MidjourneyV6}), and 2 text-rendering datasets (RenderedText~\citep{Wendler2024RenderedText} and TextAtlas~\citep{wang2025textatlas5m}). To provide a qualitative understanding of their distributions, we randomly sample 9 images from each dataset, resize each image so that its shorter edge is 256 pixels, center-crop it to a 256$\times$256 square, and visualize the resulting images in~\figref{fig:vis_image_dataset}.

\clearpage

\subsection{Image Resolution Statistics}
\label{appendix:img_res_stat}

As discussed in~\secrefc{sec:high_res_training}, we use all image datasets, excluding iNaturalist, for the 256-resolution training of \textbf{i1}. For high-resolution training at 512/1024 resolution, we filter out images whose shorter edge is smaller than 512/1024 pixels. We also remove any dataset that contains fewer than 0.3M images after filtering.

Here, we report the number of images that remain under each resolution threshold in~\tabref{tab:image_resolution_stat}. For 512-resolution training, we use all datasets except YFCC and iNaturalist. For 1024-resolution training, we use only FLUX-Reason, TextAtlas, RedCaps, GPT-Edit, and Midjourney v6 (although RenderedText satisfies the image-count requirement after resolution-based filtering, we exclude it due to its low quality).

The RedCaps dataset we use contains 5M images, which differs from the 12M images reported in the original paper. This is because RedCaps images are provided as URLs, many of which are no longer accessible. 

\begin{table}[!htb]
\centering
\begin{tabular}{llccc}
\toprule
image type & dataset & \#imgs & \#imgs {\scriptsize w/ shorter edge$\geqslant$512} & \#imgs {\scriptsize w/ shorter edge$\geqslant$1024}\\
\midrule
\multirow{7}{*}{real} & YFCC & 98,121,424 & 0 & 0 \\
& ImageNet-22K & 13,673,551 & 719,427 & 0 \\
& Megalith & 9,393,971 & 9,202,294 & 61,701 \\
& Places & 8,026,628 & 7,345,764 & 0 \\
& RedCaps & 4,817,431 & 4,705,536 & 4,134,303 \\
& iNaturalist & 4,813,543 & 0 & 0 \\
& Pexels & 2,810,634 & 2,810,621 & 0 \\
\midrule
\multirow{3}{*}{synthetic} & FLUX-Reason & 5,890,279 & 5,890,279 & 5,890,279 \\
& GPT-Edit & 1,553,575 & 1,552,821 & 357,482 \\
& Midjourney v6 & 1,240,185 & 1,240,185 & 1,240,185 \\
\midrule
\multirow{2}{*}{text-rendering} & RenderedText & 11,977,824 & 11,977,824 & 11,977,824 \\
 & TextAtlas & 5,397,762 & 4,036,973 & 919,913 \\
\bottomrule
\end{tabular}
\caption{\textbf{Number of images per dataset at each resolution threshold}. For 512- and 1024-resolution training, we use only images whose shorter edge is at least 512 and 1024 pixels, respectively.}
\label{tab:image_resolution_stat}
\end{table}

\subsection{Caption Length Distribution for Different VLM Captioners}

\begin{figure}[!htb]
    \centering
    \includegraphics[width=0.6\linewidth]{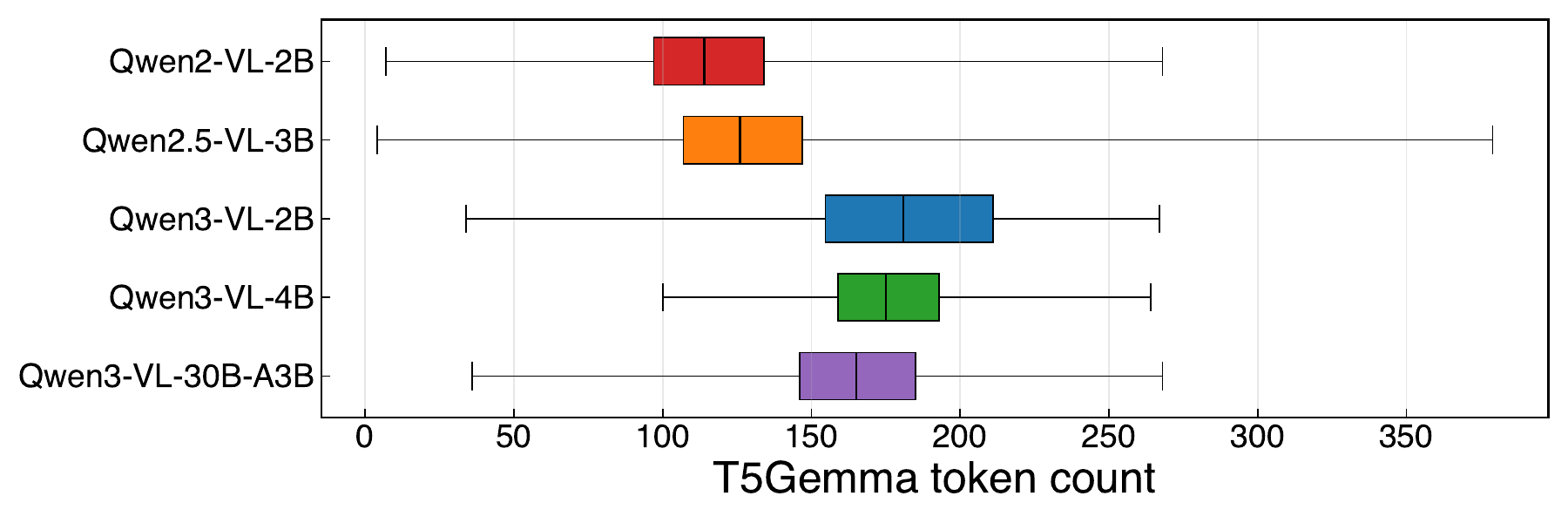}
    \caption{\textbf{ImageNet-22K caption length distributions with different captioners}, measured by token sequence length using the T5Gemma tokenizer. Prompt sets that lead to strong text-to-image performance also tend to be longer overall.}
    \label{fig:caption_length_imagenet}
\end{figure}

In~\secrefc{sec:caption_and_rewrite}, we used the same meta-prompt to prompt Qwen2-VL 2B~\citep{wang2024qwen2}, Qwen2.5-VL 3B~\citep{qwen2.5-vl}, Qwen3-VL-2B, Qwen3-VL-4B, and Qwen3-VL-30B-A3B~\citep{bai2025qwen3} to generate synthetic captions for ImageNet-22K. We then train a diffusion model on each resulting image-caption dataset. We find that the choice of VLM used for synthetic captioning has a substantial impact on downstream text-to-image performance. In~\figref{fig:caption_length_imagenet}, we further plot the sequence length distribution of the captions generated by each VLM. We observe that caption sets that lead to stronger performance also tend to be longer.

\clearpage

\subsection{Caption Length Distribution for Different Datasets}

\begin{figure}[!htb]
    \centering
    \includegraphics[width=0.6\linewidth]{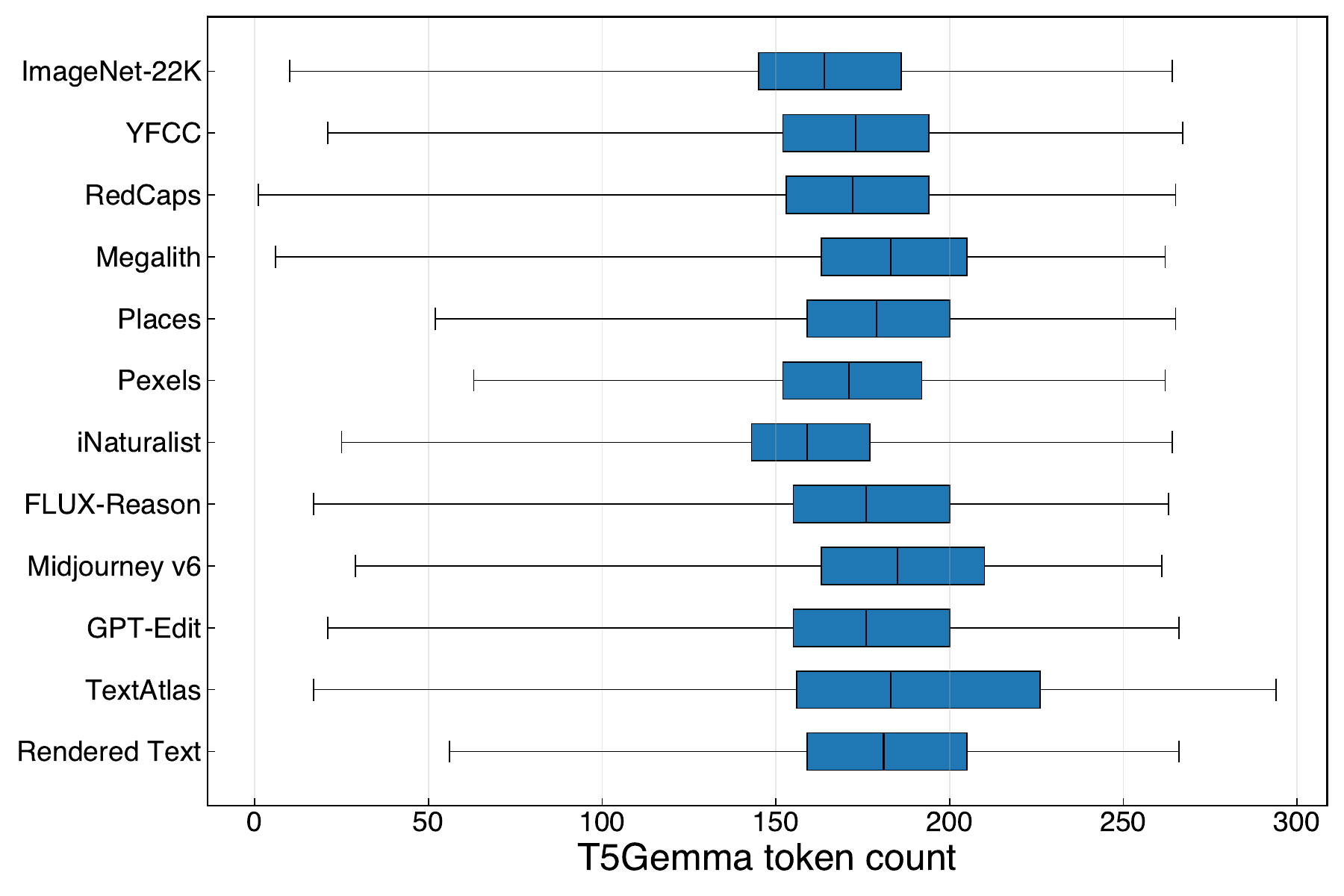}
    \caption{\textbf{Caption sequence length across datasets} for 10K random samples per dataset using the T5Gemma tokenizer.}
    \label{fig:caption_length_dataset}
\end{figure}

Here, we randomly sample 10K images from each dataset and plot the sequence length distribution of their synthetic captions in~\figref{fig:caption_length_dataset}. Although caption length distributions vary across datasets, no dataset differs substantially from the others.

\subsection{Meta-Prompt for Synthetic Captioning}

For most image datasets, we use the following minimal prompt for synthetic captioning:
\begin{promptbox}
Describe the image in detail using one paragraph.
\end{promptbox}

For text-rendering datasets, however, ground-truth text annotations are available. To reduce hallucinations in the VLM-generated captions, we include the ground-truth text in the captioning prompt.

For TextAtlas, we use:

\begin{promptbox}
Describe the image in detail in one paragraph. For reference, this is the ground truth annotation: "{annotation}"
\end{promptbox}

RenderedText consists of images of handwritten text rendered on digital 3D sheets of paper using Blender. The text varies in font size, color, and rotation, and the paper is rendered under random lighting conditions. Each ground-truth annotation contains a field, ``text'', which is a list of $n$ strings, where $n$ is the number of text lines in the image. The $i$-th element of the list corresponds to the $i$-th line of text. Therefore, for RenderedText, we use:

\begin{promptbox}
Describe the image in detail in one paragraph. For reference, there are {len(text)} lines of text in the image, and each line (comma-separated) is: {', '.join(text)}. In your description, include the transcription, font, size, color, location, and rotation angle for the text.
\end{promptbox}

\subsection{Sanity Check for Data Leakage}

For all datasets except FLUX-Reason, all captions are synthetically generated by a VLM, so overlap between the prompt sets used in our evaluation benchmarks and the captions used to train our text-to-image models is unlikely. However, for FLUX-Reason, we use five sets of captions: one set is the original ``caption\_detail'' field provided in the dataset (since it is already high-quality), and the remaining four are synthetically generated by us. According to the original paper~\citep{fang2026flux}, half of PRISM-Bench's prompts were directly selected from FLUX-Reason and then removed from the dataset. To ensure that there is no overlap between FLUX-Reason and PRISM-Bench, we perform exact string matching between each FLUX-Reason ``caption\_detail'' caption and each PRISM-Bench prompt. We find no overlap.

\end{document}